\documentclass{article} %
\usepackage{arxiv_v2,times}

\usepackage{hyperref}
\usepackage{url}
\usepackage{booktabs}
\usepackage{subcaption}

\usepackage{tikz}
\usetikzlibrary{arrows.meta, positioning, fit, backgrounds, calc, shapes.geometric}
\usepackage{amsmath}
\usepackage{amsfonts}
\usepackage{enumitem}
\usepackage{tcolorbox}
\tcbuselibrary{breakable,skins}
\usepackage{soul}
\usepackage{multirow}
\usepackage{pifont}
\usepackage{dsfont}

\newcommand{\cmark}{\text{\ding{51}}}
\newcommand{\xmark}{\text{\ding{55}}}

\newtcolorbox{findingbox}[1][]{
    enhanced,
    colback=yellow!10,
    colframe=gray!30,
    boxrule=0.5pt,
    arc=2pt,
    leftrule=3pt,
    rightrule=3pt,
    toprule=1pt,
    bottomrule=1pt,
    breakable,
    #1
}

\newtcolorbox{promptbox}[1][]{
    colback=gray!5,
    colframe=gray!60,
    boxrule=0.5pt,
    arc=2pt,
    fontupper=\small\ttfamily,
    before upper={\hyphenpenalty=10000},
    width=1.0\textwidth,
    #1
}

\newcommand{\prompthighlight}[2][yellow!30]{{%
    \colorlet{prompthighlightcolor}{#1}%
    \sethlcolor{prompthighlightcolor}%
    \hl{#2}%
}}

\title{Exploration and Exploitation Errors\\Are Measurable for Language Model Agents}

\author{Jaden Park$^{1,}$\footnote{Equal contribution} \qquad Jungtaek Kim$^{2,}$\footnotemark[1] \qquad Jongwon Jeong$^{1,}$\footnotemark[1] \qquad Robert D. Nowak$^{1,3}$\\
\textbf{Kangwook Lee}$^{3,4}$ \qquad \textbf{Yong Jae Lee}$^1$\\
$^1$University of Wisconsin--Madison \quad $^2$University of Arkansas \quad $^3$Ludo Robotics \quad $^4$KRAFTON}

\iclrfinalcopy %
\begin{document}

\maketitle

\begin{abstract}
Language Model (LM) agents are increasingly used in complex decision-making tasks, from AI coding to physical AI, where they must both explore a search space and exploit acquired knowledge effectively. Although exploration and exploitation play different roles in decision-making, it remains challenging to systematically distinguish exploration and exploitation without access to the agent's internal policy, and to determine when either constitutes an error. To address this, we design partially observable 2D grid maps paired with an unknown symbolic task directed acyclic graph (DAG), abstracting practical agentic AI scenarios. We then propose a policy-agnostic metric that detects unproductive traversals, while permitting necessary backtracking. We attribute each error to exploration or exploitation based on whether the current state admits productive exploration, exploitation, or both. Using generated maps and task DAGs with controllable difficulty, we evaluate 35 frontier LM agents and find that even state-of-the-art models struggle to reach the goal within the step budget. $\log$ exploration and exploitation errors exhibit strong relationships with task success ($R^2 = 0.885$ and $0.875$, respectively), providing empirical evidence that the metric captures behavior relevant to task success. Beyond success rate, the metric reveals distinct exploration and exploitation behaviors among models with similar success rates. We further find that a compact memory harness achieves higher task success than a reasoning baseline that retains its own reasoning traces. Finally, we extend our metric to an iconic video game, Minecraft Clone, beyond our synthetic environment. We release our code \href{https://github.com/jjj-madison/measurable-explore-exploit}{here}.
\end{abstract}

\begin{figure}[ht]
    \centering
    \begin{subfigure}[t]{0.495\textwidth}
        \centering
        \includegraphics[width=\textwidth]{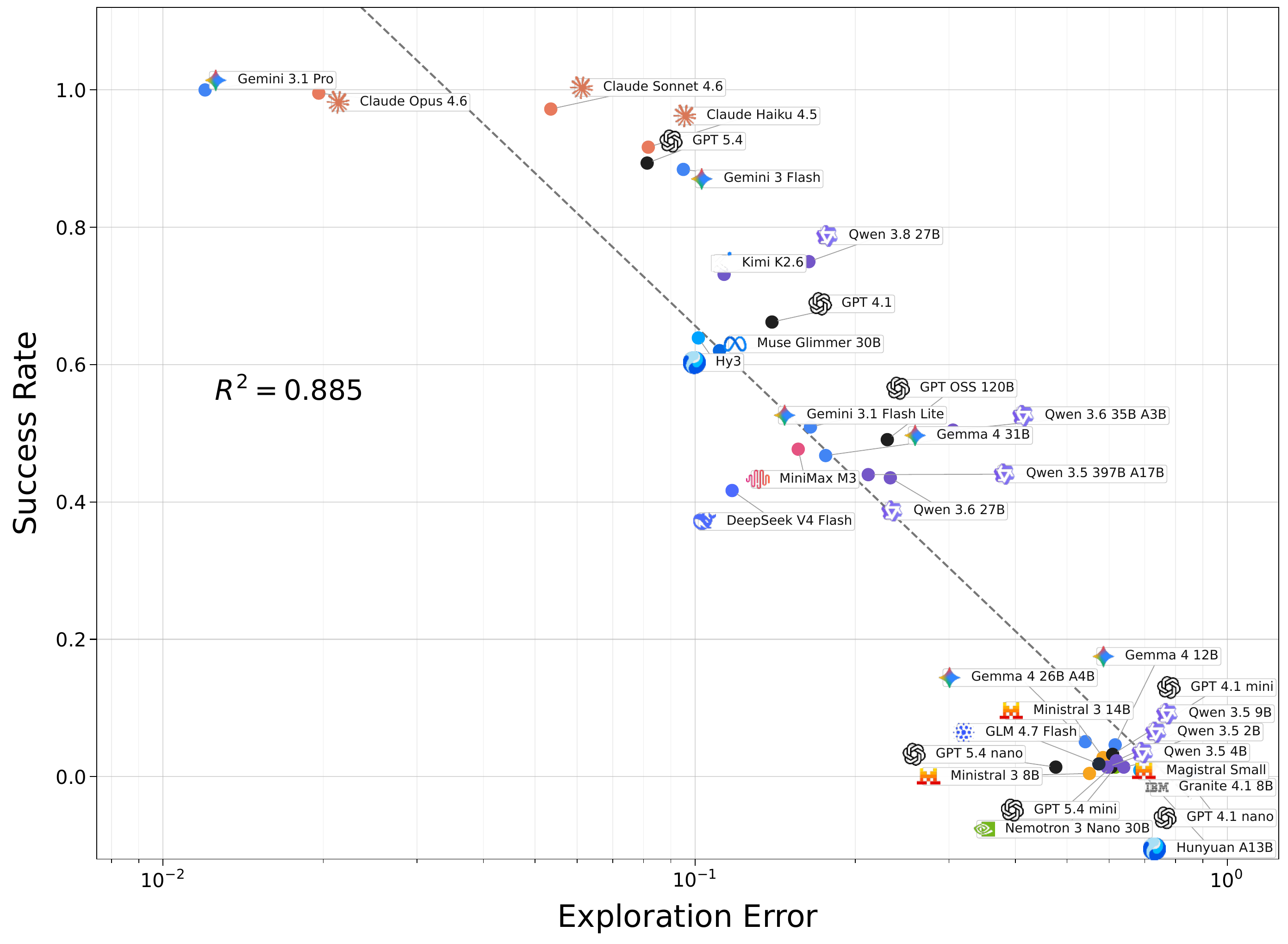}
        \caption{Success Rate vs. Exploration Error}
        \label{fig:success-vs-explore}
    \end{subfigure}
    \hfill
    \begin{subfigure}[t]{0.495\textwidth}
        \centering
        \includegraphics[width=\textwidth]{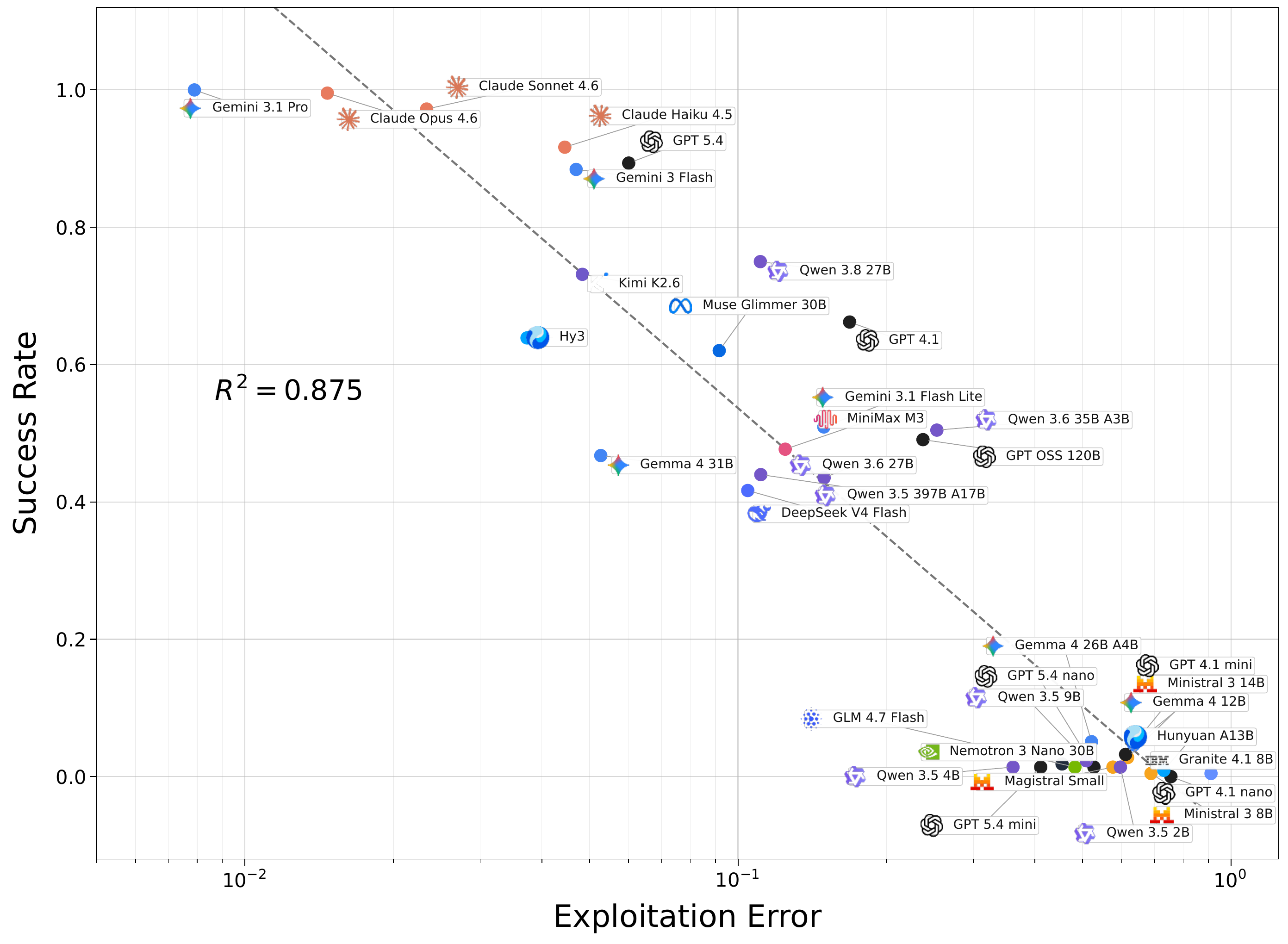}
        \caption{Success Rate vs. Exploitation Error}
        \label{fig:success-vs-exploit}
    \end{subfigure}
    \caption{Overall relationship between success rate and exploration or exploitation error across various LMs.
    In Figures~\ref{fig:success-vs-explore} and~\ref{fig:success-vs-exploit}, $\log$ exploration and exploitation errors show a strong negative linear relationship with success rate ($R^2 = 0.885$ and $0.875$, respectively). Full results, including the exploration and exploitation errors, and average budget used is reported in Appendix~\ref{apdx:full_results}.}
    \label{fig:success-vs-errors}
\end{figure}

\section{Introduction}
\label{sec:introduction}

Language model (LM) agents play a crucial role in many open-ended decision-making tasks, including AI coding~\citep{jimenez2024swe,jain2025livecodebench,merrill2026terminal}, workflow automation~\citep{anthropic2024mcp,Autoflow,Lopopolo2026,LeeY2026arxiv}, and physical AI~\citep{huang2022language,wang2023voyager,gubernatorov2025anywherevla}. By nature, these tasks require agents to explore unseen regions of the search space and exploit acquired knowledge effectively~\citep{harris2025should, KimJ2026tmlr, JeongJ2026aaiw}. Because they serve as different factors in decision-making, it is necessary to study the two factors separating them. Despite the growing adoption and strong performance of LM agents on various complex tasks, a systematic framework to assess and quantify how well these agents explore and exploit in practice does not yet exist.

In classical reinforcement learning~\citep{SuttonRS2018book}, exploration and exploitation are defined with respect to the agent's internal policy or value function. For LM agents, however, it is not straightforward because we typically only have access to the agent's observed actions. As a result, distinguishing and quantifying exploration and exploitation and identifying when either constitutes an error from behavior alone, without assuming a fixed strategy or access to the agent's internal policy, remains an open problem.

To address this, we introduce a policy-agnostic metric for quantifying exploration and exploitation errors from action trajectories alone. We instantiate tasks as partially observable 2D grid maps paired with unknown task directed acyclic graphs (DAGs) to enable systematic evaluation of exploration and exploitation from observed actions. This formulation abstracts the structure common to practical agentic tasks: navigating a partially observed space while achieving tasks with complex dependencies. A key design choice is replacing all semantic information in the task DAGs with symbolic representations. The agent must reason purely from the observed environment, effectively preventing pretrained semantic priors from confounding the evaluation.

Our error metric flags actions which no reasonable strategy would produce. Rather than specifying an optimal policy and penalizing deviation, we characterize the map state at each timestep and detect structurally redundant behavior within segments of the agent trajectory where no progress towards the task is being made. Our metric is inspired by classical graph-theoretic notions of redundancy and reuse~\citep{whitney1932, tarjan1972, edmonds1973, deng1999, panaite1999}, and attributes each error to exploration, exploitation or both.

With the proposed metric, we evaluate 35 frontier LM agents across diverse map configurations. The maps can be generated to require more exploration or exploitation by varying node density, corridor width, and task DAG size. We find that even state-of-the-art models often struggle on our task, and that log exploration and exploitation errors strongly correlate with task success. Moreover, models with similar success rates exhibit distinct exploration and exploitation behaviors. We further ablate the effects of various configurations including prompt types, semantic information, and explicit agent harnesses. Notably, we find that both exploration and exploitation can be improved through minimal harness engineering, achieving higher success than a reasoning baseline. Finally, we extend our metric to the clone version of a well-known video game, Minecraft Clone~\citep{ouyang2026gameworld}.

Summarizing, our contributions are threefold:
\begin{itemize}[leftmargin=*]
    \item We introduce a policy-agnostic metric for quantifying exploration and exploitation errors in LM agents from action trajectories, and extend it to Minecraft Clone;
    \item We design partially observable grid-map environments paired with unknown symbolic task DAGs, enabling systematic evaluation under varying exploration and exploitation demands;
    \item We evaluate 35 frontier LM agents, find that our errors strongly correlate with task success, identify distinct failure modes, and show that minimal harness engineering reduces both errors.
\end{itemize}

\section{Related Work}
\label{sec:related_work}

\paragraph{Language Model Agents.}
LM agents interact with an external environment in a multi-turn setting to solve complex problems~\citep{yao2023react, shinn2023reflexion}. To tackle a complex task, LM agents must explore the external environment to acquire new information and exploit it to achieve their goals. This has motivated a growing body of benchmarks that evaluate agent behavior across embodied interaction, software tasks, and tool-use settings~\citep{ALFWorld, jimenez2024swe, Appworld, merrill2026terminal}, while partially observable grid world environments have been adopted for more controlled and systematic analysis of the LM agents~\citep{chevalier2019babyai, chevalier2023minigrid}. More recently, \citet{zhang2026theory} assessed whether LM agents can construct and exploit spatial beliefs through exploration.

However, existing environments generally (i) rely on semantic information which confounds pretrained knowledge with in-environment reasoning, (ii) lack systematic control over task dependency structure, and (iii) do not separate and quantify exploration and exploitation errors from the LM agent trajectories. In this work, we isolate LM-agent reasoning through symbolic task DAGs, use controllable task DAG generation to vary dependency structure, and propose a metric that quantifies exploration and exploitation errors.

\paragraph{Evaluation Metrics for LM Agents' Behavior.}
Current agent evaluation methods predominantly rely on task success rates only~\citep{ALFWorld, merrill2026terminal}. While some recent studies propose more fine-grained metrics such as stepwise alignment with expected tool calls or progress compared to a reference trajectory~\citep{chen2024t, ma2024agentboard}, these methods implicitly assume access to annotated reference trajectories and thus a fixed optimal strategy. Furthermore, these works do not distinguish whether errors stem from exploration or exploitation. In contrast, our exploration and exploitation error metric, grounded in classical graph theory~\citep{whitney1932, tarjan1972, deng1999, panaite1999}, does not rely on a particular reference trajectory. Instead, our proposed metric structurally analyzes the map state at each timestep to detect actions that no reasonable strategy would produce, and attributes each error to exploration, exploitation, or both based on the map state. It is therefore policy-agnostic and can be used to evaluate diverse LM agents that may internally follow different strategies. On the other hand, the behavior of LLM reasoning has been discussed in the work by~\citet{ZengY2026icml}, devising their own metrics on the tree representation of reasoning traces.

\section{Task Formulation}
\label{sec:formulation}

\begin{figure}[t]
\begin{center}
\includegraphics[width=0.95\textwidth]{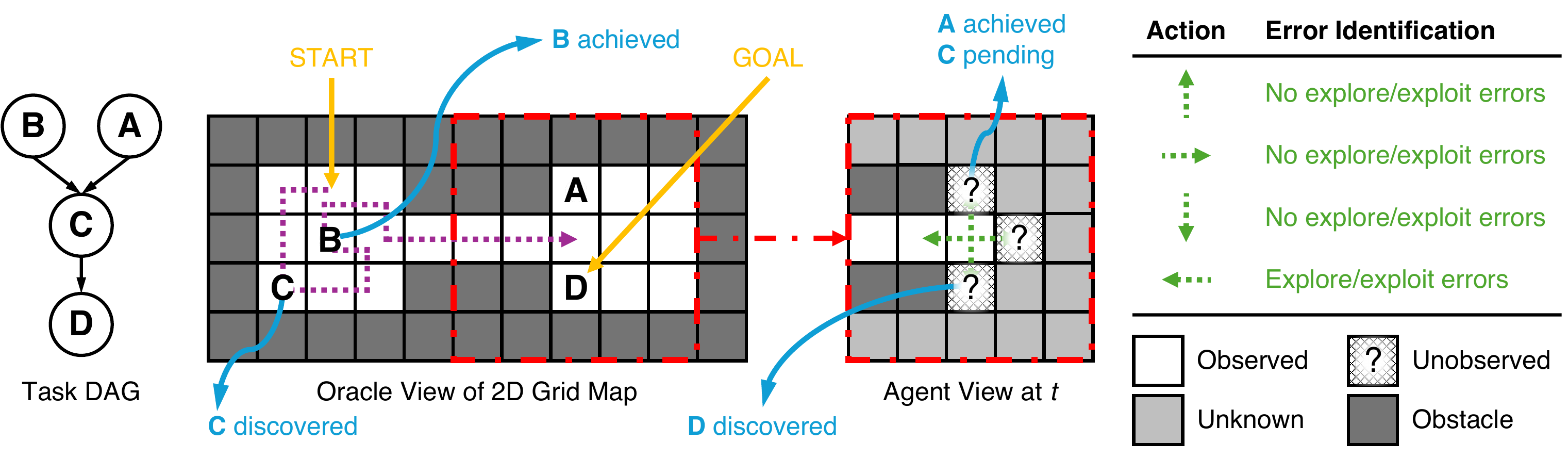}
\end{center}
\caption{Illustration of the LM agent traversing the 2D grid map to achieve the goal given by the task DAG. In each timestep, the environment returns the set of admissible moves and the information about the task node, if discovered. The LM agent must retain relevant information about the map geometry, the task DAG structure and the states of the discovered nodes to achieve the goal node.}
\label{fig:overall}
\end{figure}

\begin{figure}[t]
\begin{center}
\includegraphics[width=0.95\textwidth]{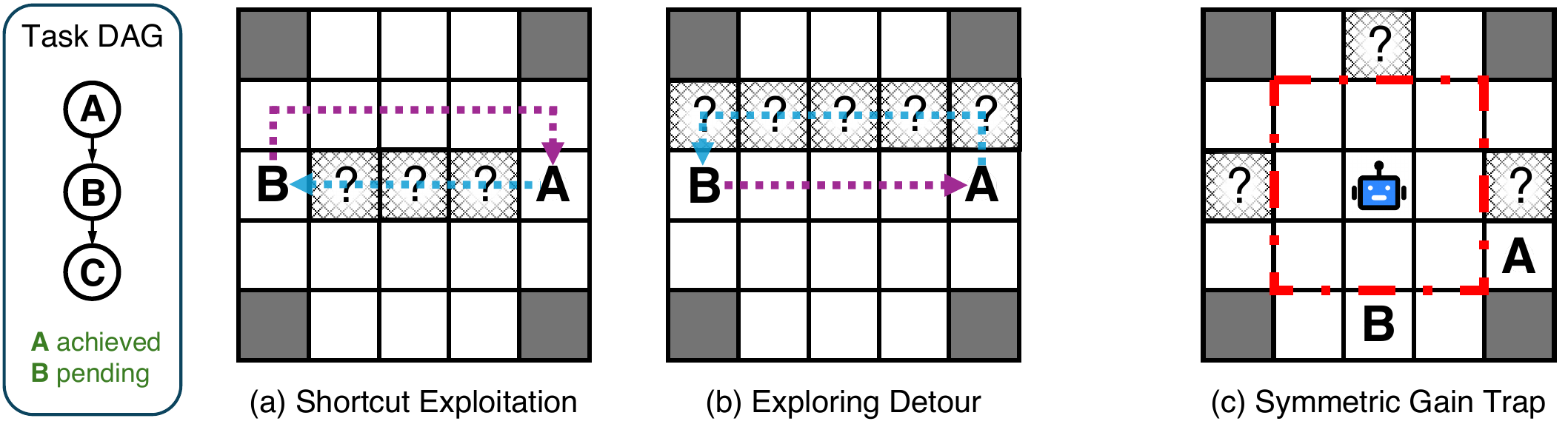}
\end{center}
\caption{Three illustrative examples demonstrating edge cases. In all three maps, node A is achieved and node B is pending, so the agent can traverse back to B, or explore to discover node C. The purple dashed line depicts the path the LM agent has taken, and the blue transparent line denotes a potentially more optimal path. In Panel~(a), the LM agent can infer the shortest path to node A and traverse unseen nodes as part of an exploitation strategy. In contrast, in Panel~(b), the agent has already taken the shortest path from node B to node A. Hence, it may be more optimal to make a short detour and observe 5 unseen cells at the cost of a short detour. Panel~(c) illustrates a case where target cells $\mathcal{T}(t)$ are placed symmetrically. Here gain will always be 1 for all four actions. A metric solely relying on gain cannot penalize the LM agent wandering indefinitely in the red dotted region.}
\label{fig:edge_case}
\end{figure}

In this section, we formally define the design of our framework and show that the specific design choices in our framework allow us to distinguish and quantify exploration and exploitation. At a high level, the LM agent must traverse the 2D grid map to discover relevant task nodes, use its accumulated knowledge to satisfy their prerequisites, and ultimately achieve the goal node. For a more mathematical definition, refer to Appendix~\ref{apdx:eqn}.

\paragraph{2D Grid Map.} We define our partially observable map $\mathcal{M}$ as the set of traversable\footnote{For example, obstacles are not included in $\mathcal{M}$.} cells in a 2D grid, analogous to real-world or video game settings where movement gradually reveals local spatial information and any task-relevant entities present. More formally, $\mathcal{M} \subset \mathbb{N}^2$ where each element in $\mathcal{M}$ is a coordinate of a cell.\footnote{We assume that $0\in \mathbb{N}$.} For example, $\mathcal{M} = \{0,1,2\}^2$ denotes a fully traversable $3\times3$ 2D grid. The map allows \texttt{up}, \texttt{down}, \texttt{left}, \texttt{right} as movements.

Each time the agent visits a cell, the environment provides a subset of \texttt{up}, \texttt{down}, \texttt{left}, \texttt{right} as admissible moves, which provide information about the neighboring cells (Figure~\ref{fig:overall}). We say a cell $c$ is \textbf{observed} if the agent has visited the cell $c$ at some previous timestep. The neighboring cells that have not been previously observed are denoted \textbf{unobserved}. Critically, information about the task DAG is only revealed when the cell is observed. The agent must traverse to unobserved cells to discover if task nodes are present in those cells. If no task node is present, the cell is simply marked as observed. Finally, cells that are neither observed nor unobserved are \textbf{unknown}; see Equation~\eqref{eqn:state} for formal definitions.

\paragraph{Task DAG.} Complex real-world tasks can be decomposed into sub-tasks with precedence constraints, and are naturally modeled as DAGs,  where sub-tasks are nodes and edges define the precedence relations between the nodes. For example, cooking tomato pasta with cheese requires finding tomato sauce, pasta, and cheese, and mixing them in the right order; see Figure~\ref{fig:semantic_example_1}. Without loss of generality, we assume a unique goal node in our setting.\footnote{If there are multiple goal nodes, we can append a dummy sink node.}

Denote $l(\cdot)$ as an injective function between the task node and its location. We say a node $u$ is \textbf{visited} if the agent is currently located at $l(u)$. If the agent has found node $u$ at some previous timestep, a node $u$ has been \textbf{seen}. With this, we define the three states each node has: $\{\text{undiscovered}, \text{discovered}, \text{achieved}\}$. A node is \textbf{undiscovered} if it has not been seen. A node is \textbf{discovered} if it has been seen but the preconditions are not met. A node is \textbf{achieved} only if the agent has visited the node after satisfying the preconditions; see Equation~\eqref{eqn:status}. Finally, each node has two types of preconditions: $\{\texttt{AND},\texttt{OR}\}$ where \texttt{AND} indicates that all of its parent nodes must be achieved, and \texttt{OR} indicates that only one of its parent nodes needs to be achieved in order to achieve the current node; see Equation~\eqref{eqn:prec}.

Upon visiting a cell in the 2D grid map, the environment provides the set of admissible movements. If a node is discovered, the environment provides information about its parent and child nodes. The locations of the revealed nodes remain unknown and must be discovered by the agent. The task is complete when the goal node is achieved.

\paragraph{Agent Harness.} In each interaction with the environment, the following information must be tracked by the LM agent to effectively solve the task.
\begin{itemize}[leftmargin=*]
    \item Map information: list of observed and unobserved cells and obstacles, if present.
    \item Task information: list of achieved and discovered task nodes, and their preconditions.
\end{itemize}
In practice, LM agents operating over long horizons benefit from structured summaries of accumulated state, rather than relying solely on raw context history~\citep{anthropic2025}. Hence, we can provide a subset of the available information and define this as the \textbf{agent harness}. Following the philosophy of ReAct~\citep{yao2023react}, we provide minimal information to the agent by default and assume that a skilled agent would maintain relevant context history and make rational decisions at each timestep. In Section~\ref{sec:app_harness_example}, we ablate explicitly providing the full harness to the agent. Providing the harness explicitly is analogous to equipping the agent with an anchored map~\citep{dessmark2004} or an advice string~\citep{fraigniaud2008, dobrev2012}, where structured information about the environment is made directly available rather than requiring reconstruction from context.

\section{Measuring Exploration and Exploitation Errors}
\label{sec:expl}

In this section, we formally define our metric for exploration and exploitation errors. 

We define \textbf{pending tasks} $\mathcal{P}(t)$ which denotes the nodes in the task DAG with prerequisites satisfied and locations known, \emph{making them ready to be achieved} (see Equation~\eqref{eqn:pending}). This reflects many real-world tasks where, once the prerequisite conditions are satisfied, progress requires traversing back to the relevant location to act. Intuitively, the LM agent can exploit its knowledge to achieve the pending tasks. The exploration counterpart to $\mathcal{P}(t)$ is the set of unobserved cells $\mathcal{U}(t)$, which the agent must traverse to in order to discover new information. We refer to the set of productive destinations as the \textbf{target set} $\mathcal{T}(t)$, which varies with the agent's state and determines the required action at timestep $t$ (Table~\ref{tab:four-cases}).

\begin{table}[t]
\caption{Four cases determining the type of required action at timestep $t$.}
\label{tab:four-cases}
\centering
\scalebox{0.90}{%
\begin{tabular}{clcc}
\toprule
\textbf{Case} & \textbf{Condition} & \textbf{Target set} $\mathcal{T}(t)$ & \textbf{Required action} \\
\midrule
1 & $\mathcal{P}(t) = \emptyset$ 
  & $\mathcal{U}(t)$ 
  & Exploration \\
2 & $g \in \mathcal{P}(t)$ 
  & $\{l(g)\}$ 
  & Exploitation \\
3 & $\mathcal{P}(t) \neq \emptyset,\; g \notin \mathcal{P}(t),\; \mathcal{U}(t) = \emptyset$ 
  & $\{l(u) : u \in \mathcal{P}(t)\}$
  & Exploitation \\
4 & $\mathcal{P}(t) \neq \emptyset,\; g \notin \mathcal{P}(t),\; \mathcal{U}(t) \neq \emptyset$ 
  & $\mathcal{U}(t)\cup \{l(u) : u \in \mathcal{P}(t)\}$ 
  & Either \\
\bottomrule
\end{tabular}%
}
\end{table}

That is, we assume that the agent must explore when $\mathcal{U}(t)\neq \emptyset$ and exploit when $\mathcal{P}(t)\neq \emptyset$, with the exception when the goal task is pending, where we only allow exploitation.

We aim to identify actions that no reasonable strategy would produce as errors. We say that an action is a \textbf{gain} (Equation~\eqref{eqn:gain}) if it steps into a target cell or reduces the minimum distance to at least one of the target cells ($\text{Gain}(t\to t+1) = 1$), and define the action as an error otherwise ($\text{Gain}(t\to t+1) = 0$). However, as shown in Figure~\ref{fig:edge_case}c, gain alone is insufficient to classify errors: when targets exist symmetrically, the agent can oscillate indefinitely.

To address this, we define a \textbf{no-progress trajectory} $\tau_{\text{np}}(t)$ as the sequence of actions since the most recent progress event, where a \textbf{progress event} is either achieving a pending task or entering an unobserved cell. $\tau_{\text{np}}(t)=\emptyset$ once progress is made. We track the history of nodes and edges the agent has traversed in each $\tau_{\text{np}}(t)$ and denote them $\mathcal{V}_{\text{np}}$ and $\mathcal{E}_{\text{np}}$ respectively, and compute the following three quantities:
\begin{itemize}[leftmargin=*]
    \item $c_t = |\mathcal{E}_{\text{np}}|-|\mathcal{V}_{\text{np}}|+1$, the cyclomatic number of the current no-progress trajectory.
    \item $e_t = \sum_{e\in \mathcal{E}_{\text{np}}} \mathrm{max}\{m_{\text{np}}(e)-2,0\}$, where $m_{\text{np}}(e)$ is the traversal count of $e$ in $\tau_{\text{np}}$.
    \item $n_t = \sum_{v\in \mathcal{V}_{\text{np}}} \mathrm{max}\{m_{\text{np}}(v)-2,0\}$, where $m_{\text{np}}(v)$ is the visit count of $v$ in $\tau_{\text{np}}$.
\end{itemize}
Intuitively, $c_t$ increases when the agent closes a new loop in explored territory~\citep{whitney1932}. $e_t$ and $n_t$ increase when the edge or node is reused beyond benign backtracking, respectively. More formally, the budget of 2 for edges is motivated by classical graph exploration, where optimal online exploration of an undirected graph traverses each undirected edge at most twice~\citep{tarjan1972, edmonds1973, panaite1999, deng1999}. The budget of 2 for nodes is the node-level analog: the tightest constant that permits a single gateway revisit without penalty. Although we do not enforce optimal traversal, we assume that a competent agent should avoid structurally redundant actions that yield no new information. Empirically, this formulation captures a wide range of failure modes, some of which are listed here in Appendix~\ref{apdx:eqn}.

Now we define the \textbf{stale score} $S_t = c_t+e_t+n_t$ and flag error when the stale score increases, i.e., $\mathds{1}\{S_t > S_{t-1}\}$.\footnote{For detailed description about this metric, please refer to Appendix~\ref{apdx:eqn}.} Combining the above, we define the error metric at timestep $t$:
\begin{equation}
    \mathrm{err}(t) = \begin{cases}
    0, & \text{if } p(t) \to p(t+1) \text{ is a progress event}, \\
    1, & \text{if } \text{Gain}(t \to t+1) = 0, \\
    0, & \text{if } |\mathcal{T}(t)| = 1 \land \text{Gain}(t \to t+1) = 1, \\
    \mathds{1}\{S_t > S_{t-1}\}, & \text{if } |\mathcal{T}(t)| > 1 \land \text{Gain}(t \to t+1) = 1.
    \end{cases}
    \label{eqn:error}
\end{equation}
The stale score is only required when more than one cell exist in the target set $\mathcal{T}(t)$;
see Figure~\ref{fig:edge_case}c. Case 2 implies $|\mathcal{T}(t)|=1$. When $\mathrm{err}(t) = 1$, we attribute the error to exploration, exploitation or both, according to the required action in Table~\ref{tab:four-cases}: Case 1 increments exploration error, Case 2 and Case 3 increment exploitation error, and Case 4 increments both errors. For more detailed discussion on what each of the four cases are, see Appendix~\ref{apdx:eqn}.

\section{Experimental Setup}\label{sec:exp_setup}

We cover detailed settings for our experiments. For full details on task generation and data statistics, refer to Appendices~\ref{sec:app_task_generation} and~\ref{sec:app_data_statistics}, respectively.

\paragraph{Task Generation.} We first generate task DAGs, then lay them out on 2D grid maps. Then, we place the nodes on the 2D grid, by varying the following parameters: (i) the number of task nodes; (ii) maximum number of nodes at each depth; (iii) the node density over grid cells. We place obstacle cells to control the width of the path from one node to another. This implicitly controls the degree of exploration and exploitation needed.

\paragraph{Models.}
We evaluate 35 LMs. The full list is in the Appendix~\ref{apdx:full_results}.

\paragraph{Prompts.}
By default, we follow ReAct~\citep{yao2023react} framework and consider four prompt variants: \textbf{base}, \textbf{exploration}, \textbf{exploitation}, and \textbf{balance}. All four variants share an identical environment description and action format; they differ only in a single strategy sentence that instructs the agent to prioritize exploration, exploitation, or balance the two. The base prompt provides no strategic guidance, leaving the exploration--exploitation tradeoff entirely to the model's internal context and reasoning capabilities. Full prompt templates and example interactions are provided in Appendix~\ref{sec:app_prompts}.

\paragraph{Evaluation.}
We report \textbf{success rate}, \textbf{exploration error}, and \textbf{exploitation error}. Success rate measures the fraction of episodes the agent achieves the goal within the step budget. Error metrics are normalized by the number of timesteps in which each action type is required: exploration error over Cases 1 and 4, and exploitation error over Cases 2, 3, and 4;
see Table~\ref{tab:four-cases}.

\section{Experimental Results}
\label{sec:experimental_results}

Figure~\ref{fig:success-vs-errors} illustrates our main results comparing the success rate versus exploration or exploitation error with various frontier LM agents. Stronger reasoning models consistently achieve higher performance, with the best models reaching up to 100\% success rate. These results exhibit strong negative relationships between success rate and both exploration and exploitation errors, with exploration error showing a marginally stronger association. This is consistent with the structure of our task: agents must first discover task-relevant nodes through exploration and subsequently exploit acquired knowledge to satisfy task dependencies and reach the goal. Persistent failures in either process can therefore limit task completion, although exploration failures may act as an earlier bottleneck by preventing relevant states from being discovered. Therefore, we argue the following:
\begin{findingbox}
    \textbf{\emph{Finding 1}}: Low exploration and exploitation error rates are strong predictors of success.
\end{findingbox}

\begin{figure}[t]
\centering
\includegraphics[width=\textwidth]{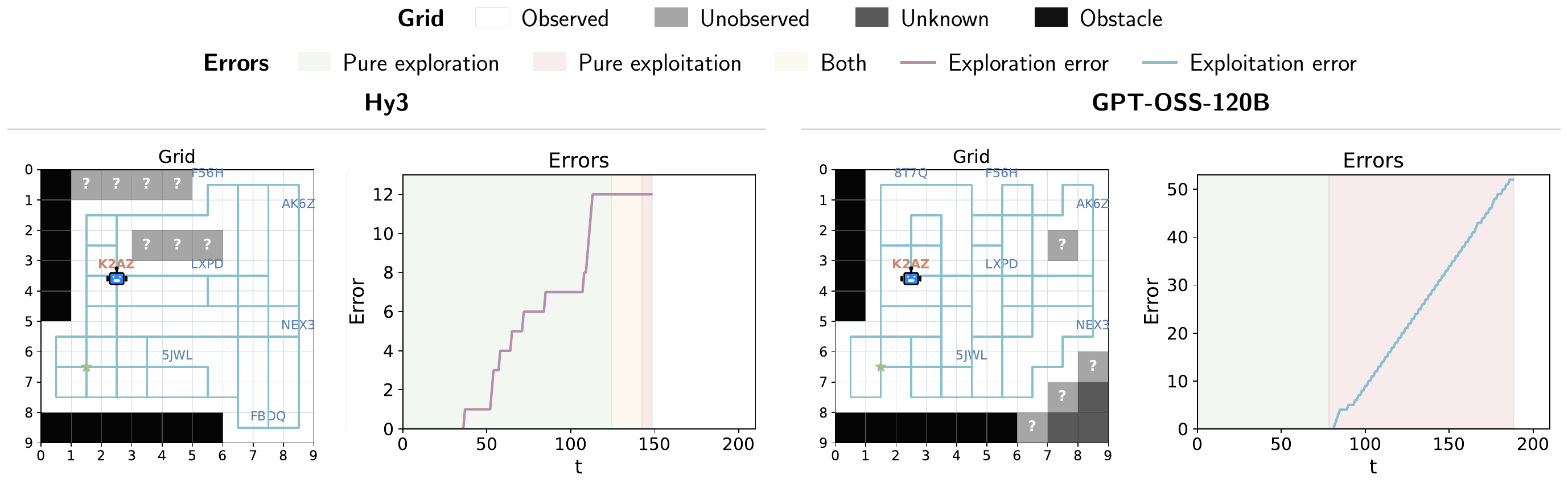}
\caption{Illustration of the trajectories of Hy3 and GPT-OSS-120B where both models succeed but demonstrate different error behavior. Hy3 reaches the goal in 148 steps with 12 no exploitation errors whereas GPT-OSS-120B takes 188 steps with no exploration errors. Each model's final grid is depicted on the left alongside the step-wise cumulative errors.}
\label{fig:behavior_pair}
\end{figure}

Table~\ref{tab:behavior_pair} and Figure~\ref{fig:behavior_pair} illustrate an example where two models show opposite error behaviors despite both succeeding on the map. Both Hy3 and GPT-OSS-120B models achieve the goal within the budget of 210 steps, yet exhibit markedly different behaviors.

We observe that Hy3 incurs an exploration error of 8.1\%, but exploits cleanly without any exploitation error, and reaches the goal in 148 steps. In contrast, GPT-OSS-120B explores without any exploration error, but exploits poorly with 47.3\% exploitation error. It discovers the goal node at step 22 and satisfies all of its prerequisites by step 78, after which the goal is pending and the only productive action is to return to it (Case 2 in Table~\ref{tab:four-cases}). Nevertheless, the agent spends another 110 steps wandering before reaching the goal at step 188, incurring an exploitation error of 47.3\%.

This example shows that two trajectories with the same outcome can be inefficient in opposite ways: Hy3 is limited by exploration, whereas GPT-OSS-120B is limited by exploitation. Our metric separates the two failure modes, which success rate alone cannot capture. Thus, we find the following:

\begin{findingbox}
    \textbf{\emph{Finding 2}}: LM agents can exhibit different exploration and exploitation behaviors beyond success.
\end{findingbox}

\begin{table}[t]
\caption{Hy3 and GPT-OSS-120B on the same map without reasoning. Both models succeed, but Hy3 incurs only exploration errors, whereas GPT-OSS-120B incurs only exploitation errors.}
\label{tab:behavior_pair}
\centering
\scalebox{0.90}{%
\begin{tabular}{lccc}
\toprule
\textbf{Model} & \textbf{Steps / Budget} & \textbf{Exploration Error} & \textbf{Exploitation Error} \\
\midrule
Hy3 & 148 / 210 & 8.1\% (12/148) & 0.0\% (0/24) \\
GPT-OSS-120B & 188 / 210 & 0.0\% (0/188) & 47.3\% (52/110) \\
\bottomrule
\end{tabular}%
}
\end{table}

\begin{figure}[t!]
    \centering
    \begin{subfigure}[b]{0.32\textwidth}
        \centering
        \includegraphics[width=\textwidth]{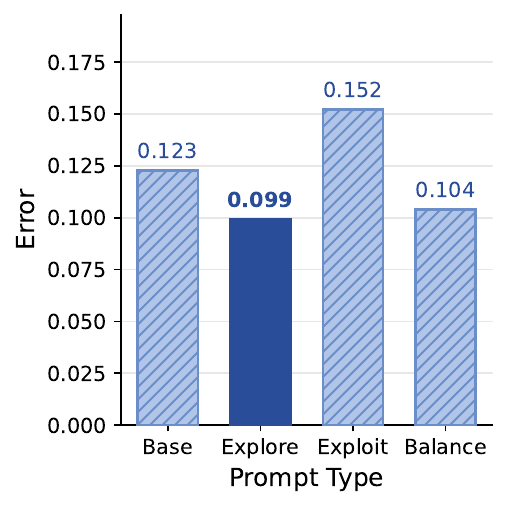}
        \caption{Exploration Error}
        \label{fig:prompt_explore_error}
    \end{subfigure}
    \hfill
    \begin{subfigure}[b]{0.32\textwidth}
        \centering
        \includegraphics[width=\textwidth]{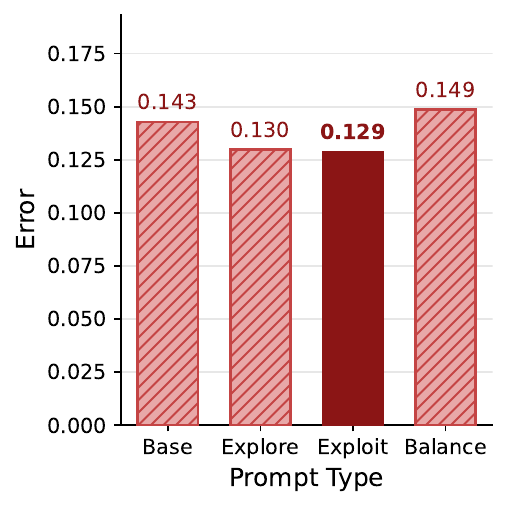}
        \caption{Exploitation Error}
        \label{fig:prompt_exploit_error}
    \end{subfigure}
    \hfill
    \begin{subfigure}[b]{0.32\textwidth}
        \centering
        \includegraphics[width=\textwidth]{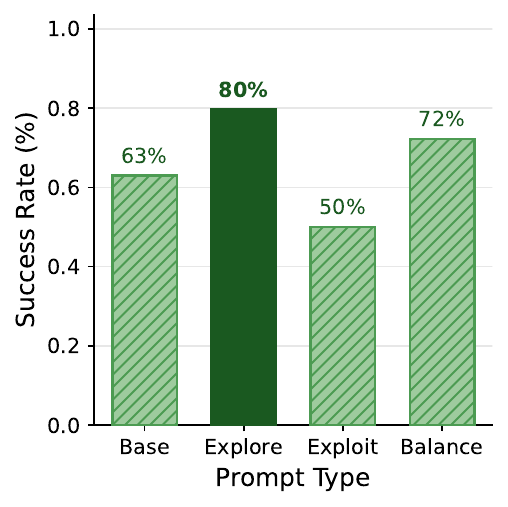}
        \caption{Success Rate}
        \label{fig:prompt_success_rate}
    \end{subfigure}
    \caption{Effect of prompts on exploration and exploitation error, and success rate for GPT-4.1. The highlighted bar in each panel indicates lowest error or highest success rate. The exploration-focused prompt achieves the lowest exploration error and the highest overall success rate, while the exploitation-focused prompt achieves the lowest exploitation error.}
    \label{fig:prompt_effects}
\end{figure}

We analyze the impact of providing prompts emphasizing exploration or exploitation by testing four prompt variants: base, exploration-focused, exploitation-focused, and balanced prompts. As depicted in Figure~\ref{fig:prompt_effects}, exploration- and exploitation-focused prompts reduce exploration and exploitation errors, respectively. The highest success rate is achieved with exploration-focused prompts, consistent with the results in Figure~\ref{fig:success-vs-errors} and Finding 1. Hence, we state the following:
\begin{findingbox}
    \textbf{\emph{Finding 3}}: Exploration-focused or exploitation-focused prompts reduce exploration and exploitation errors, respectively.
\end{findingbox}

\begin{table}[t]
\caption{Effect of harness engineering on success rates, exploration and exploitation errors, and average budget used. Harness engineering provides a structured memory summary and the model outputs actions without reasoning (Appendix~\ref{sec:app_harness_example}). Compared to reasoning, it achieves higher success rates, lower exploration errors, and lower total budget usage for both models.}
\label{tab:harness_engineering}
\centering
\resizebox{\textwidth}{!}{%
\begin{tabular}{llccccc}
\toprule
\multirow{2}{*}{\textbf{Model}} & \multirow{2}{*}{\textbf{Method}} &
\multirow{2}{*}{\textbf{Success} (\%) $\uparrow$} &
\textbf{Exploration} &
\textbf{Exploitation} &
\textbf{Avg. Budget Used} &
\textbf{Avg. Budget Used} \\
&&&
\textbf{Error} $\downarrow$ &
\textbf{Error} $\downarrow$ &
\textbf{(Total)} (\%) $\downarrow$ &
\textbf{on Success} (\%) $\downarrow$ \\
\midrule
Gemini 3.1
  & Reasoning
  & 50.0 & 0.190 & 0.193 & 77.1 & \textbf{54.2} \\
Flash Lite
  & Harness Engineering
  & \textbf{76.9} & \textbf{0.144} & \textbf{0.083} & \textbf{64.9} & 54.3 \\
\midrule
\multirow{2}{*}{GPT-4.1}
  & Reasoning
  & 66.7 & 0.159 & \textbf{0.227} & 72.6 & 58.9 \\
  & Harness Engineering
  & \textbf{72.2} & \textbf{0.121} & 0.293 & \textbf{69.2} & \textbf{57.4} \\
\bottomrule
\end{tabular}
}
\end{table}

As denoted in Section~\ref{sec:formulation}, one can design an explicit memory by acquiring relevant information the environment returns, which can be viewed as an external memory management system in the context of harness engineering; see Appendix~\ref{sec:app_harness_example} for an example, instead of relying on the LM agent's internal context and reasoning. We observe that providing structured harness to the LM agents, even without reasoning, improves success rates, exploration errors, and total budget usage compared to reasoning, except for the exploitation error of GPT-4.1; see Table~\ref{tab:harness_engineering}.

\begin{findingbox}
    \textbf{\emph{Finding 4}}: Harness engineering without reasoning outperforms reasoning by utilizing historical outcomes.
\end{findingbox}

\begin{table}[t]
\caption{Effect of semantic information for Gemini 3.1 Flash Lite and GPT-4.1. Semantic information reduces exploitation errors for both models, while exploration errors remain similar. Budget usage is the mean percentage of each map's hidden budget consumed, reported for successful trajectories and all trajectories, respectively. Each condition pools 40 evaluations over four maps, five seeds, and the default and reasoning evaluation.}
\label{tab:semantic_map}
\centering
\resizebox{\textwidth}{!}{%
\begin{tabular}{lcrrrr}
\toprule
\textbf{Model} & \textbf{Semantic} & \textbf{Success (\%)} $\uparrow$ & \textbf{Exploration Error} $\downarrow$ & \textbf{Exploitation Error} $\downarrow$ & \textbf{Avg. Budget Used (Success / Total)} $\downarrow$ \\
\midrule
  \multirow{2}{*}{Gemini 3.1 Flash Lite} & $\xmark$ & 0.0 & 0.305 & 0.405 & N/A / 100.0\% \\
   & $\cmark$ & \textbf{25.0} & \textbf{0.291} & \textbf{0.061} & \textbf{80.1\%} / \textbf{95.0\%} \\
\midrule
  \multirow{2}{*}{GPT-4.1} & $\xmark$ & 25.0 & \textbf{0.275} & 0.456 & 74.0\% / 93.5\% \\
   & $\cmark$ & \textbf{27.5} & 0.284 & \textbf{0.219} & \textbf{68.0\%} / \textbf{91.2\%} \\
\bottomrule
\end{tabular}
}
\end{table}

In our main experiments, we removed semantic information in our task DAG generation to isolate LM's semantic priors. In Table~\ref{tab:semantic_map}, we reintroduce semantic information by evaluating agents on simple pasta-cooking tasks (see Figure~\ref{fig:semantic_example_1} for an example; experimental details are provided in Appendix~\ref{sec:app_semantic_experiment}). For both models, semantic information substantially reduces exploitation error ($0.405 \to 0.061$ and $0.456 \to 0.219$), while exploration error remains similar. This suggests that semantic information helps LM agents infer and remember the task dependencies. However, semantic information does not reveal the position of the nodes, so exploration is still required:

\begin{findingbox}
    \textbf{\emph{Finding 5}}: Reintroducing semantic information reduces exploitation errors, but not exploration errors.
\end{findingbox}

\begin{figure}[ht]
  \centering
  \begin{subfigure}[b]{0.49\textwidth}
    \centering
    \includegraphics[width=\textwidth,trim={0px 24px 0px 0px},clip=true]{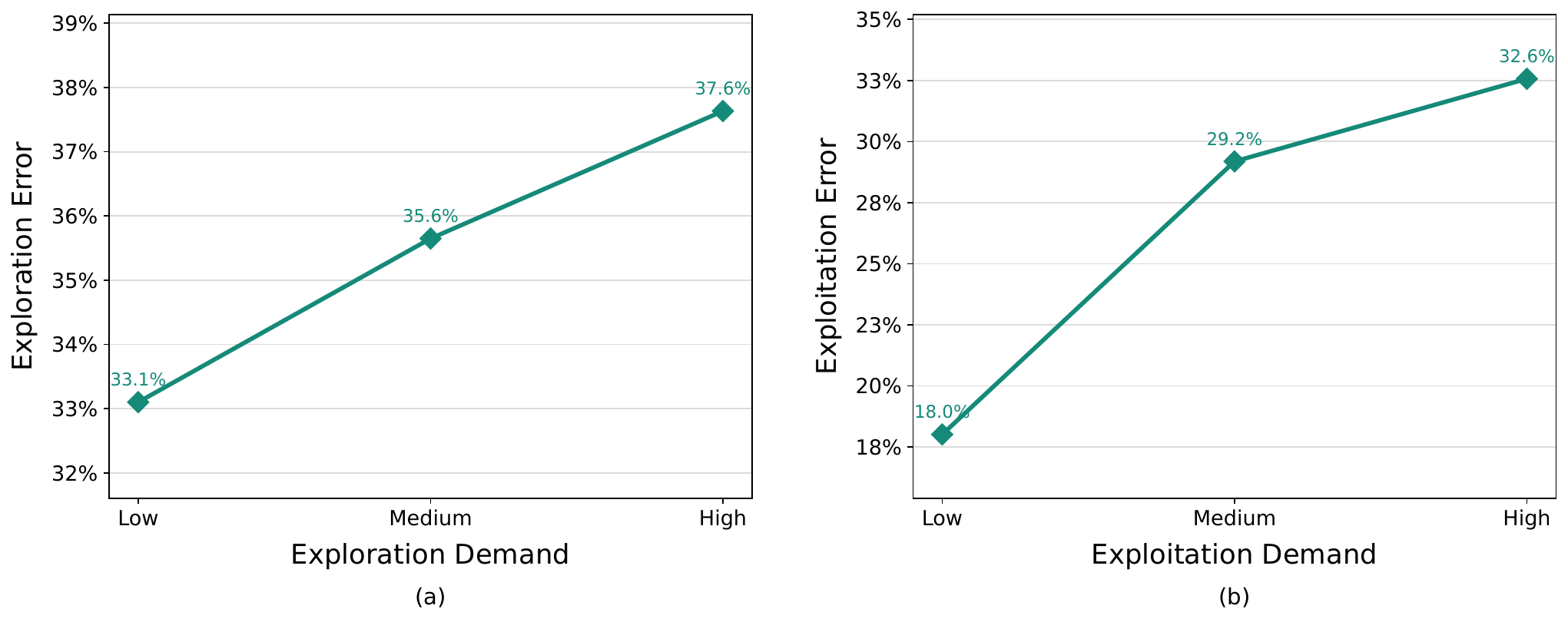}
    \caption{By Exploration Demand (by map design)}
    \label{fig:errors-by-exploitation-level}
  \end{subfigure}
  \hfill
  \begin{subfigure}[b]{0.49\textwidth}
    \centering
    \includegraphics[width=\textwidth,trim={0px 22px 0px 0px},clip=true]{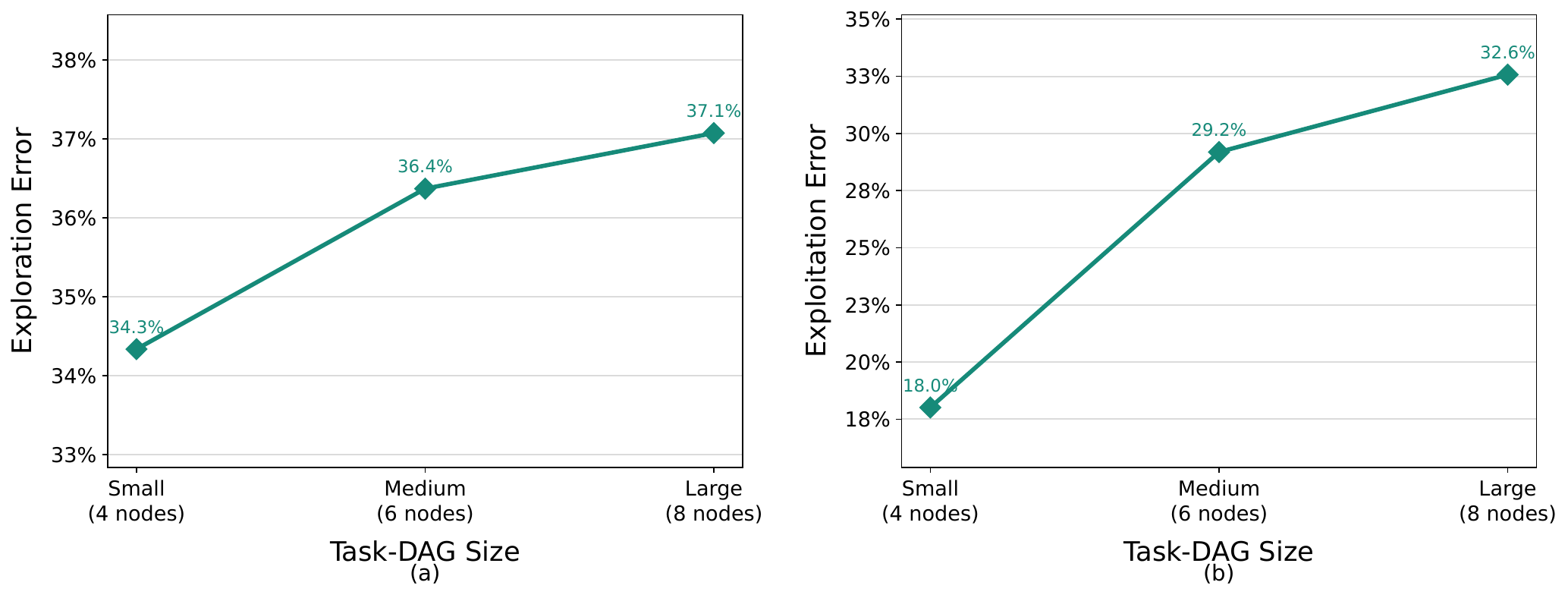}
    \caption{By Task DAG Size}
    \label{fig:errors-by-task-dag-difficulty}
  \end{subfigure}
  \caption{Exploration and exploitation errors averaged over 35 models, with varying (a) exploration demand, controlled by map design, and (b) task DAG size. Each model is evaluated on 216 episodes. Errors increase monotonically with both exploration demand and task DAG size. See Appendix~\ref{sec:app_task_generation}.}
  \label{fig:difficulty-level-errors}
\end{figure}

Figure~\ref{fig:difficulty-level-errors} shows how exploration and exploitation errors vary with map-level demand and task DAG size. Both errors increase monotonically with their corresponding demand levels: exploration error rises as exploration demand increases and exploitation error rises with exploitation demand. Similarly, increasing the task DAG size increases both exploration and exploitation errors as it requires discovering and achieving more nodes. These results indicate that the proposed metrics respond systematically to controlled increases in the exploration and exploitation requirements of the task:

\begin{findingbox}
    \textbf{\emph{Finding 6}}: Exploration and exploitation errors increase systematically with their corresponding task demands and with task DAG size.
\end{findingbox}

\section{Discussion and Limitations}\label{sec:disc}

\paragraph{On the use of symbolic abstraction.}
Our environment is intentionally not a full reproduction of real-world scenarios. As noted in Section~\ref{sec:introduction}, we remove semantic information and use symbolic task DAGs to isolate the raw exploration and exploitation capabilities of LM agents (the exact symbolic node generation process is detailed in Appendix~\ref{sec:app_task_generation}). However, a key limitation is that many real-world tasks will not appear in such a fully semantics-free form; in practice, agents are often expected to use semantic information and pretrained knowledge rather than operate in environments deliberately detached from them. This can be viewed as a limitation if the goal is to measure end-to-end real-world utility, since many practical applications allow and even require agents to leverage domain knowledge and priors to solve complex tasks more effectively. Indeed, our semantic reintroduction experiments show that such information can substantially affect agent behavior (Finding 5). At the same time, this is precisely why the symbolic abstraction is useful: it lets us test whether an agent can genuinely explore, maintain relevant memory and state information, and act on newly actionable knowledge when these behaviors must be inferred from interaction history alone rather than from semantic shortcuts. In this sense, our task is best viewed as a controllable test suite to measure raw exploration and exploitation capabilities that complement evaluations in more realistic semantic settings.

\paragraph{Per-model variance and the error metric.}
Another limitation is that our exploration and exploitation errors reported in the tables are inherently trajectory-dependent, since the per-case normalization depends on how many timesteps fall into each case, which itself is shaped by the agent's chosen path. Importantly, each LM agent shows different success rates and the average number of steps for successful trajectories which will directly influence the normalization described in Section~\ref{sec:exp_setup}. As a result, the normalized metric values should be viewed primarily as behavioral summaries rather than as complete standalone measures that can be used to compare overall agent qualities. However, this does not weaken the validity of the metric itself: for any trajectory and map state, our error function still provides a principled and consistent classification of whether an action constitutes an exploration or exploitation error. We therefore view the proposed metric as a complementary metric to success rate, offering a more fine-grained analysis of LM agent behavior. More broadly, we believe this framework provides a foundation that can be extended to more realistic settings by incorporating semantic information, richer environment structures beyond 2D grid maps, and more complicated task dependencies.

\paragraph{Extension to Minecraft Clone.}

\begin{figure}[t]
    \centering
    \begin{subfigure}[t]{0.58\textwidth}
        \centering
        \vspace{0pt}
        \includegraphics[width=0.49\linewidth,trim={0px 0px 0px 555px},clip=true]{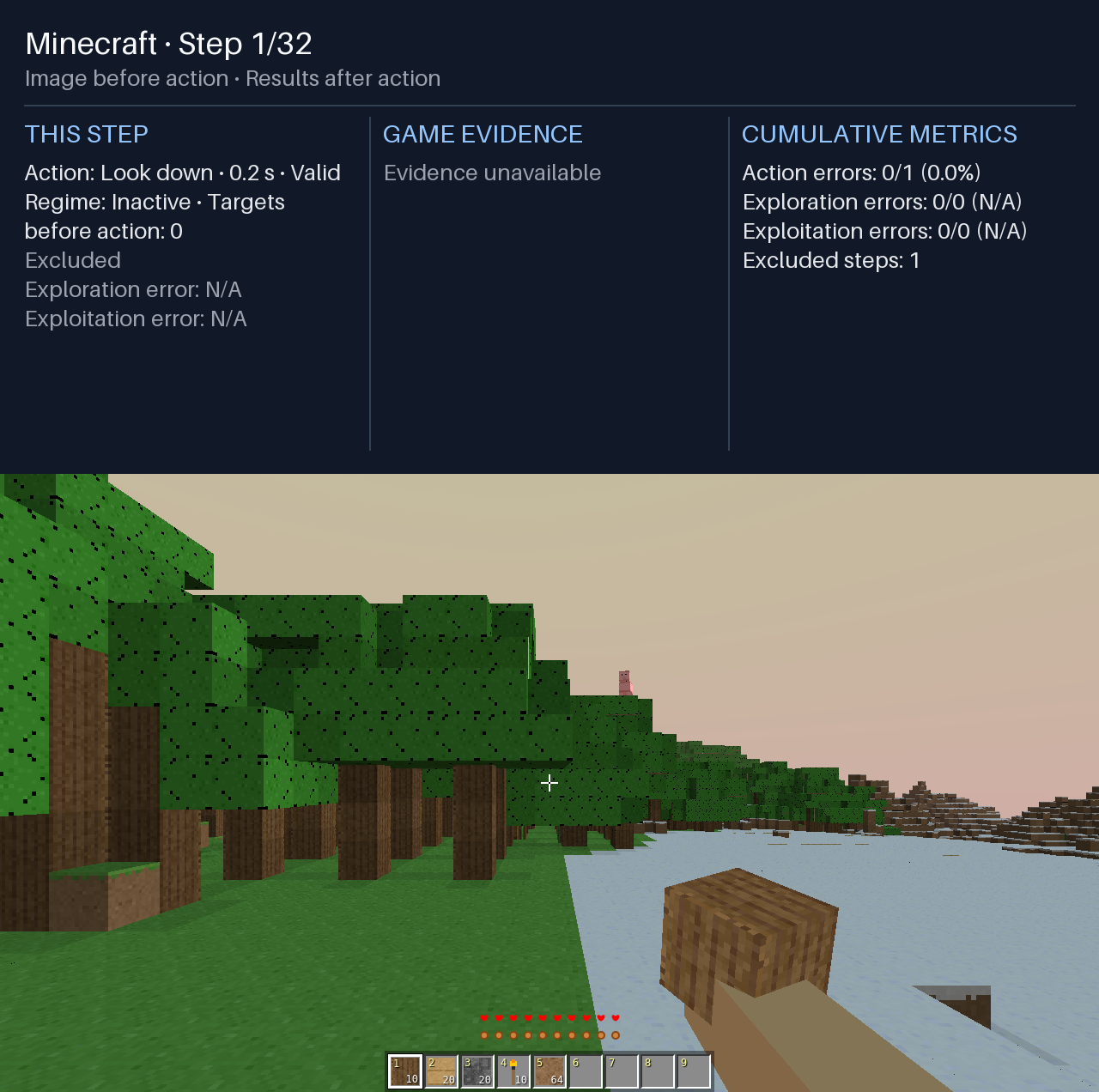}
        \includegraphics[width=0.49\linewidth,trim={0px 0px 0px 555px},clip=true]{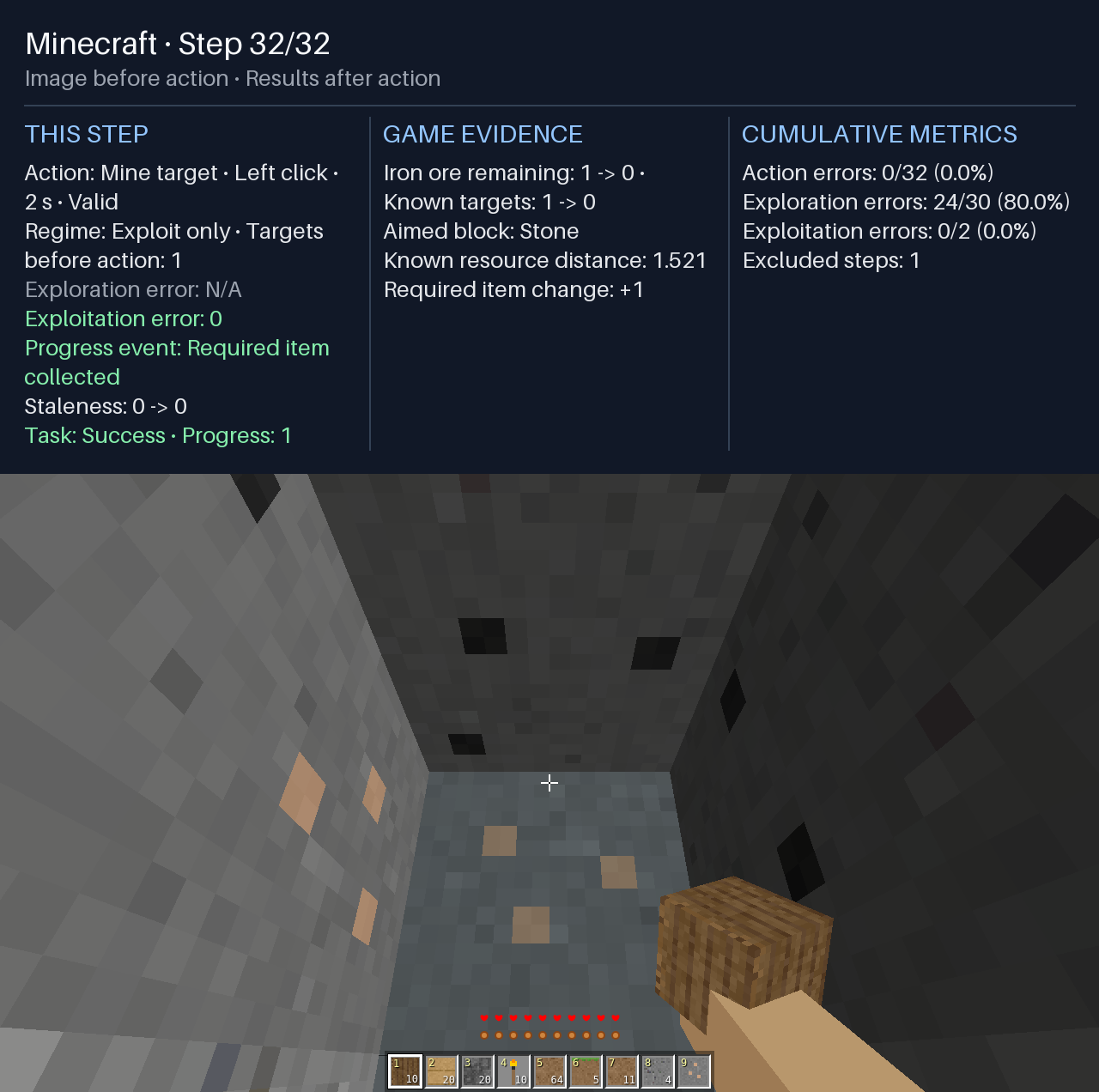}
        \caption{Observation Examples}
        \label{fig:video_games_observations}
    \end{subfigure}
    \hfill
    \begin{subfigure}[t]{0.41\textwidth}
        \centering
        \vspace{0pt}
        \setlength{\tabcolsep}{1pt}
        \renewcommand{\arraystretch}{1.15}
        \resizebox{\linewidth}{!}{%
            \begin{tabular}{@{}lcccc@{}}
                \toprule
                \multirow{2}{*}{\textbf{Model}} &
                    \textbf{Memory} & \textbf{Success} & \textbf{Exploration} & \textbf{Exploitation} \\
                & \textbf{Manage} & \textbf{Rate (\%)} & \textbf{Error} & \textbf{Error} \\
                \midrule
                Gemini 3.1 & $\xmark$ & 0.0 & 0.948 & 0.000 \\
                Flash Lite & $\cmark$ & 88.9 & 0.807 & 0.000 \\
                \midrule
                \multirow{2}{*}{GPT-4.1}
                & $\xmark$ & 11.1 & 0.877 & N/A \\        
                & $\cmark$ & 77.8 & 0.740 & 0.556 \\
                \bottomrule
            \end{tabular}
        }
        \caption{Results}
        \label{fig:video_games_results}
    \end{subfigure}
    \caption{Observation examples and results with Gemini 3.1 Flash Lite and GPT-4.1 on Minecraft Clone. Memory Manage indicates inclusion of up to two prior screenshots.}
    \label{fig:video_games}
\end{figure}

Figure~\ref{fig:video_games} extends our evaluation beyond symbolic grid maps by adapting the proposed metrics to Minecraft Clone. We discretize Minecraft Clone's keyboard and mouse controls into 21 predefined actions, with one action per decision: four movement directions, four camera adjustments, jumping, mining, block placement, nine inventory-slot selections, and waiting. Each action executes for a prescribed duration: movement for 1 second, jumping for 0.4 seconds, mining for 2 seconds, and camera adjustment for 0.2 seconds with a fixed mouse displacement. The simulation remains frozen between actions, so model inference time does not affect execution. We score the resulting transitions using position/view novelty and ore discovery for exploration, and progress toward known ore or its acquisition for exploitation. Each scored decision has equal weight regardless of action duration.  (Appendix~\ref{sec:app_video_games} details the action interface, evaluation protocol, and adaptation of the error metric).

This experiment illustrates how our framework can characterize agent behavior in a visual 3D environment from recorded interactions beyond the synthetic 2D grid setting. In the experiments, we observe that the error decomposition again reveals a distinction that task success alone cannot explain: without memory management with prior screenshots, Gemini 3.1 Flash Lite achieves zero success despite zero measured exploitation error. This is largely due to its exploration error being $0.948$ which signals a serious bottleneck in discovering the task. However, we discover that memory management reduces exploration error from $0.948$ to $0.807$ for Gemini and from $0.877$ to $0.740$ for GPT-4.1, alongside a significant increase in the success-rate from $0.0\%$ to $88.9\%$ and from $11.1\%$ to $77.8\%$, respectively. The exact details of the memory management mechanism is detailed in Appendix~\ref{sec:app_video_games}.

\section{Conclusion}

In this work, we introduce a policy-agnostic metric to quantify exploration and exploitation errors in LM agents from action trajectories. Using partially observable grid-map environments with symbolic task DAGs, we evaluate 35 frontier LM agents and show that low exploration and exploitation errors are strongly associated with success, while LM agents with similar success rates still exhibit qualitatively different behaviors. Our experiments further show that prompt design, harness engineering, and semantic information substantially affect how LM agents balance exploration and exploitation. Finally, our metric extends to Minecraft Clone, demonstrating its applicability beyond our synthetic environment. Overall, our results provide a foundation for quantifying exploration and exploitation errors in LM agents beyond success rate, offering a more informative lens for evaluating and improving LM agents in complex open-ended tasks.

\newpage
\subsection*{AI use statement}

During the preparation of this work, we used LLMs in the following:
\begin{itemize}[leftmargin=*]
    \item Writing assistance: An LLM was used for grammar correction, fluency checking, and refinement of author-written text;
    \item Code implementation: Codex and Claude Code were used to assist with implementing analysis scripts for processing experimental results and generating plots;
    \item Prompt revision: OpenAI ChatGPT, Codex, and Claude Code were used for revising prompts;
    \item Script execution: Codex and Claude Code were utilized to execute written scripts and CLI.
\end{itemize}

\subsection*{Reproducibility Statement}

We provide the details necessary to reproduce our experiments throughout the
appendix. Appendix~\ref{sec:app_task_generation} describes the procedural task
and map generation process and its parameters; subsequent sections report the
experimental configurations, episode counts, random seeds, prompt templates,
and agent harness details.

\subsubsection*{Acknowledgments}

This work was supported in part by National Science Foundation (NSF) award IIS-2404180, and Institute of Information \& communications Technology Planning \& Evaluation (IITP) grants funded by the Korea government (MSIT) (No. 2022-0-00871, Development of AI Autonomy and Knowledge Enhancement for AI Agent Collaboration), and (No. RS-2025-2543949. Environment-Aware and Domain-Adaptive Multimodal Embodied AI for Real-World Interaction). This work was also supported by NSF award DMS-2023239, NSF CAREER award CCF-2339978, the Google Cloud Research Credits Program, and a grant from FuriosaAI. In addition, it used resources through CIS251382 from the Advanced Cyberinfrastructure Coordination Ecosystem: Services \& Support (ACCESS) program, which is supported by NSF awards OAC-2138259, OAC-2138286, OAC-2138307, OAC-2137603, and OAC-2138296. We thank AMD for providing AMD Instinct GPU compute resources through AMD Dev Cloud and Mahdi Ghodsi for technical support throughout this research.

\bibliography{iclr2027_conference}
\bibliographystyle{iclr2027_conference}

\newpage
\appendix

\section*{Appendices}

Here, we provide additional details that were left out due to limited space.

\section{Additional Related Work}

\paragraph{Exploration and Exploitation.}
The exploration--exploitation tradeoff has been extensively studied in reinforcement learning (RL)~\citep{thompson1933likelihood, auer2002finite, bellemare2016unifying, pathak2017curiosity}, and has recently gained attention in the context of LM agents~\citep{tang2024code, inouewider}. Several works have shown that LMs do not explore efficiently, and have proposed methods to enhance exploration efficiency through in-context learning~\citep{krishnamurthy2024can, russo2026incontext}, prompting~\citep{ding2026calibrate}, supervised training~\citep{KimJ2026tmlr}, and RL training~\citep{szot2026expanding}. Likewise, many works report the limited exploitation capabilities of LM agents and address them through training~\citep{lehnert2024beyond}, or integrating external planners~\citep{JeongJ2026aaiw}. More recently, ~\citet{harris2025should} evaluated LMs' exploration--exploitation tradeoff in bandit settings and found that frontier models underperform simple linear regression baselines.

Prior works have focused on improving exploration or exploitation, and analyzing their tradeoffs in fixed environments where actions are independent to the environments. However, in real-world tasks, prior actions often determine what the LM agent can observe and achieve. To the best of our knowledge, no existing framework \emph{quantitatively separates and measures} exploration and exploitation errors in such environments. Our work addresses this gap through the proposed 2D grid-map environments with task DAGs and the proposed exploration and exploitation metric.

\section{Details of Task Formulation and Exploration and Exploitation Metric}\label{apdx:eqn}
In this section, we provide the detailed intuition, metric design rationale, and illustrative examples that were deferred from the main text due to limited space.

\paragraph{2D Grid Map.} For a cell $c\in\mathcal{M}$, we denote $\text{adj}(c)$ as the set of neighboring cells of $c$ - i.e. cells that can be traversed from $c$.

We use $p(t) \in \mathcal{M}$ to denote the position of the agent at timestep $t$. Each time the agent visits a cell $p(t)$, the environment provides a subset of \texttt{up, down, left, right} as admissible moves, which provide information about the neighboring cells of $p(t)$ -- i.e. $\text{adj}(p(t))$. 

We say a cell $c$ is \textbf{observed} at timestep $t$ if the agent has visited the cell $c$ at some previous timestep, i.e. $\exists t' \leq t: p(t') = c$. Neighboring cells of $p(t)$ that have not been previously observed are defined as \textbf{unobserved}. Critically, information about the task DAG is only revealed when the cell is observed. The agent must traverse to unobserved cells to discover if task nodes are present in those cells. If no task node is present, the cell is simply marked as observed. Finally, cells that are neither observed nor unobserved states are defined as \textbf{unknown}. Summarizing, we define the state function for cell $c$ at timestep $t$ as such:
\begin{equation}\label{eqn:state}
    \begin{aligned}
        \text{state}(c,t) = \begin{cases}
        \text{observed}, & \text{if } \exists t' \leq t: p(t') = c,\\
        \text{unobserved}, & \text{if } \neg\text{observed}(c,t) \land c \in \bigcup_{c': \text{observed}}\text{adj}(c'), \\
        \text{unknown}, & \text{otherwise.}
        \end{cases}        
    \end{aligned}
\end{equation}

\paragraph{Task DAG.} We define $\mathcal{G} = (N, E)$ as our task DAG with $N,E$ denoting nodes and edges, respectively. We use $\text{parent}(u)$ and $\text{child}(u)$ to denote the set of parent and child nodes of $u$, respectively. Without loss of generality,\footnote{If there are multiple goal nodes, we can append a dummy sink node.} we assume a unique sink node $g$ which denotes the goal. Each node $u \in N$ occupies a unique location in the map $l(u)$, i.e. $l: N \mapsto \mathcal{M}$ is injective.

We say node $u$ is \textbf{visited} at timestep $t$ if the agent is located at $l(u)$, i.e. $p(t) = l(u)$. If the agent has visited node $u$ at some previous timestep, i.e. $\text{state}(l(u), t) = \text{observed}$, a node $u$ has been \textbf{seen}.\footnote{Note that visited implies seen.} Each node $u\in N$ has $\text{status}(u,t) \in \{\text{undiscovered}, \text{discovered}, \text{achieved}\}$ and $\text{type}(u) = \{\texttt{AND},\texttt{OR}\} $ associated with it, which we define below:
\begin{equation}\label{eqn:status}
    \begin{aligned}
        \text{status}(u,t) = \begin{cases}
        \text{achieved}, & \text{if } \exists t' \leq t: \text{visited}(u,t') \land \text{prec}(u,t'), \\
        \text{discovered}, & \text{if } \text{seen}(u,t) \land \text{status}(u,t) \neq \text{achieved}, \\
        \text{undiscovered}, & \text{if } \neg\text{seen}(u,t),
        \end{cases}        
    \end{aligned}
\end{equation}
where $\text{prec}(u,t) = \text{true}$ if $\text{parent}(u) = \emptyset$, and when $\text{parent}(u) \neq \emptyset$,
\begin{equation}\label{eqn:prec}
    \begin{aligned}
        \text{prec}(u,t) = \begin{cases}
        \forall\, u' \in \text{parent}(u) : \text{status}(u',t) = \text{achieved}, & \text{if } \text{type}(u) = \texttt{AND}, \\
        \exists\, u' \in \text{parent}(u) : \text{status}(u',t) = \text{achieved}, & \text{if } \text{type}(u) = \texttt{OR},
        \end{cases}
    \end{aligned}
\end{equation}
i.e. a node is only achieved when it is visited and the precondition is met. Nodes with no parents are achieved immediately upon visiting. If a node is visited before the preconditions are satisfied, the agent must satisfy the preconditions and revisit the node. The precondition requires achieving either all or at least one parent node, depending on $\text{type}(u)$.

Upon visiting a cell $p(t) \in \mathcal{M}$, the environment provides the set of admissible movements. If a node $u$ is discovered at $p(t)$, the environment provides information about its depth-$1$ neighborhood $\mathcal{N}(u) = \text{parent}(u) \cup \text{child}(u)$. The locations of nodes in $\mathcal{N}(u)$ remain unknown and must be discovered by the agent. 

Throughout, we assume a movement to a neighboring cell corresponds to one timestep.\footnote{This aligns with our analysis where each timestep corresponds to one API call.} The task is complete when $\text{status}(g, t) = \text{achieved}$.

\paragraph{Measuring Exploration and Exploitation Errors.}

Using the definitions from above, we can formally define pending task $\mathcal{P}(t)$:
\begin{equation}\label{eqn:pending}
    \mathcal{P}(t) = \{u\in N: \text{status}(u,t) = \text{discovered} \land \text{prec}(u,t)\}.
\end{equation}
Achieving tasks in $\mathcal{P}(t)$ requires \textbf{exploitation} of existing knowledge since the location is already known. However, exploitation need not be constrained to the form of traversing already seen edges, as the agent may infer a shortest path from its history, as shown in Figure~\ref{fig:edge_case}(a).

We also formally define gain defined in Section~\ref{sec:expl}:
\begin{equation}\label{eqn:gain}
    \text{Gain}(t\to t+1) = 1 \Leftrightarrow \{p(t+1) \in \mathcal{T}(t)\} \lor \{\exists z \in \mathcal{T}(t): d(p(t+1), z) < d(p(t), z)\},
\end{equation}
and $\text{Gain}(t\to t+1) = 0$ otherwise. Here, $d(a,b)$ is the shortest distance between the cells $a$ and $b$. An action at timestep $t$ as an error if $\text{Gain}(t\to t+1) = 0$ (Equation~\eqref{eqn:gain}).

We refer to the set of cells that constitute  productive destinations as the \textbf{target set} $\mathcal{T}(t)$, which varies depending on the agent's current state outlined in Table~\ref{tab:four-cases}. As depicted in Table~\ref{tab:four-cases}, $\mathcal{T}(t)$ determines the required action from the LM agent at timestep $t$. Based on this, we define whether a move is a \textbf{gain} (refer to Equation~\eqref{eqn:gain}) if it steps into a target cell or reduces the minimum distance to at least one of the target cells. Note that the existential quantifier ensures a policy-agnostic evaluation: rather than assuming a specific strategy of the LM agent, we distinguish the map state at each timestep into four cases and determine whether the required action is exploration, exploitation or both, depicted in Table~\ref{tab:four-cases}.

In Table~\ref{tab:four-cases}, \textbf{Case 1} is when no tasks are pending, and the agent must explore to discover new information about the task DAG. To clarify, when the LM agent begins the game, it falls into Case 1, and both $\mathcal{P}(t)$ and $\mathcal{U}(t)$ cannot be empty at the same time. \textbf{Case 2} denotes when the goal is pending, and we assume that the only allowed strategy then is to exploit the knowledge and efficiently traverse to the goal node to finish the game. Similarly, in \textbf{Case 3} where the agent has traversed the entire map, the only viable move is to exploit its knowledge about the task DAG to complete the task. 

However, when pending tasks and unseen cells both exist (\textbf{Case 4}), we assume the LM agent can either explore or exploit. In our online setting, the space of reasonable strategies is large. For instance, as depicted in Figure~\ref{fig:edge_case}b, a path of length 7 that passes through 5 unseen cells may be more optimal for the task than a path of length 5 through seen cells. However, quantifying this tradeoff would require the full knowledge of the map and task DAG, which the agent does not have. Furthermore, since we are evaluating diverse LM agents with unknown internal reasoning, committing to any particular strategy -- for example, always exploiting the nearest pending task first -- and flagging deviations as errors would conflate strategic differences with genuine failures.

\paragraph{Challenging Edge Cases of Exploration and Exploitation Errors} Here, we list some challenging edge cases and demonstrate how our metric detects erroneous actions. All counterexamples below are instantiated on the centered $3\times 3$ grid $\mathcal M=\{-1,0,1\}^2$ with 4-neighbor moves only. We write undirected edges as $(a,b)\leftrightarrow(c,d)$. We assume that all cells have been observed, and drop the notion of task DAGs for simplicity. All edge cases are within a no-progress segment.

\begin{table}[ht]
\caption{Probe a branch and back out once.}
\centering
\small
\begin{tabular}{cccccc}
\toprule
$t$ & $p(t)$ & $c_t$ & $e_t$ & $n_t$ & $S_t$\\
\midrule
0 & $(-1,0)$ & 0 & 0 & 0 & 0\\
1 & $(0,0)$  & 0 & 0 & 0 & 0\\
2 & $(1,0)$  & 0 & 0 & 0 & 0\\
3 & $(0,0)$  & 0 & 0 & 0 & 0\\
4 & $(-1,0)$ & 0 & 0 & 0 & 0\\
\bottomrule
\end{tabular}
\end{table}
\noindent\textit{Explanation.}
This trajectory should not be penalized. It only probes a branch and backtracks once, so no cycle is formed ($c_t=0$), no edge is traversed more than twice ($e_t=0$), and no node is visited more than twice ($n_t=0$). Hence the stale score never increases.

\begin{table}[ht]
\caption{Useful gateway revisit.}
\centering
\small
\begin{tabular}{cccccc}
\toprule
$t$ & $p(t)$ & $c_t$ & $e_t$ & $n_t$ & $S_t$\\
\midrule
0 & $(-1,0)$ & 0 & 0 & 0 & 0\\
1 & $(0,0)$  & 0 & 0 & 0 & 0\\
2 & $(1,0)$  & 0 & 0 & 0 & 0\\
3 & $(0,0)$  & 0 & 0 & 0 & 0\\
4 & $(0,1)$  & 0 & 0 & 0 & 0\\
\bottomrule
\end{tabular}
\end{table}

\noindent\textit{Explanation.}
This trajectory should also remain error-free. The revisit to $(0,0)$ serves as a gateway before taking a fresh branch to $(0,1)$. Structurally, this still involves only benign backtracking: no loop is closed, and neither edge nor node reuse exceeds the allowed budget of two.

\begin{table}[ht]
\caption{Re-enter the same exhausted branch.}
\centering
\small
\begin{tabular}{cccccc}
\toprule
$t$ & $p(t)$ & $c_t$ & $e_t$ & $n_t$ & $S_t$\\
\midrule
0 & $(-1,0)$ & 0 & 0 & 0 & 0\\
1 & $(0,0)$  & 0 & 0 & 0 & 0\\
2 & $(1,0)$  & 0 & 0 & 0 & 0\\
3 & $(0,0)$  & 0 & 0 & 0 & 0\\
4 & $(-1,0)$ & 0 & 0 & 0 & 0\\
5 & $(0,0)$  & 0 & 1 & 1 & 2\\
6 & $(1,0)$  & 0 & 2 & 1 & 3\\
\bottomrule
\end{tabular}
\end{table}

\noindent\textit{Explanation.}
The prefix up to $t=4$ is still a single probe-and-return pattern and is therefore not an error. The first error appears at $t=5$, when the trajectory reuses the edge $(-1,0)\leftrightarrow(0,0)$ for the third time and visits $(0,0)$ for the third time, increasing both $e_t$ and $n_t$. At $t=6$, the edge $(0,0)\leftrightarrow(1,0)$ is also traversed for the third time, so $e_t$ increases again.

\begin{table}[ht]
\caption{Repeated use of the same cycle.}
\centering
\small
\begin{tabular}{cccccc}
\toprule
$t$ & $p(t)$ & $c_t$ & $e_t$ & $n_t$ & $S_t$\\
\midrule
0 & $(-1,-1)$ & 0 & 0 & 0 & 0\\
1 & $(0,-1)$  & 0 & 0 & 0 & 0\\
2 & $(0,0)$   & 0 & 0 & 0 & 0\\
3 & $(-1,0)$  & 0 & 0 & 0 & 0\\
4 & $(-1,-1)$ & 1 & 0 & 0 & 1\\
5 & $(0,-1)$  & 1 & 0 & 0 & 1\\
6 & $(0,0)$   & 1 & 0 & 0 & 1\\
7 & $(-1,0)$  & 1 & 0 & 0 & 1\\
8 & $(-1,-1)$ & 1 & 0 & 1 & 2\\
\bottomrule
\end{tabular}
\end{table}

\noindent\textit{Explanation.}
The first error occurs at $t=4$, when the move closes a loop and increases the cyclomatic term $c_t$ from $0$ to $1$. The next lap through the same cycle does not create a new independent cycle, so $c_t$ stays constant. However, at $t=8$ the trajectory returns to $(-1,-1)$ for the third time, increasing $n_t$ and thus the stale score again.

\begin{table}[ht]
\caption{Corridor oscillation.}
\centering
\small
\begin{tabular}{cccccc}
\toprule
$t$ & $p(t)$ & $c_t$ & $e_t$ & $n_t$ & $S_t$\\
\midrule
0 & $(0,0)$  & 0 & 0 & 0 & 0\\
1 & $(0,1)$  & 0 & 0 & 0 & 0\\
2 & $(0,0)$  & 0 & 0 & 0 & 0\\
3 & $(0,-1)$ & 0 & 0 & 0 & 0\\
4 & $(0,0)$  & 0 & 0 & 1 & 1\\
5 & $(0,1)$  & 0 & 1 & 1 & 2\\
6 & $(0,0)$  & 0 & 2 & 2 & 4\\
7 & $(0,-1)$ & 0 & 3 & 2 & 5\\
\bottomrule
\end{tabular}
\end{table}

\noindent\textit{Explanation.}
The initial oscillation up to $t=3$ is tolerated as benign movement between two directions. The first error occurs at $t=4$, when $(0,0)$ is visited for the third time, increasing $n_t$. Subsequent steps repeatedly reuse the same corridor edges beyond twice, so $e_t$ increases at each additional oscillation.

\begin{table}[ht]
\caption{Comb / broom graph.}
\centering
\small
\begin{tabular}{cccccc}
\toprule
$t$ & $p(t)$ & $c_t$ & $e_t$ & $n_t$ & $S_t$\\
\midrule
0 & $(-1,0)$ & 0 & 0 & 0 & 0\\
1 & $(0,0)$  & 0 & 0 & 0 & 0\\
2 & $(1,0)$  & 0 & 0 & 0 & 0\\
3 & $(1,1)$  & 0 & 0 & 0 & 0\\
4 & $(1,0)$  & 0 & 0 & 0 & 0\\
5 & $(0,0)$  & 0 & 0 & 0 & 0\\
6 & $(-1,0)$ & 0 & 0 & 0 & 0\\
7 & $(0,0)$  & 0 & 1 & 1 & 2\\
8 & $(0,1)$  & 0 & 1 & 1 & 2\\
\bottomrule
\end{tabular}
\end{table}

\noindent\textit{Explanation.}
The trajectory first explores one tooth of the broom and fully backtracks along the shared stem, which remains within the benign budget. The first error occurs at $t=7$, when the shared stem edge $(-1,0)\leftrightarrow(0,0)$ is traversed for the third time and $(0,0)$ is visited for the third time. The final move to $(0,1)$ opens a fresh branch, so the stale score does not increase further.

\section{Full Results Across Prompt Modes}\label{apdx:full_results}
\subsection{Models}
We evaluate the following 35 models.
GPT-4.1, GPT-4.1 mini, GPT-4.1 nano~\citep{openai2025gpt41},
GPT-5.4, GPT-5.4 mini, GPT-5.4 nano~\citep{openai2026gpt54thinking},
Gemini 3.1 Pro~\citep{google2025gemini31pro},
Gemini 3 Flash~\citep{google2025gemini3flash},
Gemini 3.1 Flash Lite~\citep{google2025gemini31flashlite},
Claude Opus 4.6~\citep{anthropic2026claude46opus},
Claude Sonnet 4.6~\citep{anthropic2026claude46sonnet},
Claude Haiku 4.5~\citep{anthropic2025claude45haiku},
GPT-OSS-120B~\citep{agarwal2025gpt},
Muse Glimmer 30B~\citep{muse_glimmer_30b},
Gemma 4 31B~\citep{gemma4_31b},
Gemma 4 26B A4B~\citep{gemma4_26b_a4b},
Gemma 4 12B~\citep{gemma4_12b},
Qwen 3.8 27B~\citep{qwen38_27b},
Qwen 3.6 35B A3B~\citep{qwen36_35b_a3b},
Qwen 3.6 27B~\citep{qwen36_27b},
Qwen 3.5 397B A17B~\citep{qwen35_397b_a17b},
Qwen 3.5 9B~\citep{qwen35_9b},
Qwen 3.5 4B~\citep{qwen35_4b},
Qwen 3.5 2B~\citep{qwen35_2b},
Mistral 3 14B~\citep{ministral3_14b},
Mistral 3 8B~\citep{ministral3_8b},
Magistral Small~\citep{magistral_small},
DeepSeek V4 Flash~\citep{deepseek_v4_flash},
Kimi K2.6~\citep{kimi_k26},
Granite 4.1 8B~\citep{granite41_8b},
Nemotron 3 Nano 30B~\citep{nemotron3_nano_30b},
GLM 4.7 Flash~\citep{glm47_flash},
MiniMax M3~\citep{minimax_m3},
Hy3~\citep{hy3},
Hunyuan A13B~\citep{hunyuan_a13b}.
For all models, we set the temperature as $0$.

\subsection{Full results in Explore-Exploit setting}
Tables~\ref{tab:combined_8_prompts}, \ref{tab:default_4_prompts}, and \ref{tab:reasoning_4_prompts} report results in our main setting, where the map is partially observable and the task DAG is unknown, so the agent must both explore and exploit. We report results aggregated over all eight prompt sets, and separately for the four prompt variants without and with reasoning. In reasoning mode, some episodes terminated before reaching the step budget because the model exhausted its output-token allowance while reasoning, without producing an action (Qwen3.6-27B: 51/108 episodes; DeepSeek-V4-Flash: 64/108 episodes). These are counted as failures. Budget used is always computed as the number of steps taken divided by the full step budget, so for these models the average budget used over all episodes can be lower than over successful episodes. Due to cost, we exclude the 13 API-based models from the exploit-only setting.

\begin{table*}[t]
\caption{Results aggregated over default and reasoning prompts (216 episodes per model). Budget used is the percentage of the step budget, averaged over all episodes (Total) or successful episodes only (Success); N/A indicates no successful episodes. Models are sorted by accuracy.}
\label{tab:combined_8_prompts}
\centering
\setlength{\tabcolsep}{1pt}
\resizebox{\textwidth}{!}{%
\begin{tabular}{lrrrrr}
\toprule
\textbf{Model} & \textbf{Accuracy} $\uparrow$ & \textbf{Exploration Error} $\downarrow$ & \textbf{Exploitation Error} $\downarrow$ & \textbf{Avg. Budget Used (Total)} $\downarrow$ & \textbf{Avg. Budget Used (Success)} $\downarrow$ \\
\midrule
Gemini 3.1 Pro~\citep{google2025gemini31pro} & 100.00\% & 1.20\% & 0.79\% & 44.70\% & 44.70\% \\
Claude Opus 4.6~\citep{anthropic2026claude46opus} & 99.54\% & 1.96\% & 1.47\% & 47.48\% & 47.23\% \\
Claude Sonnet 4.6~\citep{anthropic2026claude46sonnet} & 97.22\% & 5.35\% & 2.34\% & 52.90\% & 51.55\% \\
Claude Haiku 4.5~\citep{anthropic2025claude45haiku} & 91.67\% & 8.16\% & 4.45\% & 61.27\% & 57.75\% \\
GPT-5.4~\citep{openai2026gpt54thinking} & 89.35\% & 8.13\% & 6.00\% & 57.45\% & 52.38\% \\
Gemini 3 Flash~\citep{google2025gemini3flash} & 88.43\% & 9.50\% & 4.70\% & 58.16\% & 52.68\% \\
Qwen3.8-27B~\citep{qwen38_27b} & 75.00\% & 16.36\% & 11.10\% & 68.31\% & 57.75\% \\
Kimi-K2.6~\citep{kimi_k26} & 73.15\% & 11.34\% & 4.83\% & 67.90\% & 56.12\% \\
GPT-4.1~\citep{openai2025gpt41} & 66.20\% & 13.95\% & 16.85\% & 74.73\% & 61.82\% \\
Hy3~\citep{hy3} & 63.89\% & 10.14\% & 3.73\% & 73.90\% & 59.15\% \\
Muse-Glimmer-30B~\citep{muse_glimmer_30b} & 62.04\% & 11.11\% & 9.16\% & 77.14\% & 63.15\% \\
Gemini 3.1 Flash Lite~\citep{google2025gemini31flashlite} & 50.93\% & 16.47\% & 14.94\% & 77.19\% & 55.21\% \\
Qwen3.6-35B-A3B~\citep{qwen36_35b_a3b} & 50.46\% & 30.52\% & 25.32\% & 80.31\% & 64.02\% \\
GPT-OSS-120B~\citep{agarwal2025gpt} & 49.07\% & 22.97\% & 23.73\% & 82.50\% & 64.34\% \\
MiniMax-M3~\citep{minimax_m3} & 47.69\% & 15.62\% & 12.47\% & 85.03\% & 69.08\% \\
Gemma-4-31B-IT~\citep{gemma4_31b} & 46.76\% & 17.59\% & 5.27\% & 84.33\% & 66.49\% \\
Qwen3.5-397B-A17B~\citep{qwen35_397b_a17b} & 43.98\% & 21.15\% & 11.12\% & 79.83\% & 60.64\% \\
Qwen3.6-27B~\citep{qwen36_27b} & 43.52\% & 23.27\% & 14.95\% & 59.62\% & 54.34\% \\
DeepSeek-V4-Flash~\citep{deepseek_v4_flash} & 41.67\% & 11.74\% & 10.47\% & 62.87\% & 58.13\% \\
Gemma-4-26B-A4B-IT~\citep{gemma4_26b_a4b} & 5.09\% & 54.12\% & 52.12\% & 98.02\% & 61.21\% \\
Gemma-4-12B-IT~\citep{gemma4_12b} & 4.63\% & 61.58\% & 63.51\% & 98.52\% & 67.96\% \\
GPT-4.1 mini~\citep{openai2025gpt41} & 3.24\% & 60.92\% & 61.12\% & 98.72\% & 60.37\% \\
Ministral-3-14B-Reasoning-2512~\citep{ministral3_14b_reasoning} & 2.78\% & 58.44\% & 61.71\% & 99.23\% & 72.19\% \\
Qwen3.5-9B~\citep{qwen35_9b} & 2.31\% & 61.92\% & 50.88\% & 98.92\% & 70.12\% \\
GLM-4.7-Flash~\citep{glm47_flash} & 1.85\% & 57.39\% & 45.43\% & 93.71\% & 62.98\% \\
Magistral-Small-2509~\citep{magistral_small} & 1.39\% & 71.46\% & 57.62\% & 99.70\% & 78.09\% \\
NVIDIA-Nemotron-3-Nano-30B-A3B-BF16~\citep{nemotron3_nano_30b} & 1.39\% & 61.69\% & 48.24\% & 98.49\% & 77.81\% \\
GPT-5.4 mini~\citep{openai2026gpt54thinking} & 1.39\% & 60.53\% & 41.12\% & 99.58\% & 69.43\% \\
GPT-5.4 nano~\citep{openai2026gpt54thinking} & 1.39\% & 47.62\% & 52.72\% & 99.76\% & 82.46\% \\
Qwen3.5-2B~\citep{qwen35_2b} & 1.39\% & 59.49\% & 59.67\% & 96.69\% & 60.47\% \\
Qwen3.5-4B~\citep{qwen35_4b} & 1.39\% & 63.91\% & 36.13\% & 98.75\% & 59.06\% \\
Hunyuan-A13B-Instruct~\citep{hunyuan_a13b} & 0.93\% & 68.51\% & 73.08\% & 98.66\% & 70.62\% \\
Granite-4.1-8B~\citep{granite41_8b} & 0.46\% & 85.60\% & 91.09\% & 99.77\% & 50.88\% \\
Ministral-3-8B-Reasoning-2512~\citep{ministral3_8b_reasoning} & 0.46\% & 55.10\% & 68.83\% & 99.97\% & 93.33\% \\
GPT-4.1 nano~\citep{openai2025gpt41} & 0.00\% & 83.20\% & 75.54\% & 100.00\% & N/A \\
\bottomrule
\end{tabular}%
}
\end{table*}
 
\begin{table*}[t]
\caption{Results without reasoning, aggregated over the four prompt variants (108 episodes per model). Budget used is the percentage of the step budget, averaged over all episodes (Total) or successful episodes only (Success); N/A indicates no successful episodes. Models are sorted by accuracy.}
\label{tab:default_4_prompts}
\centering
\setlength{\tabcolsep}{1pt}
\setlength{\tabcolsep}{1pt}
\resizebox{\textwidth}{!}{%
\begin{tabular}{lrrrrr}
\toprule
\textbf{Model} & \textbf{Accuracy} $\uparrow$ & \textbf{Exploration Error} $\downarrow$ & \textbf{Exploitation Error} $\downarrow$ & \textbf{Avg. Budget Used (Total)} $\downarrow$ & \textbf{Avg. Budget Used (Success)} $\downarrow$ \\
\midrule
Gemini 3.1 Pro~\citep{google2025gemini31pro} & 100.00\% & 1.82\% & 1.27\% & 45.00\% & 45.00\% \\
Claude Opus 4.6~\citep{anthropic2026claude46opus} & 99.07\% & 2.65\% & 2.01\% & 48.96\% & 48.48\% \\
Claude Sonnet 4.6~\citep{anthropic2026claude46sonnet} & 94.44\% & 6.47\% & 2.70\% & 55.29\% & 52.66\% \\
Claude Haiku 4.5~\citep{anthropic2025claude45haiku} & 90.74\% & 9.47\% & 4.79\% & 60.13\% & 56.06\% \\
Gemini 3 Flash~\citep{google2025gemini3flash} & 87.96\% & 8.78\% & 6.98\% & 59.67\% & 54.15\% \\
GPT-5.4~\citep{openai2026gpt54thinking} & 87.04\% & 8.31\% & 7.53\% & 59.21\% & 53.13\% \\
GPT-4.1~\citep{openai2025gpt41} & 65.74\% & 12.15\% & 10.46\% & 76.86\% & 64.80\% \\
Muse-Glimmer-30B~\citep{muse_glimmer_30b} & 53.70\% & 11.71\% & 11.76\% & 81.13\% & 64.86\% \\
Qwen3.8-27B~\citep{qwen38_27b} & 52.78\% & 24.94\% & 18.55\% & 76.30\% & 55.09\% \\
Gemini 3.1 Flash Lite~\citep{google2025gemini31flashlite} & 51.85\% & 13.88\% & 11.14\% & 77.29\% & 56.21\% \\
Kimi-K2.6~\citep{kimi_k26} & 47.22\% & 17.01\% & 9.66\% & 83.37\% & 64.79\% \\
DeepSeek-V4-Flash~\citep{deepseek_v4_flash} & 45.37\% & 13.97\% & 10.63\% & 84.56\% & 65.97\% \\
MiniMax-M3~\citep{minimax_m3} & 42.59\% & 14.66\% & 9.18\% & 86.21\% & 67.62\% \\
Qwen3.6-27B~\citep{qwen36_27b} & 34.26\% & 29.42\% & 21.66\% & 86.43\% & 60.38\% \\
Gemma-4-31B-IT~\citep{gemma4_31b} & 32.41\% & 16.68\% & 6.90\% & 91.38\% & 73.40\% \\
Qwen3.5-397B-A17B~\citep{qwen35_397b_a17b} & 31.48\% & 20.02\% & 13.33\% & 89.15\% & 65.55\% \\
Hy3~\citep{hy3} & 29.63\% & 14.43\% & 7.40\% & 92.68\% & 75.30\% \\
GPT-OSS-120B~\citep{agarwal2025gpt} & 27.78\% & 26.16\% & 28.89\% & 93.05\% & 74.97\% \\
Qwen3.6-35B-A3B~\citep{qwen36_35b_a3b} & 17.59\% & 44.73\% & 43.80\% & 93.68\% & 64.07\% \\
GPT-4.1 mini~\citep{openai2025gpt41} & 6.48\% & 52.07\% & 55.31\% & 97.43\% & 60.37\% \\
Gemma-4-12B-IT~\citep{gemma4_12b} & 2.78\% & 65.98\% & 71.39\% & 98.70\% & 53.27\% \\
Gemma-4-26B-A4B-IT~\citep{gemma4_26b_a4b} & 2.78\% & 52.03\% & 43.84\% & 98.98\% & 63.29\% \\
Magistral-Small-2509~\citep{magistral_small} & 1.85\% & 69.66\% & 22.73\% & 99.72\% & 84.85\% \\
GPT-5.4 mini~\citep{openai2026gpt54thinking} & 1.85\% & 53.78\% & 24.14\% & 99.78\% & 87.88\% \\
NVIDIA-Nemotron-3-Nano-30B-A3B-BF16~\citep{nemotron3_nano_30b} & 0.93\% & 76.16\% & 76.92\% & 99.82\% & 80.70\% \\
Qwen3.5-9B~\citep{qwen35_9b} & 0.93\% & 80.83\% & 73.77\% & 99.38\% & 33.33\% \\
Granite-4.1-8B~\citep{granite41_8b} & 0.00\% & 85.17\% & 91.75\% & 100.00\% & N/A \\
Ministral-3-14B-Reasoning-2512~\citep{ministral3_14b_reasoning} & 0.00\% & 55.82\% & 25.00\% & 100.00\% & N/A \\
Ministral-3-8B-Reasoning-2512~\citep{ministral3_8b_reasoning} & 0.00\% & 39.13\% & 19.15\% & 100.00\% & N/A \\
GPT-4.1 nano~\citep{openai2025gpt41} & 0.00\% & 80.63\% & 73.22\% & 100.00\% & N/A \\
GPT-5.4 nano~\citep{openai2026gpt54thinking} & 0.00\% & 53.25\% & 57.40\% & 100.00\% & N/A \\
Qwen3.5-2B~\citep{qwen35_2b} & 0.00\% & 66.20\% & 60.84\% & 100.00\% & N/A \\
Qwen3.5-4B~\citep{qwen35_4b} & 0.00\% & 81.57\% & 0.00\% & 100.00\% & N/A \\
Hunyuan-A13B-Instruct~\citep{hunyuan_a13b} & 0.00\% & 75.64\% & 73.63\% & 100.00\% & N/A \\
GLM-4.7-Flash~\citep{glm47_flash} & 0.00\% & 69.51\% & 58.50\% & 100.00\% & N/A \\
\bottomrule
\end{tabular}%
}
\end{table*}
 
\begin{table*}[t]
\caption{Results with reasoning, aggregated over the four prompt variants (108 episodes per model). Budget used is the percentage of the step budget, averaged over all episodes (Total) or successful episodes only (Success); N/A indicates no successful episodes. Models are sorted by accuracy.}
\label{tab:reasoning_4_prompts}
\centering
\resizebox{\textwidth}{!}{%
\begin{tabular}{lrrrrr}
\toprule
\textbf{Model} & \textbf{Accuracy} $\uparrow$ & \textbf{Exploration Error} $\downarrow$ & \textbf{Exploitation Error} $\downarrow$ & \textbf{Avg. Budget Used (Total)} $\downarrow$ & \textbf{Avg. Budget Used (Success)} $\downarrow$ \\
\midrule
Claude Opus 4.6~\citep{anthropic2026claude46opus} & 100.00\% & 1.24\% & 0.90\% & 46.00\% & 46.00\% \\
Claude Sonnet 4.6~\citep{anthropic2026claude46sonnet} & 100.00\% & 4.14\% & 1.98\% & 50.51\% & 50.51\% \\
Gemini 3.1 Pro~\citep{google2025gemini31pro} & 100.00\% & 0.55\% & 0.37\% & 44.40\% & 44.40\% \\
Kimi-K2.6~\citep{kimi_k26} & 99.07\% & 1.80\% & 1.02\% & 52.43\% & 51.98\% \\
Hy3~\citep{hy3} & 98.15\% & 2.39\% & 1.86\% & 55.12\% & 54.27\% \\
Qwen3.8-27B~\citep{qwen38_27b} & 97.22\% & 4.97\% & 2.43\% & 60.33\% & 59.20\% \\
Claude Haiku 4.5~\citep{anthropic2025claude45haiku} & 92.59\% & 6.87\% & 4.07\% & 62.42\% & 59.41\% \\
GPT-5.4~\citep{openai2026gpt54thinking} & 91.67\% & 7.93\% & 4.01\% & 55.70\% & 51.67\% \\
Gemini 3 Flash~\citep{google2025gemini3flash} & 88.89\% & 10.30\% & 2.16\% & 56.65\% & 51.23\% \\
Qwen3.6-35B-A3B~\citep{qwen36_35b_a3b} & 83.33\% & 9.90\% & 11.42\% & 66.95\% & 64.01\% \\
Muse-Glimmer-30B~\citep{muse_glimmer_30b} & 70.37\% & 10.49\% & 4.54\% & 73.16\% & 61.85\% \\
GPT-OSS-120B~\citep{agarwal2025gpt} & 70.37\% & 18.81\% & 19.09\% & 71.96\% & 60.15\% \\
GPT-4.1~\citep{openai2025gpt41} & 66.67\% & 15.89\% & 22.69\% & 72.59\% & 58.89\% \\
Gemma-4-31B-IT~\citep{gemma4_31b} & 61.11\% & 18.71\% & 4.23\% & 77.28\% & 62.83\% \\
Qwen3.5-397B-A17B~\citep{qwen35_397b_a17b} & 56.48\% & 22.60\% & 9.32\% & 70.51\% & 57.90\% \\
MiniMax-M3~\citep{minimax_m3} & 52.78\% & 16.62\% & 15.38\% & 83.85\% & 70.26\% \\
Qwen3.6-27B~\citep{qwen36_27b} & 52.78\% & 3.97\% & 0.76\% & 32.81\% & 50.42\% \\
Gemini 3.1 Flash Lite~\citep{google2025gemini31flashlite} & 50.00\% & 19.01\% & 19.33\% & 77.08\% & 54.17\% \\
DeepSeek-V4-Flash~\citep{deepseek_v4_flash} & 37.96\% & 6.21\% & 9.85\% & 41.19\% & 48.75\% \\
Gemma-4-26B-A4B-IT~\citep{gemma4_26b_a4b} & 7.41\% & 56.25\% & 57.33\% & 97.07\% & 60.43\% \\
Gemma-4-12B-IT~\citep{gemma4_12b} & 6.48\% & 57.16\% & 53.17\% & 98.33\% & 74.26\% \\
Ministral-3-14B-Reasoning-2512~\citep{ministral3_14b_reasoning} & 5.56\% & 61.07\% & 61.90\% & 98.46\% & 72.19\% \\
Qwen3.5-9B~\citep{qwen35_9b} & 3.70\% & 42.87\% & 37.22\% & 98.46\% & 79.32\% \\
GLM-4.7-Flash~\citep{glm47_flash} & 3.70\% & 43.17\% & 37.25\% & 87.42\% & 62.98\% \\
GPT-5.4 nano~\citep{openai2026gpt54thinking} & 2.78\% & 41.97\% & 46.22\% & 99.51\% & 82.46\% \\
Qwen3.5-2B~\citep{qwen35_2b} & 2.78\% & 52.26\% & 58.85\% & 93.38\% & 60.47\% \\
Qwen3.5-4B~\citep{qwen35_4b} & 2.78\% & 45.76\% & 38.58\% & 97.50\% & 59.06\% \\
NVIDIA-Nemotron-3-Nano-30B-A3B-BF16~\citep{nemotron3_nano_30b} & 1.85\% & 46.64\% & 39.37\% & 97.15\% & 76.36\% \\
Hunyuan-A13B-Instruct~\citep{hunyuan_a13b} & 1.85\% & 61.07\% & 72.66\% & 97.32\% & 70.62\% \\
Granite-4.1-8B~\citep{granite41_8b} & 0.93\% & 86.04\% & 90.37\% & 99.55\% & 50.88\% \\
Magistral-Small-2509~\citep{magistral_small} & 0.93\% & 73.27\% & 71.96\% & 99.67\% & 64.58\% \\
Ministral-3-8B-Reasoning-2512~\citep{ministral3_8b_reasoning} & 0.93\% & 71.06\% & 78.06\% & 99.94\% & 93.33\% \\
GPT-5.4 mini~\citep{openai2026gpt54thinking} & 0.93\% & 67.32\% & 58.40\% & 99.38\% & 32.54\% \\
GPT-4.1 mini~\citep{openai2025gpt41} & 0.00\% & 69.58\% & 71.73\% & 100.00\% & N/A \\
GPT-4.1 nano~\citep{openai2025gpt41} & 0.00\% & 85.77\% & 87.05\% & 100.00\% & N/A \\
\bottomrule
\end{tabular}%
}
\end{table*}

\subsection{Full results in Exploit-only setting}
To isolate exploitation, we make the entire map and the task DAG structure fully observable to the agent, so it only needs to plan and exploit its knowledge to achieve the goal. Since no exploration is required, we report only exploitation error (Tables~\ref{tab:exploit_only_combined_prompts}, \ref{tab:exploit_only_default_prompts}, and \ref{tab:exploit_only_reasoning_prompts}). As in the explore--exploit setting, some reasoning-mode episodes terminated early because the model exhausted its output-token allowance without producing an action (DeepSeek-V4-Flash: 86/108 episodes); these are counted as failures, and budget used is still normalized by the full step budget.

\begin{table*}[t]
\caption{Results in the exploitation-only setting, aggregated over default and reasoning variants (216 episodes per model). Since no exploration is required, only exploitation error is reported. Budget used is the percentage of the step budget, averaged over all episodes (Total) or successful episodes only (Success); N/A indicates no successful episodes. Models are sorted by accuracy.}
\label{tab:exploit_only_combined_prompts}
\centering
\setlength{\tabcolsep}{1pt}
\resizebox{\textwidth}{!}{%
\begin{tabular}{lrrrr}
\toprule
\textbf{Model} & \textbf{Accuracy} $\uparrow$ & \textbf{Exploitation Error} $\downarrow$ & \textbf{Avg. Budget Used (Total)} $\downarrow$ & \textbf{Avg. Budget Used (Success)} $\downarrow$ \\
\midrule
Kimi-K2.6~\citep{kimi_k26} & 87.04\% & 23.02\% & 46.06\% & 38.56\% \\
MiniMax-M3~\citep{minimax_m3} & 84.72\% & 27.56\% & 52.90\% & 44.94\% \\
Hy3~\citep{hy3} & 78.70\% & 31.52\% & 52.18\% & 39.25\% \\
Muse-Glimmer-30B~\citep{muse_glimmer_30b} & 74.54\% & 30.80\% & 60.47\% & 46.97\% \\
Qwen3.5-397B-A17B~\citep{qwen35_397b_a17b} & 71.76\% & 34.99\% & 59.31\% & 44.91\% \\
Qwen3.6-35B-A3B~\citep{qwen36_35b_a3b} & 57.41\% & 48.94\% & 63.53\% & 38.07\% \\
Gemma-4-31B-IT~\citep{gemma4_31b} & 52.78\% & 57.39\% & 70.93\% & 44.92\% \\
Qwen3.8-27B~\citep{qwen38_27b} & 52.31\% & 68.72\% & 65.45\% & 34.83\% \\
DeepSeek-V4-Flash~\citep{deepseek_v4_flash} & 44.44\% & 28.53\% & 41.52\% & 54.93\% \\
Qwen3.6-27B~\citep{qwen36_27b} & 43.06\% & 52.57\% & 55.95\% & 36.81\% \\
Qwen3.5-9B~\citep{qwen35_9b} & 26.85\% & 64.52\% & 85.56\% & 51.71\% \\
NVIDIA-Nemotron-3-Nano-30B-A3B-BF16~\citep{nemotron3_nano_30b} & 13.89\% & 69.42\% & 90.96\% & 47.13\% \\
Hunyuan-A13B-Instruct~\citep{hunyuan_a13b} & 12.04\% & 74.68\% & 92.69\% & 51.05\% \\
Gemma-4-26B-A4B-IT~\citep{gemma4_26b_a4b} & 10.19\% & 72.31\% & 95.63\% & 57.06\% \\
Gemma-4-12B-IT~\citep{gemma4_12b} & 9.72\% & 75.40\% & 94.65\% & 45.01\% \\
Qwen3.5-4B~\citep{qwen35_4b} & 7.87\% & 72.42\% & 95.81\% & 53.16\% \\
GLM-4.7-Flash~\citep{glm47_flash} & 6.94\% & 71.77\% & 95.22\% & 57.09\% \\
Magistral-Small-2509~\citep{magistral_small} & 6.48\% & 82.45\% & 96.79\% & 50.47\% \\
Ministral-3-14B-Reasoning-2512~\citep{ministral3_14b_reasoning} & 4.63\% & 72.65\% & 97.90\% & 54.67\% \\
Ministral-3-8B-Reasoning-2512~\citep{ministral3_8b_reasoning} & 3.70\% & 72.59\% & 98.56\% & 61.07\% \\
Qwen3.5-2B~\citep{qwen35_2b} & 1.39\% & 74.23\% & 85.87\% & 75.84\% \\
Granite-4.1-8B~\citep{granite41_8b} & 0.00\% & 89.83\% & 100.00\% & N/A \\
\bottomrule
\end{tabular}%
}
\end{table*}

\begin{table*}[t]
\caption{Results in the exploitation-only setting without reasoning (108 episodes per model). Budget used is the percentage of the step budget, averaged over all episodes (Total) or successful episodes only (Success); N/A indicates no successful episodes. Models are sorted by accuracy.}
\label{tab:exploit_only_default_prompts}
\centering
\setlength{\tabcolsep}{1pt}
\resizebox{\textwidth}{!}{%
\begin{tabular}{lrrrr}
\toprule
\textbf{Model} & \textbf{Accuracy} $\uparrow$ & \textbf{Exploitation Error} $\downarrow$ & \textbf{Avg. Budget Used (Total)} $\downarrow$ & \textbf{Avg. Budget Used (Success)} $\downarrow$ \\
\midrule
Kimi-K2.6~\citep{kimi_k26} & 75.00\% & 30.89\% & 63.01\% & 50.68\% \\
MiniMax-M3~\citep{minimax_m3} & 70.37\% & 35.39\% & 70.24\% & 57.70\% \\
DeepSeek-V4-Flash~\citep{deepseek_v4_flash} & 68.52\% & 28.98\% & 72.22\% & 59.46\% \\
Hy3~\citep{hy3} & 57.41\% & 41.98\% & 74.73\% & 55.97\% \\
Muse-Glimmer-30B~\citep{muse_glimmer_30b} & 56.48\% & 35.95\% & 79.53\% & 63.76\% \\
Qwen3.5-397B-A17B~\citep{qwen35_397b_a17b} & 54.63\% & 41.85\% & 74.23\% & 52.83\% \\
Qwen3.6-27B~\citep{qwen36_27b} & 24.07\% & 62.47\% & 88.29\% & 51.35\% \\
Qwen3.6-35B-A3B~\citep{qwen36_35b_a3b} & 16.67\% & 61.88\% & 93.26\% & 59.56\% \\
Gemma-4-31B-IT~\citep{gemma4_31b} & 15.74\% & 71.27\% & 93.04\% & 55.80\% \\
Gemma-4-12B-IT~\citep{gemma4_12b} & 8.33\% & 73.83\% & 96.08\% & 53.00\% \\
Gemma-4-26B-A4B-IT~\citep{gemma4_26b_a4b} & 8.33\% & 67.67\% & 97.23\% & 66.76\% \\
Magistral-Small-2509~\citep{magistral_small} & 5.56\% & 78.86\% & 97.76\% & 59.68\% \\
Qwen3.8-27B~\citep{qwen38_27b} & 5.56\% & 87.10\% & 96.60\% & 38.73\% \\
Ministral-3-14B-Reasoning-2512~\citep{ministral3_14b_reasoning} & 4.63\% & 62.18\% & 98.70\% & 72.00\% \\
Qwen3.5-9B~\citep{qwen35_9b} & 4.63\% & 76.24\% & 97.77\% & 51.85\% \\
Ministral-3-8B-Reasoning-2512~\citep{ministral3_8b_reasoning} & 1.85\% & 56.87\% & 99.67\% & 82.21\% \\
Qwen3.5-2B~\citep{qwen35_2b} & 0.93\% & 81.11\% & 99.58\% & 55.17\% \\
Qwen3.5-4B~\citep{qwen35_4b} & 0.93\% & 82.14\% & 99.76\% & 73.81\% \\
Hunyuan-A13B-Instruct~\citep{hunyuan_a13b} & 0.93\% & 84.63\% & 99.35\% & 29.82\% \\
Granite-4.1-8B~\citep{granite41_8b} & 0.00\% & 88.91\% & 100.00\% & N/A \\
NVIDIA-Nemotron-3-Nano-30B-A3B-BF16~\citep{nemotron3_nano_30b} & 0.00\% & 81.52\% & 100.00\% & N/A \\
GLM-4.7-Flash~\citep{glm47_flash} & 0.00\% & 78.41\% & 100.00\% & N/A \\
\bottomrule
\end{tabular}%
}
\end{table*}

\begin{table*}[t]
\caption{Results in the exploitation-only setting with reasoning (108 episodes per model). Budget used is the percentage of the step budget, averaged over all episodes (Total) or successful episodes only (Success); N/A indicates no successful episodes. Models are sorted by accuracy.}
\label{tab:exploit_only_reasoning_prompts}
\centering
\setlength{\tabcolsep}{1pt}
\resizebox{\textwidth}{!}{%
\begin{tabular}{lrrrr}
\toprule
\textbf{Model} & \textbf{Accuracy} $\uparrow$ & \textbf{Exploitation Error} $\downarrow$ & \textbf{Avg. Budget Used (Total)} $\downarrow$ & \textbf{Avg. Budget Used (Success)} $\downarrow$ \\
\midrule
Hy3~\citep{hy3} & 100.00\% & 2.91\% & 29.64\% & 29.64\% \\
MiniMax-M3~\citep{minimax_m3} & 99.07\% & 10.51\% & 35.56\% & 35.88\% \\
Kimi-K2.6~\citep{kimi_k26} & 99.07\% & 3.72\% & 29.12\% & 29.38\% \\
Qwen3.8-27B~\citep{qwen38_27b} & 99.07\% & 8.16\% & 34.30\% & 34.62\% \\
Qwen3.6-35B-A3B~\citep{qwen36_35b_a3b} & 98.15\% & 7.24\% & 33.80\% & 34.42\% \\
Muse-Glimmer-30B~\citep{muse_glimmer_30b} & 92.59\% & 19.89\% & 41.41\% & 36.73\% \\
Gemma-4-31B-IT~\citep{gemma4_31b} & 89.81\% & 29.34\% & 48.82\% & 43.02\% \\
Qwen3.5-397B-A17B~\citep{qwen35_397b_a17b} & 88.89\% & 22.03\% & 44.38\% & 40.04\% \\
Qwen3.6-27B~\citep{qwen36_27b} & 62.04\% & 5.85\% & 23.61\% & 31.17\% \\
Qwen3.5-9B~\citep{qwen35_9b} & 49.07\% & 47.50\% & 73.35\% & 51.69\% \\
NVIDIA-Nemotron-3-Nano-30B-A3B-BF16~\citep{nemotron3_nano_30b} & 27.78\% & 54.56\% & 81.92\% & 47.13\% \\
Hunyuan-A13B-Instruct~\citep{hunyuan_a13b} & 23.15\% & 62.70\% & 86.04\% & 51.90\% \\
DeepSeek-V4-Flash~\citep{deepseek_v4_flash} & 20.37\% & 24.01\% & 10.82\% & 39.71\% \\
Qwen3.5-4B~\citep{qwen35_4b} & 14.81\% & 61.59\% & 91.85\% & 51.87\% \\
GLM-4.7-Flash~\citep{glm47_flash} & 13.89\% & 64.27\% & 90.43\% & 57.09\% \\
Gemma-4-26B-A4B-IT~\citep{gemma4_26b_a4b} & 12.04\% & 77.15\% & 94.02\% & 50.35\% \\
Gemma-4-12B-IT~\citep{gemma4_12b} & 11.11\% & 77.06\% & 93.22\% & 39.02\% \\
Magistral-Small-2509~\citep{magistral_small} & 7.41\% & 86.07\% & 95.82\% & 43.57\% \\
Ministral-3-8B-Reasoning-2512~\citep{ministral3_8b_reasoning} & 5.56\% & 88.53\% & 97.45\% & 54.02\% \\
Ministral-3-14B-Reasoning-2512~\citep{ministral3_14b_reasoning} & 4.63\% & 83.53\% & 97.10\% & 37.35\% \\
Qwen3.5-2B~\citep{qwen35_2b} & 1.85\% & 63.55\% & 72.15\% & 86.17\% \\
Granite-4.1-8B~\citep{granite41_8b} & 0.00\% & 90.76\% & 100.00\% & N/A \\
\bottomrule
\end{tabular}%
}
\end{table*}

\clearpage
\subsection[Sensitivity to Budget Constants]{Sensitivity to Budget Constants}
\label{apdx:budget_sensitivity}
We evaluate sensitivity to the episode-budget multiplier $\alpha$ and the revisit allowance $k$.

\paragraph[Episode budget $\alpha$.]{Episode budget.} We rescore each saved trajectory at $B_\alpha=\alpha|\mathcal{O}|$ for $\alpha\in\{1,2,3\}$, with $k=2$ fixed. An episode succeeds only if its goal is activated by the cutoff. Errors and eligible-step denominators include only actions through that cutoff; early terminal failures remain failures.

\paragraph[Revisit allowance $k$.]{Revisit allowance.} We hold $\alpha=3$ fixed and replace the edge and node reuse terms by $e_t(k)=\sum_e\max(m_t(e)-k,0)$ and $n_t(k)=\sum_v\max(n_t(v)-k,0)$, for $k\in\{1,2,3\}$. The cycle term, progress resets, gain test, and error attribution remain unchanged.

\paragraph[Results.]{Results.} Across the tested episode budgets, exploration--success $R^2$ ranges from 0.616 to 0.907, and exploitation--success $R^2$ ranges from 0.501 to 0.875. Across revisit allowances, the corresponding ranges are 0.872--0.892 and 0.874--0.876. The revisit-allowance sweep does not significantly alter the $R^2$ value, proving robustness against this choice. In contrast, the $R^2$ values change significantly when $\alpha=1$. However, this is expected as with $\alpha=1$, we are allowing the agent to visit the cells exactly once, which will significantly limit the success of the models. We observe that with increasd $\alpha$, the strong correlation is recovered. Tables~\ref{tab:sensitivity_alpha_correlations}--\ref{tab:sensitivity_k_models} report the regressions and per-model results.

\begin{table}[htbp]
\centering
\caption{Sensitivity of the success--error relationships to $\alpha$; $k=2$. Both columns report $R^2$ for the log-error regression. $\alpha=1$ is equivalent to visiting the cells exactly once, which significantly limit success, and hence the metric's $R^2$ values are affected.}
\label{tab:sensitivity_alpha_correlations}

\begin{tabular}{lcc}
\toprule
$\alpha$ & Exploration vs. success & Exploitation vs. success \\
\midrule
1 & 0.616 & 0.501 \\
2 & 0.907 & 0.861 \\
3 (baseline) & 0.885 & 0.875 \\
\bottomrule
\end{tabular}%

\end{table}
\begin{table}[htbp]
\centering
\caption{Sensitivity of the success--error relationships to $k$; $\alpha=3$. Both columns report $R^2$ for the log-error regression. The metric is not sensitive to the budget hyperparameter $k$.}
\label{tab:sensitivity_k_correlations}

\begin{tabular}{lcc}
\toprule
$k$ & Exploration vs. success & Exploitation vs. success \\
\midrule
1 & 0.872 & 0.874 \\
2 (baseline) & 0.885 & 0.875 \\
3 & 0.892 & 0.876 \\
\bottomrule
\end{tabular}%

\end{table}
\begin{table}[p]
\centering
\caption{Per-model success rate (\%) under different episode budgets, with $k=2$. Each value uses 216 evaluations. Model order follows success at $\alpha=3$.}
\label{tab:sensitivity_alpha_models}
\small

\begin{tabular}{lrrr}
\toprule
Model & $\alpha=1$ & $\alpha=2$ & $\alpha=3$ \\
\midrule
Gemini 3.1 Pro & 20.4 & 92.1 & 100.0 \\
Claude Opus 4.6 & 22.2 & 82.9 & 99.5 \\
Claude Sonnet 4.6 & 10.2 & 81.5 & 97.2 \\
Claude Haiku 4.5 & 13.4 & 60.6 & 91.7 \\
GPT 5.4 & 19.4 & 66.7 & 89.4 \\
Gemini 3 Flash & 17.1 & 64.4 & 88.4 \\
Qwen 3.8 27B & 4.2 & 50.9 & 75.0 \\
Kimi K2.6 & 6.5 & 53.2 & 73.1 \\
GPT 4.1 & 8.8 & 39.4 & 66.2 \\
Hy3 & 3.2 & 45.8 & 63.9 \\
Muse Glimmer 30B & 3.2 & 37.5 & 62.0 \\
Gemini 3.1 Flash Lite & 9.7 & 36.1 & 50.9 \\
Qwen 3.6 35B A3B & 1.4 & 28.2 & 50.5 \\
GPT OSS 120B & 1.4 & 26.9 & 49.1 \\
MiniMax M3 & 0.5 & 23.1 & 47.7 \\
Gemma 4 31B & 2.8 & 23.1 & 46.8 \\
Qwen 3.5 397B A17B & 2.8 & 28.7 & 44.0 \\
Qwen 3.6 27B & 4.6 & 33.8 & 43.5 \\
DeepSeek V4 Flash & 4.2 & 26.4 & 41.7 \\
Gemma 4 26B A4B & 0.5 & 3.2 & 5.1 \\
Gemma 4 12B & 0.0 & 2.8 & 4.6 \\
GPT 4.1 mini & 0.0 & 3.2 & 3.2 \\
Ministral 3 14B & 0.0 & 0.9 & 2.8 \\
Qwen 3.5 9B & 0.5 & 0.5 & 2.3 \\
GLM 4.7 Flash & 0.5 & 0.9 & 1.9 \\
Magistral Small & 0.0 & 0.5 & 1.4 \\
Nemotron 3 Nano 30B & 0.0 & 0.0 & 1.4 \\
Qwen 3.5 2B & 0.0 & 1.4 & 1.4 \\
Qwen 3.5 4B & 0.0 & 0.9 & 1.4 \\
GPT 5.4 mini & 0.5 & 0.5 & 1.4 \\
GPT 5.4 nano & 0.0 & 0.0 & 1.4 \\
Hunyuan A13B & 0.0 & 0.5 & 0.9 \\
Ministral 3 8B & 0.0 & 0.0 & 0.5 \\
Granite 4.1 8B & 0.0 & 0.5 & 0.5 \\
GPT 4.1 nano & 0.0 & 0.0 & 0.0 \\
\bottomrule
\end{tabular}%

\end{table}
\begin{table}[p]
\centering
\caption{Per-model exploration and exploitation error rates (\%) for revisit allowances $k\in\{1,2,3\}$, while fixing $\alpha=3$. Rates pool errors over eligible steps; success rates are unchanged by $k$.}
\label{tab:sensitivity_k_models}
\scriptsize
\setlength{\tabcolsep}{4pt}

\begin{tabular}{lrrrrrr}
\toprule
 & \multicolumn{3}{c}{Exploration error} & \multicolumn{3}{c}{Exploitation error} \\
\cmidrule(lr){2-4}\cmidrule(lr){5-7}
Model & $k=1$ & $k=2$ & $k=3$ & $k=1$ & $k=2$ & $k=3$ \\
\midrule
Gemini 3.1 Pro & 1.6 & 1.2 & 1.1 & 0.9 & 0.8 & 0.8 \\
Claude Opus 4.6 & 2.4 & 2.0 & 1.9 & 2.0 & 1.5 & 1.3 \\
Claude Sonnet 4.6 & 6.9 & 5.4 & 5.1 & 2.9 & 2.3 & 2.2 \\
Claude Haiku 4.5 & 9.9 & 8.2 & 7.8 & 5.4 & 4.5 & 4.3 \\
GPT 5.4 & 9.7 & 8.1 & 7.8 & 7.2 & 6.0 & 5.7 \\
Gemini 3 Flash & 11.2 & 9.5 & 8.7 & 5.5 & 4.7 & 4.4 \\
Qwen 3.8 27B & 22.1 & 16.4 & 13.0 & 14.6 & 11.1 & 9.4 \\
Kimi K2.6 & 16.2 & 11.3 & 9.4 & 6.0 & 4.8 & 4.7 \\
GPT 4.1 & 18.0 & 13.9 & 12.6 & 19.2 & 16.8 & 16.2 \\
Hy3 & 14.4 & 10.1 & 8.9 & 4.6 & 3.7 & 3.6 \\
Muse Glimmer 30B & 15.9 & 11.1 & 9.8 & 11.5 & 9.2 & 8.6 \\
Gemini 3.1 Flash Lite & 22.9 & 16.5 & 14.2 & 17.9 & 14.9 & 14.2 \\
Qwen 3.6 35B A3B & 37.2 & 30.5 & 25.4 & 31.0 & 25.3 & 21.1 \\
GPT OSS 120B & 28.3 & 23.0 & 18.6 & 27.5 & 23.7 & 20.5 \\
MiniMax M3 & 21.4 & 15.6 & 13.1 & 16.5 & 12.5 & 11.2 \\
Gemma 4 31B & 24.8 & 17.6 & 14.4 & 6.4 & 5.3 & 5.2 \\
Qwen 3.5 397B A17B & 28.8 & 21.2 & 17.2 & 16.0 & 11.1 & 9.8 \\
Qwen 3.6 27B & 31.9 & 23.3 & 17.6 & 20.2 & 15.0 & 11.5 \\
DeepSeek V4 Flash & 16.3 & 11.7 & 10.4 & 12.9 & 10.5 & 10.2 \\
Gemma 4 26B A4B & 61.1 & 54.1 & 48.7 & 57.8 & 52.1 & 47.3 \\
Gemma 4 12B & 67.3 & 61.6 & 56.5 & 67.5 & 63.5 & 59.5 \\
GPT 4.1 mini & 67.9 & 60.9 & 53.9 & 67.1 & 61.1 & 55.3 \\
Ministral 3 14B & 65.1 & 58.4 & 53.1 & 68.3 & 61.7 & 56.3 \\
Qwen 3.5 9B & 67.4 & 61.9 & 56.6 & 56.7 & 50.9 & 44.8 \\
GLM 4.7 Flash & 63.5 & 57.4 & 51.9 & 54.3 & 45.4 & 38.4 \\
Magistral Small & 76.9 & 71.5 & 66.0 & 65.7 & 57.6 & 50.5 \\
Nemotron 3 Nano 30B & 67.2 & 61.7 & 56.4 & 55.6 & 48.2 & 42.3 \\
Qwen 3.5 2B & 65.2 & 59.5 & 53.9 & 66.9 & 59.7 & 51.9 \\
Qwen 3.5 4B & 69.4 & 63.9 & 58.8 & 42.9 & 36.1 & 31.4 \\
GPT 5.4 mini & 68.1 & 60.5 & 54.0 & 51.0 & 41.1 & 36.9 \\
GPT 5.4 nano & 55.6 & 47.6 & 41.8 & 58.5 & 52.7 & 47.8 \\
Hunyuan A13B & 73.2 & 68.5 & 63.5 & 79.6 & 73.1 & 66.0 \\
Ministral 3 8B & 62.6 & 55.1 & 49.1 & 72.5 & 68.8 & 66.2 \\
Granite 4.1 8B & 87.6 & 85.6 & 82.9 & 92.1 & 91.1 & 89.3 \\
GPT 4.1 nano & 86.0 & 83.2 & 79.9 & 81.6 & 75.5 & 68.8 \\
\bottomrule
\end{tabular}%
\end{table}

\clearpage

\section{Additional Experimental Details}
\label{sec:exp_details}

In this section we describe the task generation process, data statistics, and prompt examples with more detail.

\subsection{Task Generation}
\label{sec:app_task_generation}

We first generate each task instance programmatically with a fixed random seed, by controlling two preset controls: the size of the task DAG and the exploitation demand implied by the map. The task DAG size determines the structure of the graph by fixing the number of nodes in the DAG, for example, 4, 6, and 8 nodes for small, medium, and large size, respectively. We keep at most 3 nodes per layer (i.e. the same depth from the primitive nodes), and sample the number of prerequisites and dependencies (i.e. the edge structure) from predefined distributions. While there can be more than one primitive node of depth 0, we assume, without loss of generality, that task DAG contains exactly one goal node. All prerequisite edges point from shallower nodes to deeper nodes to ensure that the sampled graph is acyclic.

To prevent agents from exploiting semantics in node names, each node is named with a random four-character visible token, e.g. D7UX and 9J7T, rather than a meaningful name. Note that using simple alphanumerics such as A, B, C, D or 1, 2, 3, 4 can also erroneously inject information regarding the precedence. If the sampled graph contains any node disconnected from the goal node, the goal's prerequisite is repaired so that every node remains relevant to task completion.

After the task DAG is sampled, we map it onto a 2D grid. The 2D grid's size is determined with the ratio of the number of task nodes in the task DAG to the number of cells in the grid -- i.e. setting a lower density produces a larger map, unless the map size is fixed a priori.

The exploitation demand controls how the 2D grid map is generated, and we use two knobs to control the exploitation demand: (i) map density and (ii) corridor widths. Lower map density implies that the task nodes are placed in a more sparse fashion, implicitly increasing the distance between the task nodes, and hence requires more exploration. Similarly, increasing the corridor width not only increases the effective map size but also generates more admissible moves on each cell, compared to having long narrow paths. When generating the maps, we use densities of 0.1, 0.25, and 0.4 and corridor widths of 2--3, 1--3, and 1 cells for low, medium, and high exploitation demand, respectively. For corridor widths, we use 1, 1-3, 2-3 cells for low, medium, and high exploration. The cell in which the LM agent starts do not contain any task node, and corridors are generated between the starting position and task nodes to enforce that the traversable region is fully connected.

Beyond the task DAG and exploitation presets, the environment generator controls the spatial layout and budget per episode. First, the aspect ratio of 2D grid map specifies the target width-to-height ratio of the grid map, after the 2D grid map size is determined from the node budget and exploitation demand. The generator selects integer dimensions that best match this ratio. A budget parameter $\alpha$ defines the hidden interaction budget as the following:
\begin{equation}
    B = \alpha |\mathcal{O}|,
\end{equation}
where $|\mathcal{O}|$ denotes the number of traversable cells in the generated map. Importantly, $\alpha$ does not alter the sampled DAG or the grid geometry itself; it simply sets the maximum number of turns the LM agent can take to solve the task. In our setup, we fix the aspect ratio of 2D grid map as $1$ and a budget parameter as $3$. Appendix~\ref{apdx:budget_sensitivity} examines sensitivity to the episode budget and the edge/node revisit allowance.

Finally, we use three categorical distributions to determine the logical complexity of the symbolic task graph. For each non-primitive node, we use \texttt{option\_count\_probs} to control the number of alternative prerequisite sets and therefore the amount of \texttt{OR}-type branching, while \texttt{dependency\_count\_probs} controls the number of parents within each prerequisite set and therefore the \texttt{AND}-type arity of each option. The parameter \texttt{goal\_dependency\_count\_probs} plays the same role for the unique goal node, determining how many predecessors must be jointly satisfied at the top of the graph. Parent nodes are sampled by controlling parent-depth bias: a candidate parent at depth $d$ for a child at depth $D$ receives weight proportional to $\exp(-\beta((D-1)-d))$,
where $\beta$ is the bias parameter, where larger $\beta$ favors dependencies from more recent layers. We use $\beta = 1$ (i.e., \texttt{parent\_depth\_bias}$=1$) for all tasks. For easy, medium, and hard maps, respectively, we set \texttt{option\_count\_probs} to $\{1:1.0\}$, $\{1:0.8, 2:0.2\}$, and $\{1:0.6, 2:0.4\}$, and set both \texttt{dependency\_count\_probs} and \texttt{goal\_dependency\_count\_probs} to uniform distributions over $\{1,2\}$, $\{1,2\}$, and $\{1,2,3\}$.

\subsection{Statistics on Experiments}
\label{sec:app_data_statistics}

We report the number of episodes and seed configurations used in each experiment.

\paragraph{Main Experiment.}
For the main experiment, we use procedurally generated maps parameterized by three exploitation demand (\texttt{low}, \texttt{medium}, \texttt{high}) and three task-DAG size (\texttt{small}, \texttt{medium}, \texttt{large}), yielding $3 \times 3 = 9$ distinct map configurations.
Each configuration is run with 3 seeds (seeds 0, 1, 2), producing 27 episodes per prompt set.
We evaluate 8 prompt sets (4 base types $\times$ 2 reasoning modes: \texttt{default}, \texttt{default-balance}, \texttt{default-exploration}, \texttt{default-exploitation}, \texttt{reasoning}, \texttt{reasoning-balance}, \texttt{reasoning-exploration}, \texttt{reasoning-exploitation}), resulting in $8 \times 27 = 216$ total episodes per model. The harness engineering experiments use the same procedurally generated dataset of 9 map configurations with 3 seeds per configuration, identical to the main experiment.

\paragraph{Semantic Experiment.}
For the semantic experiments, we use 4 hand-crafted custom maps (custom-201 through custom-204), each available in both a semantic variant (with meaningful state names) and a symbolic variant (with abstract identifiers).
Each map is evaluated with 5 seeds (seeds 0--4), yielding $4 \times 5 = 20$ episodes per prompt-variant combination.
We test 2 prompt sets (\texttt{default}, \texttt{reasoning}) across both semantic and symbolic variants, for a total of $2 \times 2 \times 20 = 80$ episodes per model.

\subsection{Prompts}
\label{sec:app_prompts}

Each system prompt consists of three parts: (1)~an environment description, (2)~an optional strategy prompt, and (3)~an action format specification. Parts~(1) and~(3) are identical across all four variants. The four variants differ only in the strategy prompt (part~2). Below we show the full system prompt for each variant, with the strategy sentence highlighted in \colorbox{yellow!30}{yellow}. We also provide example turns illustrating the observation format.

\begin{itemize}[leftmargin=*]
    \item \textbf{Base Prompt.} No strategy sentence is included. The agent receives only the environment description and action format, leaving the exploration--exploitation tradeoff entirely to the model's own reasoning.
    \item \textbf{Exploration Prompt.} A single strategy sentence is inserted between the environment description and the action format, instructing the agent to prioritize visiting unvisited cells.
    \item \textbf{Exploitation Prompt.} The strategy sentence instructs the agent to prioritize achieving already-discovered states whose prerequisites are satisfied.
    \item \textbf{Balance Prompt.} The strategy sentence combines both the exploration and exploitation definitions and asks the agent to choose the balance that minimizes total steps.

\end{itemize}

\paragraph{Example Turns.}
At each turn, the agent receives an observation and must produce a single JSON action. Below are two example turns: one where the agent finds nothing, and one where a state is discovered.

\begin{tcolorbox}[colback=blue!3, colframe=blue!40, boxrule=0.5pt, arc=2pt, fontupper=\small\ttfamily, title=Example Observation (Empty Cell)]
OBSERVATION: You are at [7, 0]. You found nothing here.
Available directions: up
\end{tcolorbox}

\begin{tcolorbox}[colback=green!3, colframe=green!40, boxrule=0.5pt, arc=2pt, fontupper=\small\ttfamily, title=Example Agent Response]
\{"action":"up"\}
\end{tcolorbox}

\begin{tcolorbox}[colback=blue!3, colframe=blue!40, boxrule=0.5pt, arc=2pt, fontupper=\small\ttfamily, title=Example Observation (State Discovery)]
OBSERVATION: You are at [0, 1]. You discovered state U\_02. U\_02 has no prerequisites and is immediately activated! U\_02 has ancestors: R\_01, G\_00.
Available directions: up, down, right
\end{tcolorbox}

\begin{promptbox}[title=System Prompt (Base)]
You are controlling an agent in a partially observed symbolic grid environment.
Your objective is to activate the goal state.
At each step, you are given your current position, the directions you can legally move, and any newly discovered symbolic states at your current cell.
Newly discovered states may include prerequisite information and ancestor hints.
A state can be activated when you are on its cell and its prerequisites are satisfied.
The full map, hidden budget, and undiscovered states are not available to you.
Reply with exactly one JSON object containing one valid action from available\_directions like this: \{"action":"up"\}, \{"action":"down"\}, \{"action":"left"\}, \{"action":"right"\}
\end{promptbox}

\begin{promptbox}[title=System Prompt (Exploration)]
You are controlling an agent in a partially observed symbolic grid environment.
Your objective is to activate the goal state.
At each step, you are given your current position, the directions you can legally move, and any newly discovered symbolic states at your current cell.
Newly discovered states may include prerequisite information and ancestor hints.
A state can be activated when you are on its cell and its prerequisites are satisfied.
The full map, hidden budget, and undiscovered states are not available to you.
\prompthighlight{Prioritize exploration when deciding where to move. Treat exploration as deliberately moving toward cells you have not visited yet and roaming to uncover cells and symbolic states that you have not discovered yet.}
Reply with exactly one JSON object containing one valid action from available\_directions like this: \{"action":"up"\}, \{"action":"down"\}, \{"action":"left"\}, \{"action":"right"\}
\end{promptbox}

\begin{promptbox}[title=System Prompt (Exploitation)]
You are controlling an agent in a partially observed symbolic grid environment.
Your objective is to activate the goal state.
At each step, you are given your current position, the directions you can legally move, and any newly discovered symbolic states at your current cell.
Newly discovered states may include prerequisite information and ancestor hints.
A state can be activated when you are on its cell and its prerequisites are satisfied.
The full map, hidden budget, and undiscovered states are not available to you.
\prompthighlight{Prioritize exploitation when deciding where to move. Among the symbolic states you have already discovered, first target states whose prerequisites are already satisfied, and move along the shortest available path to activate them.}
Reply with exactly one JSON object containing one valid action from available\_directions like this: \{"action":"up"\}, \{"action":"down"\}, \{"action":"left"\}, \{"action":"right"\}
\end{promptbox}

\begin{promptbox}[title=System Prompt (Balance)]
You are controlling an agent in a partially observed symbolic grid environment.
Your objective is to activate the goal state.
At each step, you are given your current position, the directions you can legally move, and any newly discovered symbolic states at your current cell.
Newly discovered states may include prerequisite information and ancestor hints.
A state can be activated when you are on its cell and its prerequisites are satisfied.
The full map, hidden budget, and undiscovered states are not available to you.
\prompthighlight{Balance exploration and exploitation when deciding where to move. Treat exploration as deliberately moving toward cells you have not visited yet and roaming to uncover cells and symbolic states that you have not discovered yet. Treat exploitation as targeting already discovered symbolic states whose prerequisites are already satisfied and moving along the shortest available path to activate them. Choose the balance between these two behaviors based on which actions are most likely to solve the task in the fewest steps.}
Reply with exactly one JSON object containing one valid action from available\_directions like this: \{"action":"up"\}, \{"action":"down"\}, \{"action":"left"\}, \{"action":"right"\}
\end{promptbox}

\begin{figure}[t]
\centering
\resizebox{\textwidth}{!}{%
\begin{tikzpicture}[
    box/.style={draw, rounded corners=3pt, minimum height=0.9cm, align=center, font=\small},
    envbox/.style={box, fill=green!12, minimum width=2.0cm},
    llmbox/.style={box, fill=blue!12, minimum width=1.6cm},
    membox/.style={box, fill=orange!15, minimum width=1.8cm},
    sumbox/.style={draw, rounded corners=2pt, fill=orange!8, align=left,
                   font=\scriptsize\ttfamily, text=orange!30!black, minimum width=2.8cm},
    arr/.style={-{Stealth[length=5pt]}, thick},
    memarrow/.style={arr, orange!70!black},
    lbl/.style={font=\large},
]

\node[llmbox] (llmA) at (3.0, 1.8) {LLM};

\begin{scope}[on background layer]
    \node[draw=gray!50, dashed, rounded corners=6pt, fill=gray!4,
          inner xsep=18pt, inner ysep=18pt,
          fit=(llmA)] (harnessA) {};
\end{scope}
\node[font=\scriptsize\itshape, text=gray!55, anchor=south] at (harnessA.north) {Agent Harness};

\node[envbox] (envA) at (-0.5, 1.8) {Environment};

\draw[arr] ([yshift=3pt]envA.east) -- node[above, font=\scriptsize] {observation} ([yshift=3pt]llmA.west);
\draw[arr] ([yshift=-3pt]llmA.west) -- node[below, font=\scriptsize] {action} ([yshift=-3pt]envA.east);

\node[llmbox] (llmB) at (12.5, 1.8) {LLM};
\node[membox] (mem) at (9.6, -0.3) {Memory\\[-2pt]Manager};
\node[sumbox] (summary) at (12.5, -0.3) {
    Visited: [0,0]...[2,3]\\
    Frontier: [1,3], [3,3]\\
    Activated: R\_01\\
    Activatable: B\_02
};

\begin{scope}[on background layer]
    \node[draw=gray!50, dashed, rounded corners=6pt, fill=gray!4,
          inner xsep=12pt, inner ysep=14pt,
          fit=(llmB)(mem)(summary)] (harnessB) {};
\end{scope}
\node[font=\scriptsize\itshape, text=gray!55, anchor=south] at (harnessB.north) {Agent Harness};

\node[envbox] (envB) at (6.0, 1.8) {Environment};

\draw[arr] ([yshift=3pt]envB.east) -- node[above, font=\scriptsize] {observation} ([yshift=3pt]llmB.west);
\draw[arr] ([yshift=-3pt]llmB.west) -- node[below, font=\scriptsize] {action} ([yshift=-3pt]envB.east);

\coordinate (fork) at (8.5, 1.90);
\fill[black] (fork) circle (1.5pt);
\draw[memarrow] (fork) |- (mem.west);

\draw[memarrow] (mem.east) -- (summary.west);
\draw[memarrow] (summary.north) -- (llmB.south);

\node[lbl] at (1.8, -1.8) {(a) Baseline};
\node[lbl] at (10.5, -1.8) {(b) Harness Engineering};

\end{tikzpicture}%
}%
\caption{Harness Design Comparison. (a)~In the baseline, the agent harness passes observations directly to the LLM and returns actions as in ~\citet{yao2023react}. (b)~With harness engineering, each observation is also routed through a rule-based memory manager that produces a structured summary (visited cells, frontier, activated and activatable states, ...) which is injected into the LLM prompt. The information content is identical but the relevant information is highlighted explicitly through harness engineering.}
\label{fig:harness_engineering}
\end{figure}

\section{Harness Engineering}
\label{sec:app_harness_example}

Figure~\ref{fig:harness_engineering} illustrates the information flow in the baseline agent harness and explicit harness engineering. Specifically, we inject a \textbf{structured memory summary} alongside the observation at every turn, which is a rule-based memory manager that aggregates past observations into a fixed-format summary and appends it to the turn prompt. Crucially, the memory summary contains \emph{no new information}: every field is derivable from the history of observations the agent has already received. The engineering lies entirely in \emph{how} the information is organized and presented.

The memory summary consists of the following components:
\begin{itemize}[leftmargin=*]
    \item \textbf{Learned coordinate system:} inferred from movement outcomes (e.g., ``up = y+1, right = x+1'').
    \item \textbf{Known goal state:} the goal state name, once discovered.
    \item \textbf{Visited cells:} cumulative list of all cells the agent has stepped on.
    \item \textbf{Traversable frontier:} cells known to be reachable (observed as neighbors) but not yet visited, directly supporting exploration.
    \item \textbf{Obstacle cells:} cells inferred to be impassable.
    \item \textbf{Discovered states:} each discovered state with its position, prerequisite conditions, and successor states.
    \item \textbf{Activated / activatable states:} which states have been activated, and which states currently have all prerequisites satisfied and are ready for activation, directly supporting exploitation.
\end{itemize}

The system prompt is updated to inform the model that this summary is available (lines highlighted in \colorbox{yellow!30}{yellow}):

\begin{promptbox}[title=System Prompt (with Harness Engineering)]
You are controlling an agent in a partially observed symbolic grid environment.
Your objective is to activate the goal state.
At each step, you are given your current observation that contains your current position, the directions you can legally move, and any newly discovered symbolic states at your current cell.
\prompthighlight{Also, you are given a summary of your explored map, visited cells, reachable frontier cells, discovered states, activated states, and prerequisite relations.}
Newly discovered states may include prerequisite information and ancestor hints.
A state can be activated when you are on its cell and its prerequisites are satisfied.
The full map, hidden budget, and undiscovered states are not available to you.
Reply with exactly one JSON object containing one valid action from available\_directions like this: \{"action":"up"\}, \{"action":"down"\}, \{"action":"left"\}, \{"action":"right"\}
\end{promptbox}

Below is an example turn prompt after 12 steps. The observation (top) is identical to the baseline; the memory summary (bottom) is the injected harness.

\begin{tcolorbox}[colback=blue!3, colframe=blue!40, boxrule=0.5pt, arc=2pt, fontupper=\small\ttfamily, title=Example Observation with Harness Engineering (Step 12)]
OBSERVATION: You are at [2, 3]. You found nothing here. Your available action is up, down, left, or right. You spent 12 steps.\medskip\\
From your movements so far, coordinate system: [x, y] where up=y+1, down=y-1, left=x-1, right=x+1. The goal state is G\_00. You have visited [0, 0], [0, 1], [1, 1], [1, 2], [2, 2], [2, 3]. You have not visited these cells yet, but you know you can pass through them: [1, 3], [3, 3]. You know you cannot pass through these cells: [2, 4]. You discovered R\_01 at [0, 0]. It is already activated. Once it is activated, you can next pursue B\_02. You discovered B\_02 at [1, 2]. It is not activated yet. To activate it, you should find R\_01 first. Once it is activated, you can next pursue G\_00. Activated states: R\_01. The discovered states whose prerequisites are already satisfied: B\_02.
\end{tcolorbox}

In this example, the baseline agent would need to recall from conversation history that B\_02's prerequisite R\_01 has already been activated and that B\_02 is therefore ready to achieve. With the harness, this information is stated explicitly (``The discovered states whose prerequisites are already satisfied: B\_02''), allowing the model to exploit this information without relying on its memory retrieval. Similarly, the frontier cells ([1, 3], [3, 3]) are explicitly listed, guiding exploration toward unvisited but reachable cells.

\section{Semantic Experiment}\label{sec:app_semantic_experiment}

\begin{figure}[t]
    \begin{center}
        \includegraphics[width=0.95\textwidth]{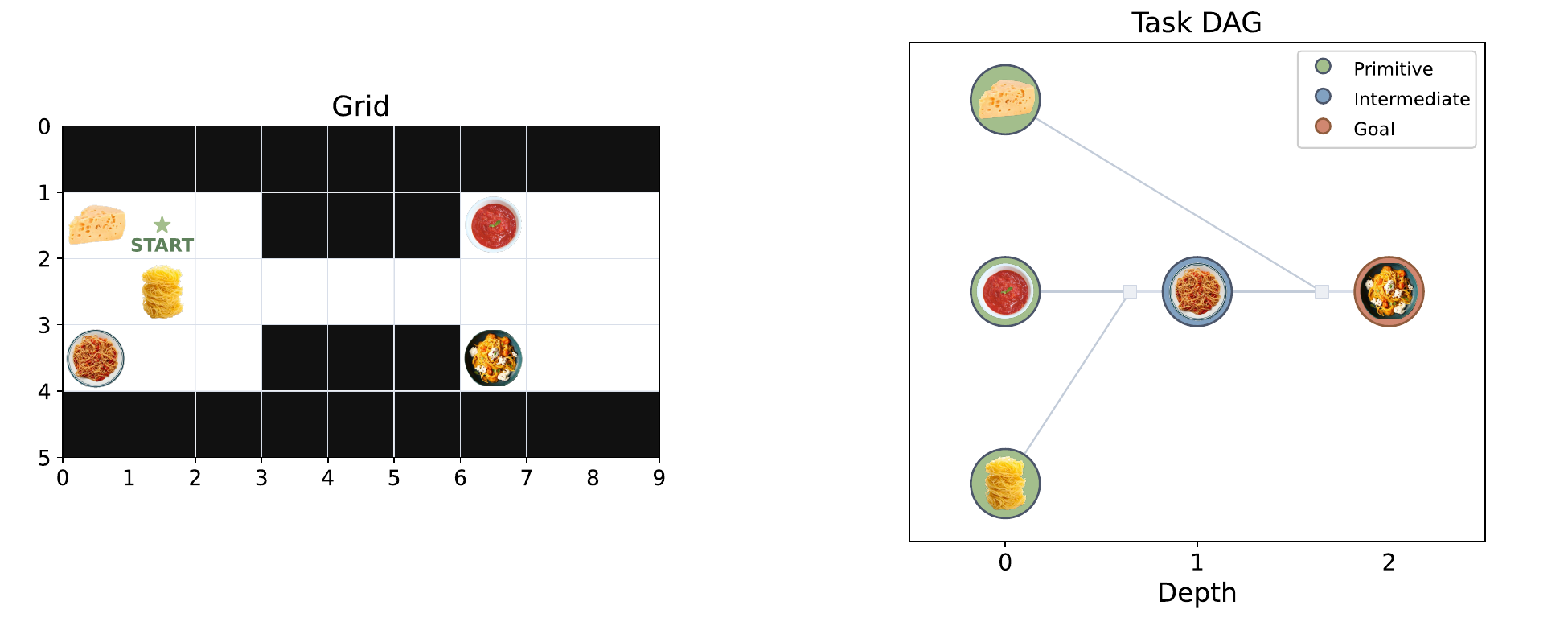}
    \end{center}
    \caption{An oracle view of the 2D grid map and the corresponding task DAG for experiments with semantic information injection. This example illustrates cooking tomato pasta with cheese. First, tomato pasta is made from pasta and tomato sauce. Second, tomato pasta with cheese is made by adding cheese to the tomato pasta. In this case, the LM agent may overly rely on semantic priors. For example, it might erroneously assume that the pasta sauce and cheese must be placed in proximity.}
    \label{fig:semantic_example_1}
\end{figure}

For the semantic information injection experiments, we generate four task instances based on pasta-cooking scenarios. As shown in Figure~\ref{fig:semantic_example_1}, the task DAG generation is motivated from real-world cooking tasks. To prepare the final dish, the agent must visit all required cells on the 2D grid to find the ingredients and perform intermediate steps.

\paragraph{Observation Format.}
The grid map, task DAG structure, and observation format remain identical between the symbolic and semantic experiments. The only difference lies in the \emph{task DAG's node names} presented to the agent in each observation. In symbolic experiments, state names are replaced with randomly generated four-character alphanumeric codes (e.g., \texttt{B4KD}, \texttt{H2NZ}, \texttt{Q7M2}), designed to carry no semantic meaning. In the semantic experiments, however, the task nodes are labeled with meaningful cooking-related names (e.g., \texttt{Pasta}, \texttt{Tomato Pasta}, \texttt{Tomato Pasta with Cheese}).

Below we show two example observations from the same 2D grid cell (i.e. what the environment returns to the LM agent), illustrating how the agent perceives a discovered state under each state.

\begin{tcolorbox}[colback=blue!3, colframe=blue!40, boxrule=0.5pt, arc=2pt, fontupper=\small\ttfamily, title=Observation --- Symbolic Condition]
OBSERVATION: You are at [1, 2]. You found B4KD which is now activated. Now, you can go to H2NZ. Your available action is down, left, right, or up. You spent 21 steps.
\end{tcolorbox}

\begin{tcolorbox}[colback=green!3, colframe=green!40, boxrule=0.5pt, arc=2pt, fontupper=\small\ttfamily, title=Observation --- Semantic Condition]
OBSERVATION: You are at [1, 2]. You found Pasta which is now activated. Now, you can go to Tomato Pasta. Your available action is down, left, right, or up. You spent 1 steps.
\end{tcolorbox}

\section{Examples of LM Agent's Runs in Our Environments}

We visualize the action trajectories of Claude Haiku 4.5,
Claude Sonnet 4.6,
Claude Opus 4.6,
Gemini 3 Flash,
Gemini 3.1 Flash Lite,
Gemini 3.1 Pro,
GPT-4.1 nano,
GPT-4.1 mini,
GPT-4.1,
GPT-5.4 nano,
GPT-5.4 mini,
GPT-5.4,
and GPT-OSS-120B,
and our metric values in Figures~\ref{fig:claude-haiku-4-5-1} to~\ref{fig:gpt-oss-120b-3} over each timestep.

\newcommand{\framewidth}{0.95\textwidth}

\newcommand{\geminiflash}{Gemini 3 Flash}
\newcommand{\geminipro}{Gemini 3.1 Pro}
\newcommand{\geminiflashlite}{Gemini 3.1 Flash Lite}

\newcommand{\gptfouronenano}{GPT-4.1 nano}
\newcommand{\gptfouronemini}{GPT-4.1 mini}
\newcommand{\gptfourone}{GPT-4.1}

\newcommand{\gptfivefournano}{GPT-5.4 nano}
\newcommand{\gptfivefourmini}{GPT-5.4 mini}
\newcommand{\gptfivefour}{GPT-5.4}

\newcommand{\gptoss}{GPT-OSS-120B}

\newcommand{\claudehaiku}{Claude Haiku 4.5}
\newcommand{\claudeopus}{Claude Opus 4.6}
\newcommand{\claudesonnet}{Claude Sonnet 4.6}

\newcommand{\currentmodel}{\claudeopus}

\renewcommand{\currentmodel}{\claudehaiku}

\begin{figure}[ht]
    \begin{center}
        \begin{subfigure}[t]{\framewidth}
        \centering
        \includegraphics[width=\textwidth,trim={0 0 0 33pt},clip]{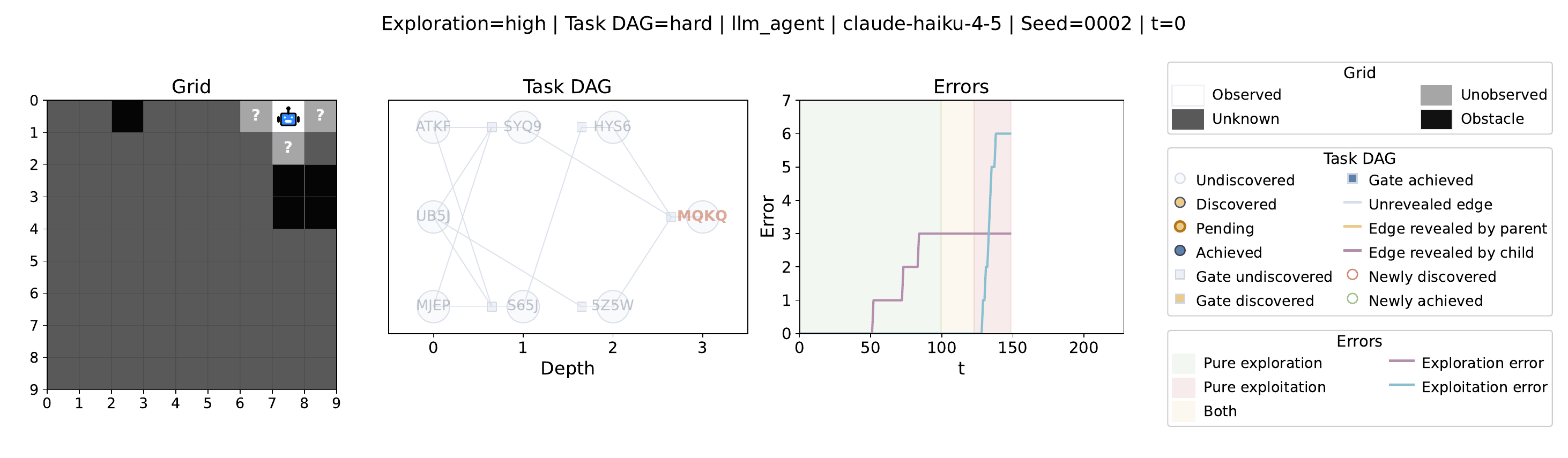}
        \caption{\currentmodel, $t = 0$}
        \end{subfigure}
        \begin{subfigure}[t]{\framewidth}
        \centering
        \includegraphics[width=\textwidth,trim={0 0 0 33pt},clip]{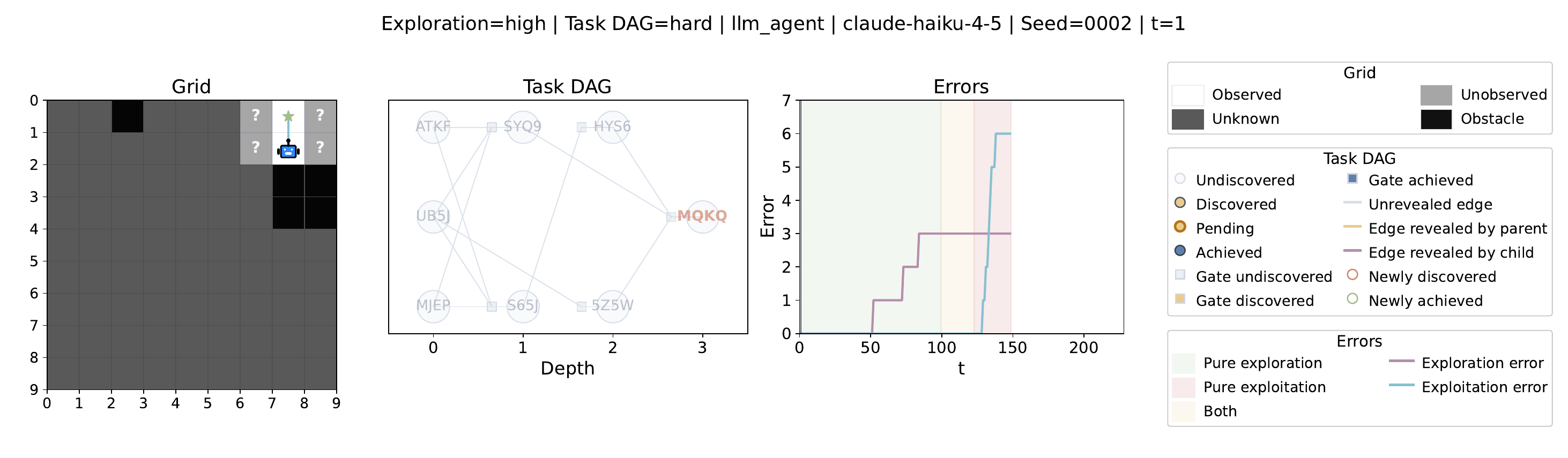}
        \caption{\currentmodel, $t = 1$}
        \end{subfigure}
        \begin{subfigure}[t]{\framewidth}
        \centering
        \includegraphics[width=\textwidth,trim={0 0 0 33pt},clip]{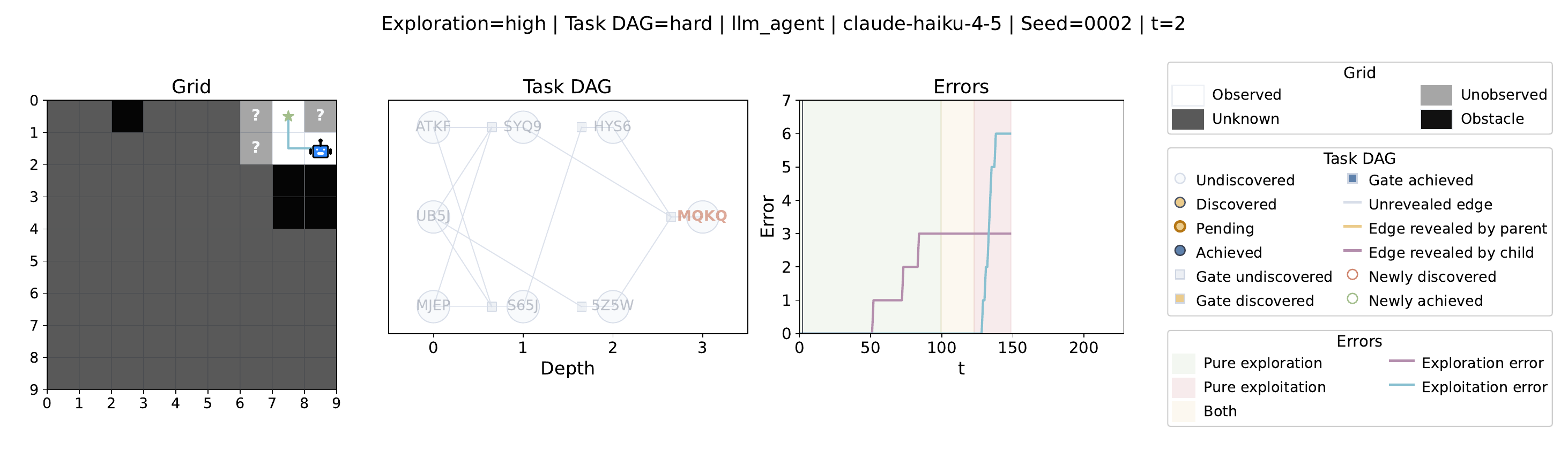}
        \caption{\currentmodel, $t = 2$}
        \end{subfigure}
        \begin{subfigure}[t]{\framewidth}
        \centering
        \includegraphics[width=\textwidth,trim={0 0 0 33pt},clip]{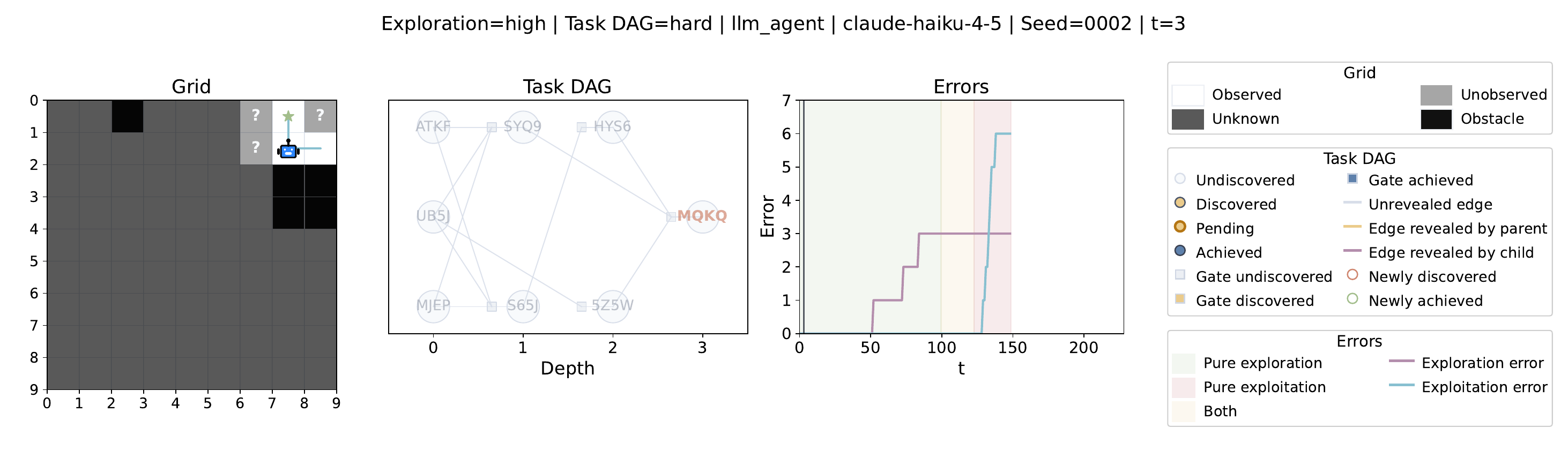}
        \caption{\currentmodel, $t = 3$}
        \end{subfigure}
        \begin{subfigure}[t]{\framewidth}
        \centering
        \includegraphics[width=\textwidth,trim={0 0 0 33pt},clip]{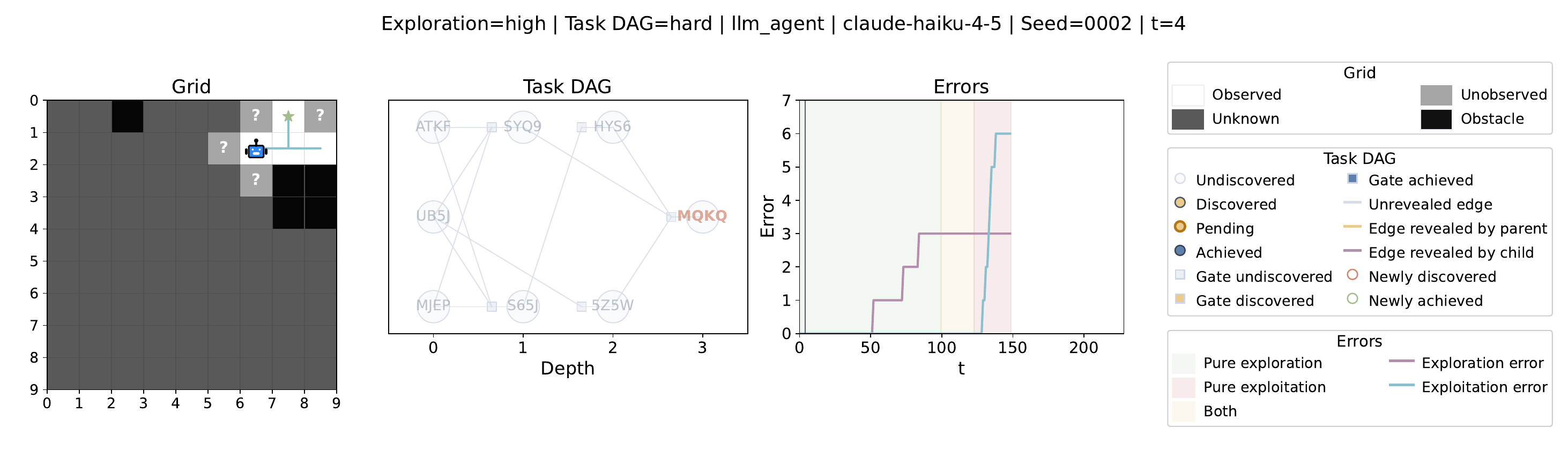}
        \caption{\currentmodel, $t = 4$}
        \end{subfigure}
    \end{center}
    \caption{Results of action trajectories and metric values over each timestep for~\currentmodel.}
    \label{fig:claude-haiku-4-5-1}
\end{figure}

\begin{figure}[ht]
    \begin{center}
        \begin{subfigure}[t]{\framewidth}
        \centering
        \includegraphics[width=\textwidth,trim={0 0 0 33pt},clip]{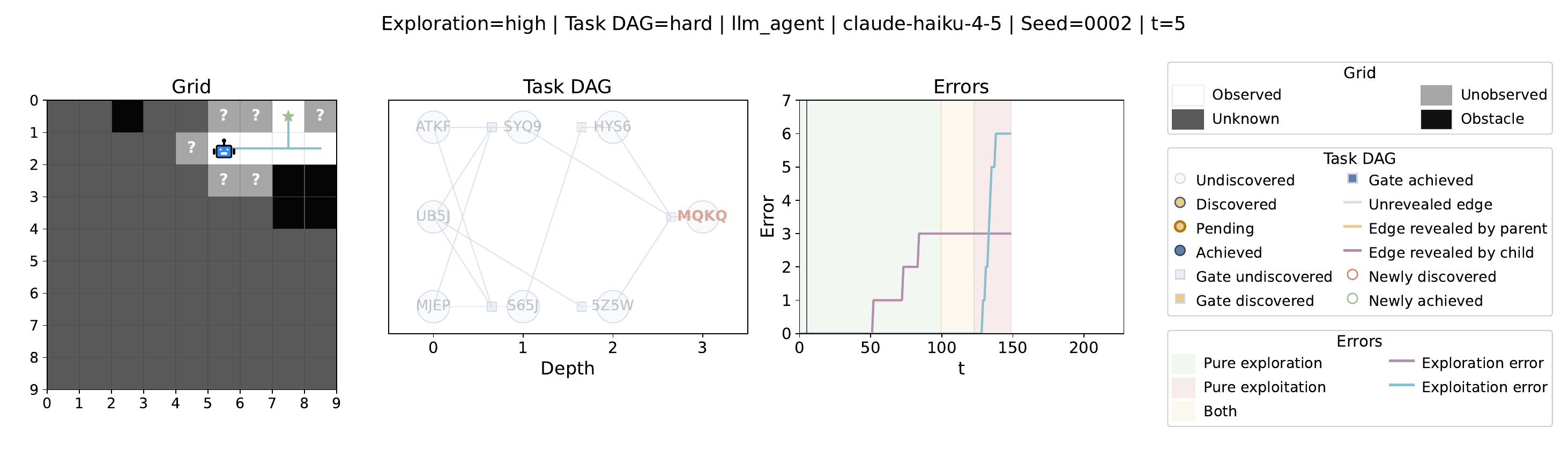}
        \caption{\currentmodel, $t = 5$}
        \end{subfigure}
        \begin{subfigure}[t]{\framewidth}
        \centering
        \includegraphics[width=\textwidth,trim={0 0 0 33pt},clip]{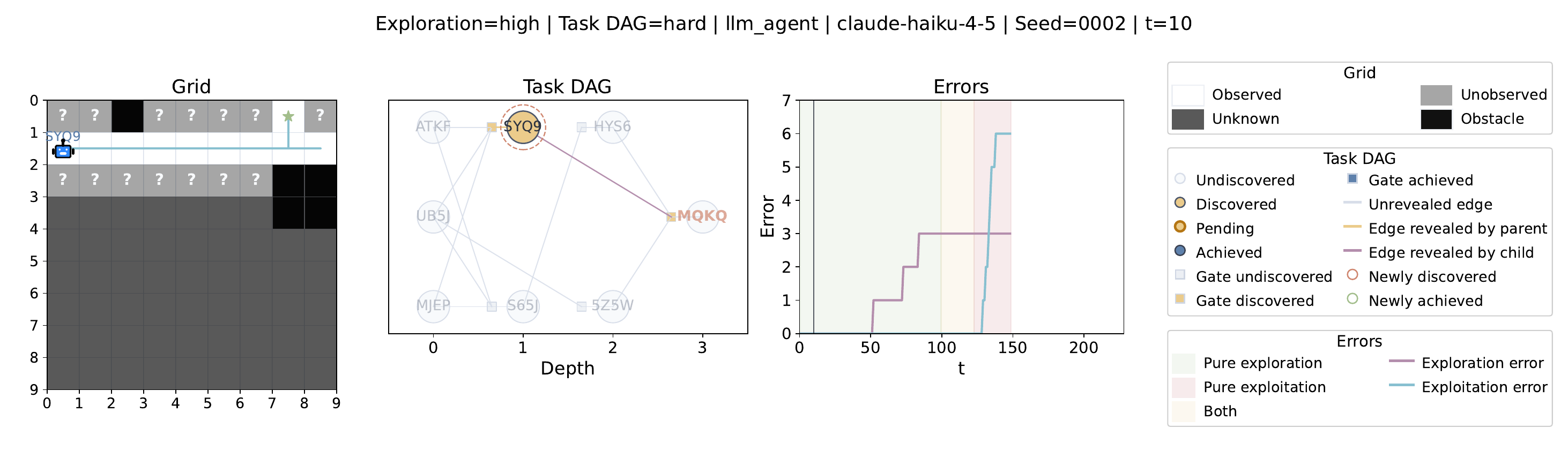}
        \caption{\currentmodel, $t = 10$}
        \end{subfigure}
        \begin{subfigure}[t]{\framewidth}
        \centering
        \includegraphics[width=\textwidth,trim={0 0 0 33pt},clip]{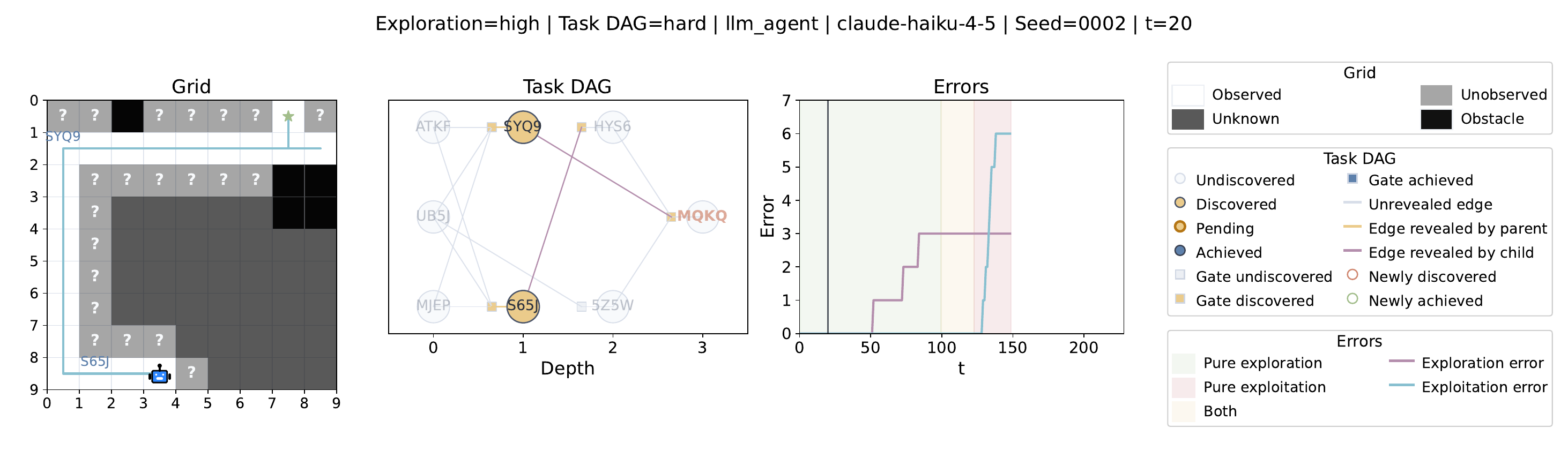}
        \caption{\currentmodel, $t = 20$}
        \end{subfigure}
        \begin{subfigure}[t]{\framewidth}
        \centering
        \includegraphics[width=\textwidth,trim={0 0 0 33pt},clip]{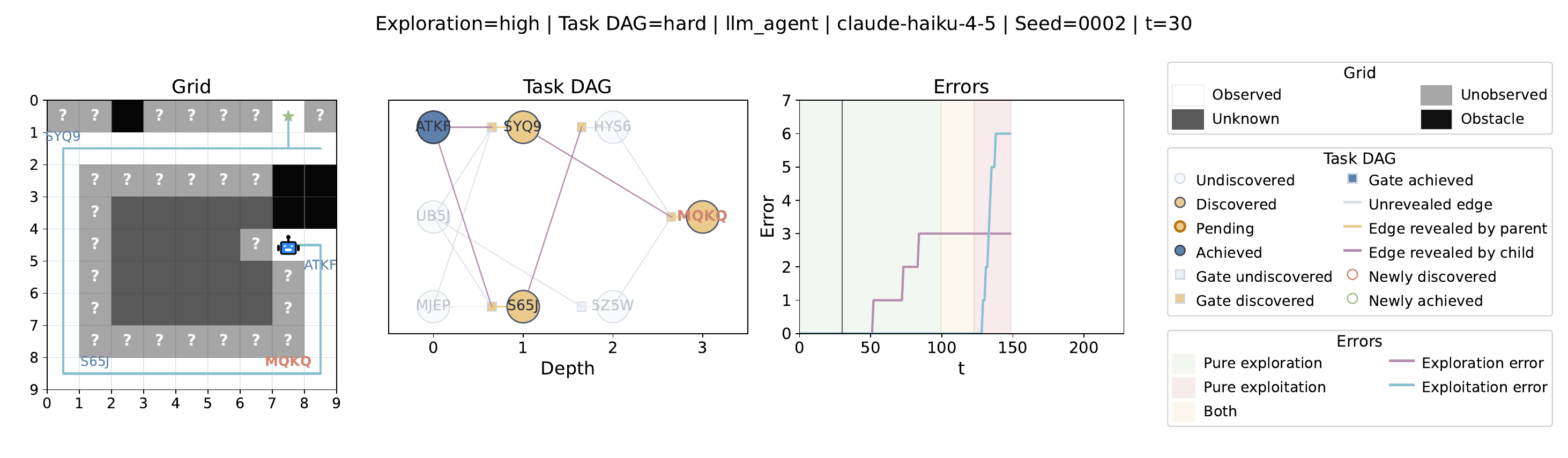}
        \caption{\currentmodel, $t = 30$}
        \end{subfigure}
        \begin{subfigure}[t]{\framewidth}
        \centering
        \includegraphics[width=\textwidth,trim={0 0 0 33pt},clip]{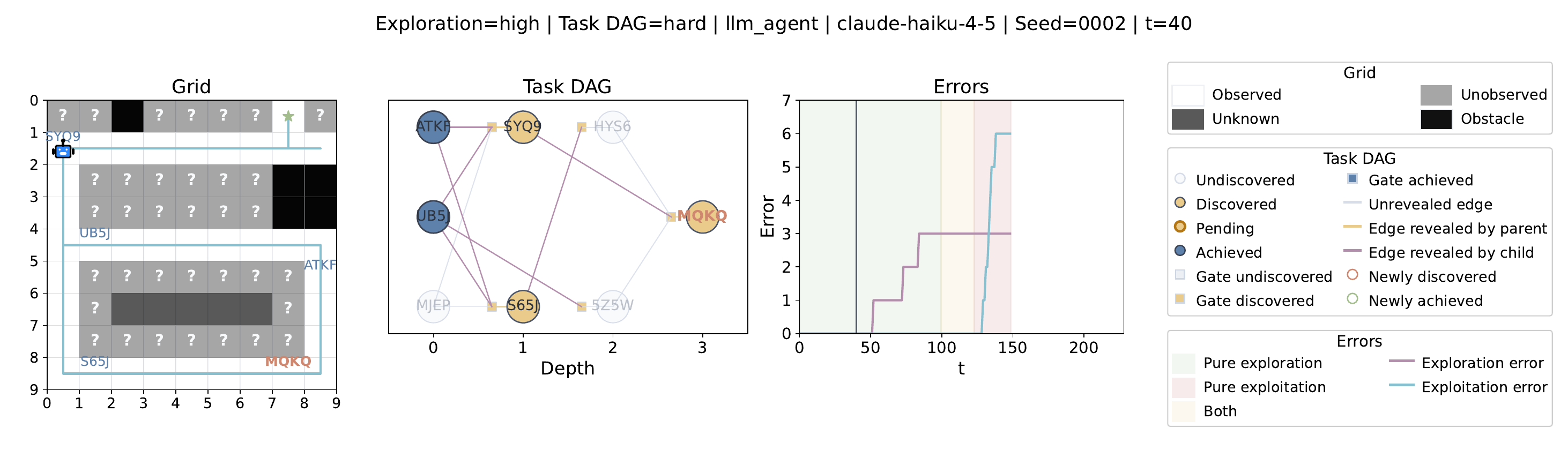}
        \caption{\currentmodel, $t = 40$}
        \end{subfigure}
    \end{center}
    \caption{Results of action trajectories and metric values over each timestep for~\currentmodel.}
\end{figure}

\begin{figure}[ht]
    \begin{center}
        \begin{subfigure}[t]{\framewidth}
        \centering
        \includegraphics[width=\textwidth,trim={0 0 0 33pt},clip]{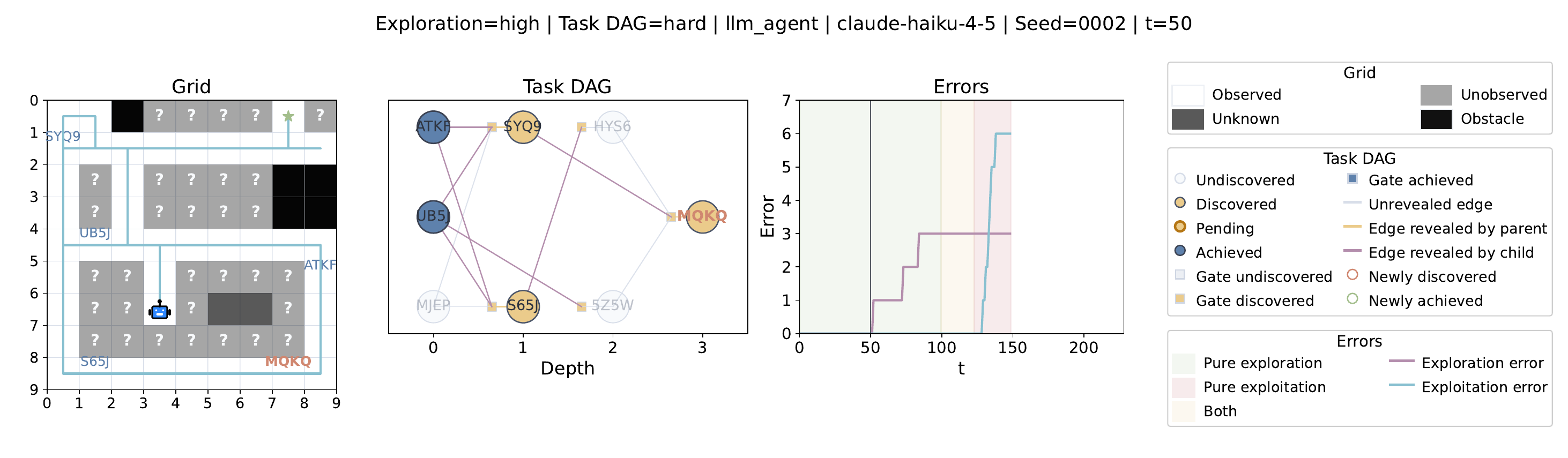}
        \caption{\currentmodel, $t = 50$}
        \end{subfigure}
        \begin{subfigure}[t]{\framewidth}
        \centering
        \includegraphics[width=\textwidth,trim={0 0 0 33pt},clip]{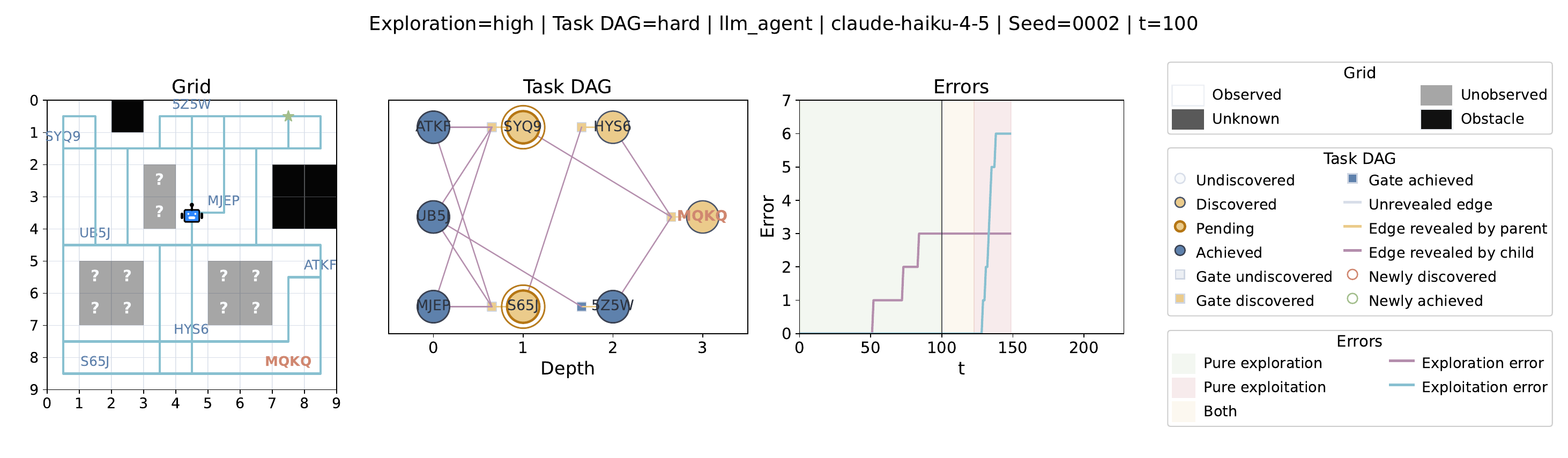}
        \caption{\currentmodel, $t = 100$}
        \end{subfigure}
        \begin{subfigure}[t]{\framewidth}
        \centering
        \includegraphics[width=\textwidth,trim={0 0 0 33pt},clip]{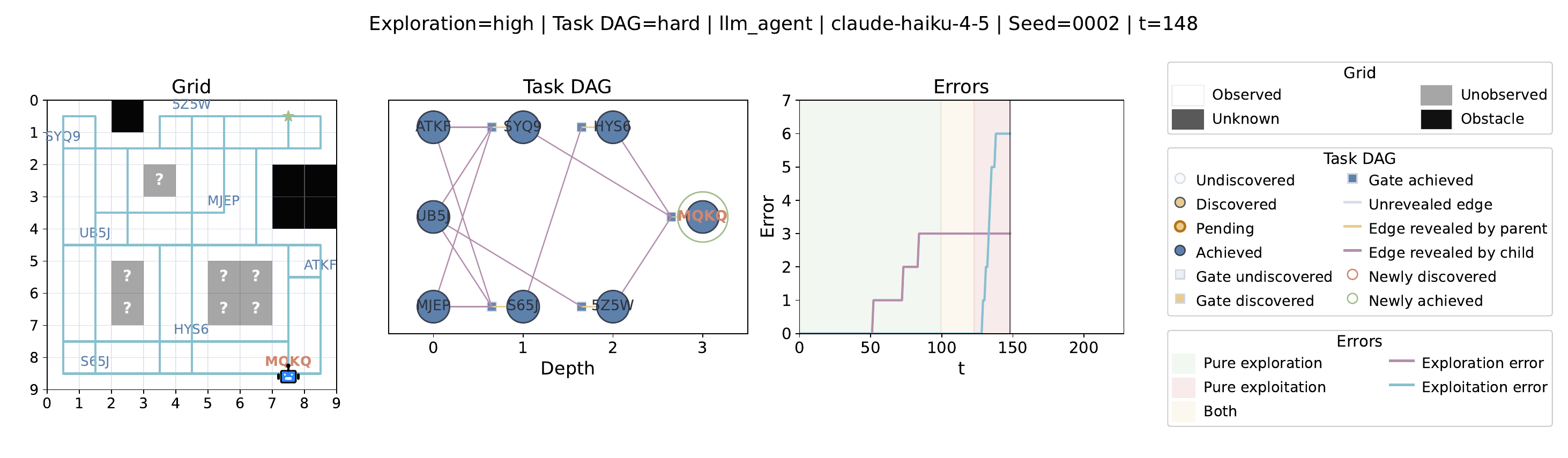}
        \caption{\currentmodel, $t = 148$}
        \end{subfigure}
    \end{center}
    \caption{Results of action trajectories and metric values over each timestep for~\currentmodel.}
\end{figure}

\clearpage
\renewcommand{\currentmodel}{\claudesonnet}

\begin{figure}[ht]
    \begin{center}
        \begin{subfigure}[t]{\framewidth}
        \centering
        \includegraphics[width=\textwidth,trim={0 0 0 33pt},clip]{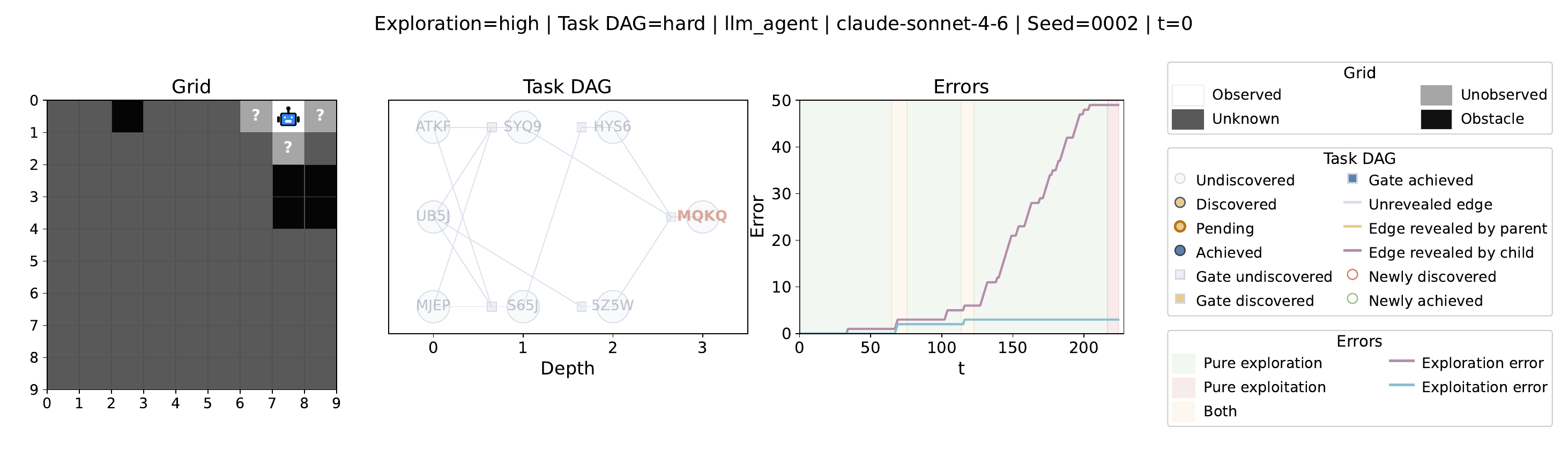}
        \caption{\currentmodel, $t = 0$}
        \end{subfigure}
        \begin{subfigure}[t]{\framewidth}
        \centering
        \includegraphics[width=\textwidth,trim={0 0 0 33pt},clip]{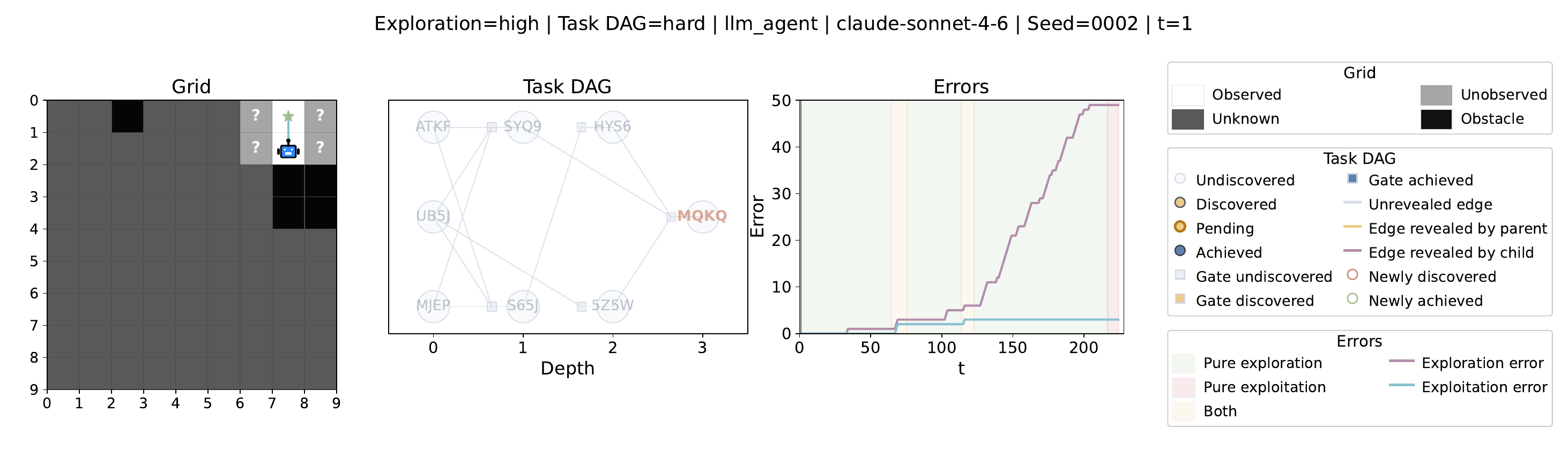}
        \caption{\currentmodel, $t = 1$}
        \end{subfigure}
        \begin{subfigure}[t]{\framewidth}
        \centering
        \includegraphics[width=\textwidth,trim={0 0 0 33pt},clip]{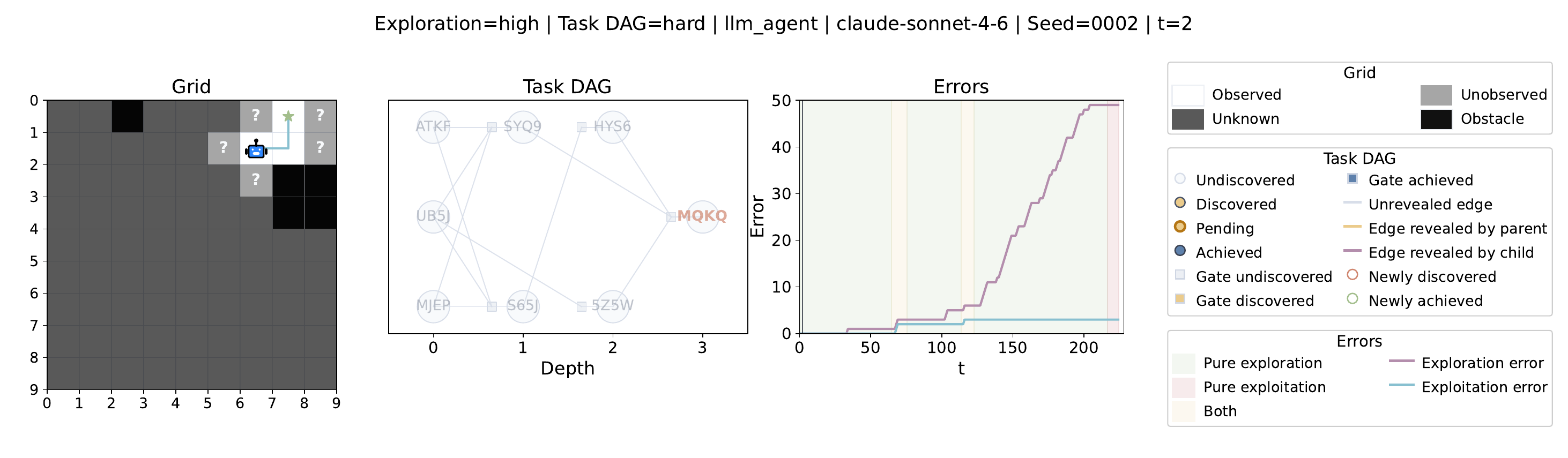}
        \caption{\currentmodel, $t = 2$}
        \end{subfigure}
        \begin{subfigure}[t]{\framewidth}
        \centering
        \includegraphics[width=\textwidth,trim={0 0 0 33pt},clip]{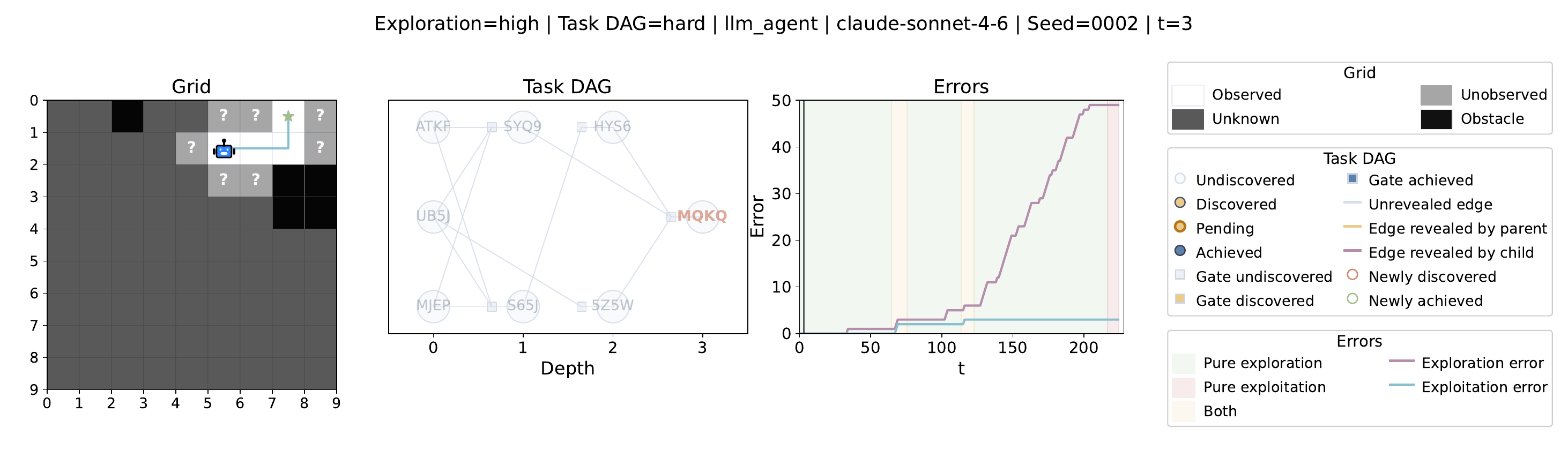}
        \caption{\currentmodel, $t = 3$}
        \end{subfigure}
        \begin{subfigure}[t]{\framewidth}
        \centering
        \includegraphics[width=\textwidth,trim={0 0 0 33pt},clip]{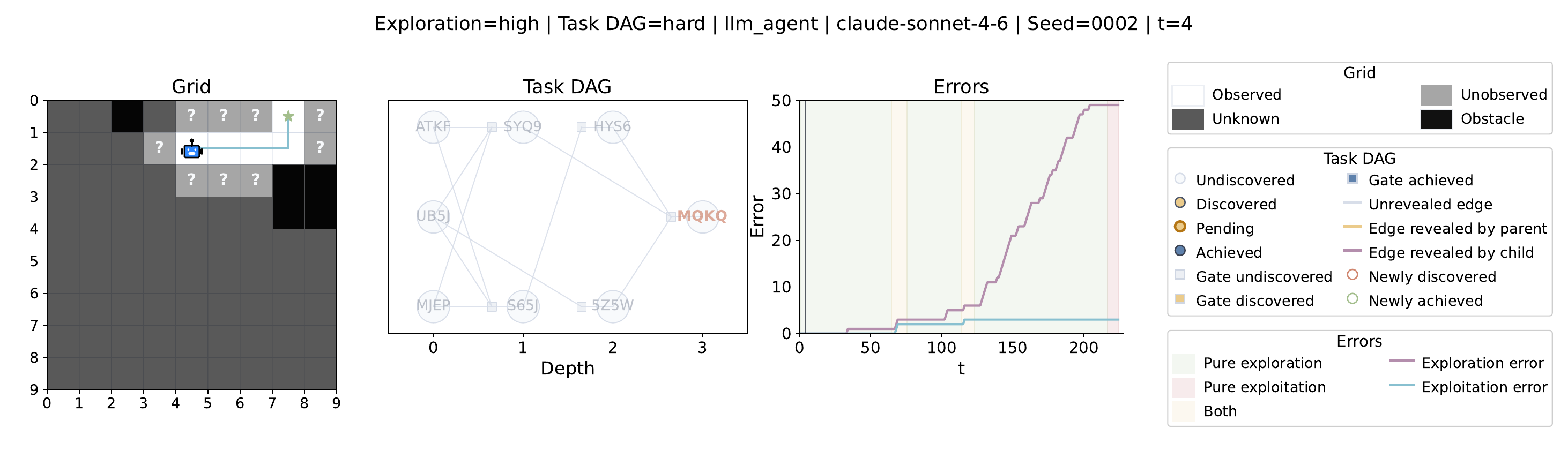}
        \caption{\currentmodel, $t = 4$}
        \end{subfigure}
    \end{center}
    \caption{Results of action trajectories and metric values over each timestep for~\currentmodel.}
\end{figure}

\begin{figure}[ht]
    \begin{center}
        \begin{subfigure}[t]{\framewidth}
        \centering
        \includegraphics[width=\textwidth,trim={0 0 0 33pt},clip]{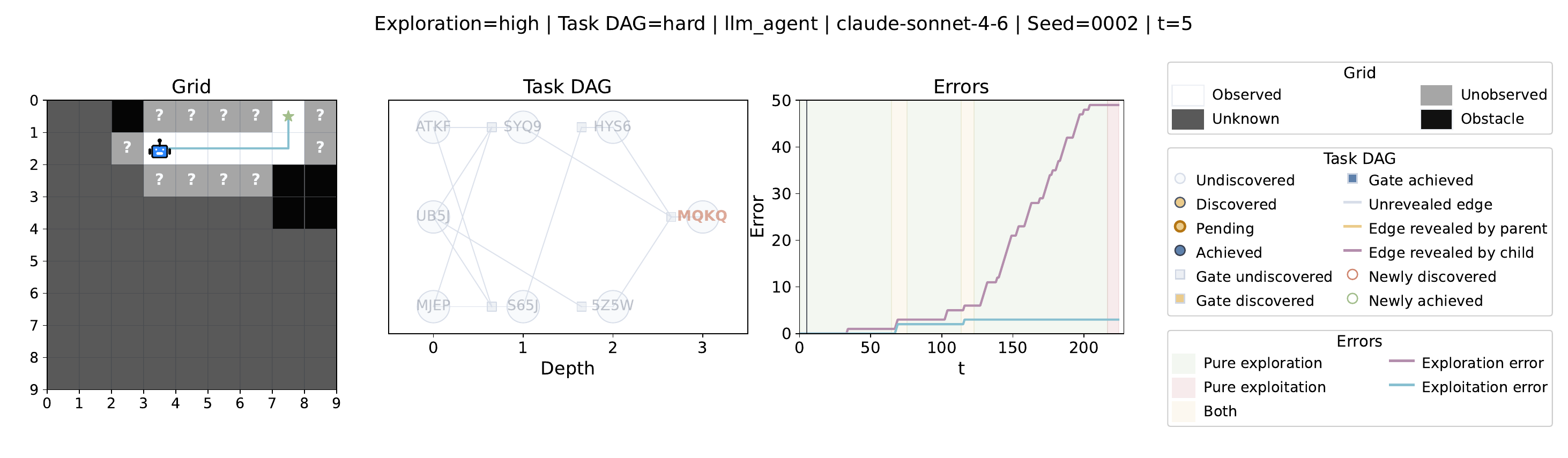}
        \caption{\currentmodel, $t = 5$}
        \end{subfigure}
        \begin{subfigure}[t]{\framewidth}
        \centering
        \includegraphics[width=\textwidth,trim={0 0 0 33pt},clip]{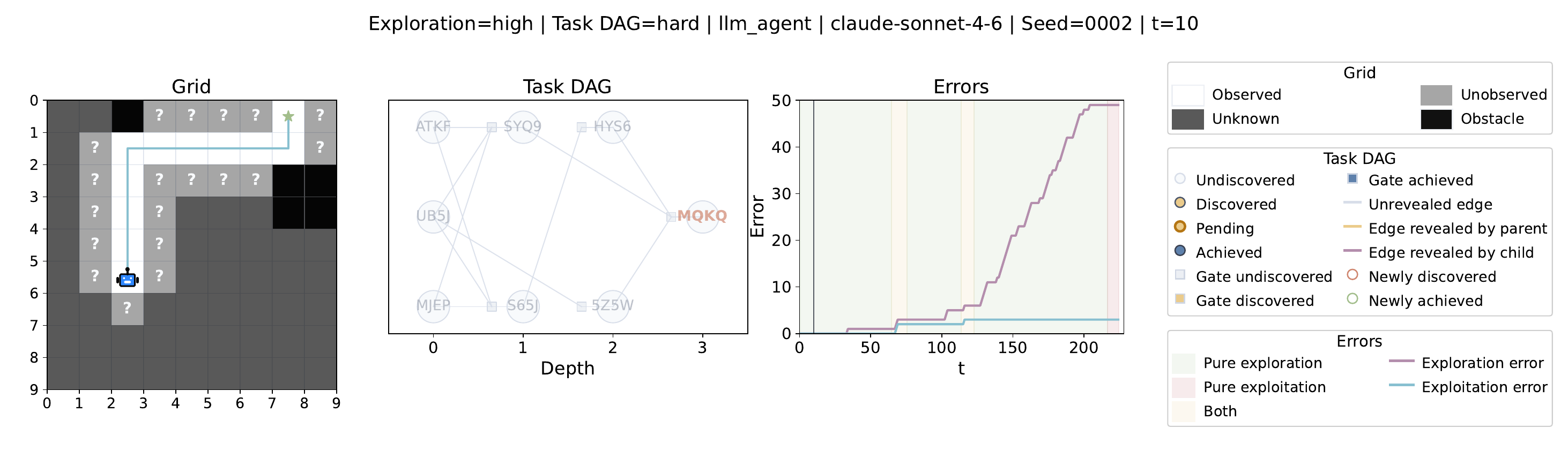}
        \caption{\currentmodel, $t = 10$}
        \end{subfigure}
        \begin{subfigure}[t]{\framewidth}
        \centering
        \includegraphics[width=\textwidth,trim={0 0 0 33pt},clip]{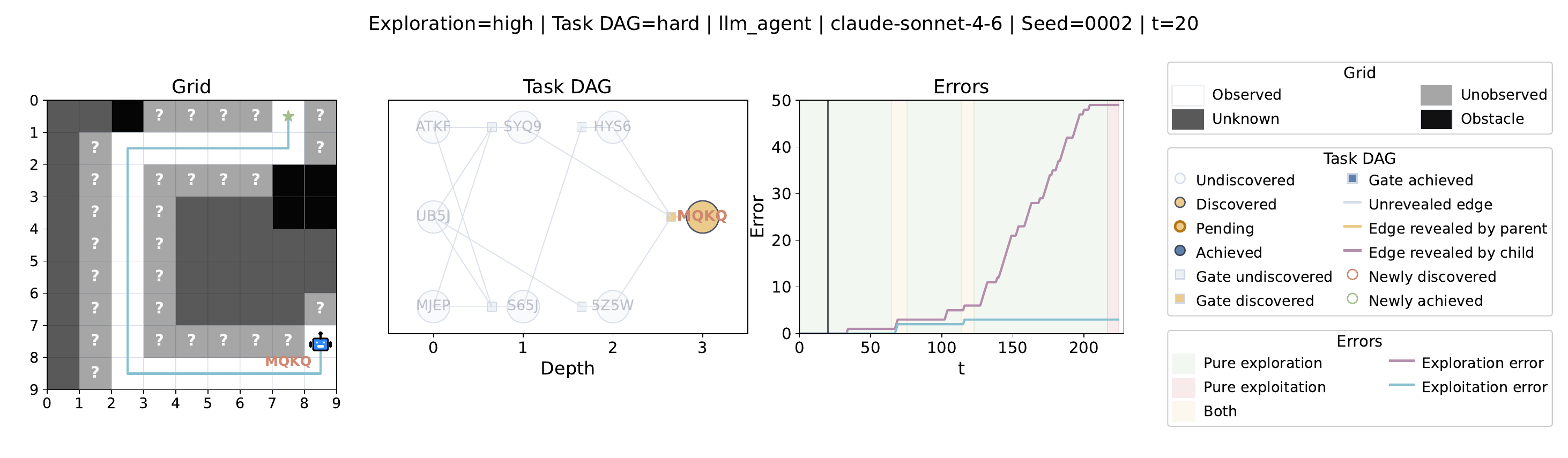}
        \caption{\currentmodel, $t = 20$}
        \end{subfigure}
        \begin{subfigure}[t]{\framewidth}
        \centering
        \includegraphics[width=\textwidth,trim={0 0 0 33pt},clip]{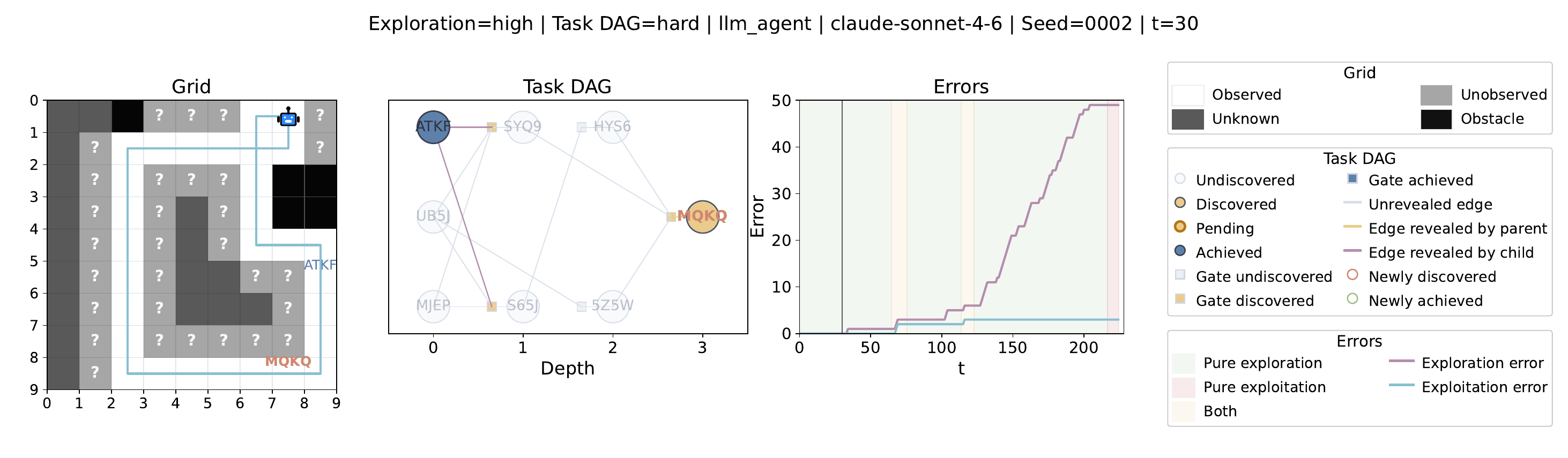}
        \caption{\currentmodel, $t = 30$}
        \end{subfigure}
        \begin{subfigure}[t]{\framewidth}
        \centering
        \includegraphics[width=\textwidth,trim={0 0 0 33pt},clip]{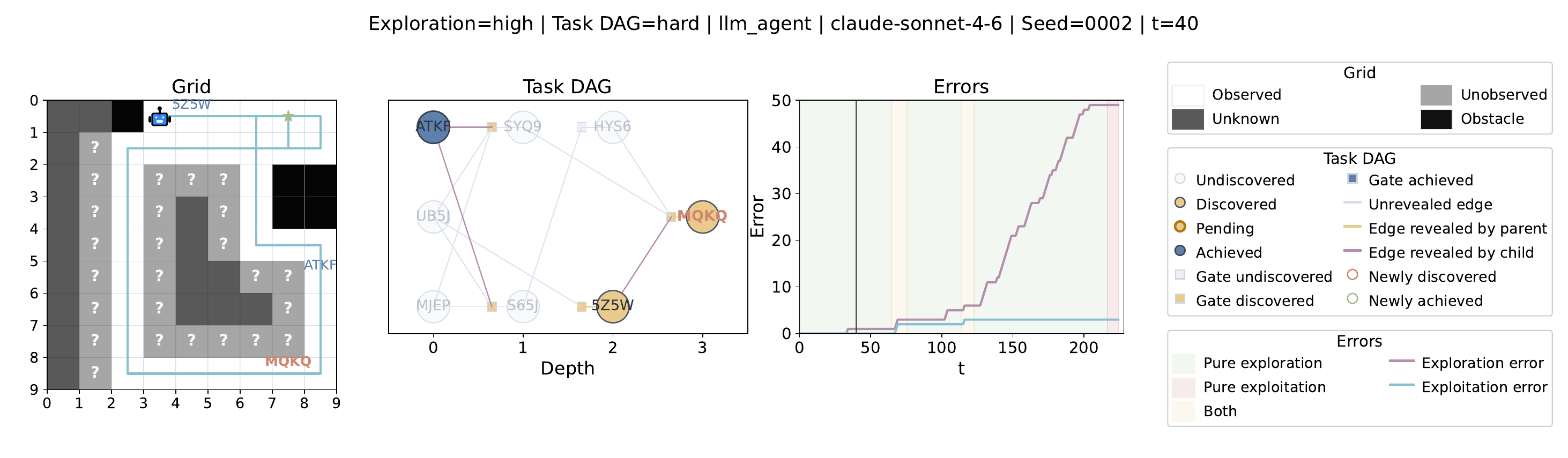}
        \caption{\currentmodel, $t = 40$}
        \end{subfigure}
    \end{center}
    \caption{Results of action trajectories and metric values over each timestep for~\currentmodel.}
\end{figure}

\begin{figure}[ht]
    \begin{center}
        \begin{subfigure}[t]{\framewidth}
        \centering
        \includegraphics[width=\textwidth,trim={0 0 0 33pt},clip]{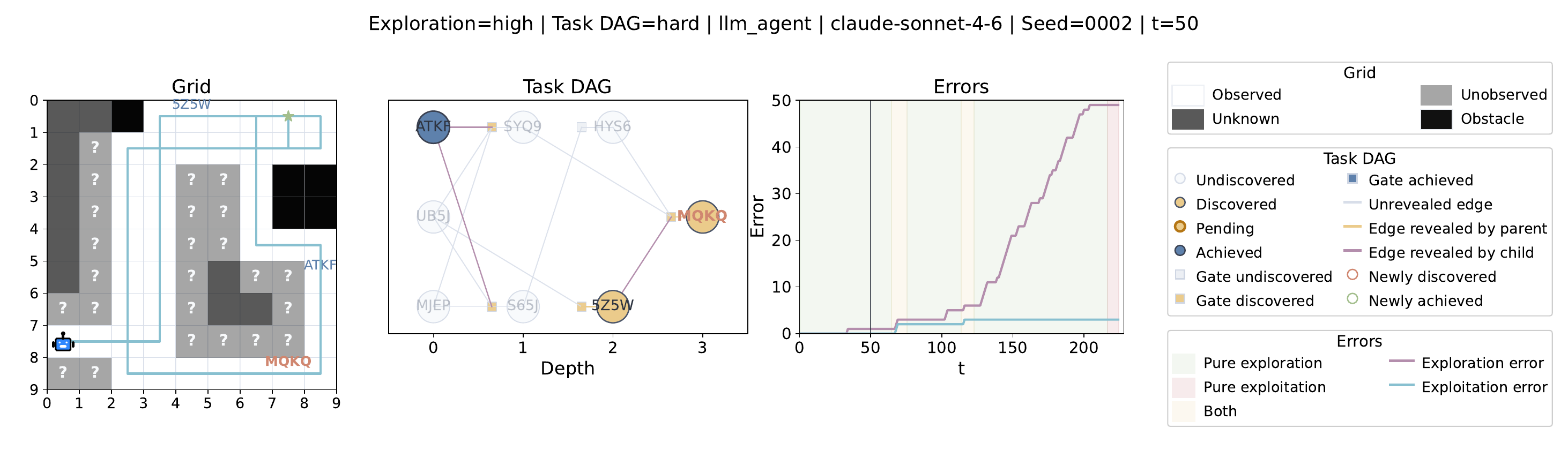}
        \caption{\currentmodel, $t = 50$}
        \end{subfigure}
        \begin{subfigure}[t]{\framewidth}
        \centering
        \includegraphics[width=\textwidth,trim={0 0 0 33pt},clip]{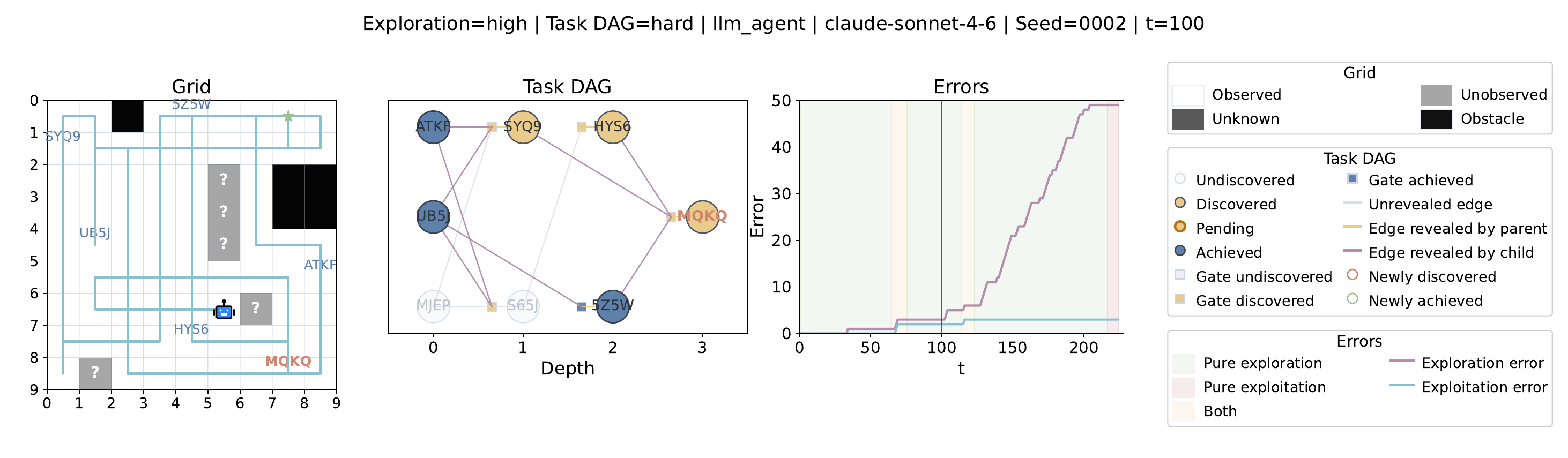}
        \caption{\currentmodel, $t = 100$}
        \end{subfigure}
        \begin{subfigure}[t]{\framewidth}
        \centering
        \includegraphics[width=\textwidth,trim={0 0 0 33pt},clip]{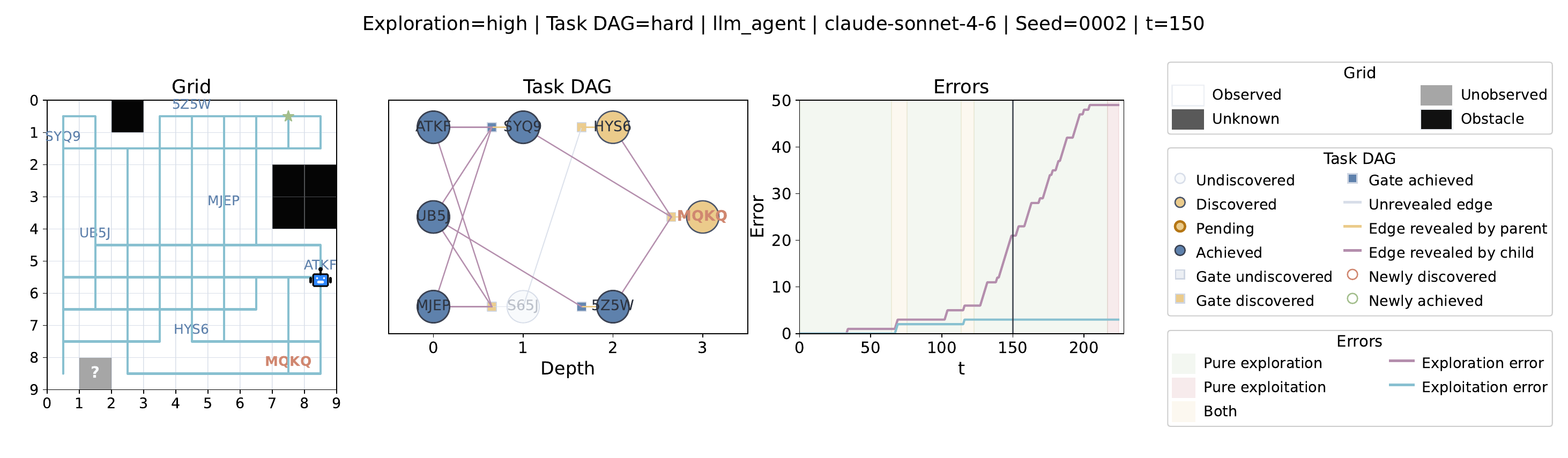}
        \caption{\currentmodel, $t = 150$}
        
        \end{subfigure}
        \begin{subfigure}[t]{\framewidth}
        \centering
        \includegraphics[width=\textwidth,trim={0 0 0 33pt},clip]{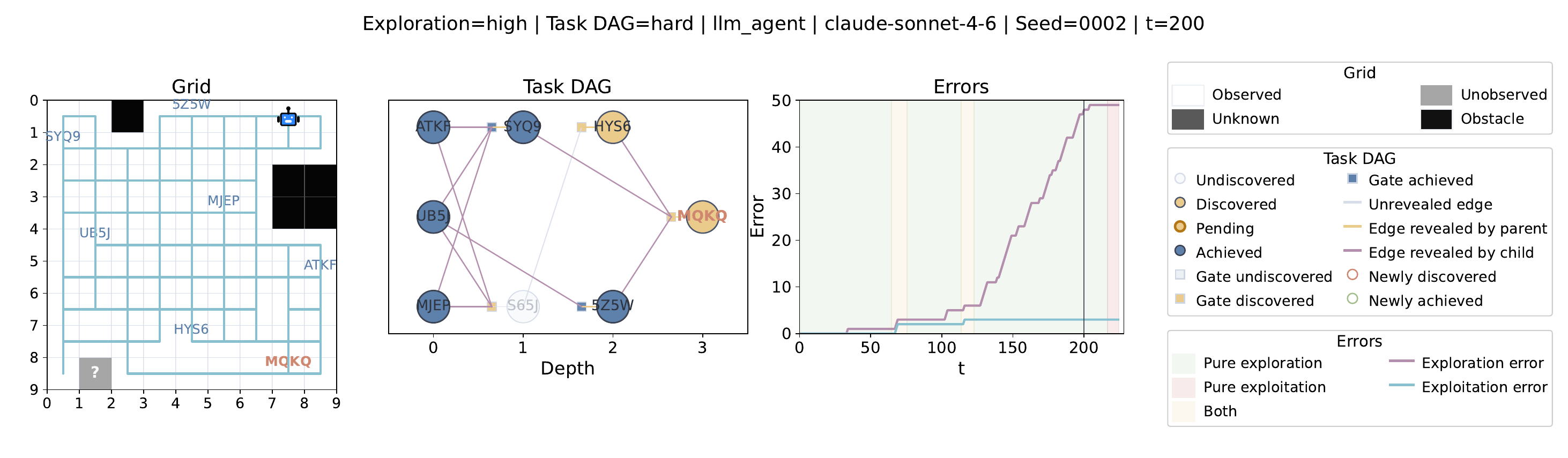}
        \caption{\currentmodel, $t = 200$}
        
        \end{subfigure}
        \begin{subfigure}[t]{\framewidth}
        \centering
        \includegraphics[width=\textwidth,trim={0 0 0 33pt},clip]{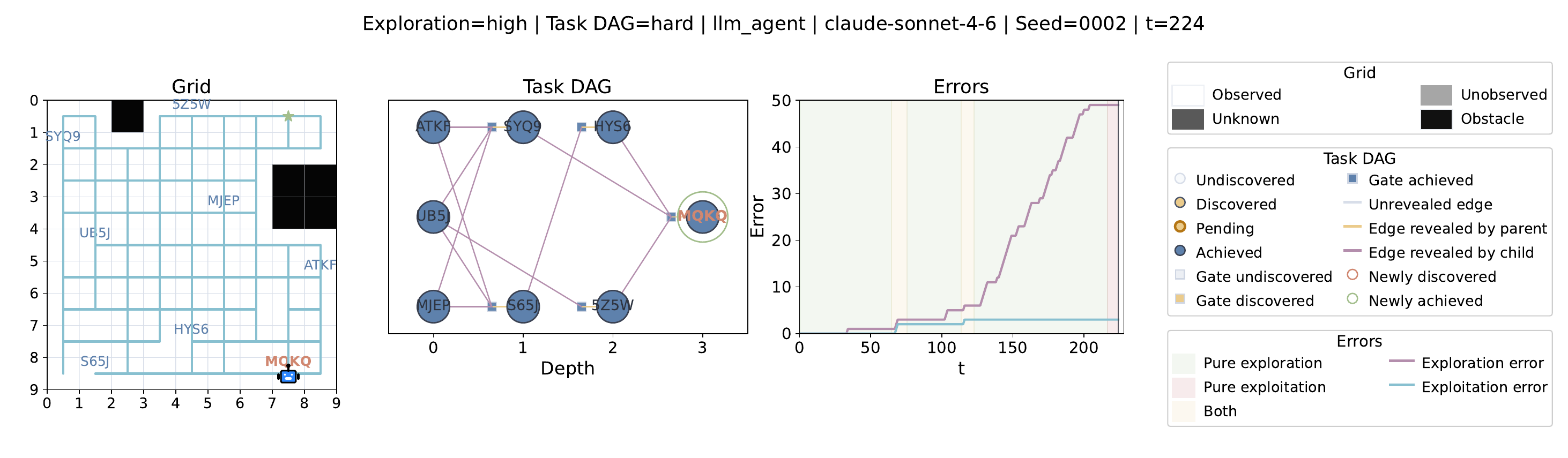}
        \caption{\currentmodel, $t = 224$}
        
        \end{subfigure}
    \end{center}
    \caption{Results of action trajectories and metric values over each timestep for~\currentmodel.}
\end{figure}

\clearpage
\renewcommand{\currentmodel}{\claudeopus}

\begin{figure}[ht]
    \begin{center}
        \begin{subfigure}[t]{\framewidth}
        \centering
        \includegraphics[width=\textwidth,trim={0 0 0 33pt},clip]{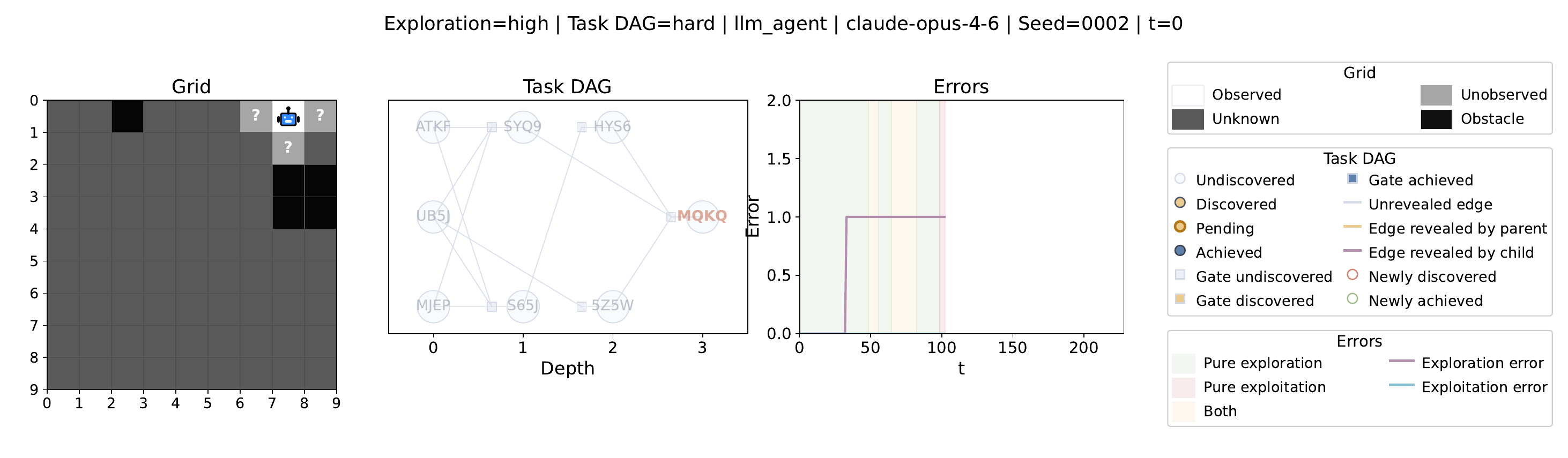}
        \caption{\currentmodel, $t = 0$}
        \end{subfigure}
        \begin{subfigure}[t]{\framewidth}
        \centering
        \includegraphics[width=\textwidth,trim={0 0 0 33pt},clip]{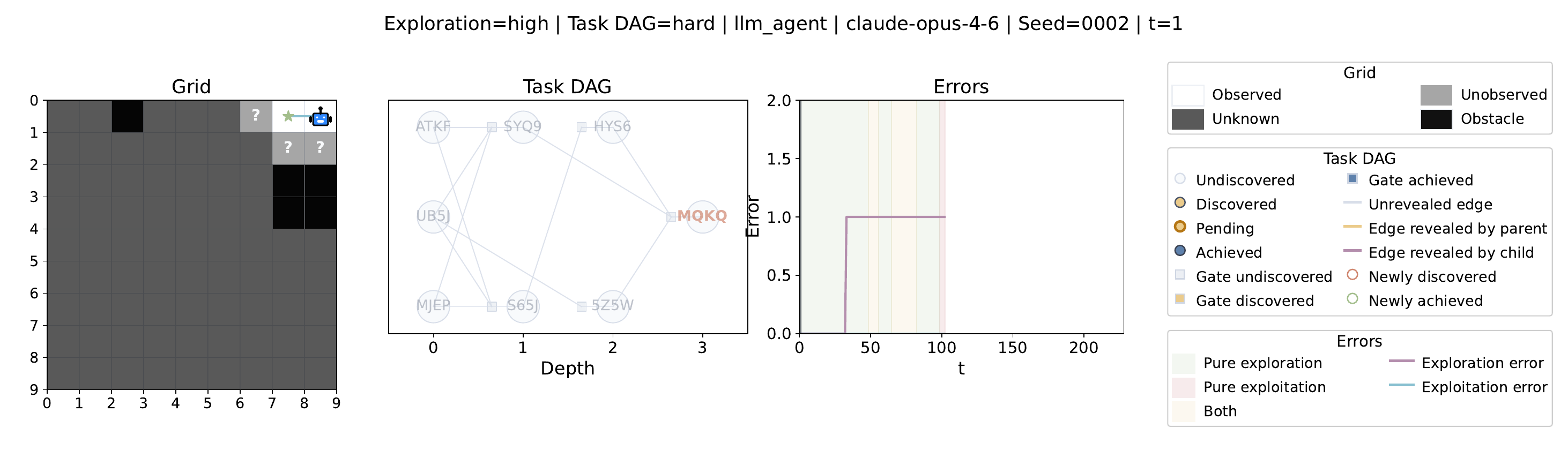}
        \caption{\currentmodel, $t = 1$}
        \end{subfigure}
        \begin{subfigure}[t]{\framewidth}
        \centering
        \includegraphics[width=\textwidth,trim={0 0 0 33pt},clip]{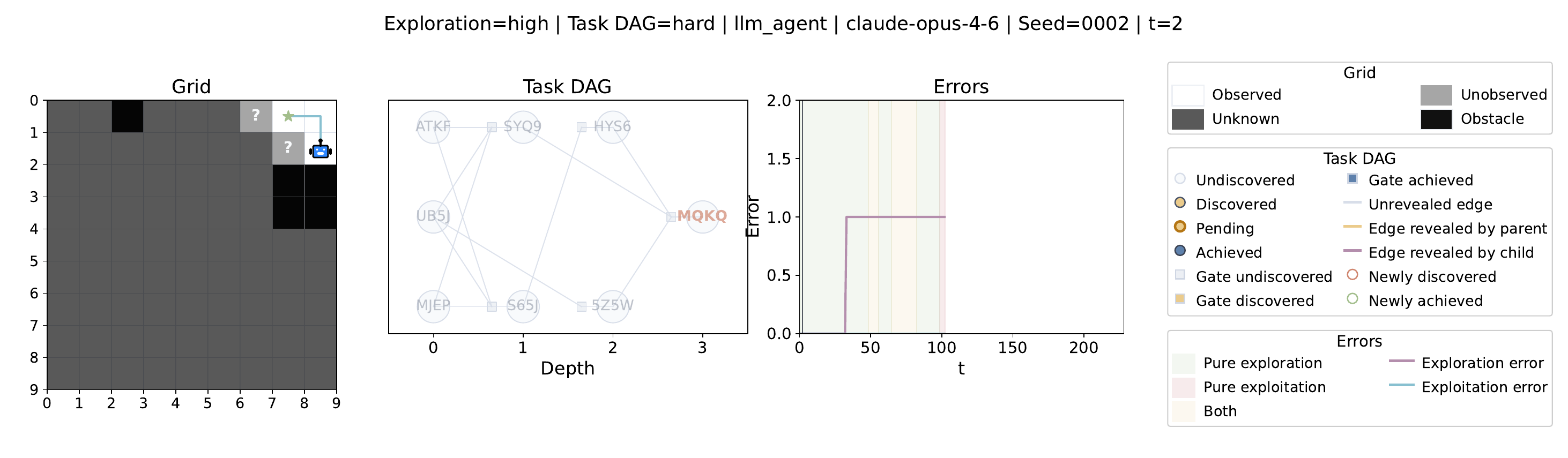}
        \caption{\currentmodel, $t = 2$}
        \end{subfigure}
        \begin{subfigure}[t]{\framewidth}
        \centering
        \includegraphics[width=\textwidth,trim={0 0 0 33pt},clip]{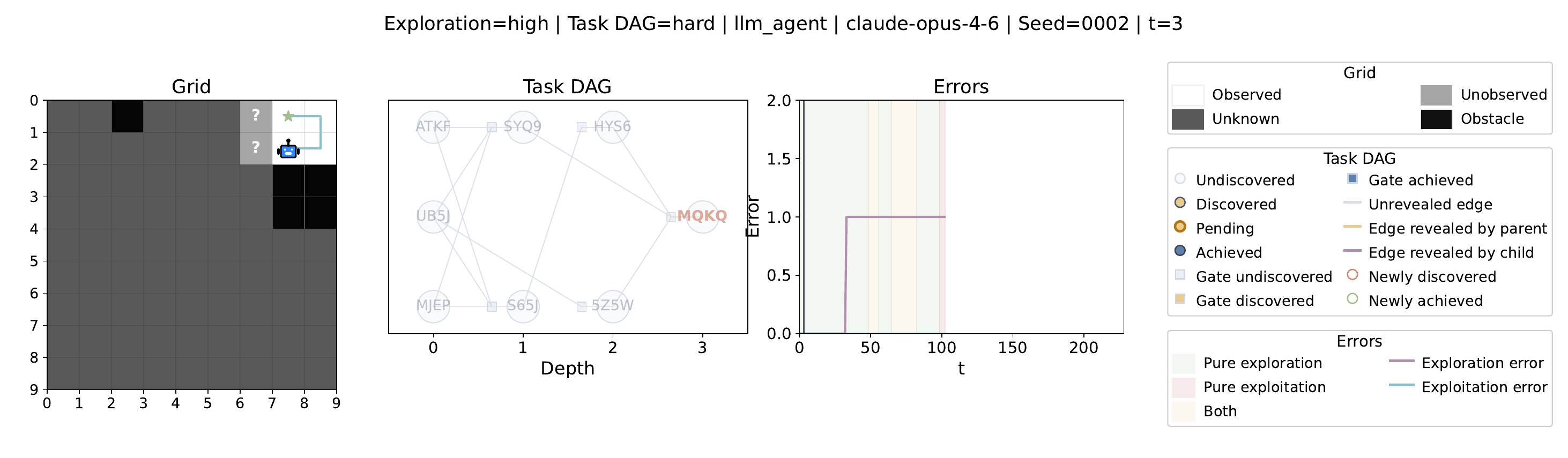}
        \caption{\currentmodel, $t = 3$}
        \end{subfigure}
        \begin{subfigure}[t]{\framewidth}
        \centering
        \includegraphics[width=\textwidth,trim={0 0 0 33pt},clip]{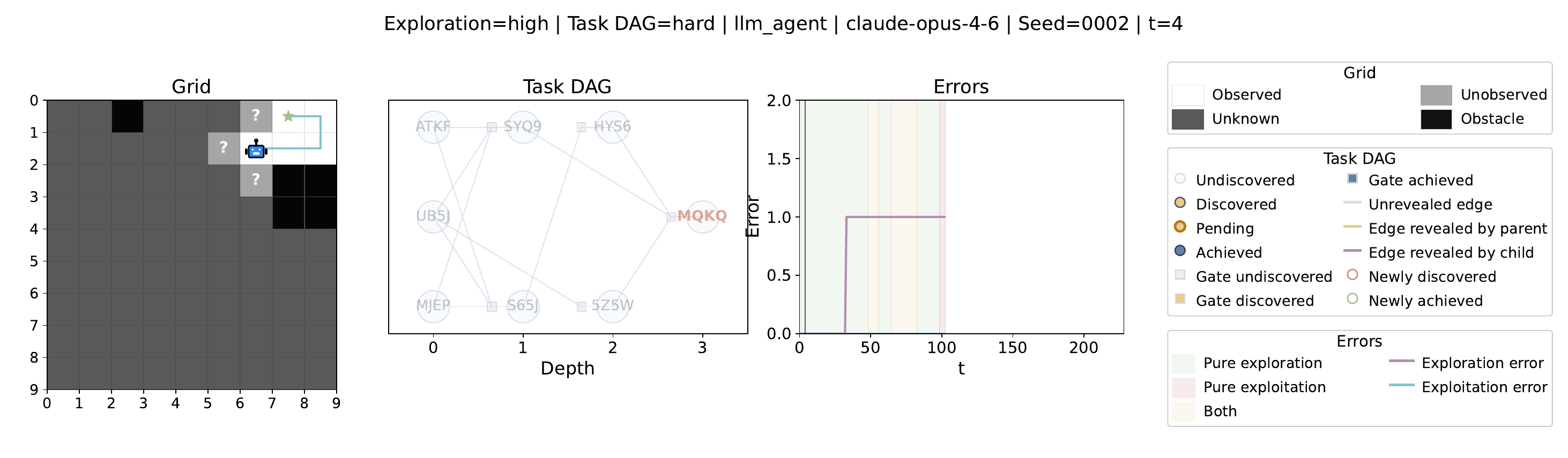}
        \caption{\currentmodel, $t = 4$}
        \end{subfigure}
    \end{center}
    \caption{Results of action trajectories and metric values over each timestep for~\currentmodel.}
\end{figure}

\begin{figure}[ht]
    \begin{center}
        \begin{subfigure}[t]{\framewidth}
        \centering
        \includegraphics[width=\textwidth,trim={0 0 0 33pt},clip]{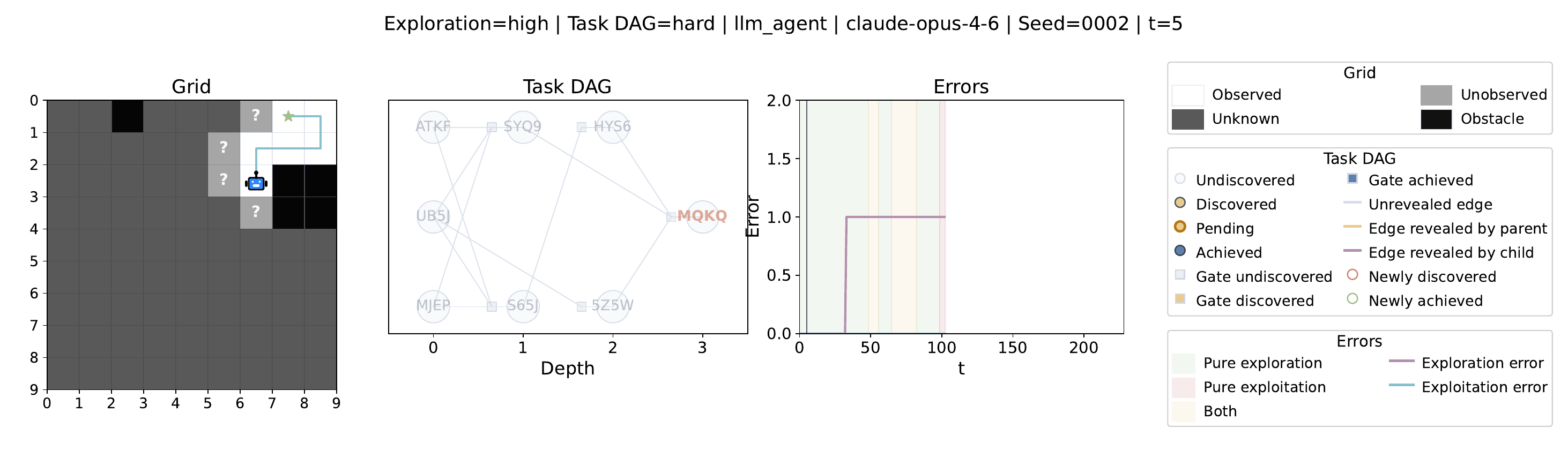}
        \caption{\currentmodel, $t = 5$}
        \end{subfigure}
        \begin{subfigure}[t]{\framewidth}
        \centering
        \includegraphics[width=\textwidth,trim={0 0 0 33pt},clip]{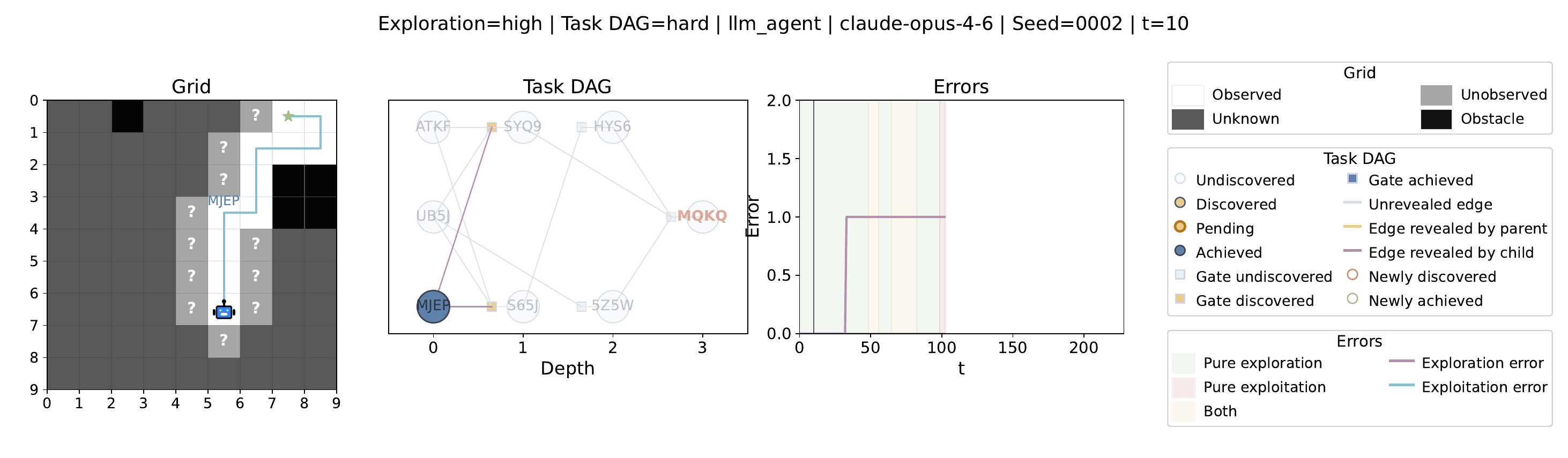}
        \caption{\currentmodel, $t = 10$}
        \end{subfigure}
        \begin{subfigure}[t]{\framewidth}
        \centering
        \includegraphics[width=\textwidth,trim={0 0 0 33pt},clip]{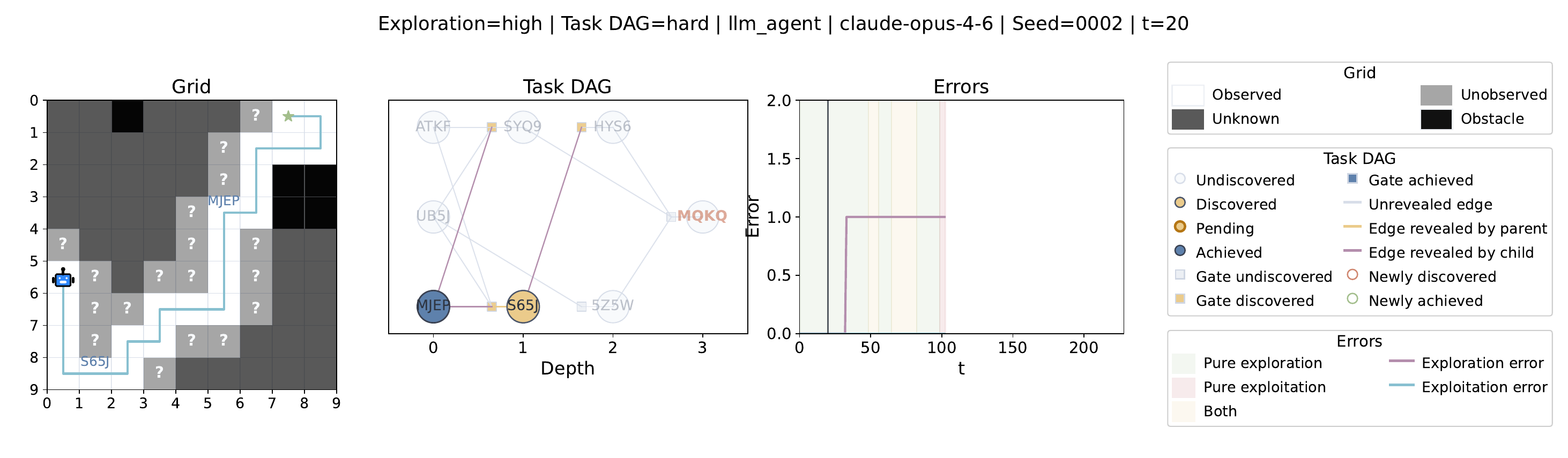}
        \caption{\currentmodel, $t = 20$}
        \end{subfigure}
        \begin{subfigure}[t]{\framewidth}
        \centering
        \includegraphics[width=\textwidth,trim={0 0 0 33pt},clip]{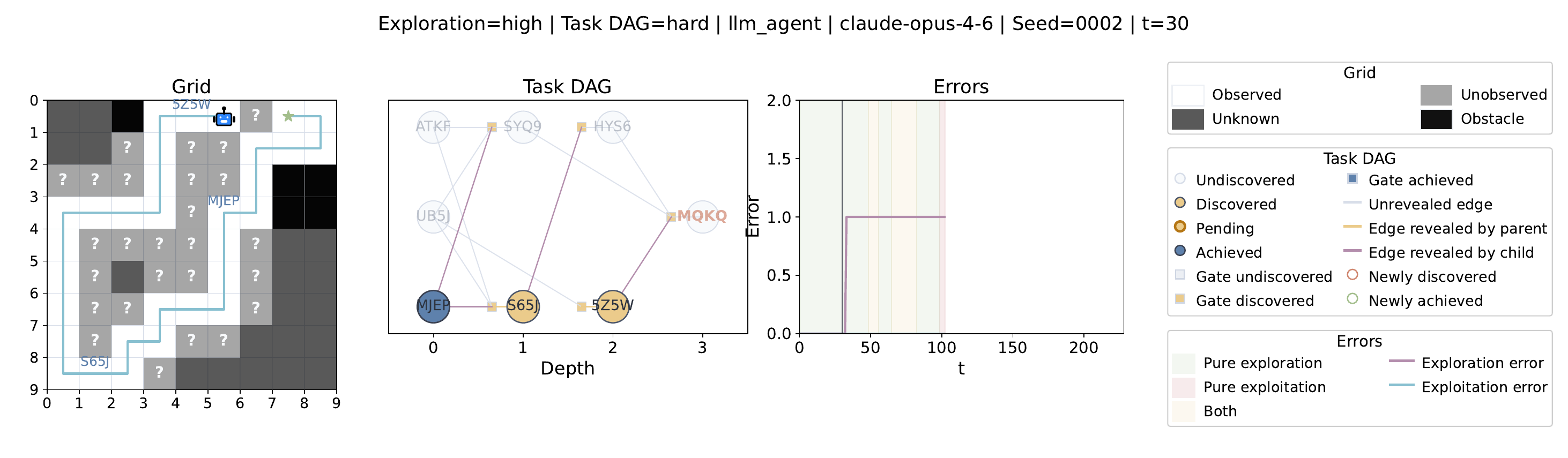}
        \caption{\currentmodel, $t = 30$}
        \end{subfigure}
        \begin{subfigure}[t]{\framewidth}
        \centering
        \includegraphics[width=\textwidth,trim={0 0 0 33pt},clip]{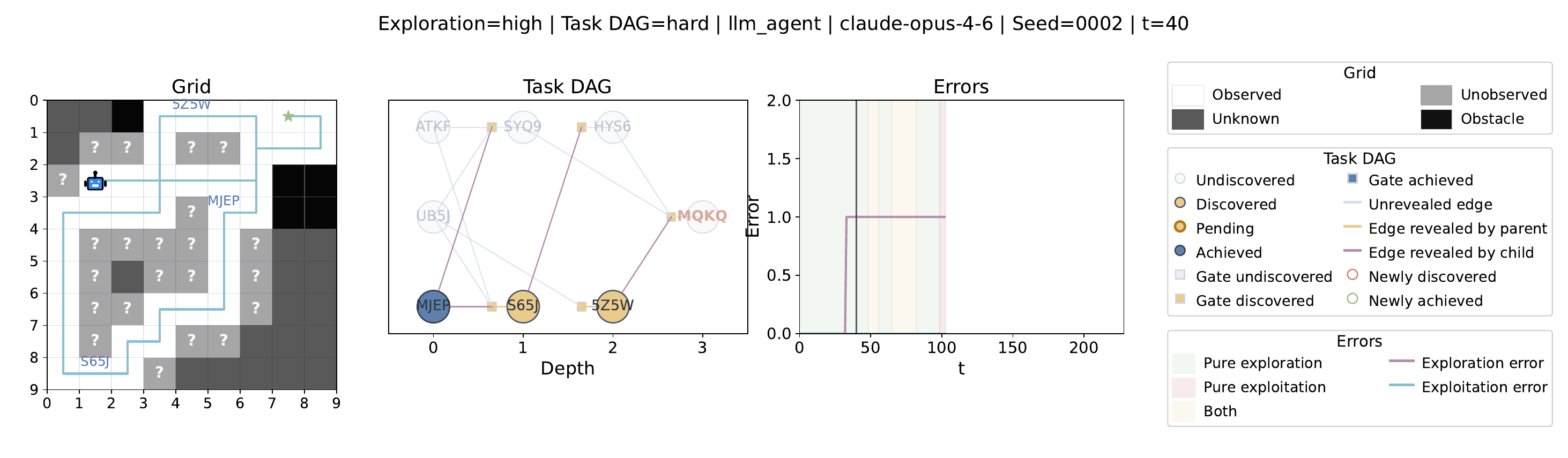}
        \caption{\currentmodel, $t = 40$}
        \end{subfigure}
    \end{center}
    \caption{Results of action trajectories and metric values over each timestep for~\currentmodel.}
\end{figure}

\begin{figure}[ht]
    \begin{center}
        \begin{subfigure}[t]{\framewidth}
        \centering
        \includegraphics[width=\textwidth,trim={0 0 0 33pt},clip]{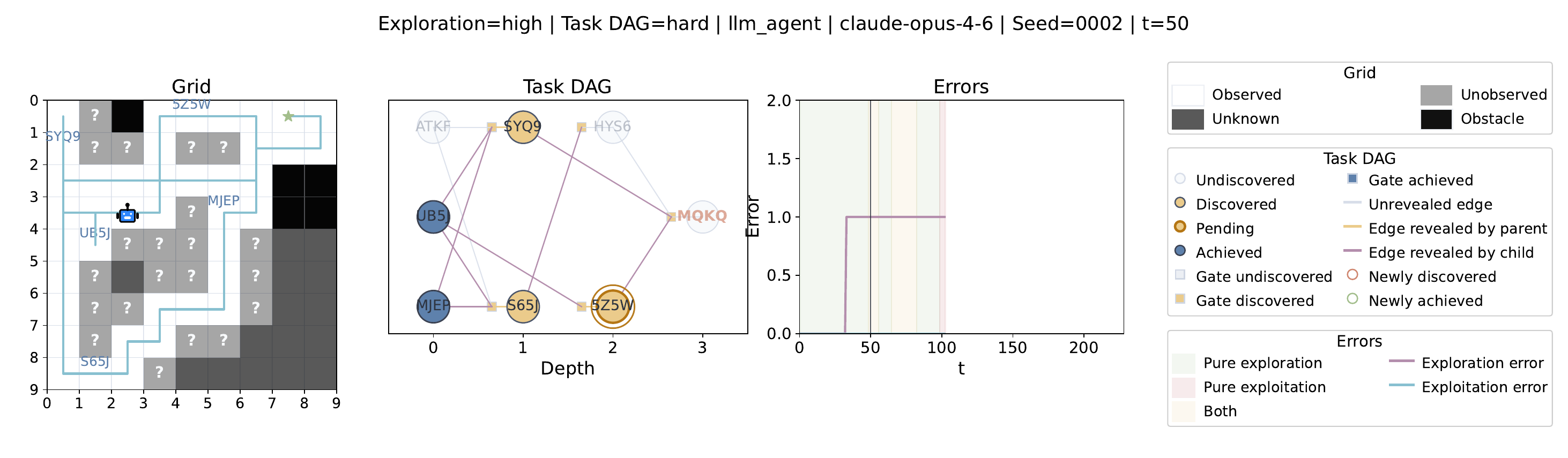}
        \caption{\currentmodel, $t = 50$}
        \end{subfigure}
        \begin{subfigure}[t]{\framewidth}
        \centering
        \includegraphics[width=\textwidth,trim={0 0 0 33pt},clip]{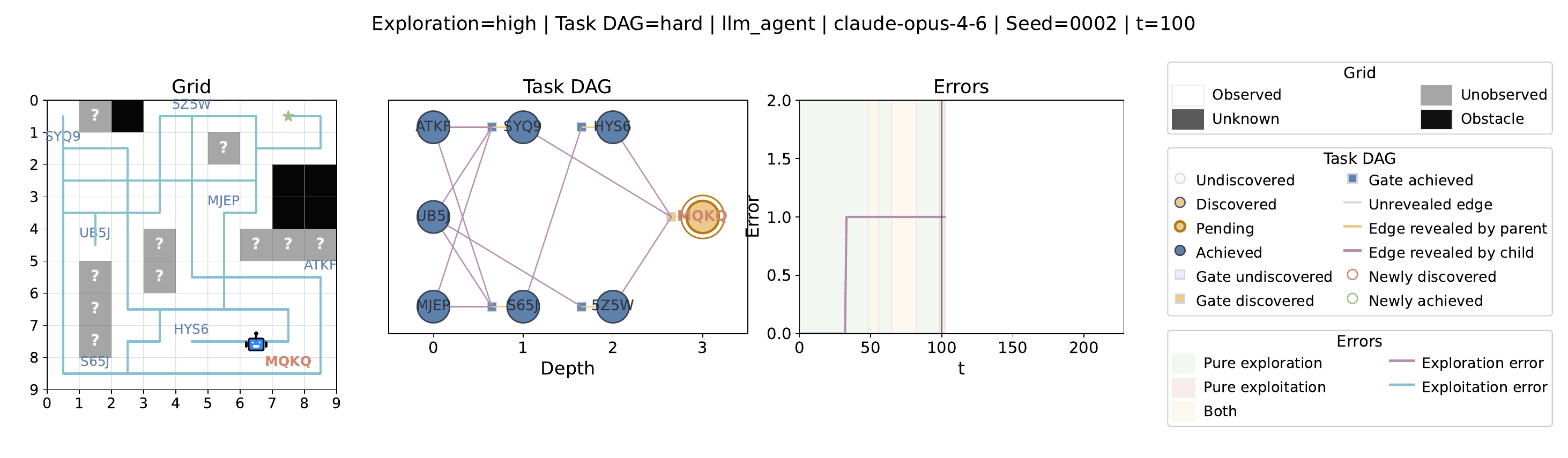}
        \caption{\currentmodel, $t = 100$}
        \end{subfigure}
        \begin{subfigure}[t]{\framewidth}
        \centering
        \includegraphics[width=\textwidth,trim={0 0 0 33pt},clip]{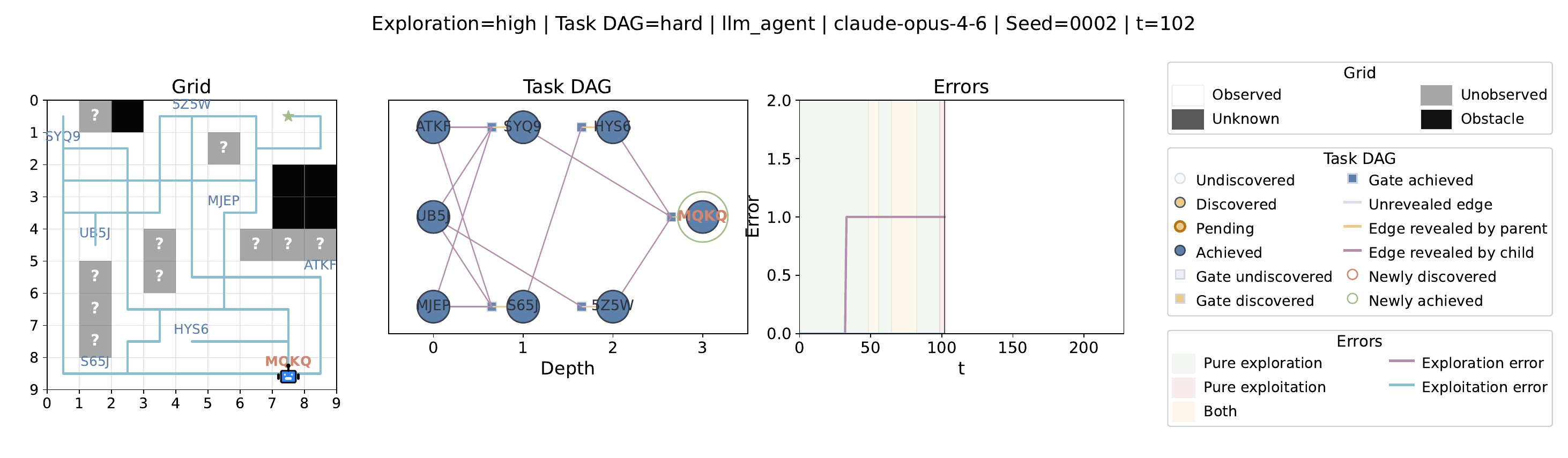}
        \caption{\currentmodel, $t = 102$}
        \end{subfigure}
    \end{center}
    \caption{Results of action trajectories and metric values over each timestep for~\currentmodel.}
\end{figure}

\clearpage

\renewcommand{\currentmodel}{\geminiflash}

\begin{figure}[ht]
    \begin{center}
        \begin{subfigure}[t]{\framewidth}
        \centering
        \includegraphics[width=\textwidth,trim={0 0 0 33pt},clip]{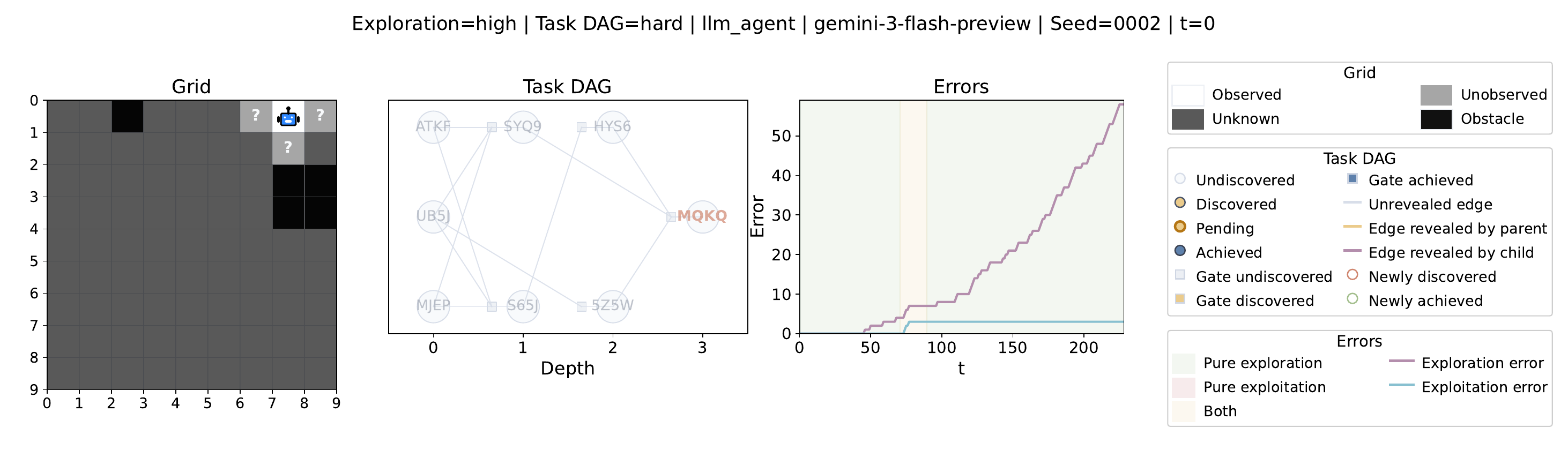}
        \caption{\currentmodel, $t = 0$}
        \end{subfigure}
        \begin{subfigure}[t]{\framewidth}
        \centering
        \includegraphics[width=\textwidth,trim={0 0 0 33pt},clip]{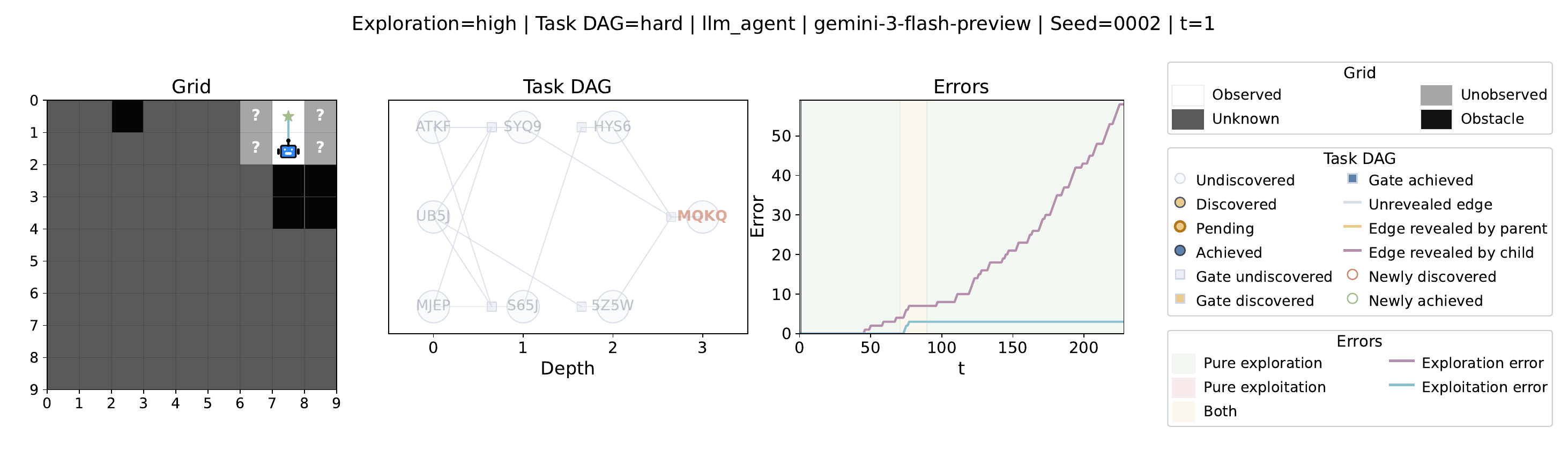}
        \caption{\currentmodel, $t = 1$}
        \end{subfigure}
        \begin{subfigure}[t]{\framewidth}
        \centering
        \includegraphics[width=\textwidth,trim={0 0 0 33pt},clip]{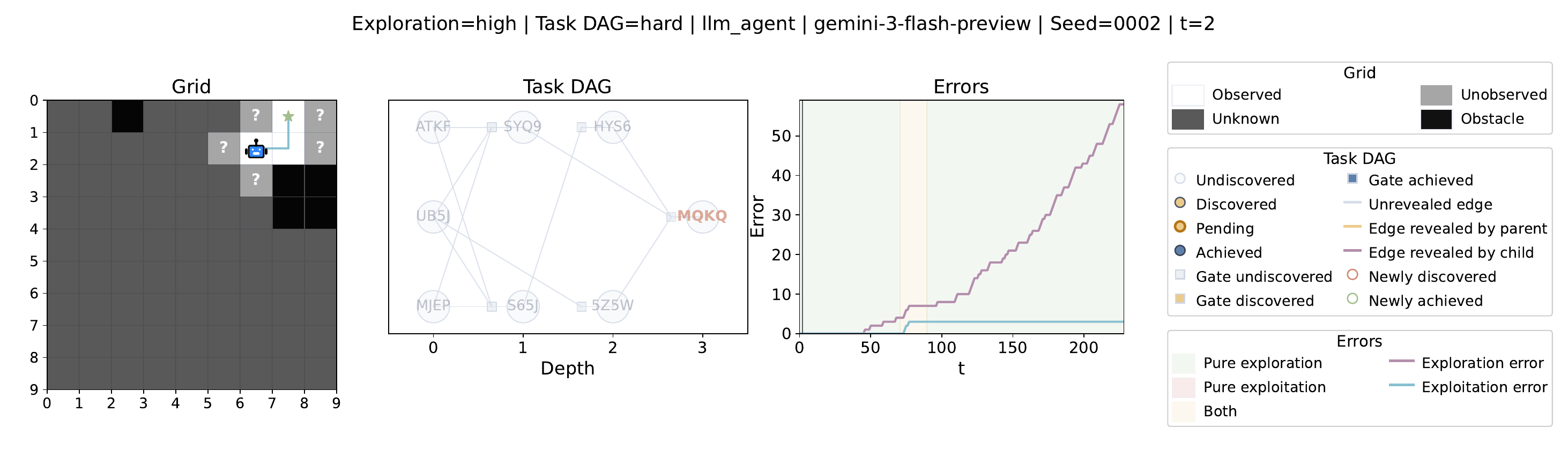}
        \caption{\currentmodel, $t = 2$}
        \end{subfigure}
        \begin{subfigure}[t]{\framewidth}
        \centering
        \includegraphics[width=\textwidth,trim={0 0 0 33pt},clip]{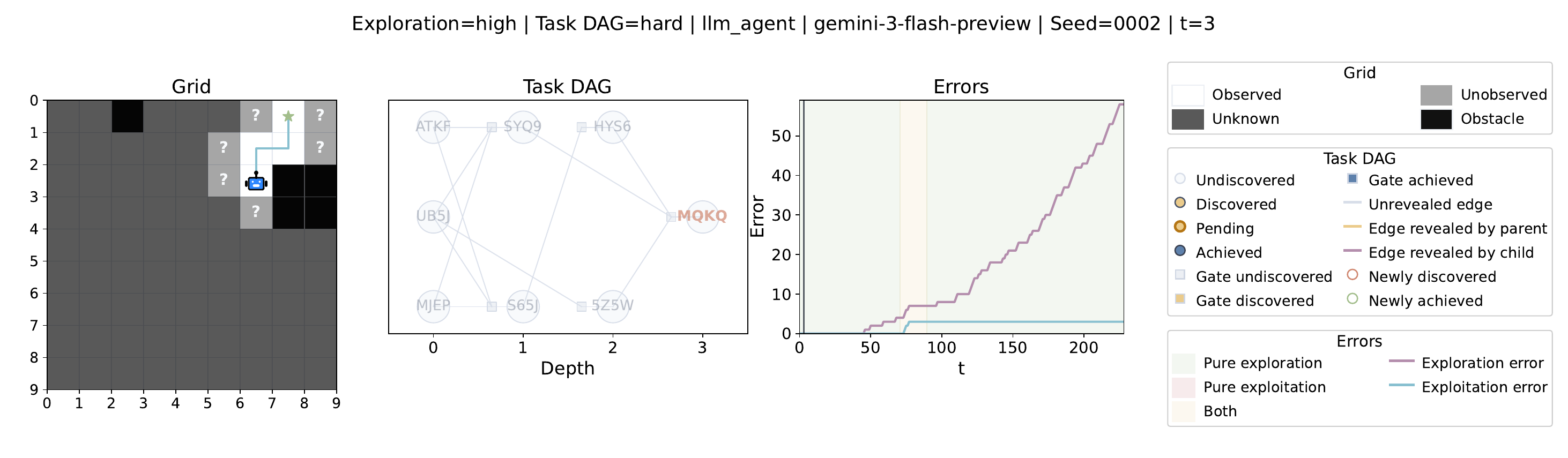}
        \caption{\currentmodel, $t = 3$}
        \end{subfigure}
        \begin{subfigure}[t]{\framewidth}
        \centering
        \includegraphics[width=\textwidth,trim={0 0 0 33pt},clip]{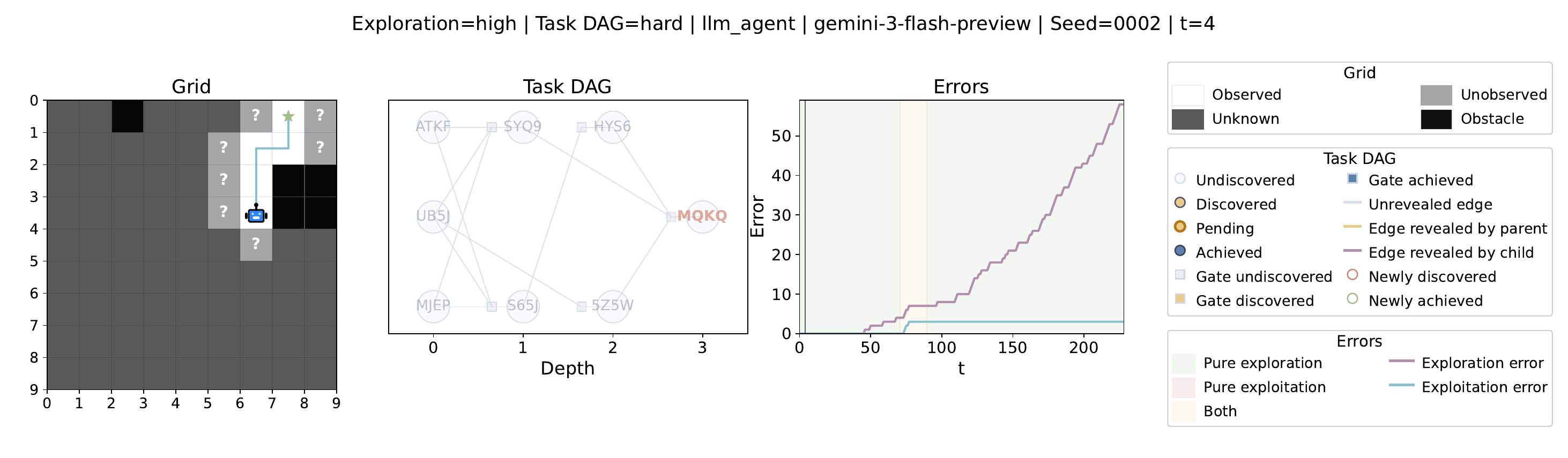}
        \caption{\currentmodel, $t = 4$}
        \end{subfigure}
    \end{center}
    \caption{Results of action trajectories and metric values over each timestep for~\currentmodel.}
\end{figure}

\begin{figure}[ht]
    \begin{center}
        \begin{subfigure}[t]{\framewidth}
        \centering
        \includegraphics[width=\textwidth,trim={0 0 0 33pt},clip]{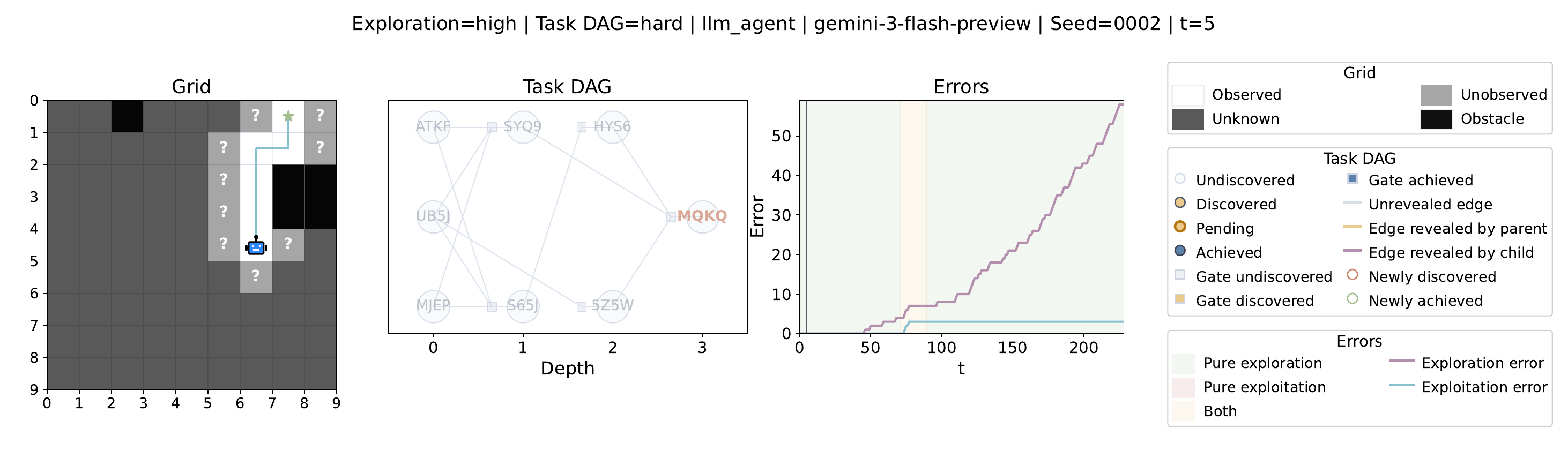}
        \caption{\currentmodel, $t = 5$}
        \end{subfigure}
        \begin{subfigure}[t]{\framewidth}
        \centering
        \includegraphics[width=\textwidth,trim={0 0 0 33pt},clip]{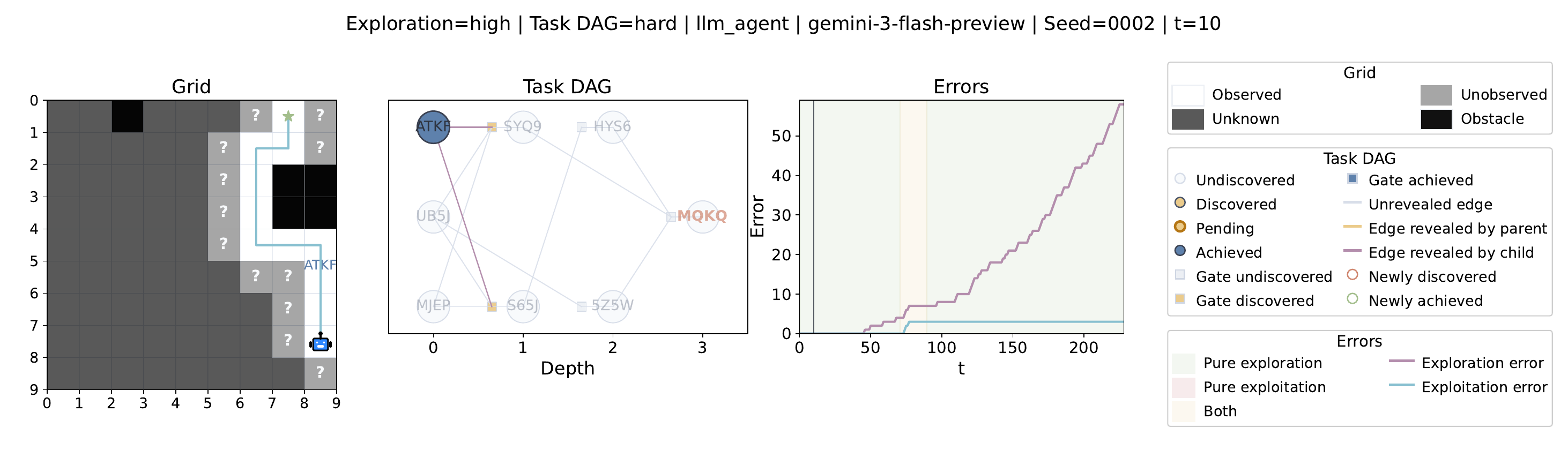}
        \caption{\currentmodel, $t = 10$}
        \end{subfigure}
        \begin{subfigure}[t]{\framewidth}
        \centering
        \includegraphics[width=\textwidth,trim={0 0 0 33pt},clip]{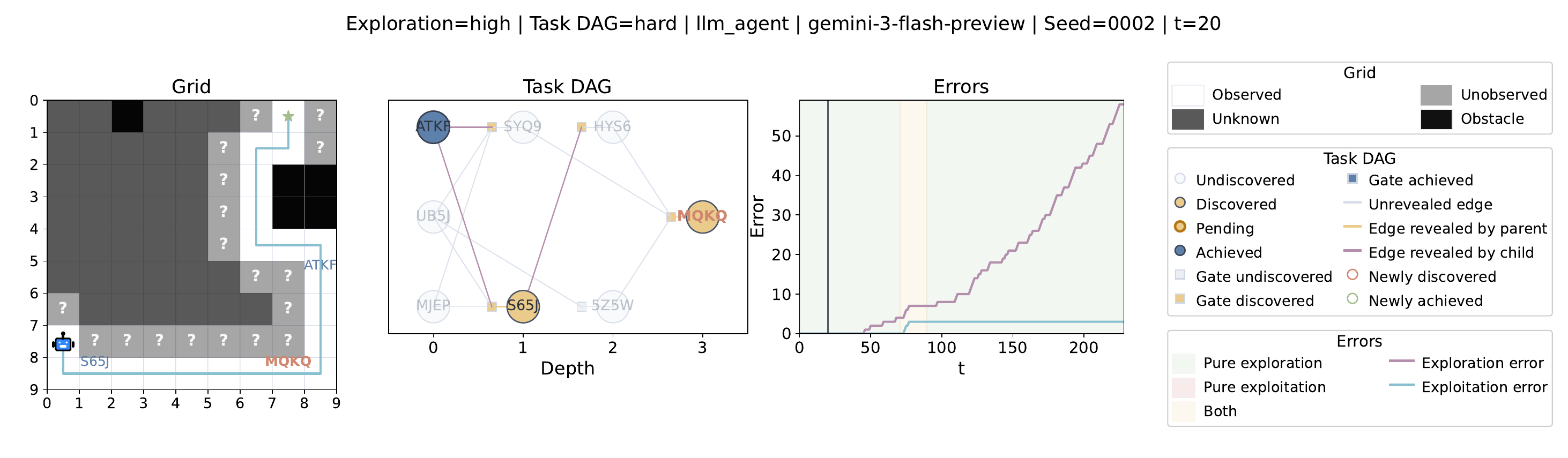}
        \caption{\currentmodel, $t = 20$}
        \end{subfigure}
        \begin{subfigure}[t]{\framewidth}
        \centering
        \includegraphics[width=\textwidth,trim={0 0 0 33pt},clip]{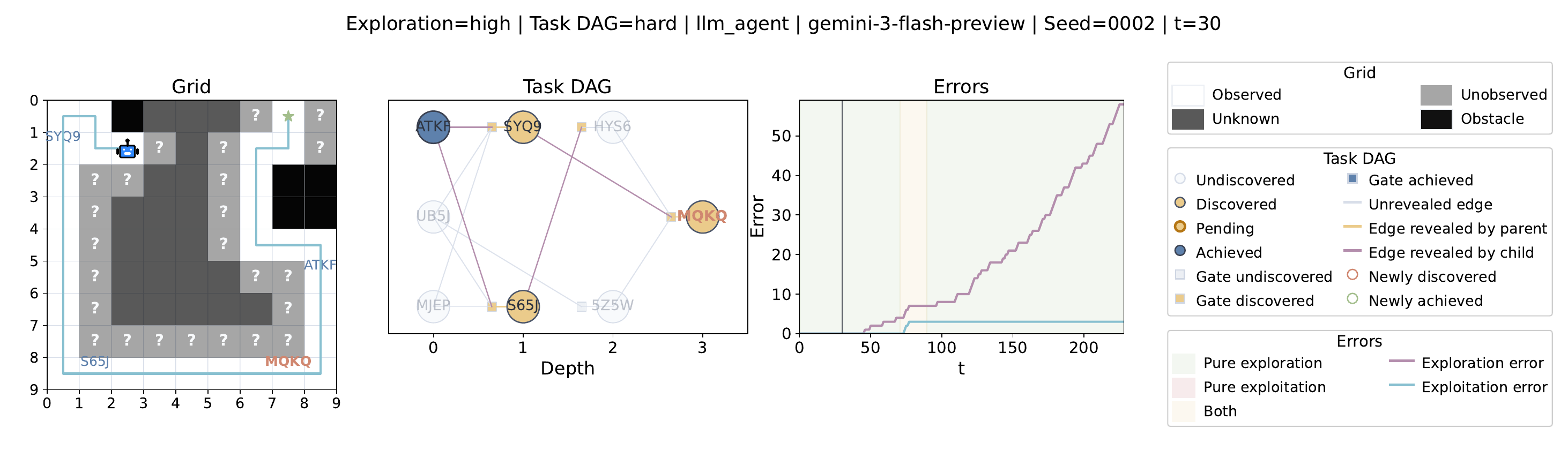}
        \caption{\currentmodel, $t = 30$}
        \end{subfigure}
        \begin{subfigure}[t]{\framewidth}
        \centering
        \includegraphics[width=\textwidth,trim={0 0 0 33pt},clip]{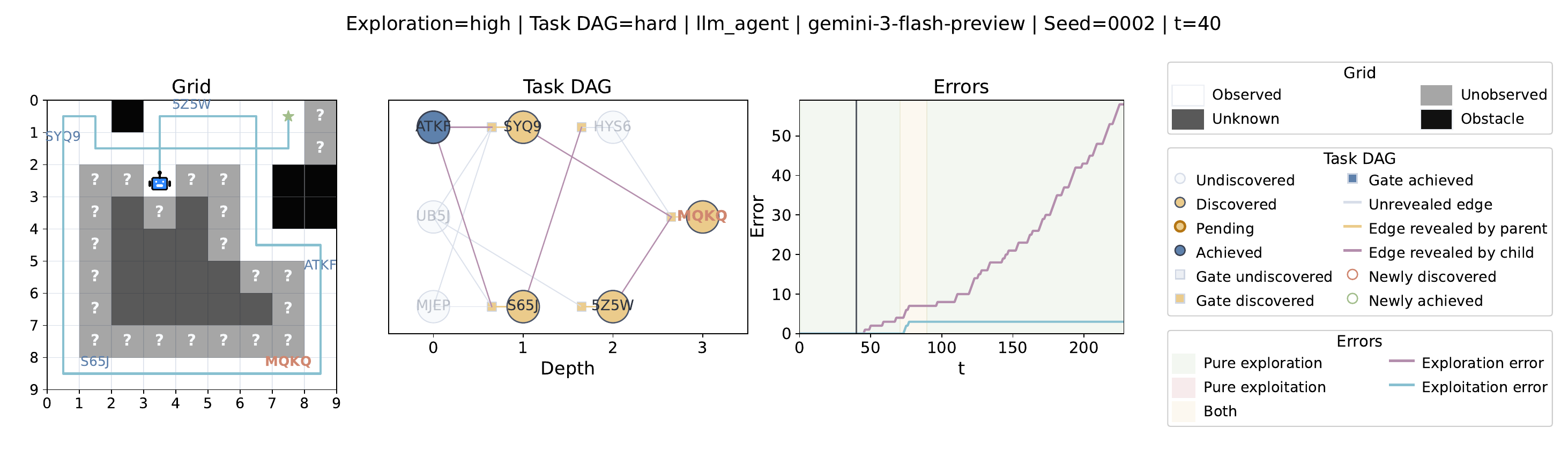}
        \caption{\currentmodel, $t = 40$}
        \end{subfigure}
    \end{center}
    \caption{Results of action trajectories and metric values over each timestep for~\currentmodel.}
\end{figure}

\begin{figure}[ht]
    \begin{center}
        \begin{subfigure}[t]{\framewidth}
        \centering
        \includegraphics[width=\textwidth,trim={0 0 0 33pt},clip]{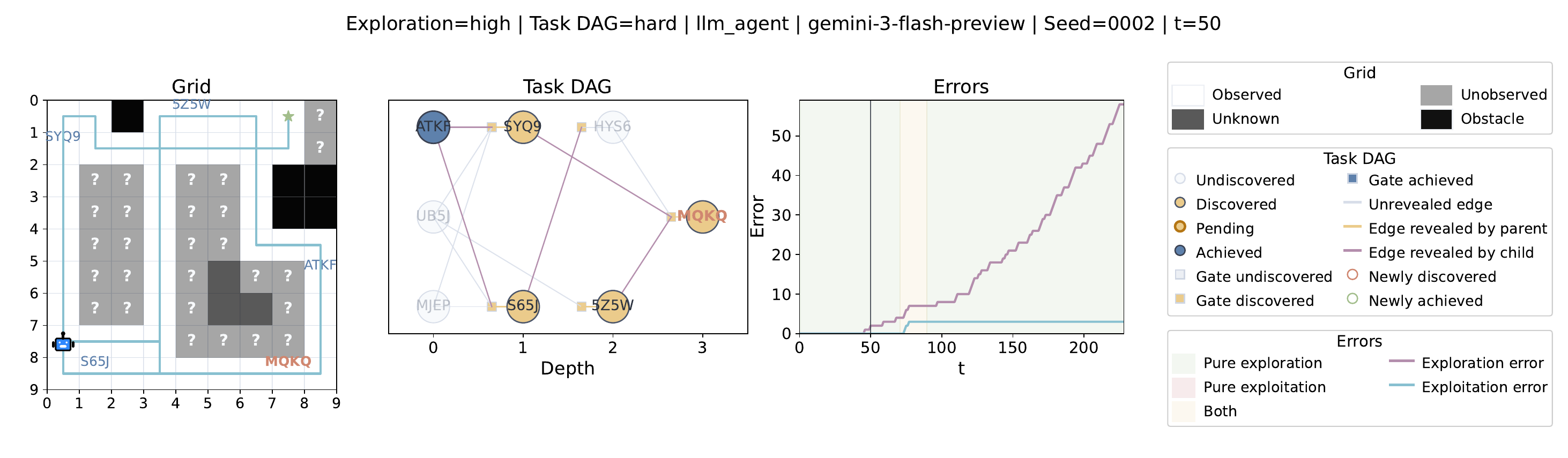}
        \caption{\currentmodel, $t = 50$}
        \end{subfigure}
        \begin{subfigure}[t]{\framewidth}
        \centering
        \includegraphics[width=\textwidth,trim={0 0 0 33pt},clip]{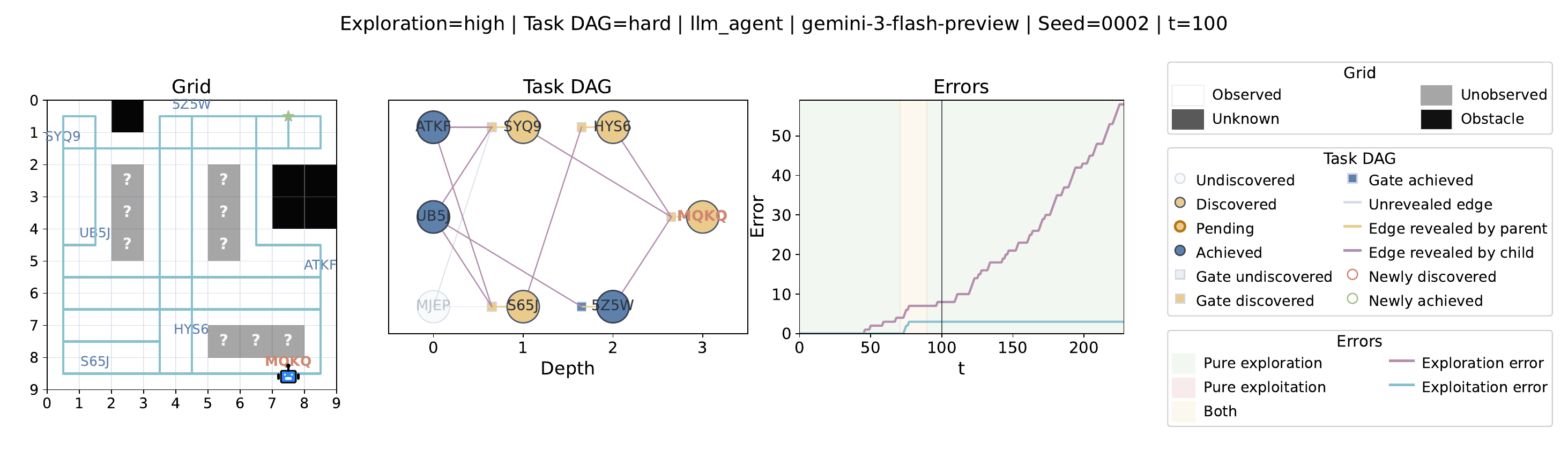}
        \caption{\currentmodel, $t = 100$}
        \end{subfigure}
        \begin{subfigure}[t]{\framewidth}
        \centering
        \includegraphics[width=\textwidth,trim={0 0 0 33pt},clip]{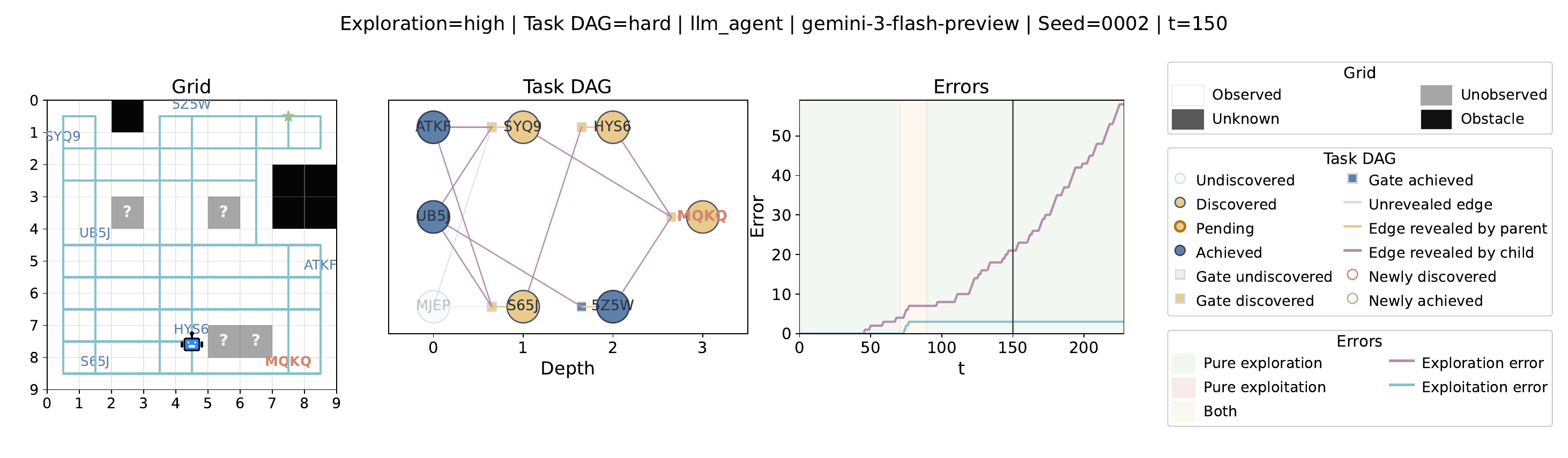}
        \caption{\currentmodel, $t = 150$}
        \end{subfigure}
        \begin{subfigure}[t]{\framewidth}
        \centering
        \includegraphics[width=\textwidth,trim={0 0 0 33pt},clip]{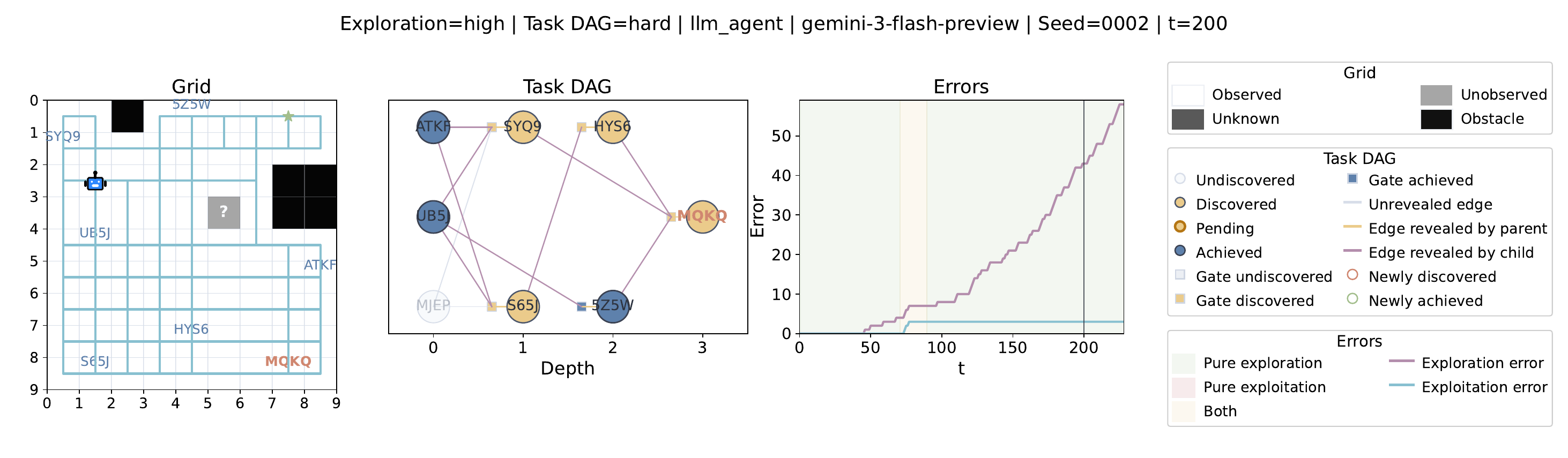}
        \caption{\currentmodel, $t = 200$}
        \end{subfigure}
        \begin{subfigure}[t]{\framewidth}
        \centering
        \includegraphics[width=\textwidth,trim={0 0 0 33pt},clip]{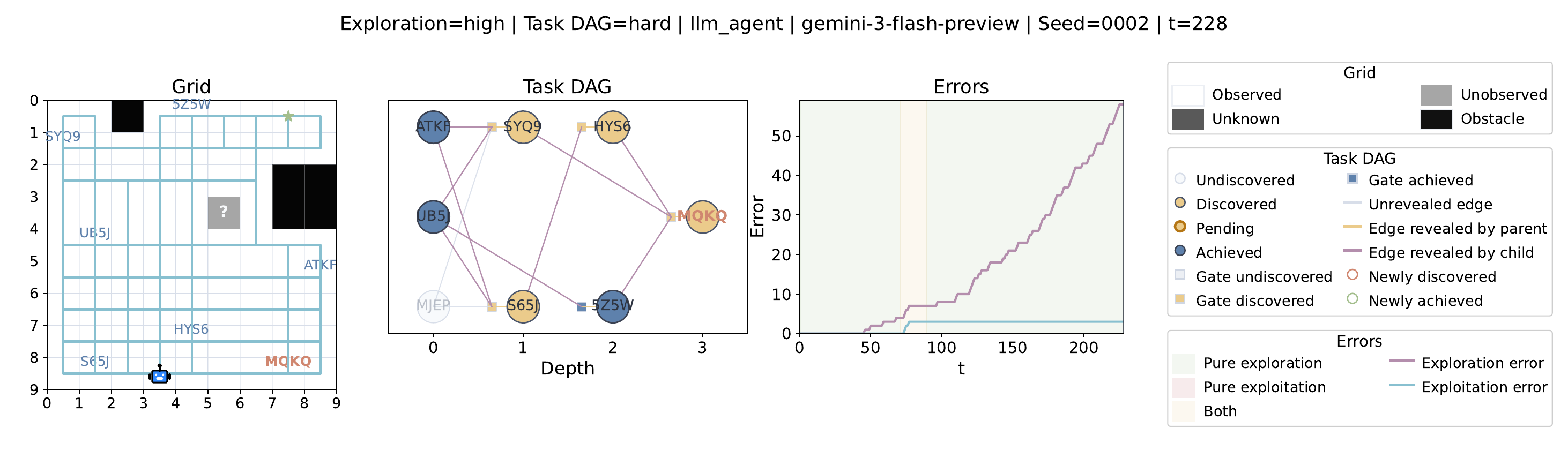}
        \caption{\currentmodel, $t = 228$}
        \end{subfigure}
    \end{center}
    \caption{Results of action trajectories and metric values over each timestep for~\currentmodel.}
\end{figure}

\clearpage
\renewcommand{\currentmodel}{\geminiflashlite}

\begin{figure}[ht]
    \begin{center}
        \begin{subfigure}[t]{\framewidth}
        \centering
        \includegraphics[width=\textwidth,trim={0 0 0 33pt},clip]{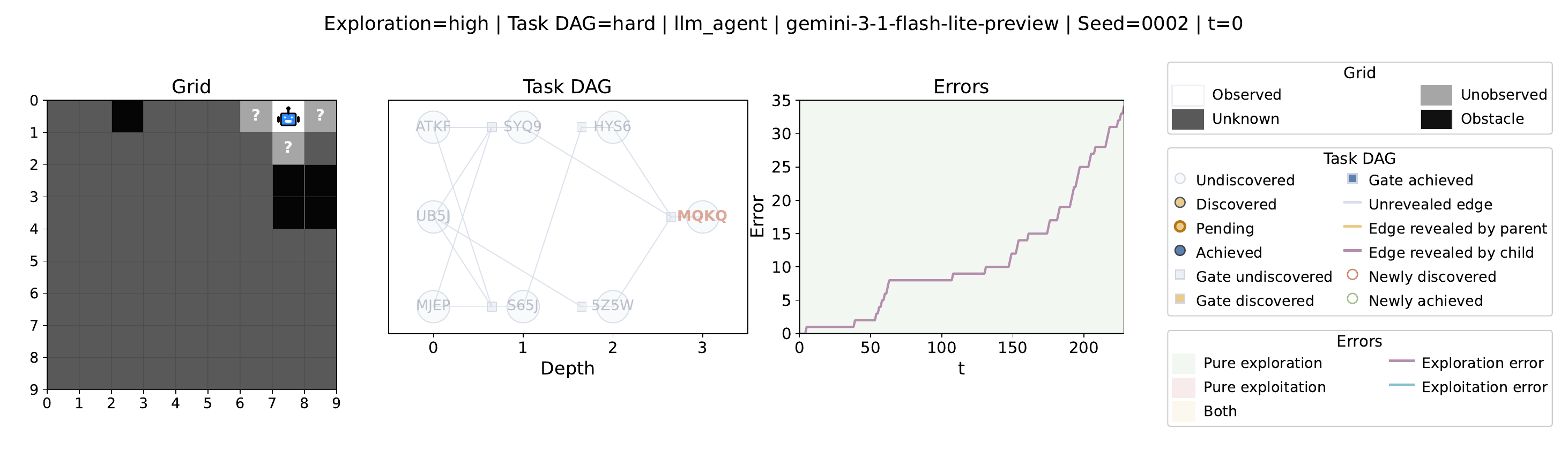}
        \caption{\currentmodel, $t = 0$}
        \end{subfigure}
        \begin{subfigure}[t]{\framewidth}
        \centering
        \includegraphics[width=\textwidth,trim={0 0 0 33pt},clip]{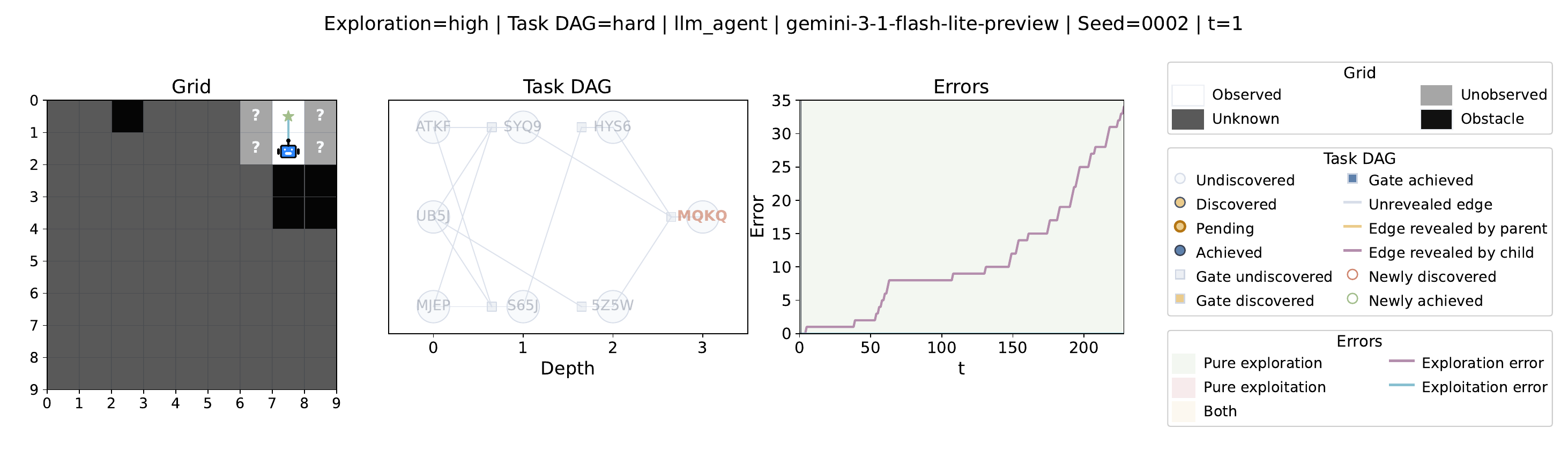}
        \caption{\currentmodel, $t = 1$}
        \end{subfigure}
        \begin{subfigure}[t]{\framewidth}
        \centering
        \includegraphics[width=\textwidth,trim={0 0 0 33pt},clip]{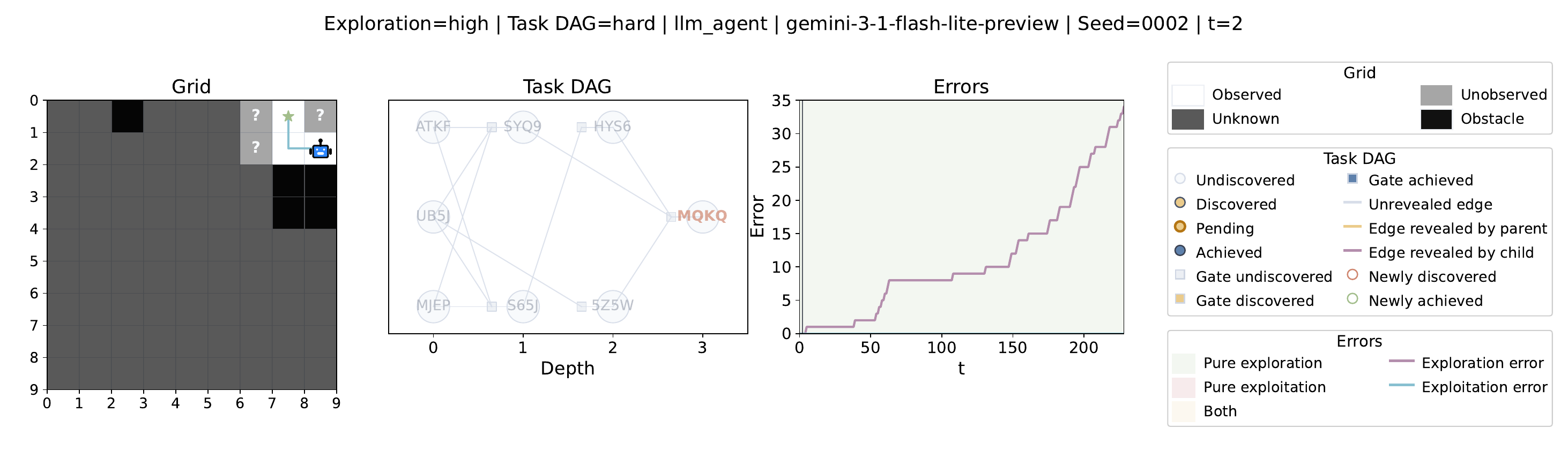}
        \caption{\currentmodel, $t = 2$}
        \end{subfigure}
        \begin{subfigure}[t]{\framewidth}
        \centering
        \includegraphics[width=\textwidth,trim={0 0 0 33pt},clip]{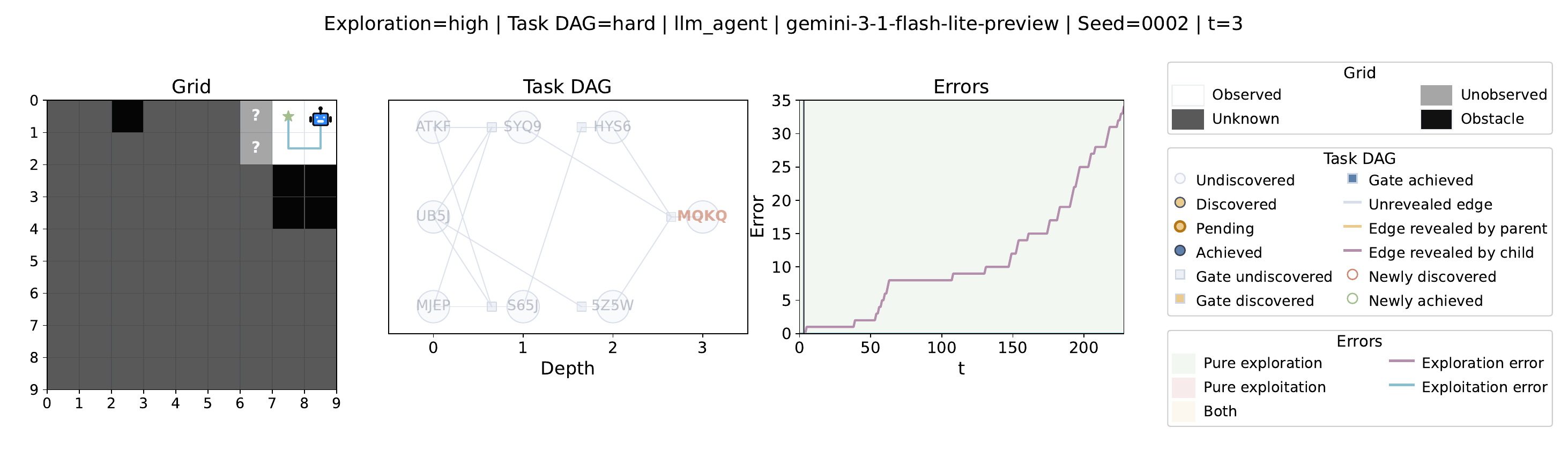}
        \caption{\currentmodel, $t = 3$}
        \end{subfigure}
        \begin{subfigure}[t]{\framewidth}
        \centering
        \includegraphics[width=\textwidth,trim={0 0 0 33pt},clip]{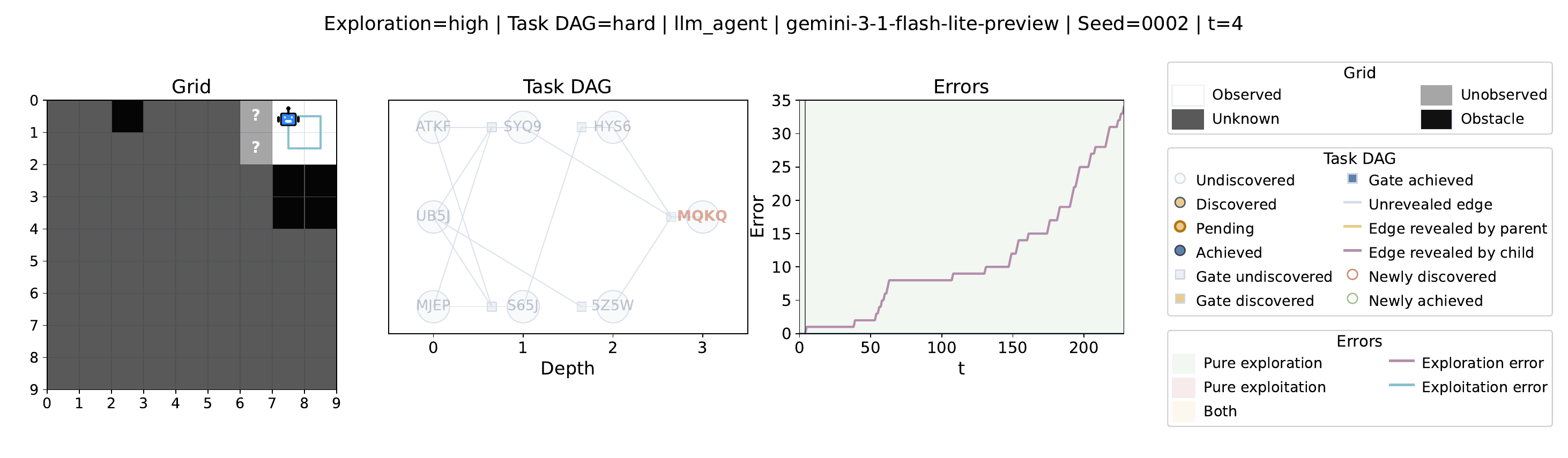}
        \caption{\currentmodel, $t = 4$}
        \end{subfigure}
    \end{center}
    \caption{Results of action trajectories and metric values over each timestep for~\currentmodel.}
\end{figure}

\begin{figure}[ht]
    \begin{center}
        \begin{subfigure}[t]{\framewidth}
        \centering
        \includegraphics[width=\textwidth,trim={0 0 0 33pt},clip]{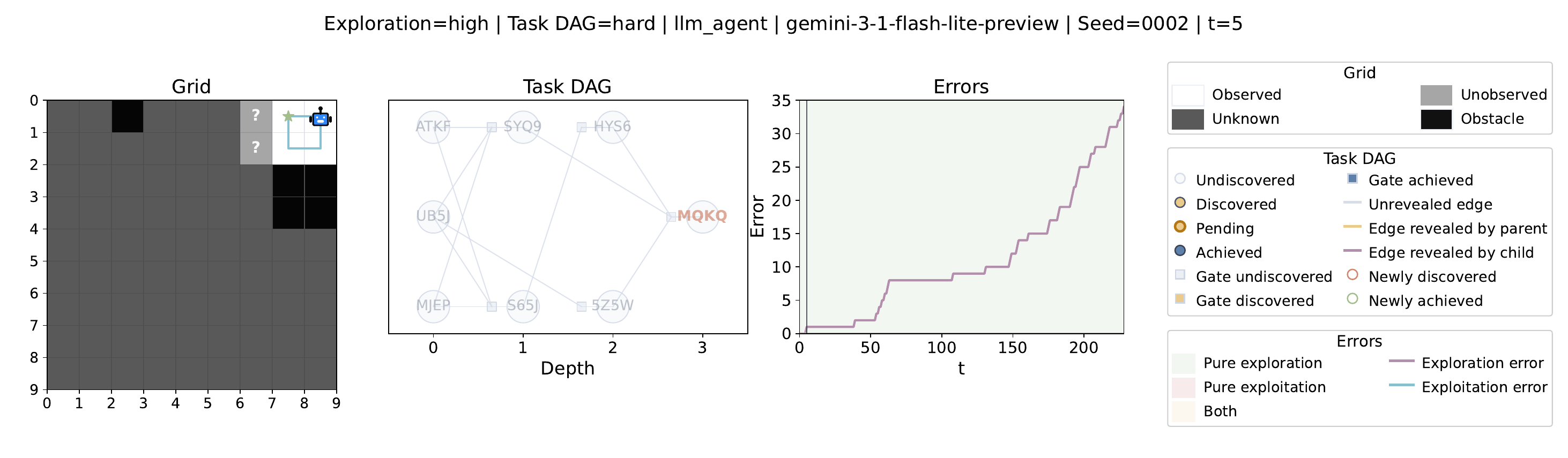}
        \caption{\currentmodel, $t = 5$}
        \end{subfigure}
        \begin{subfigure}[t]{\framewidth}
        \centering
        \includegraphics[width=\textwidth,trim={0 0 0 33pt},clip]{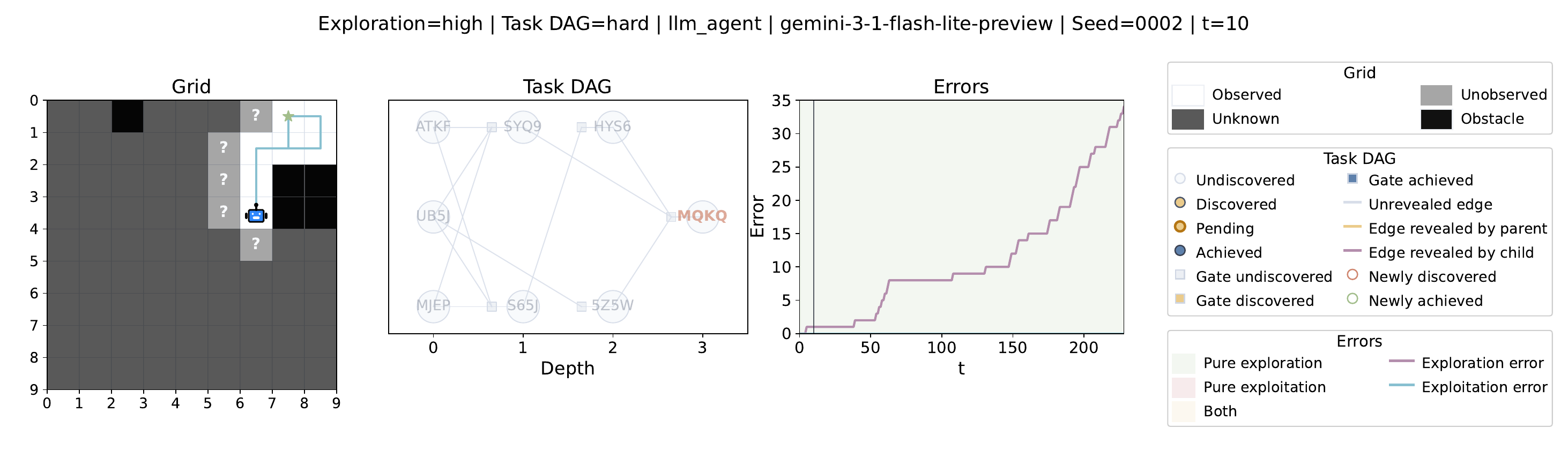}
        \caption{\currentmodel, $t = 10$}
        \end{subfigure}
        \begin{subfigure}[t]{\framewidth}
        \centering
        \includegraphics[width=\textwidth,trim={0 0 0 33pt},clip]{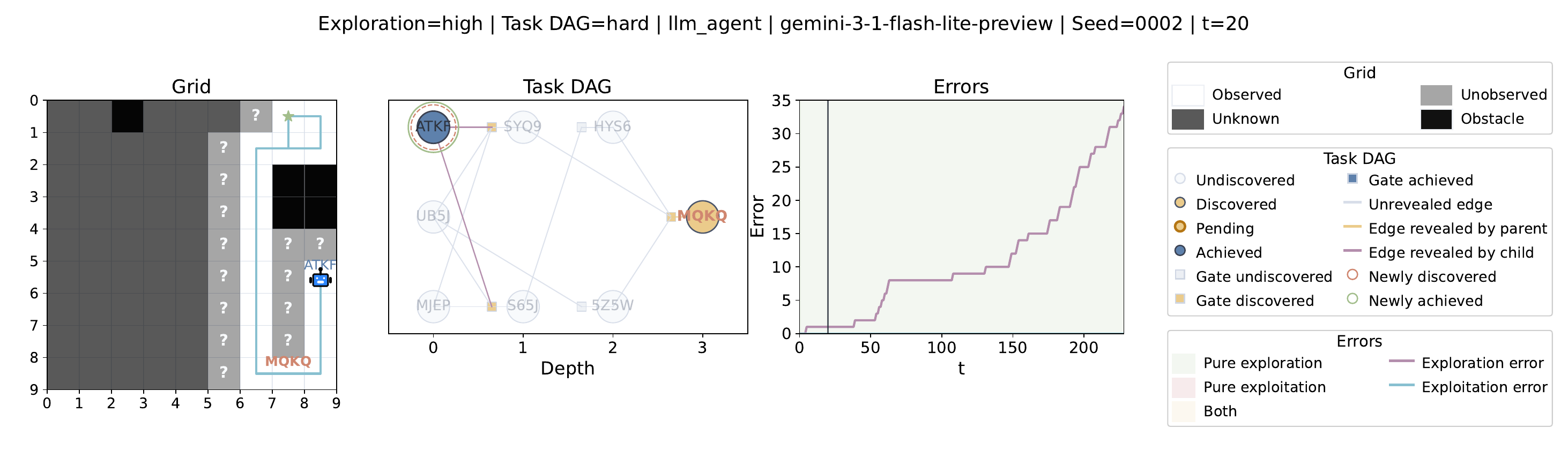}
        \caption{\currentmodel, $t = 20$}
        \end{subfigure}
        \begin{subfigure}[t]{\framewidth}
        \centering
        \includegraphics[width=\textwidth,trim={0 0 0 33pt},clip]{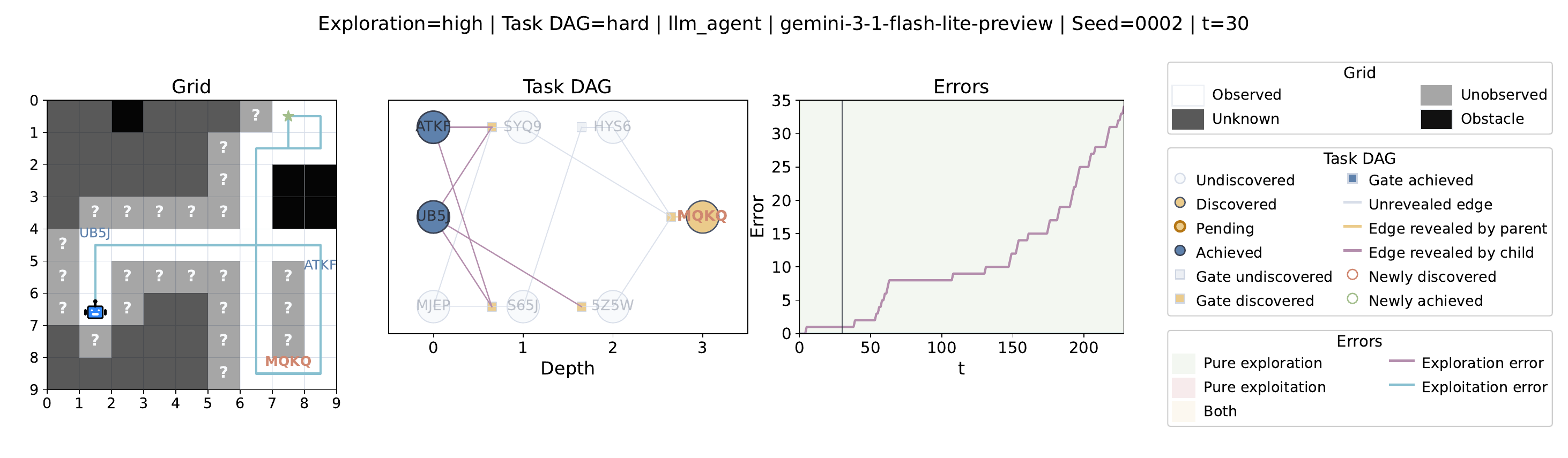}
        \caption{\currentmodel, $t = 30$}
        \end{subfigure}
        \begin{subfigure}[t]{\framewidth}
        \centering
        \includegraphics[width=\textwidth,trim={0 0 0 33pt},clip]{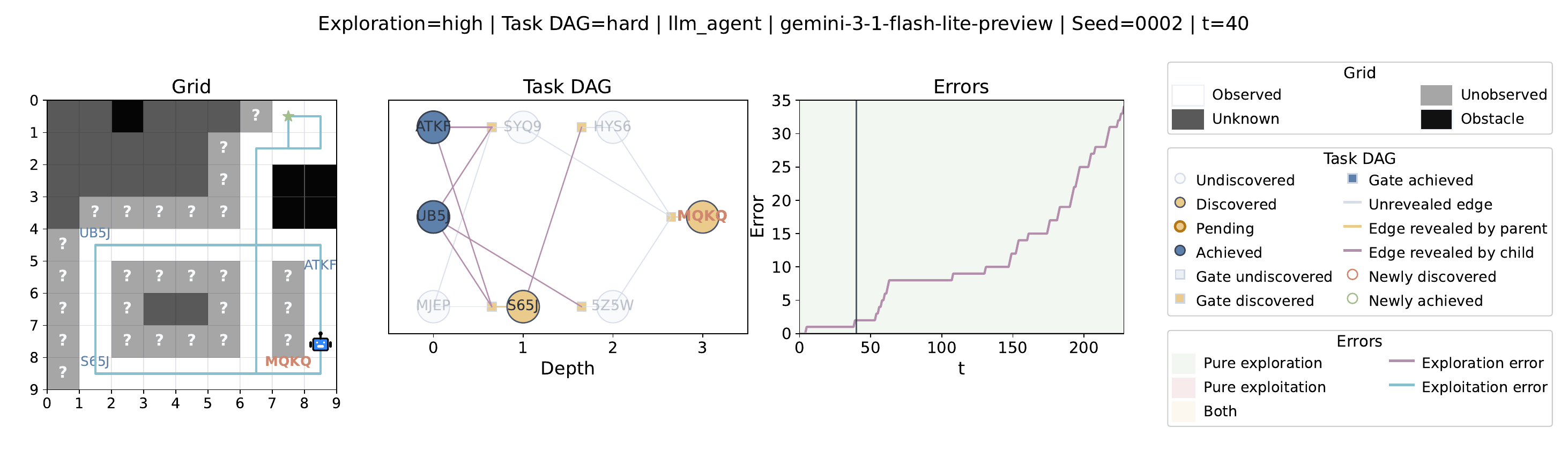}
        \caption{\currentmodel, $t = 40$}
        \end{subfigure}
    \end{center}
    \caption{Results of action trajectories and metric values over each timestep for~\currentmodel.}
\end{figure}

\begin{figure}[ht]
    \begin{center}
        \begin{subfigure}[t]{\framewidth}
        \centering
        \includegraphics[width=\textwidth,trim={0 0 0 33pt},clip]{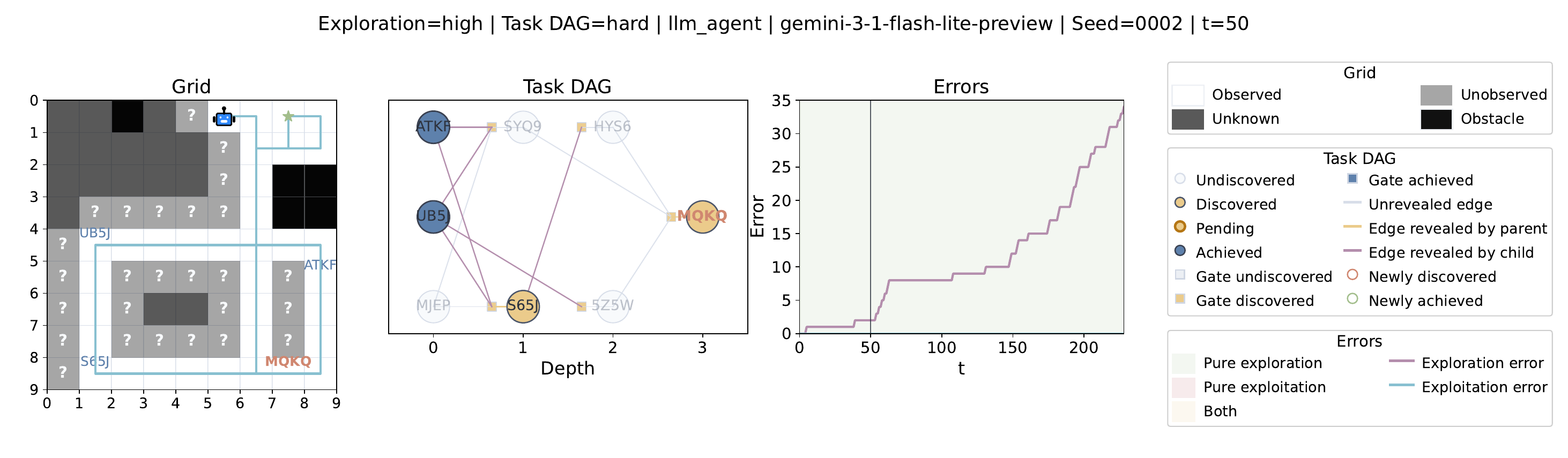}
        \caption{\currentmodel, $t = 50$}
        \end{subfigure}
        \begin{subfigure}[t]{\framewidth}
        \centering
        \includegraphics[width=\textwidth,trim={0 0 0 33pt},clip]{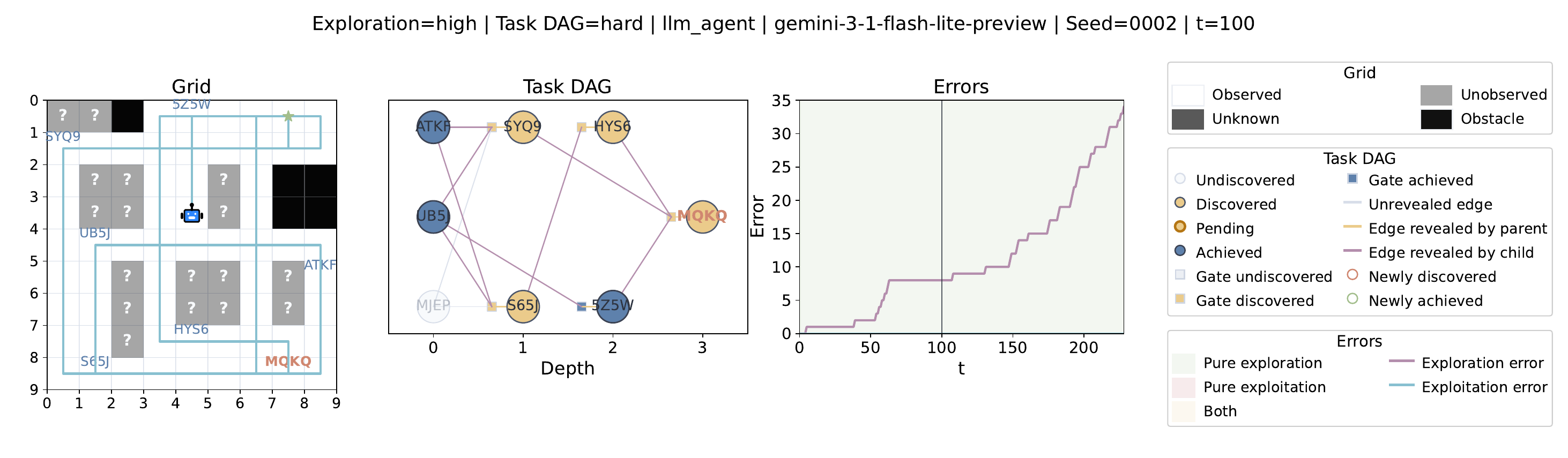}
        \caption{\currentmodel, $t = 100$}
        \end{subfigure}
        \begin{subfigure}[t]{\framewidth}
        \centering
        \includegraphics[width=\textwidth,trim={0 0 0 33pt},clip]{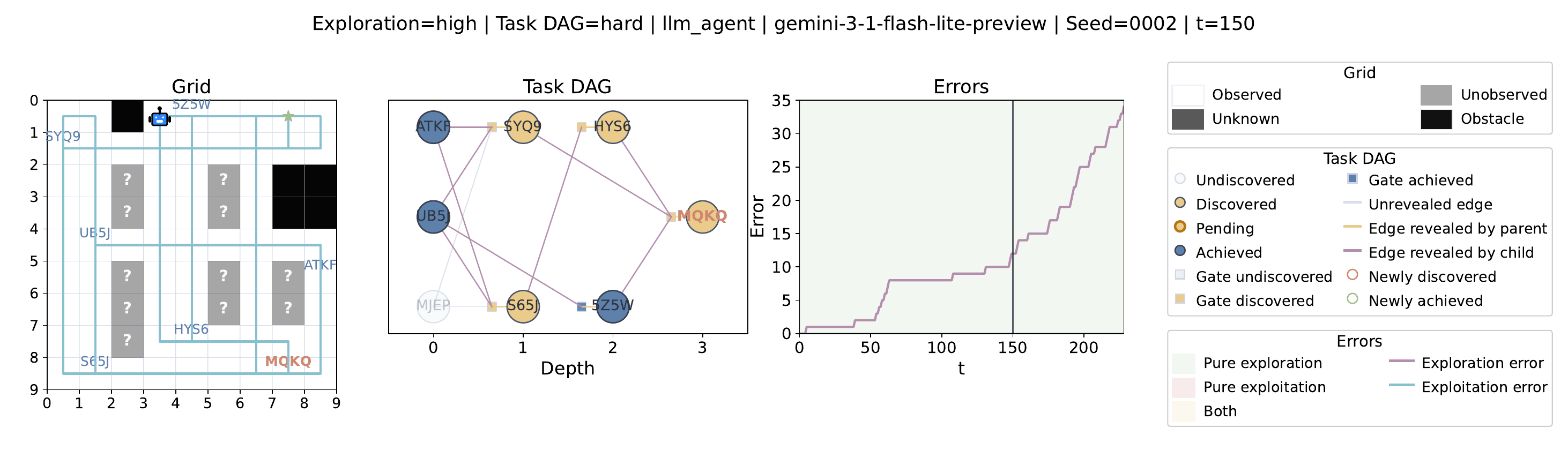}
        \caption{\currentmodel, $t = 150$}
        \end{subfigure}
        \begin{subfigure}[t]{\framewidth}
        \centering
        \includegraphics[width=\textwidth,trim={0 0 0 33pt},clip]{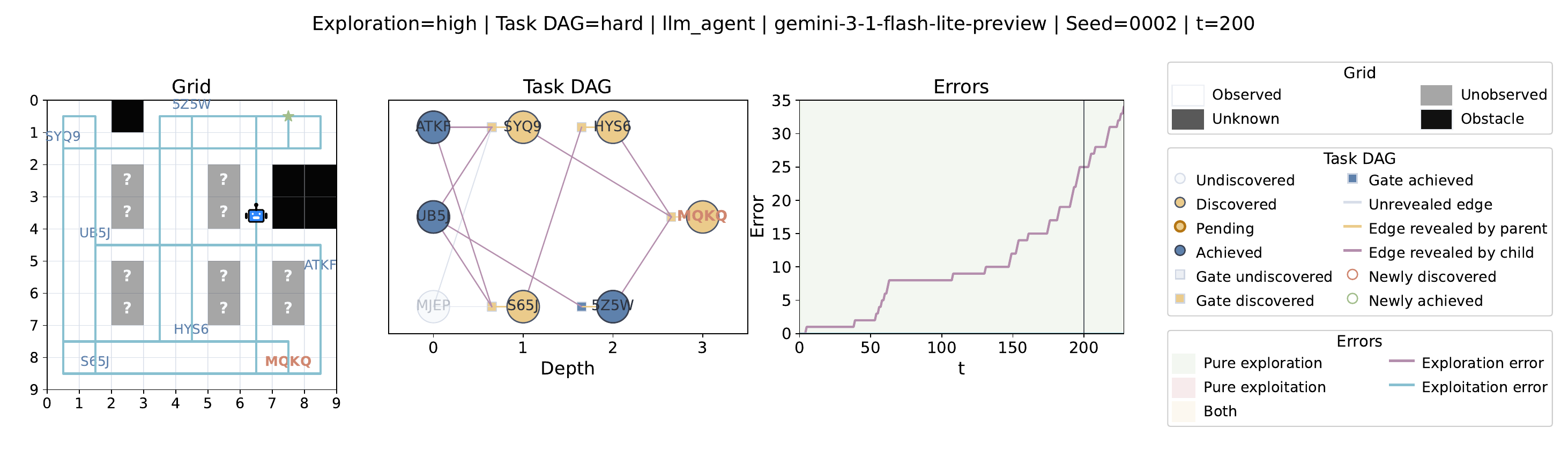}
        \caption{\currentmodel, $t = 200$}
        \end{subfigure}
        \begin{subfigure}[t]{\framewidth}
        \centering
        \includegraphics[width=\textwidth,trim={0 0 0 33pt},clip]{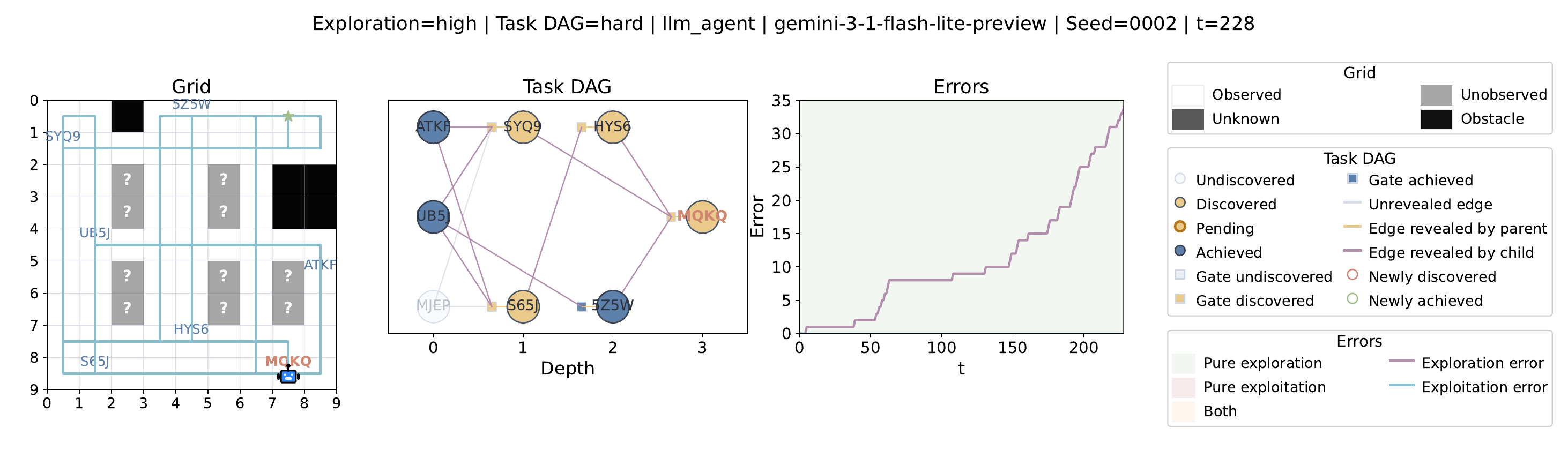}
        \caption{\currentmodel, $t = 228$}
        \end{subfigure}
    \end{center}
    \caption{Results of action trajectories and metric values over each timestep for~\currentmodel.}
\end{figure}

\clearpage
\renewcommand{\currentmodel}{\geminipro}

\begin{figure}[ht]
    \begin{center}
        \begin{subfigure}[t]{\framewidth}
        \centering
        \includegraphics[width=\textwidth,trim={0 0 0 33pt},clip]{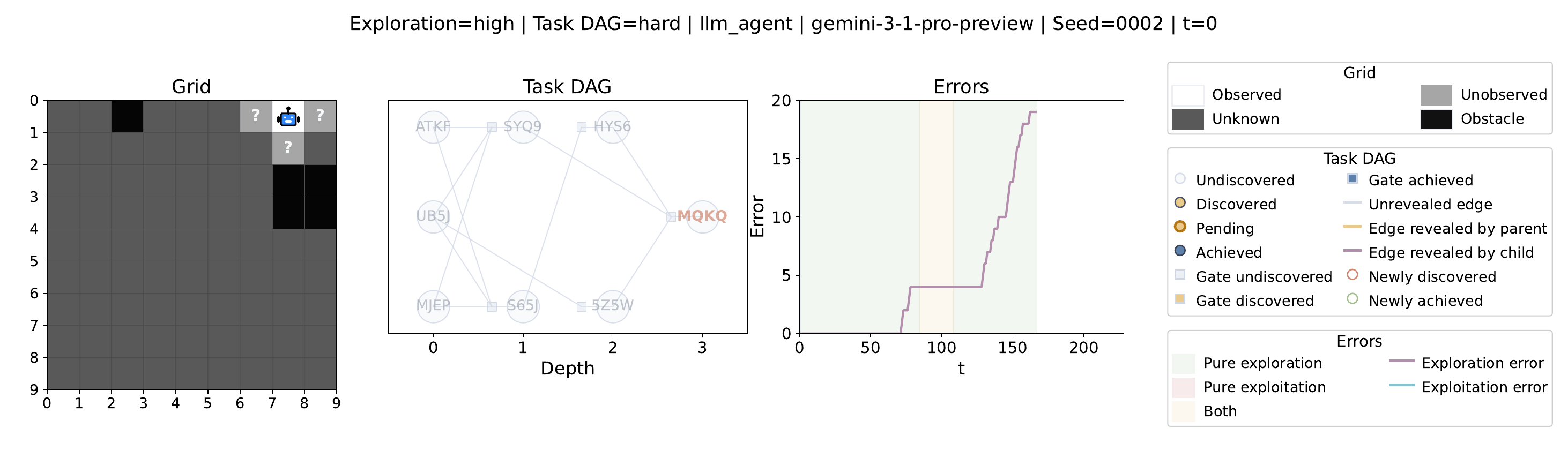}
        \caption{\currentmodel, $t = 0$}
        \end{subfigure}
        \begin{subfigure}[t]{\framewidth}
        \centering
        \includegraphics[width=\textwidth,trim={0 0 0 33pt},clip]{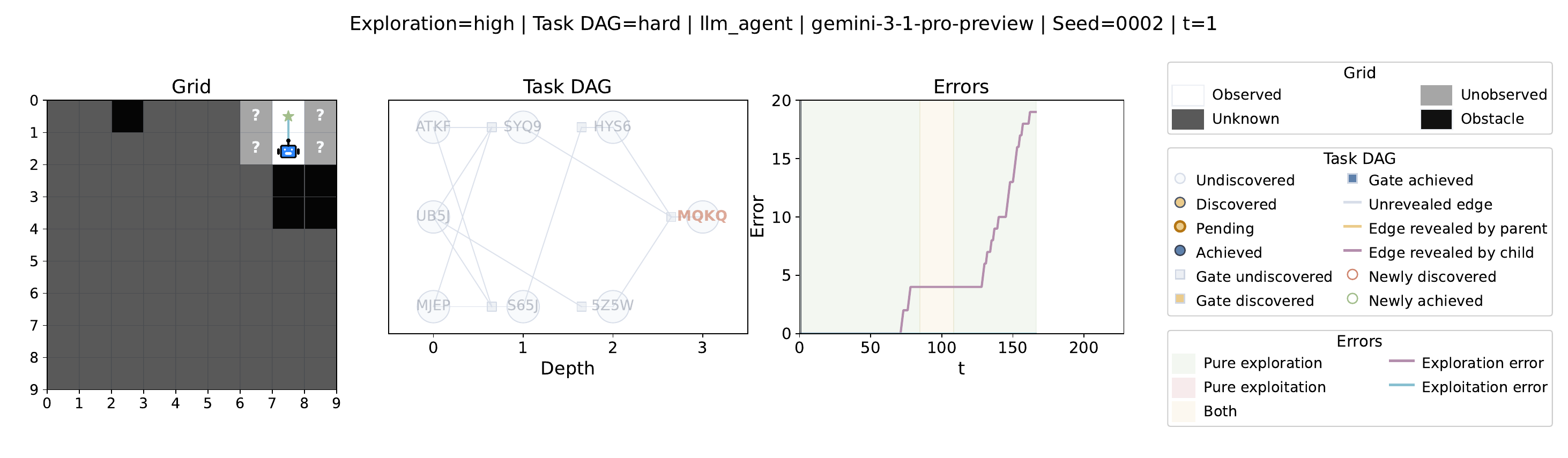}
        \caption{\currentmodel, $t = 1$}
        \end{subfigure}
        \begin{subfigure}[t]{\framewidth}
        \centering
        \includegraphics[width=\textwidth,trim={0 0 0 33pt},clip]{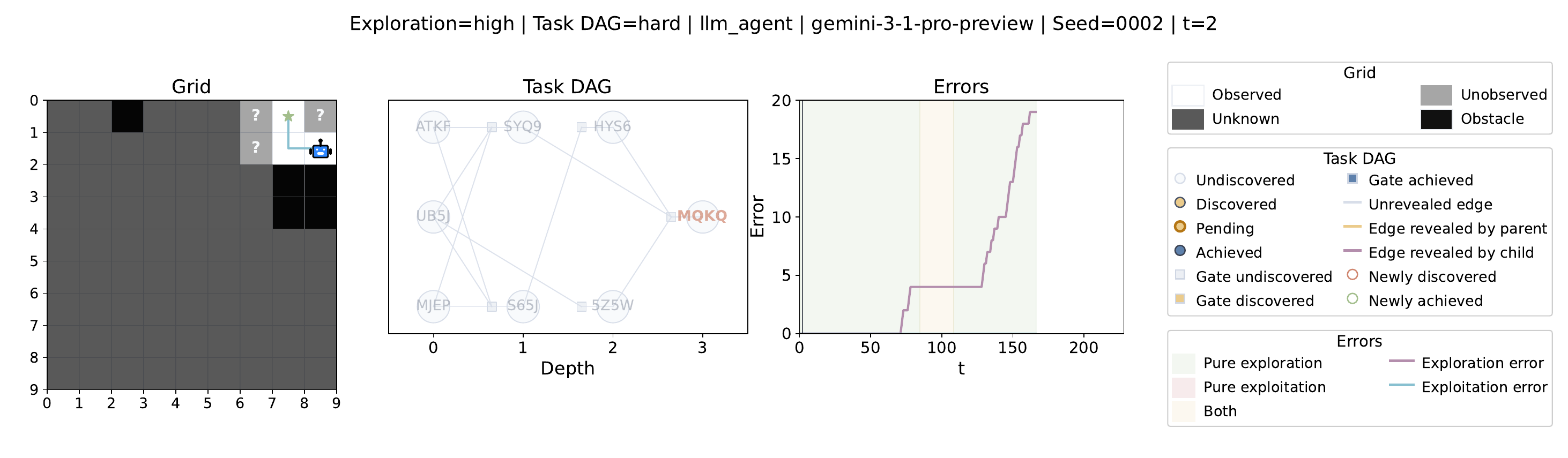}
        \caption{\currentmodel, $t = 2$}
        \end{subfigure}
        \begin{subfigure}[t]{\framewidth}
        \centering
        \includegraphics[width=\textwidth,trim={0 0 0 33pt},clip]{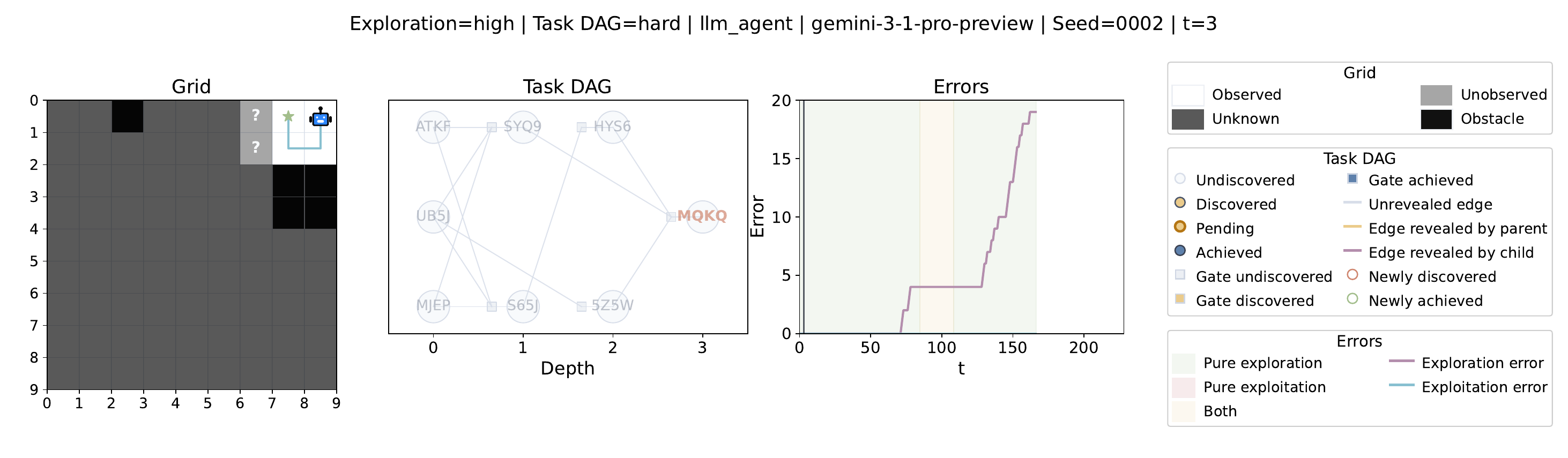}
        \caption{\currentmodel, $t = 3$}
        \end{subfigure}
        \begin{subfigure}[t]{\framewidth}
        \centering
        \includegraphics[width=\textwidth,trim={0 0 0 33pt},clip]{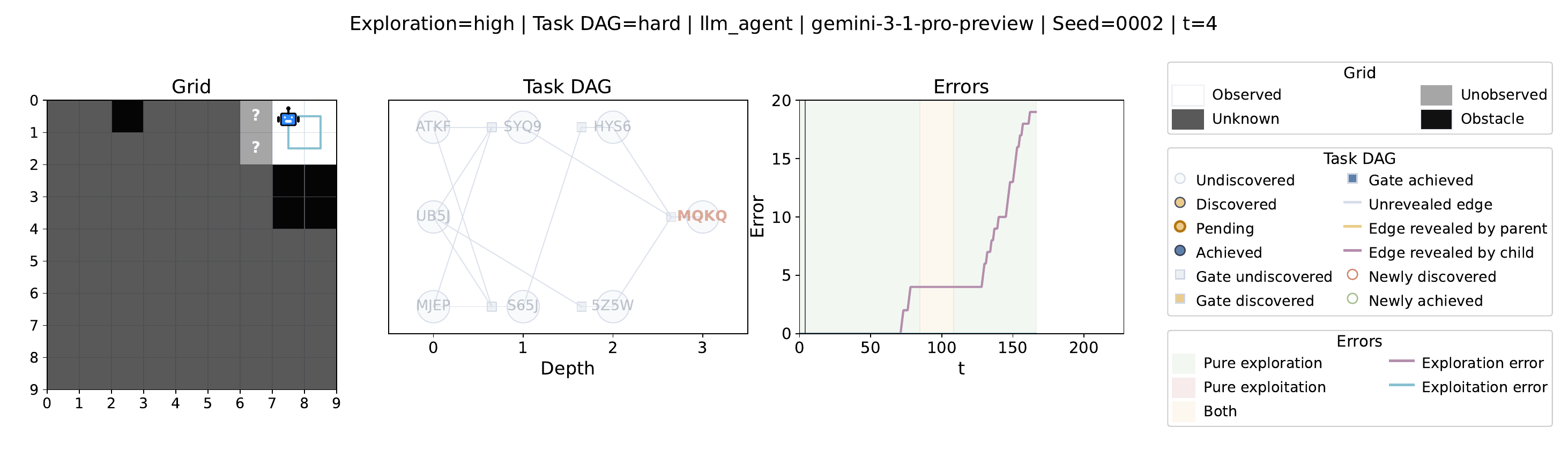}
        \caption{\currentmodel, $t = 4$}
        \end{subfigure}
    \end{center}
    \caption{Results of action trajectories and metric values over each timestep for~\currentmodel.}
\end{figure}

\begin{figure}[ht]
    \begin{center}
        \begin{subfigure}[t]{\framewidth}
        \centering
        \includegraphics[width=\textwidth,trim={0 0 0 33pt},clip]{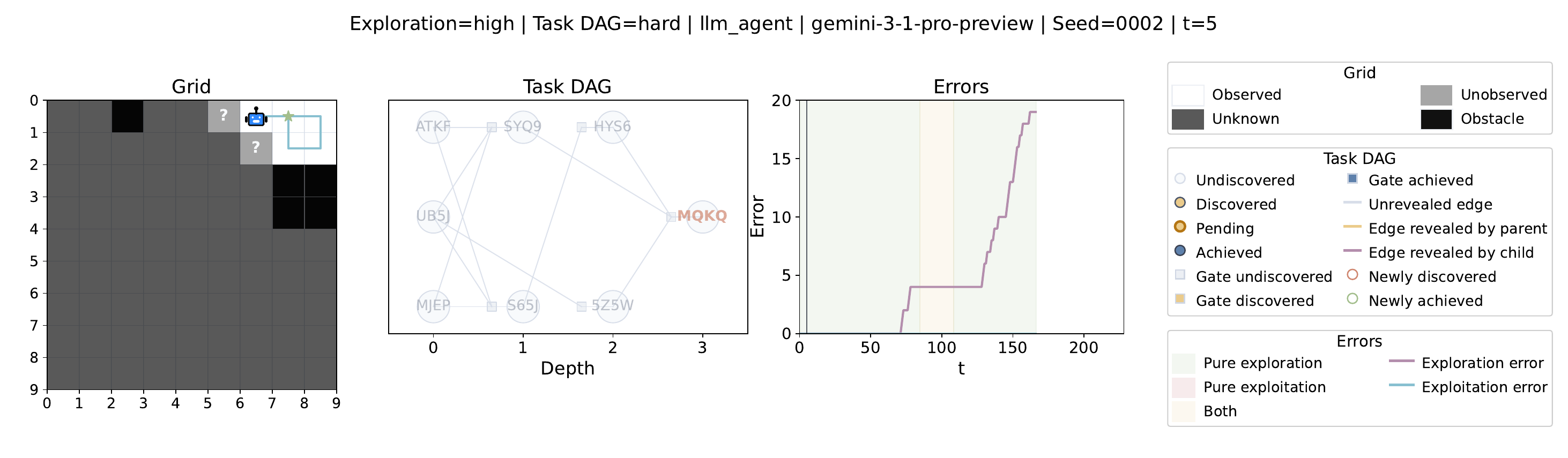}
        \caption{\currentmodel, $t = 5$}
        \end{subfigure}
        \begin{subfigure}[t]{\framewidth}
        \centering
        \includegraphics[width=\textwidth,trim={0 0 0 33pt},clip]{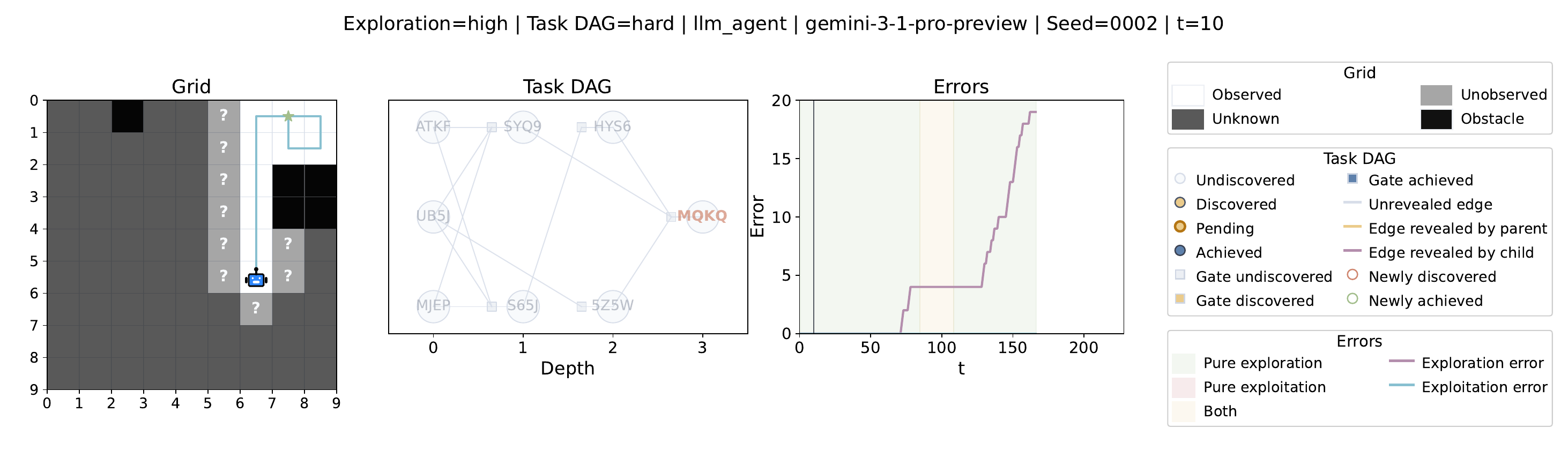}
        \caption{\currentmodel, $t = 10$}
        \end{subfigure}
        \begin{subfigure}[t]{\framewidth}
        \centering
        \includegraphics[width=\textwidth,trim={0 0 0 33pt},clip]{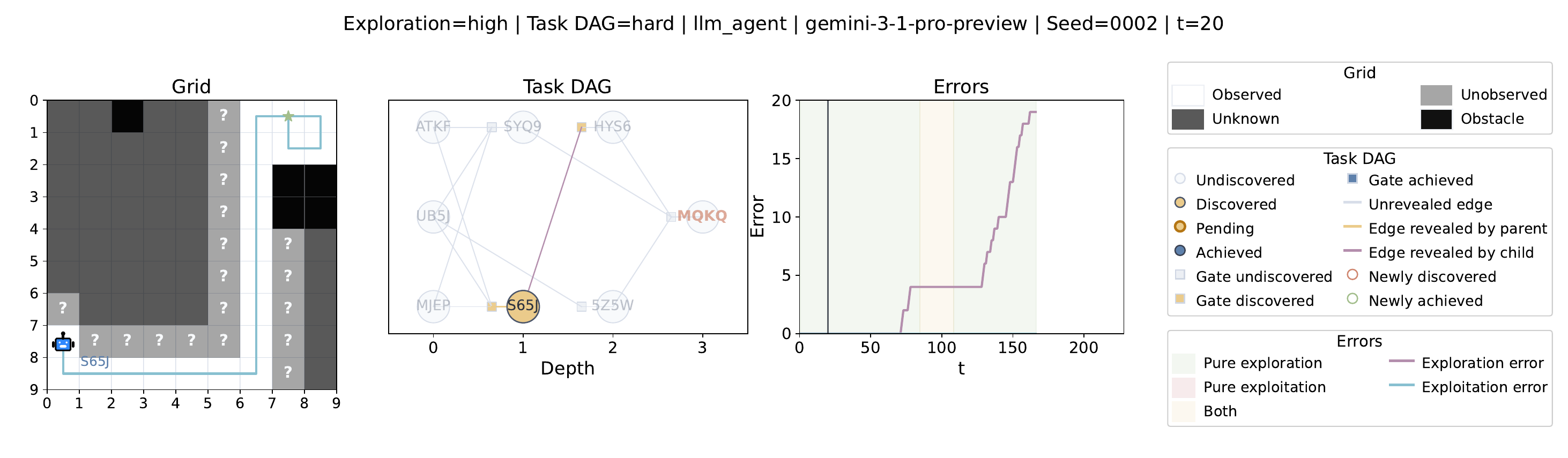}
        \caption{\currentmodel, $t = 20$}
        \end{subfigure}
        \begin{subfigure}[t]{\framewidth}
        \centering
        \includegraphics[width=\textwidth,trim={0 0 0 33pt},clip]{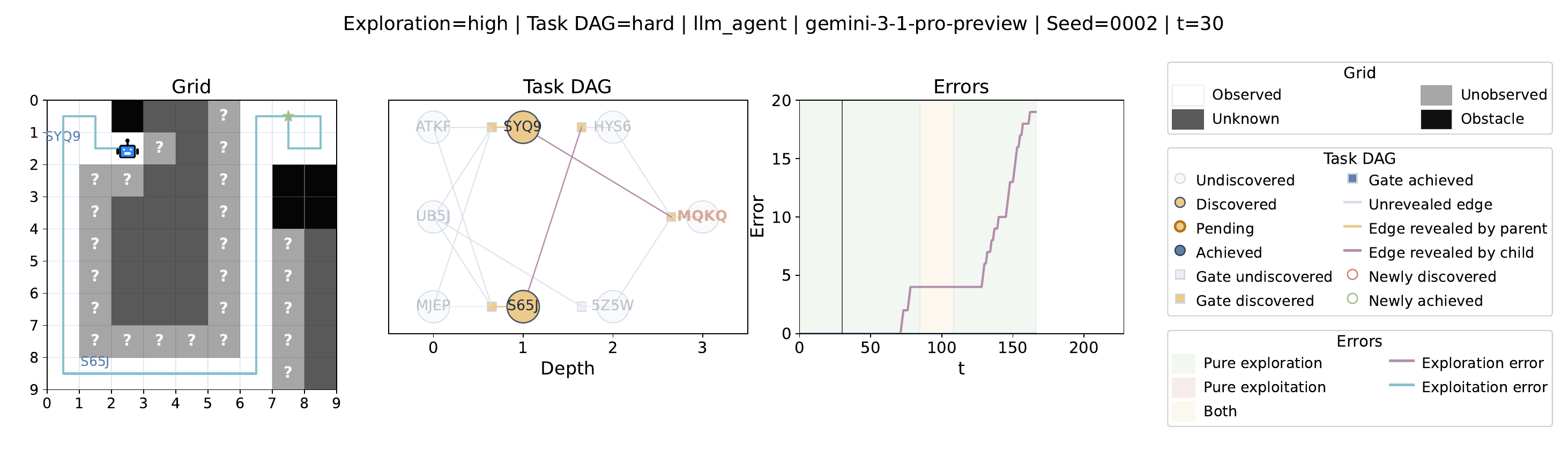}
        \caption{\currentmodel, $t = 30$}
        \end{subfigure}
        \begin{subfigure}[t]{\framewidth}
        \centering
        \includegraphics[width=\textwidth,trim={0 0 0 33pt},clip]{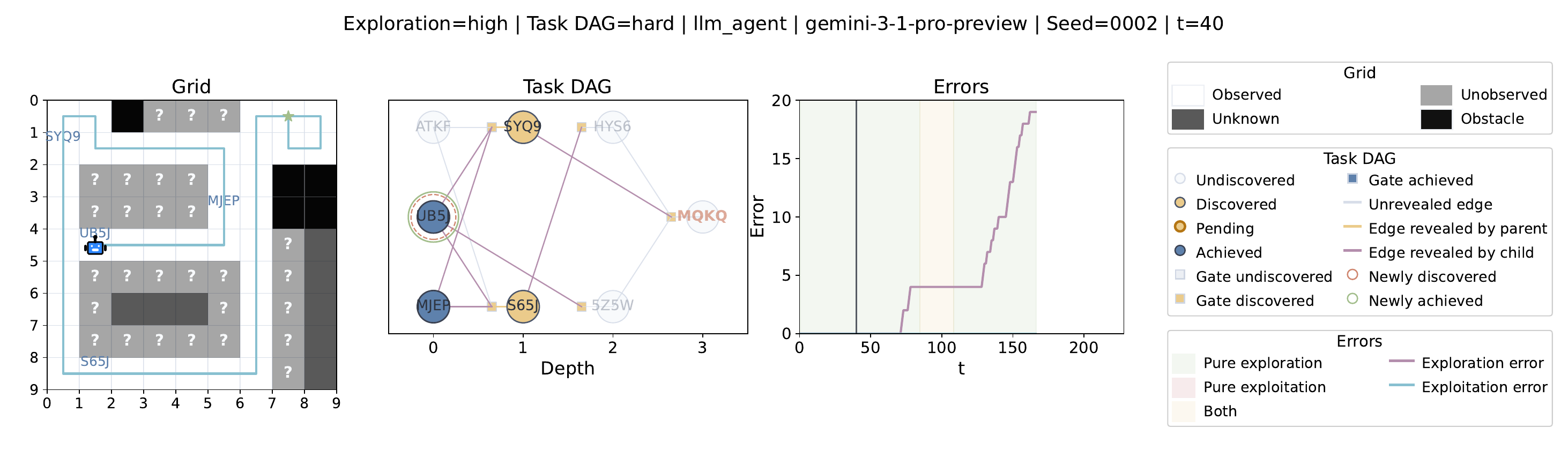}
        \caption{\currentmodel, $t = 40$}
        \end{subfigure}
    \end{center}
    \caption{Results of action trajectories and metric values over each timestep for~\currentmodel.}
\end{figure}

\begin{figure}[ht]
    \begin{center}
        \begin{subfigure}[t]{\framewidth}
        \centering
        \includegraphics[width=\textwidth,trim={0 0 0 33pt},clip]{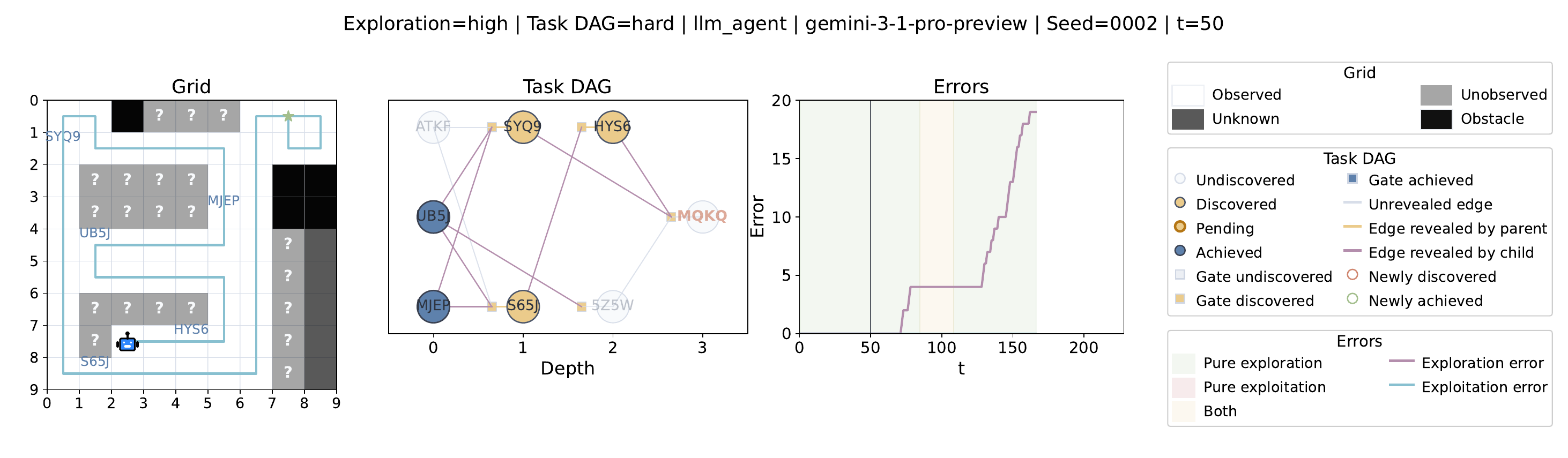}
        \caption{\currentmodel, $t = 50$}
        \end{subfigure}
        \begin{subfigure}[t]{\framewidth}
        \centering
        \includegraphics[width=\textwidth,trim={0 0 0 33pt},clip]{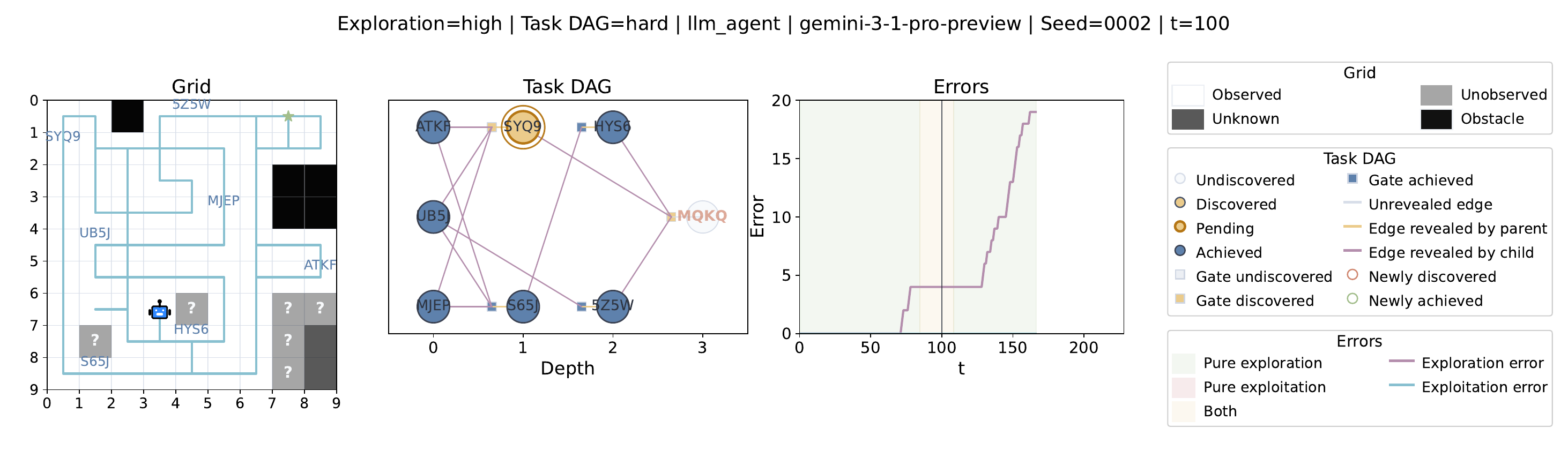}
        \caption{\currentmodel, $t = 100$}
        \end{subfigure}
        \begin{subfigure}[t]{\framewidth}
        \centering
        \includegraphics[width=\textwidth,trim={0 0 0 33pt},clip]{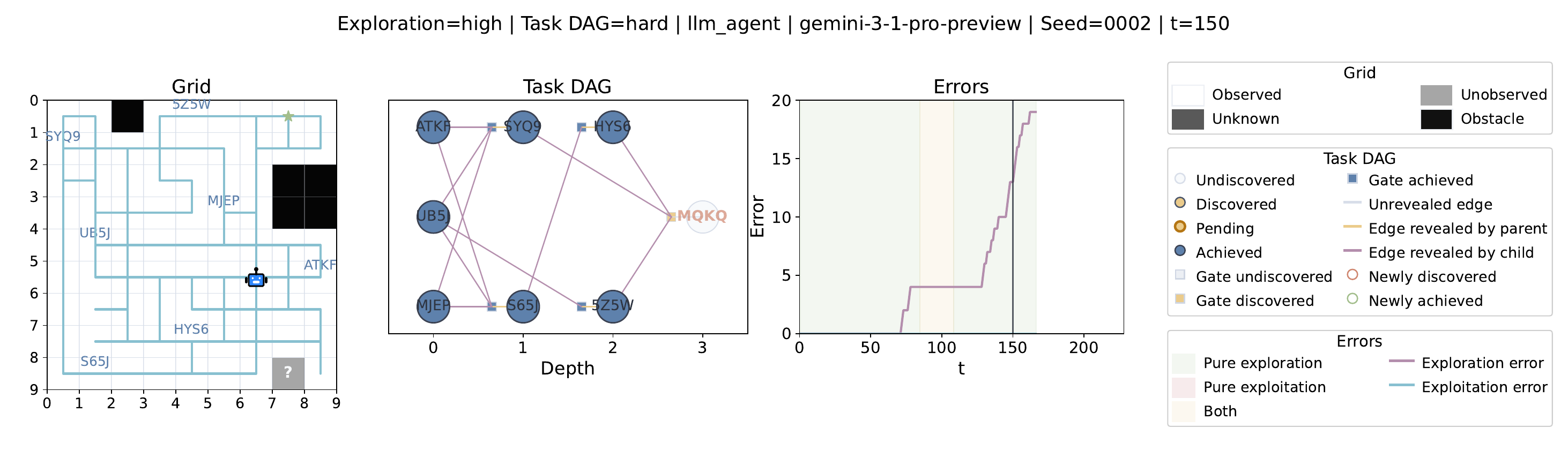}
        \caption{\currentmodel, $t = 150$}
        \end{subfigure}
        \begin{subfigure}[t]{\framewidth}
        \centering
        \includegraphics[width=\textwidth,trim={0 0 0 33pt},clip]{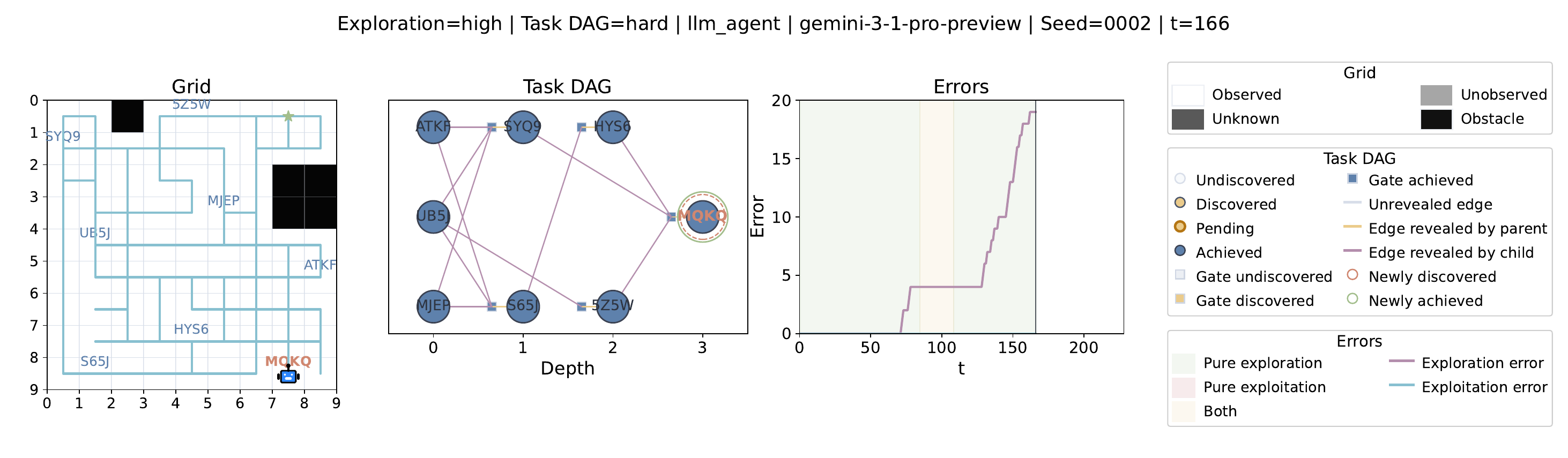}
        \caption{\currentmodel, $t = 166$}
        \end{subfigure}
    \end{center}
    \caption{Results of action trajectories and metric values over each timestep for~\currentmodel.}
\end{figure}

\clearpage

\renewcommand{\currentmodel}{\gptfouronenano}

\begin{figure}[ht]
    \begin{center}
        \begin{subfigure}[t]{\framewidth}
        \centering
        \includegraphics[width=\textwidth,trim={0 0 0 33pt},clip]{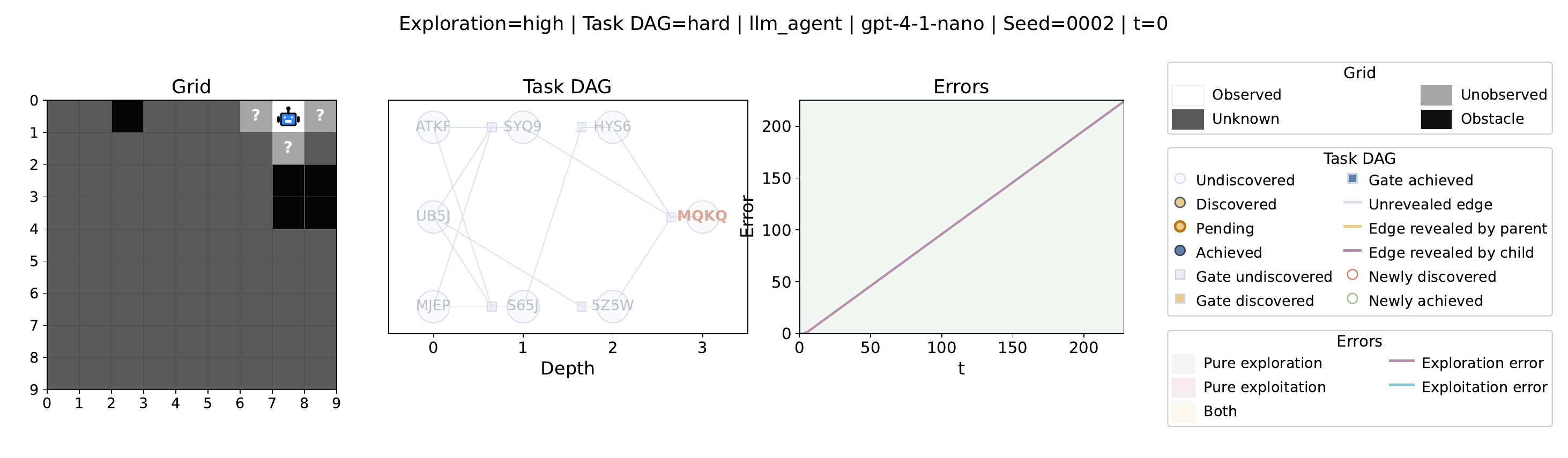}
        \caption{\currentmodel, $t = 0$}
        \end{subfigure}
        \begin{subfigure}[t]{\framewidth}
        \centering
        \includegraphics[width=\textwidth,trim={0 0 0 33pt},clip]{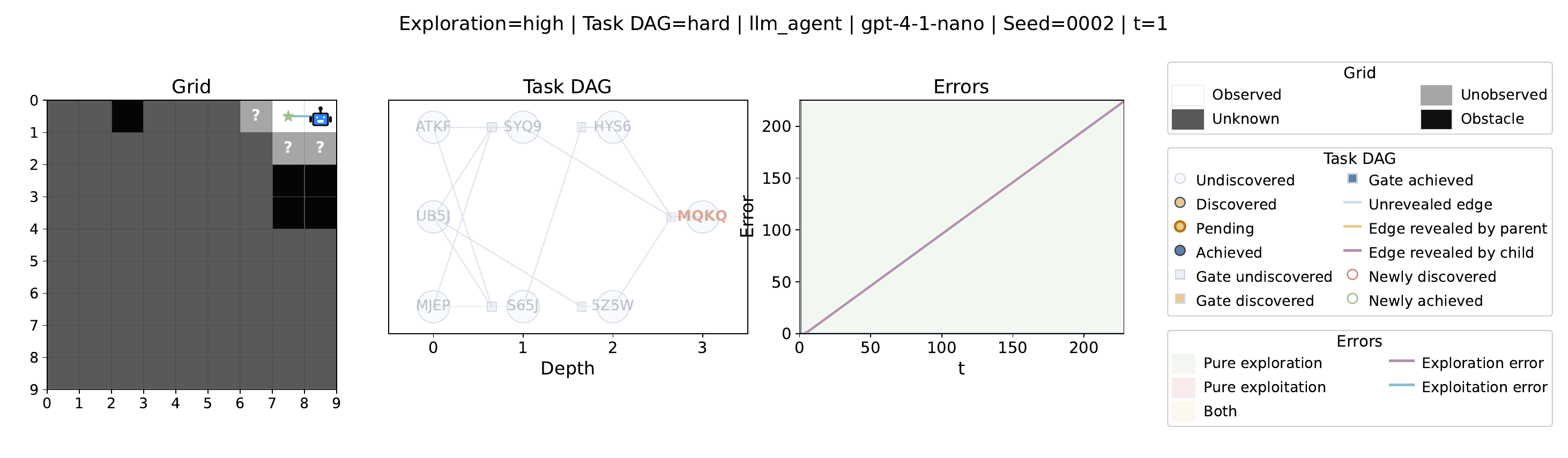}
        \caption{\currentmodel, $t = 1$}
        \end{subfigure}
        \begin{subfigure}[t]{\framewidth}
        \centering
        \includegraphics[width=\textwidth,trim={0 0 0 33pt},clip]{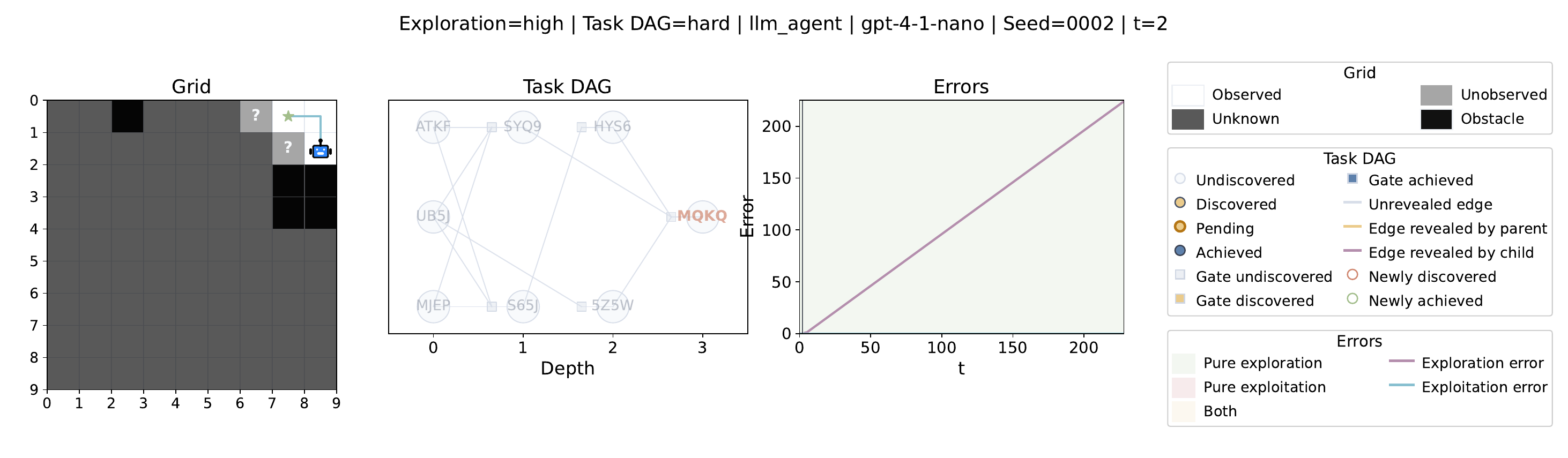}
        \caption{\currentmodel, $t = 2$}
        \end{subfigure}
        \begin{subfigure}[t]{\framewidth}
        \centering
        \includegraphics[width=\textwidth,trim={0 0 0 33pt},clip]{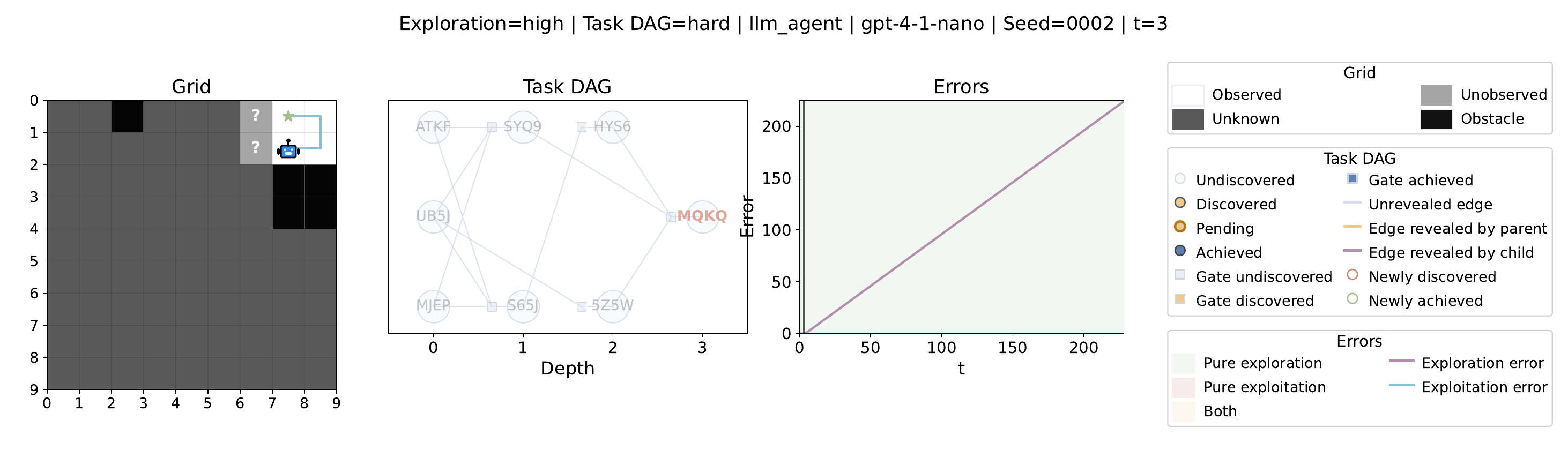}
        \caption{\currentmodel, $t = 3$}
        \end{subfigure}
        \begin{subfigure}[t]{\framewidth}
        \centering
        \includegraphics[width=\textwidth,trim={0 0 0 33pt},clip]{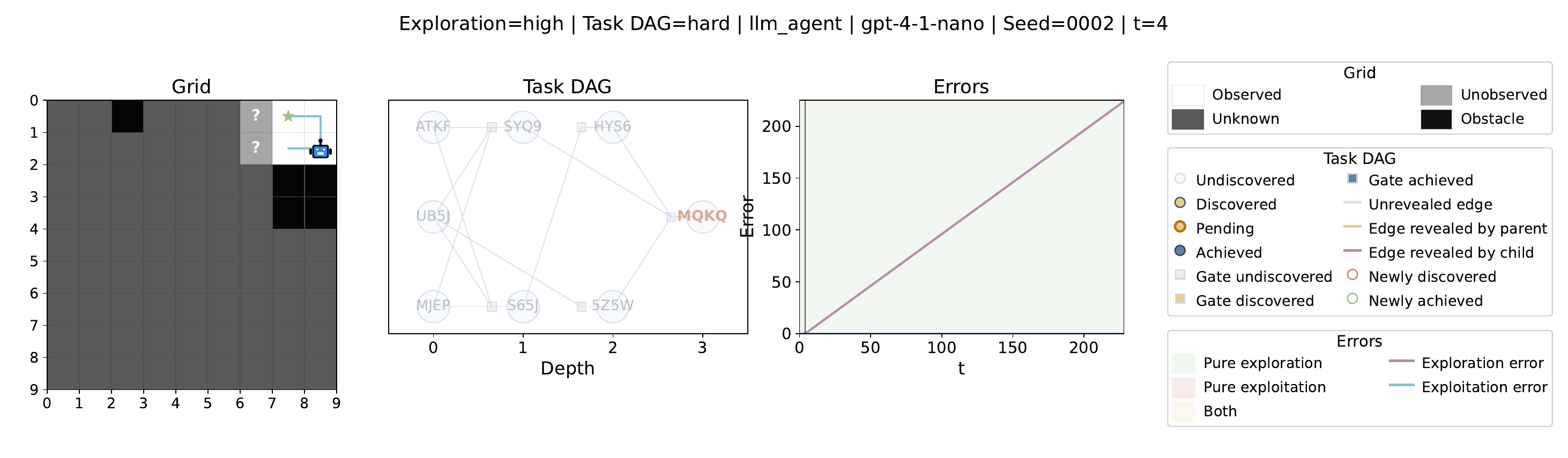}
        \caption{\currentmodel, $t = 4$}
        \end{subfigure}
    \end{center}
    \caption{Results of action trajectories and metric values over each timestep for~\currentmodel.}
\end{figure}

\begin{figure}[ht]
    \begin{center}
        \begin{subfigure}[t]{\framewidth}
        \centering
        \includegraphics[width=\textwidth,trim={0 0 0 33pt},clip]{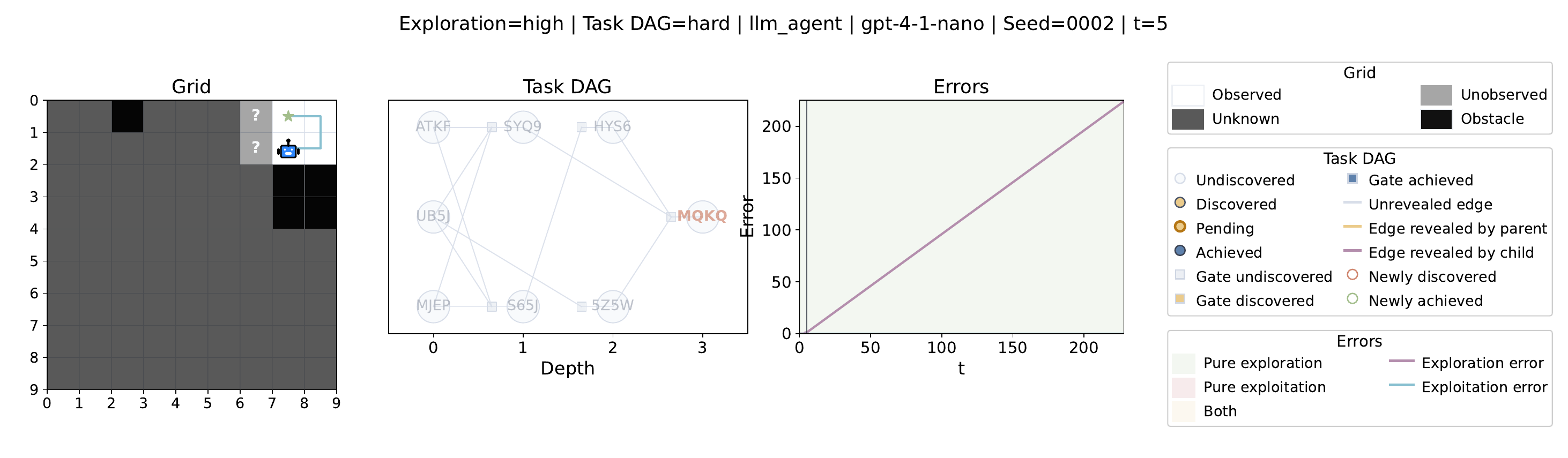}
        \caption{\currentmodel, $t = 5$}
        \end{subfigure}
        \begin{subfigure}[t]{\framewidth}
        \centering
        \includegraphics[width=\textwidth,trim={0 0 0 33pt},clip]{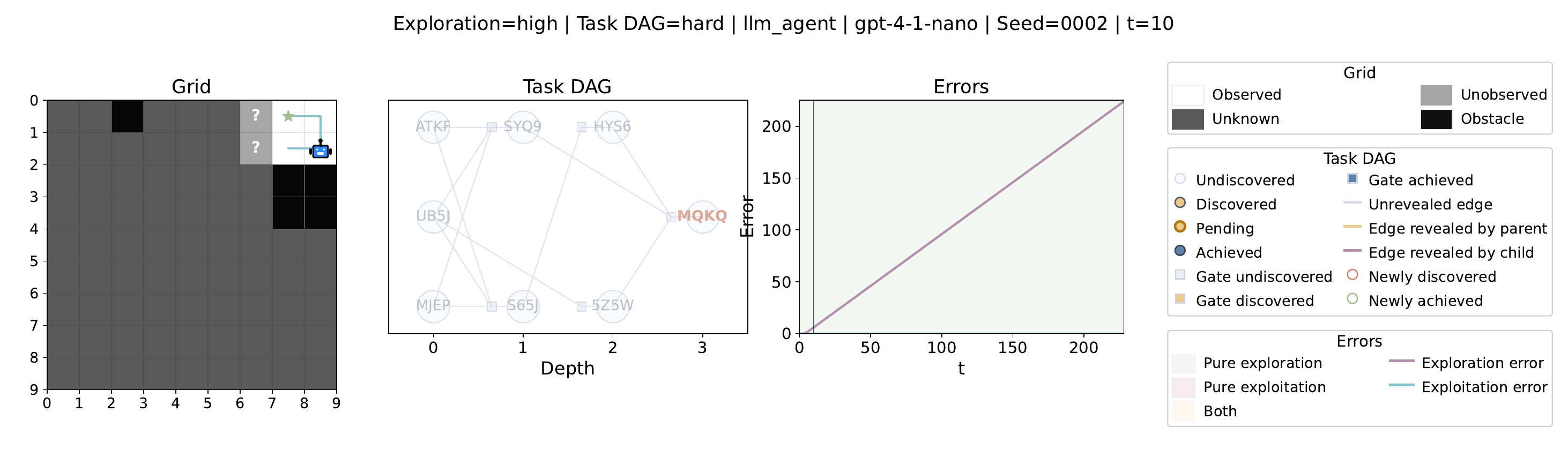}
        \caption{\currentmodel, $t = 10$}
        \end{subfigure}
        \begin{subfigure}[t]{\framewidth}
        \centering
        \includegraphics[width=\textwidth,trim={0 0 0 33pt},clip]{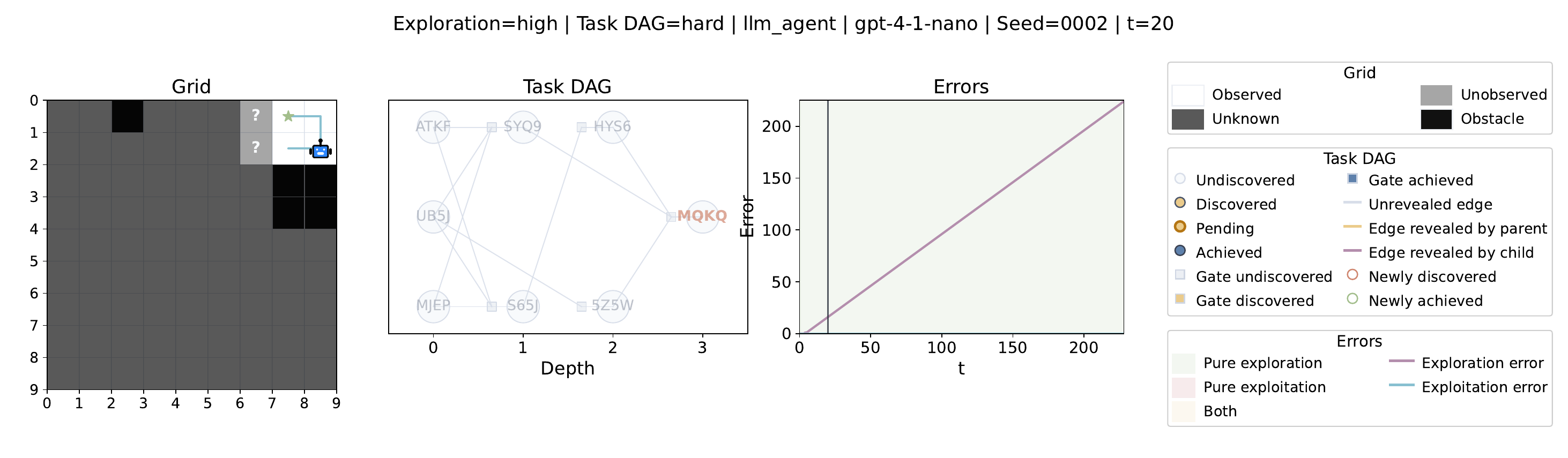}
        \caption{\currentmodel, $t = 20$}
        \end{subfigure}
        \begin{subfigure}[t]{\framewidth}
        \centering
        \includegraphics[width=\textwidth,trim={0 0 0 33pt},clip]{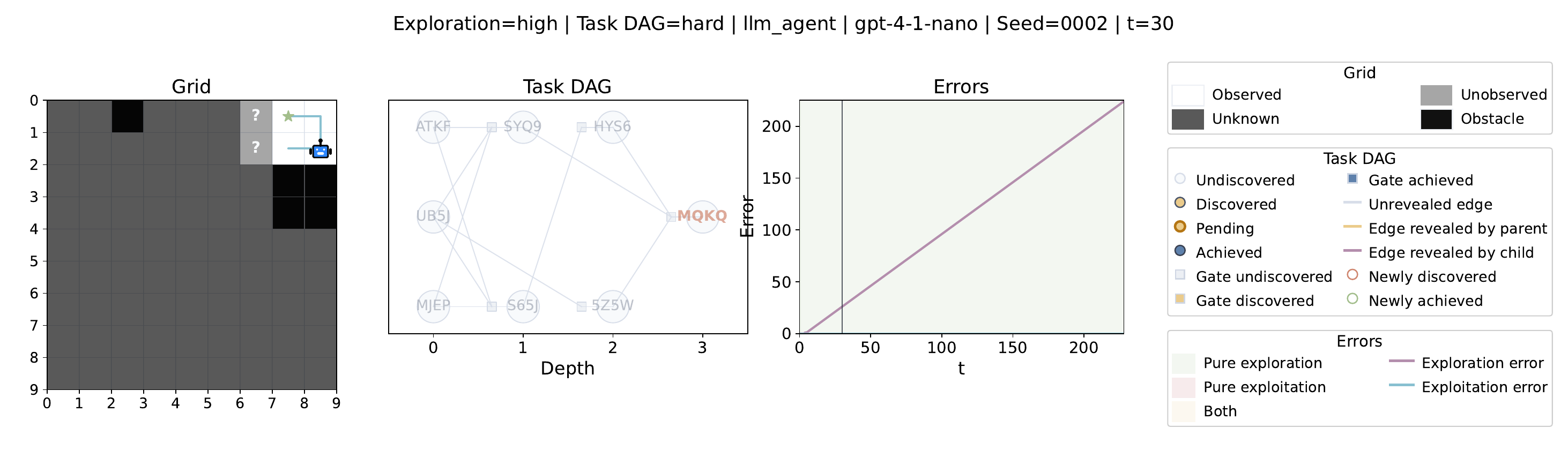}
        \caption{\currentmodel, $t = 30$}
        \end{subfigure}
        \begin{subfigure}[t]{\framewidth}
        \centering
        \includegraphics[width=\textwidth,trim={0 0 0 33pt},clip]{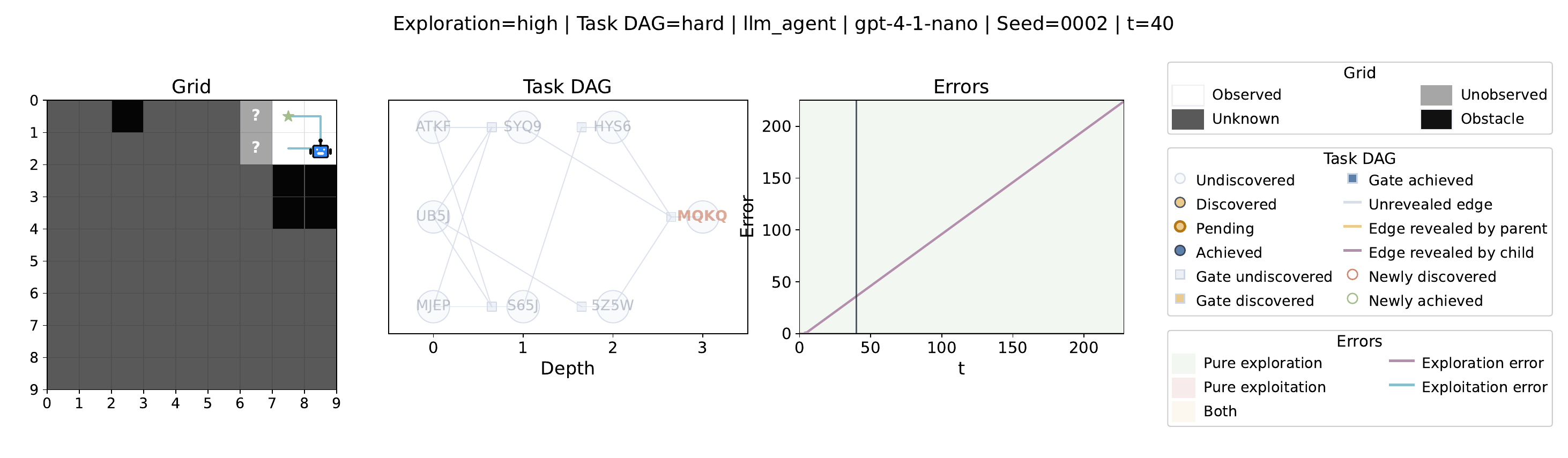}
        \caption{\currentmodel, $t = 40$}
        \end{subfigure}
    \end{center}
    \caption{Results of action trajectories and metric values over each timestep for~\currentmodel.}
\end{figure}

\begin{figure}[ht]
    \begin{center}
        \begin{subfigure}[t]{\framewidth}
        \centering
        \includegraphics[width=\textwidth,trim={0 0 0 33pt},clip]{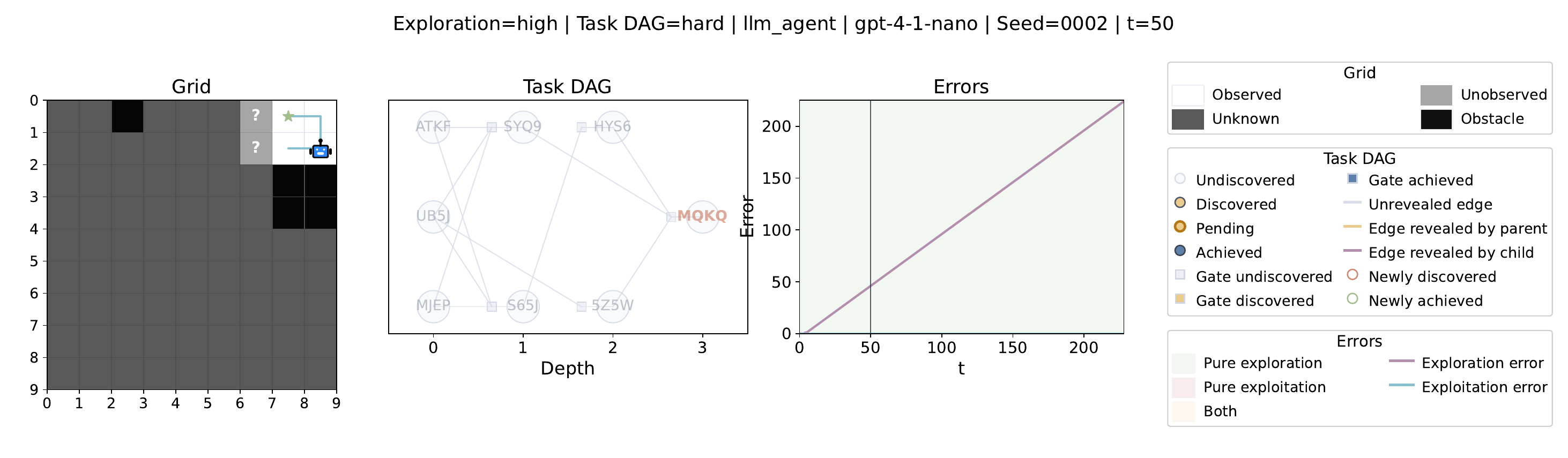}
        \caption{\currentmodel, $t = 50$}
        \end{subfigure}
        \begin{subfigure}[t]{\framewidth}
        \centering
        \includegraphics[width=\textwidth,trim={0 0 0 33pt},clip]{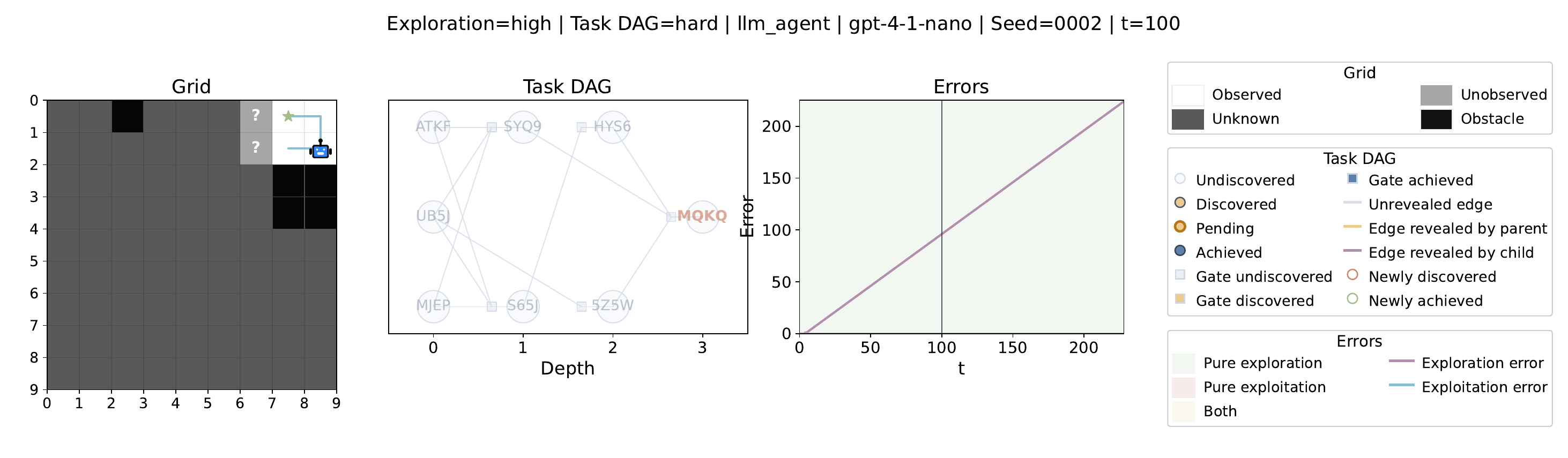}
        \caption{\currentmodel, $t = 100$}
        \end{subfigure}
        \begin{subfigure}[t]{\framewidth}
        \centering
        \includegraphics[width=\textwidth,trim={0 0 0 33pt},clip]{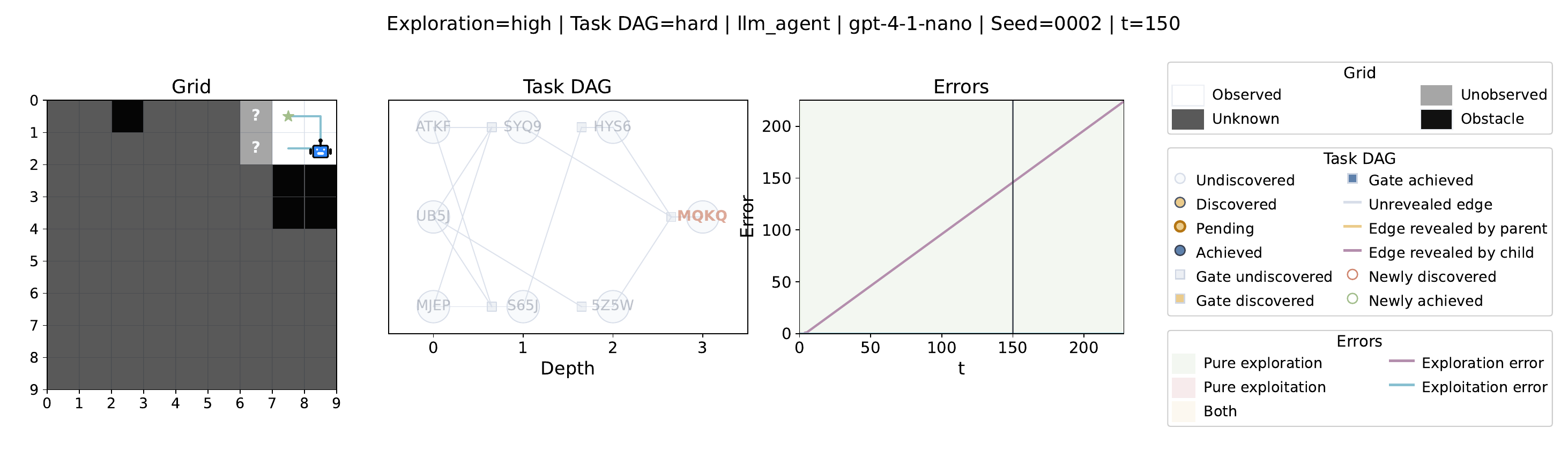}
        \caption{\currentmodel, $t = 150$}
        \end{subfigure}
        \begin{subfigure}[t]{\framewidth}
        \centering
        \includegraphics[width=\textwidth,trim={0 0 0 33pt},clip]{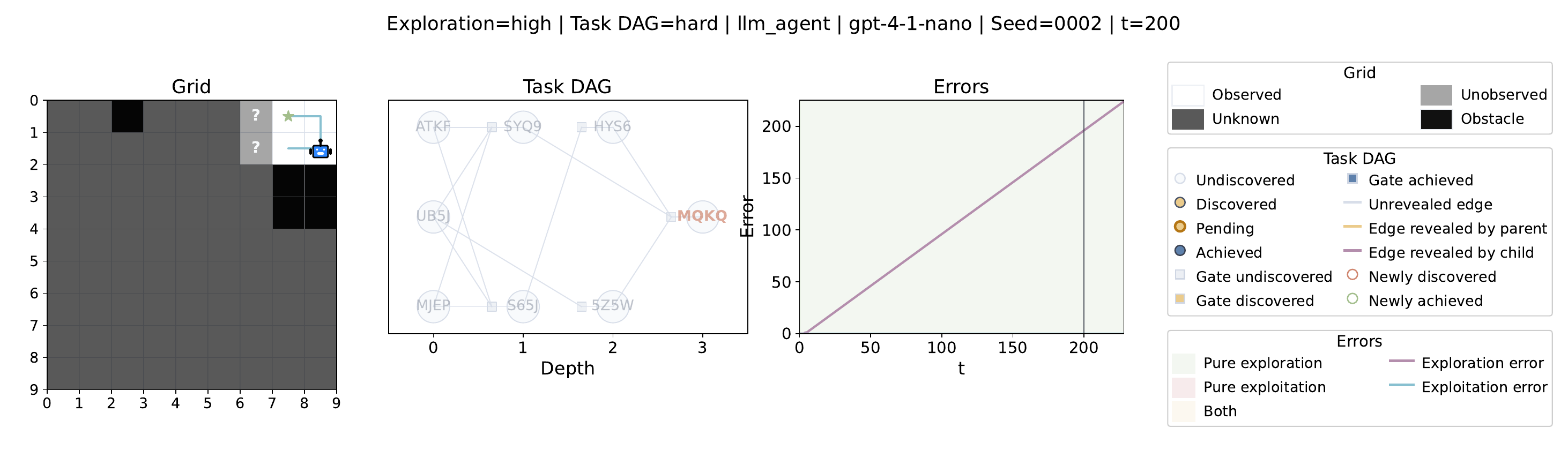}
        \caption{\currentmodel, $t = 200$}
        \end{subfigure}
        \begin{subfigure}[t]{\framewidth}
        \centering
        \includegraphics[width=\textwidth,trim={0 0 0 33pt},clip]{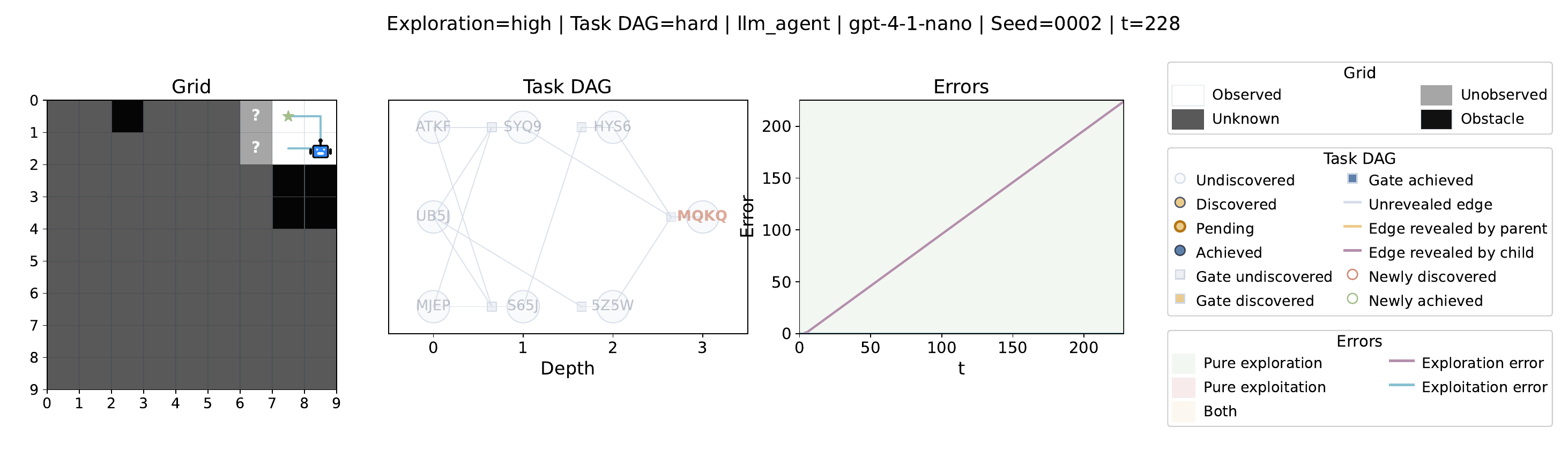}
        \caption{\currentmodel, $t = 228$}
        \end{subfigure}
    \end{center}
    \caption{Results of action trajectories and metric values over each timestep for~\currentmodel.}
\end{figure}

\clearpage
\renewcommand{\currentmodel}{\gptfouronemini}

\begin{figure}[ht]
    \begin{center}
        \begin{subfigure}[t]{\framewidth}
        \centering
        \includegraphics[width=\textwidth,trim={0 0 0 33pt},clip]{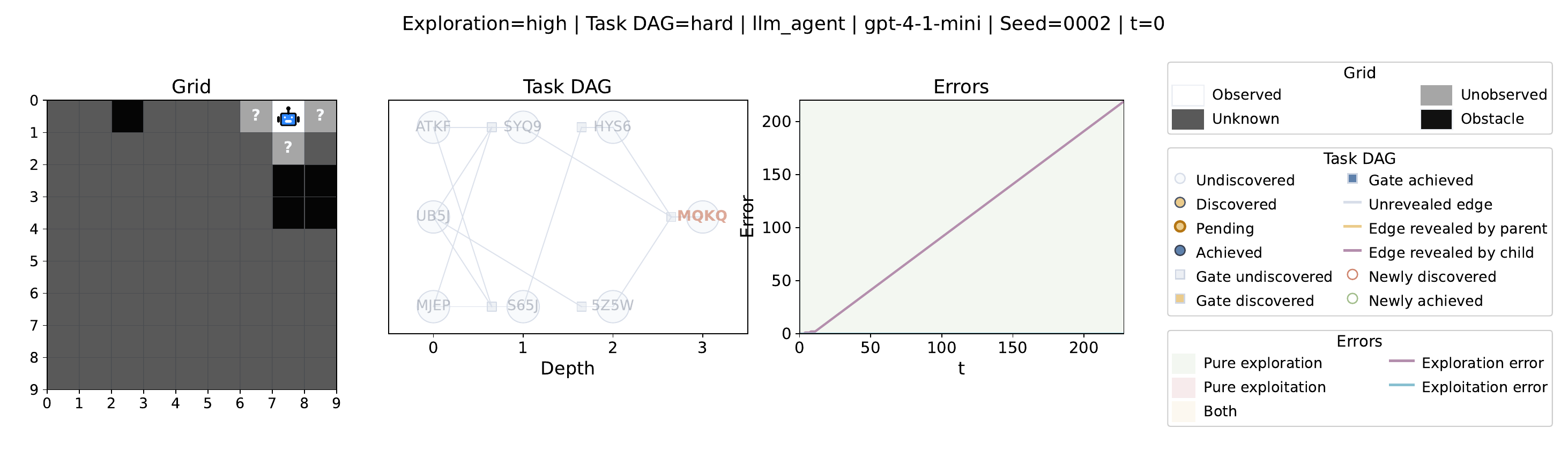}
        \caption{\currentmodel, $t = 0$}
        \end{subfigure}
        \begin{subfigure}[t]{\framewidth}
        \centering
        \includegraphics[width=\textwidth,trim={0 0 0 33pt},clip]{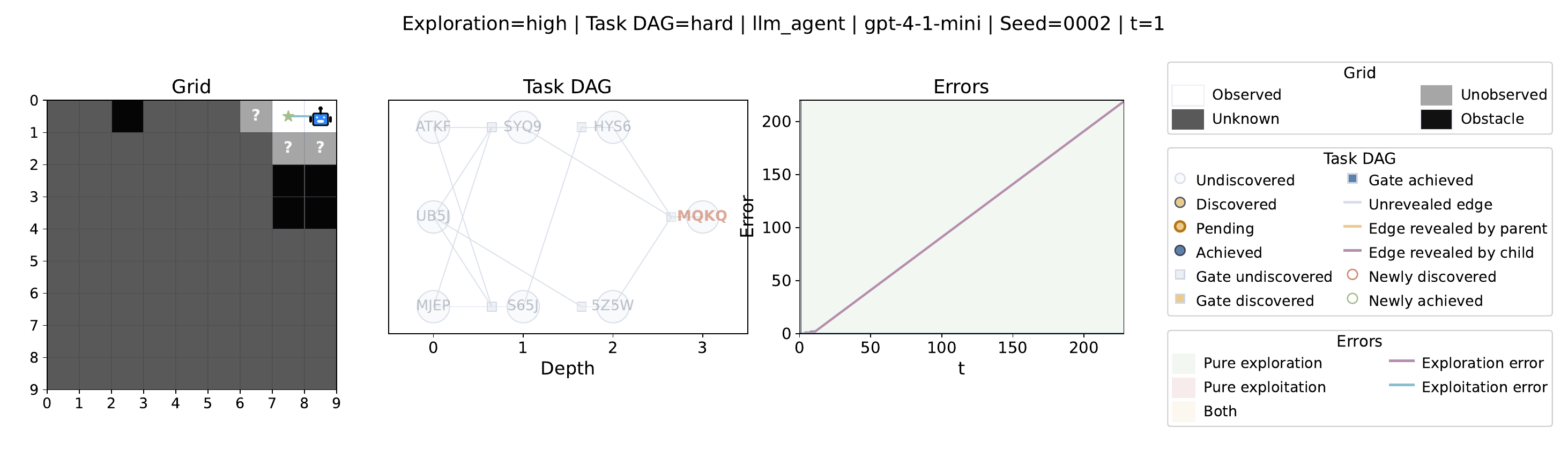}
        \caption{\currentmodel, $t = 1$}
        \end{subfigure}
        \begin{subfigure}[t]{\framewidth}
        \centering
        \includegraphics[width=\textwidth,trim={0 0 0 33pt},clip]{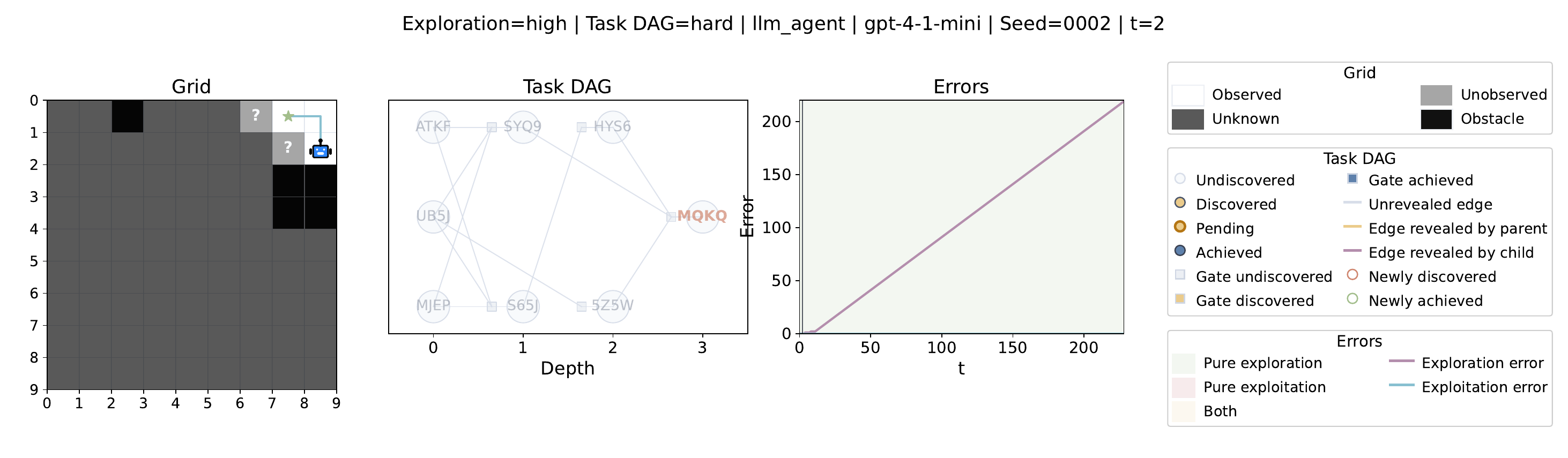}
        \caption{\currentmodel, $t = 2$}
        \end{subfigure}
        \begin{subfigure}[t]{\framewidth}
        \centering
        \includegraphics[width=\textwidth,trim={0 0 0 33pt},clip]{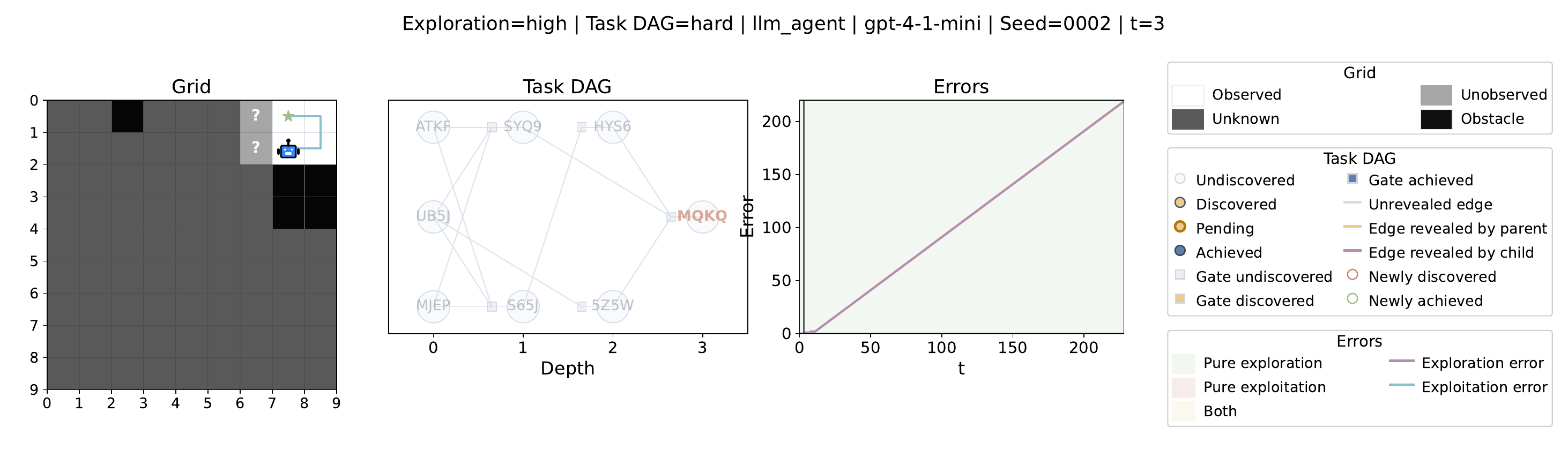}
        \caption{\currentmodel, $t = 3$}
        \end{subfigure}
        \begin{subfigure}[t]{\framewidth}
        \centering
        \includegraphics[width=\textwidth,trim={0 0 0 33pt},clip]{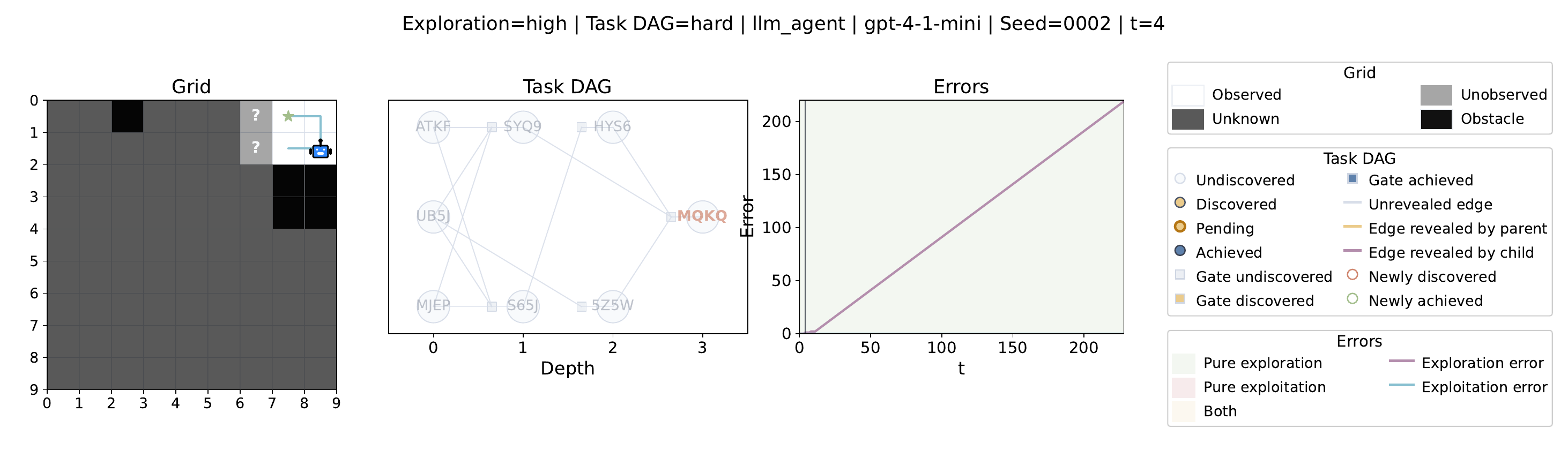}
        \caption{\currentmodel, $t = 4$}
        \end{subfigure}
    \end{center}
    \caption{Results of action trajectories and metric values over each timestep for~\currentmodel.}
\end{figure}

\begin{figure}[ht]
    \begin{center}
        \begin{subfigure}[t]{\framewidth}
        \centering
        \includegraphics[width=\textwidth,trim={0 0 0 33pt},clip]{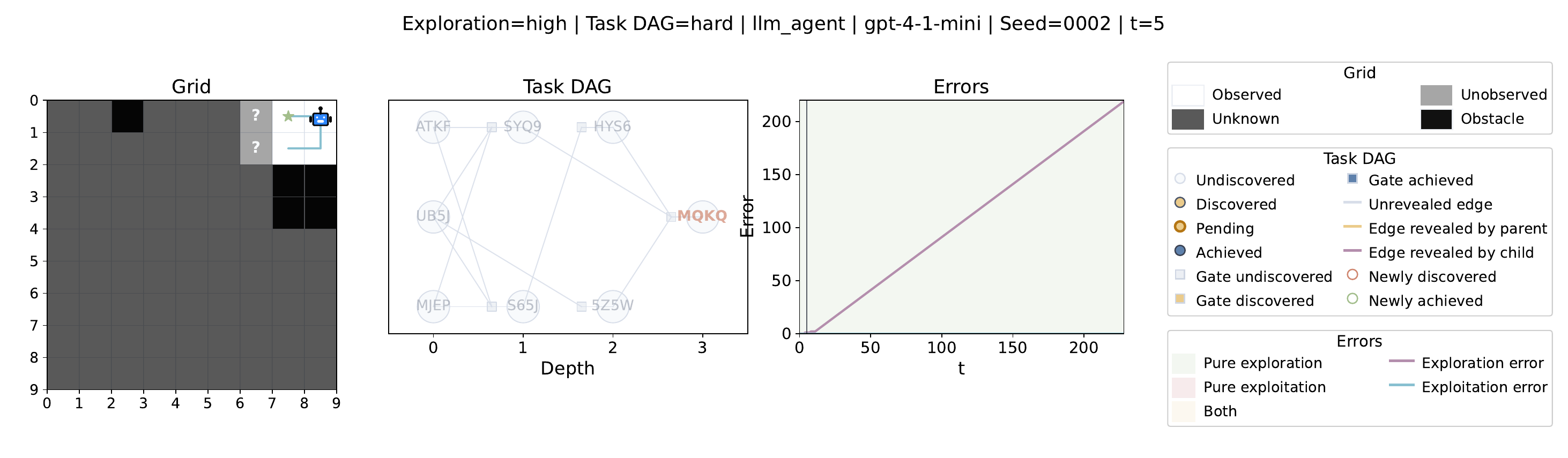}
        \caption{\currentmodel, $t = 5$}
        \end{subfigure}
        \begin{subfigure}[t]{\framewidth}
        \centering
        \includegraphics[width=\textwidth,trim={0 0 0 33pt},clip]{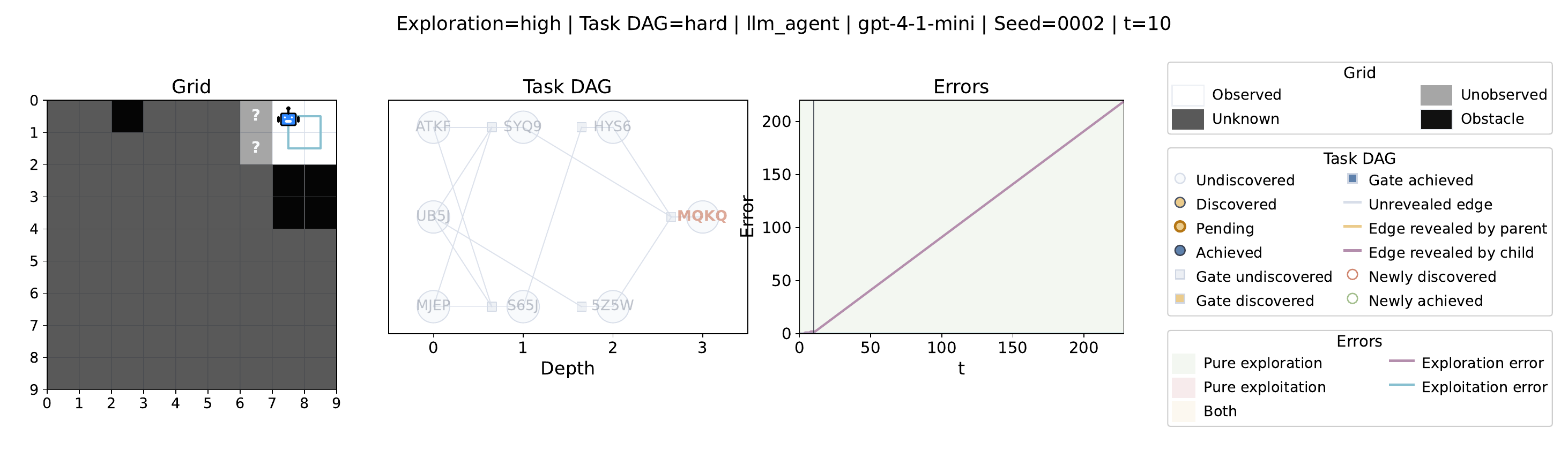}
        \caption{\currentmodel, $t = 10$}
        \end{subfigure}
        \begin{subfigure}[t]{\framewidth}
        \centering
        \includegraphics[width=\textwidth,trim={0 0 0 33pt},clip]{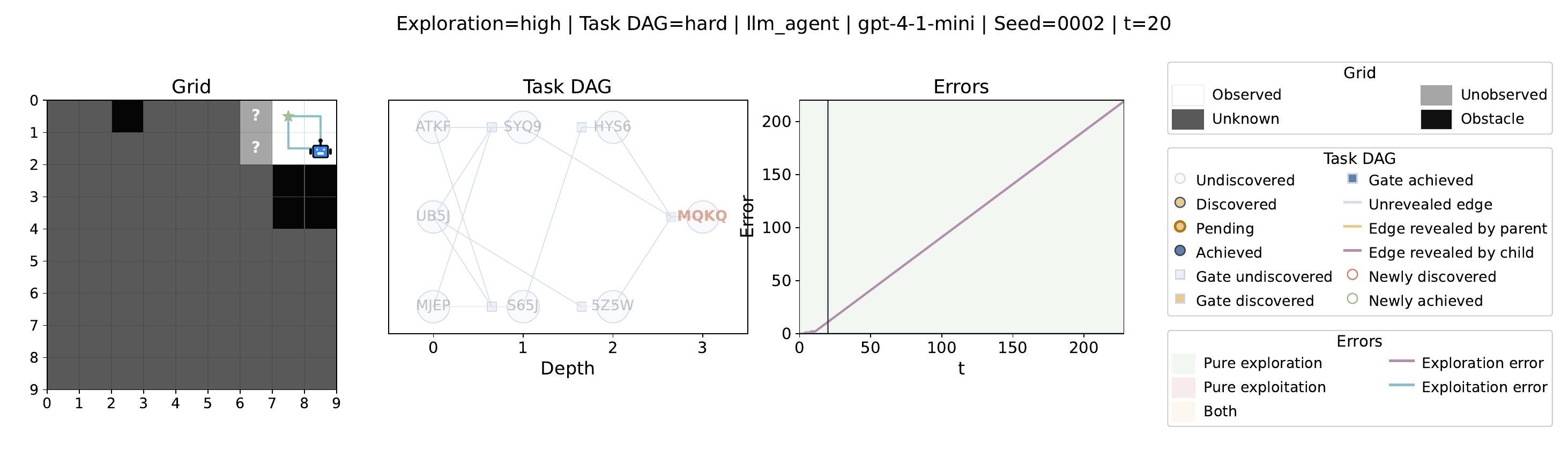}
        \caption{\currentmodel, $t = 20$}
        \end{subfigure}
        \begin{subfigure}[t]{\framewidth}
        \centering
        \includegraphics[width=\textwidth,trim={0 0 0 33pt},clip]{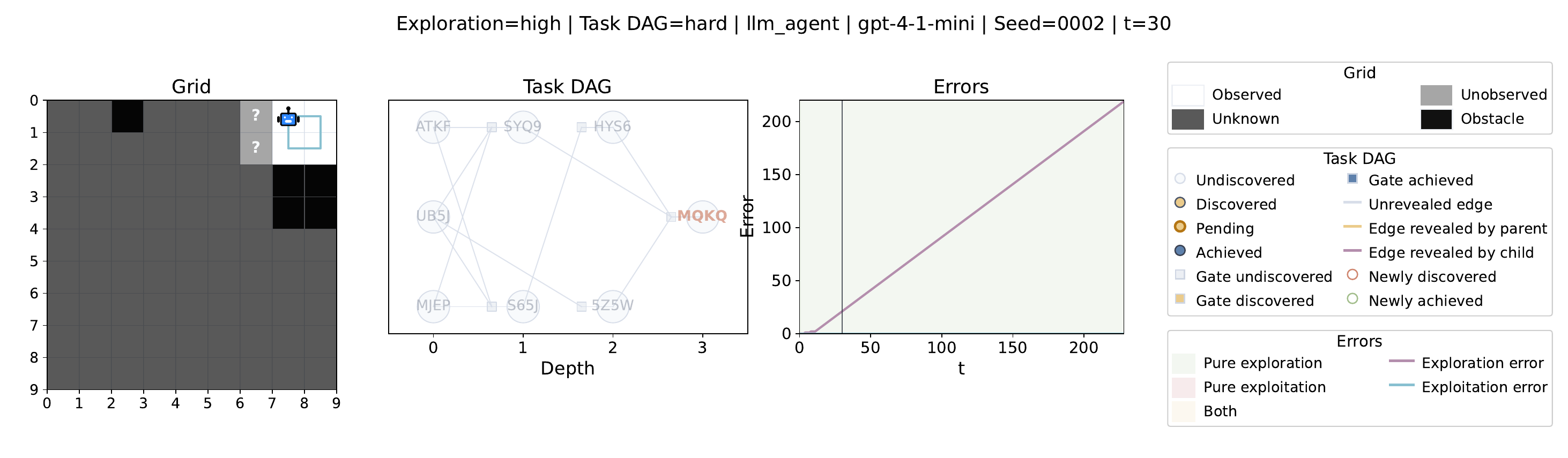}
        \caption{\currentmodel, $t = 30$}
        \end{subfigure}
        \begin{subfigure}[t]{\framewidth}
        \centering
        \includegraphics[width=\textwidth,trim={0 0 0 33pt},clip]{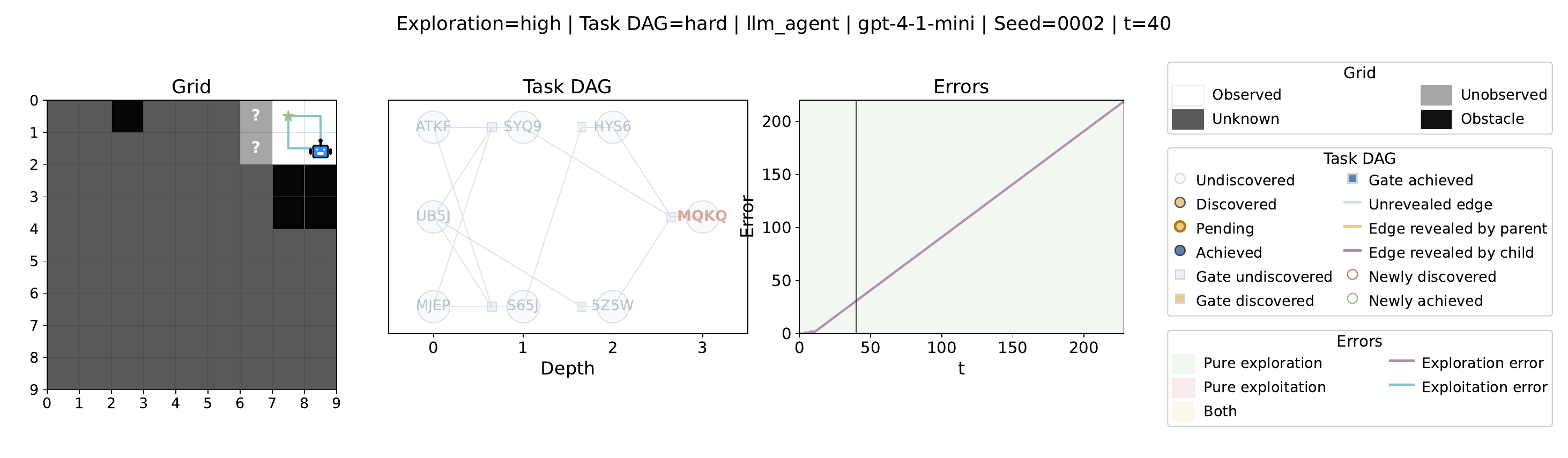}
        \caption{\currentmodel, $t = 40$}
        \end{subfigure}
    \end{center}
    \caption{Results of action trajectories and metric values over each timestep for~\currentmodel.}
\end{figure}

\begin{figure}[ht]
    \begin{center}
        \begin{subfigure}[t]{\framewidth}
        \centering
        \includegraphics[width=\textwidth,trim={0 0 0 33pt},clip]{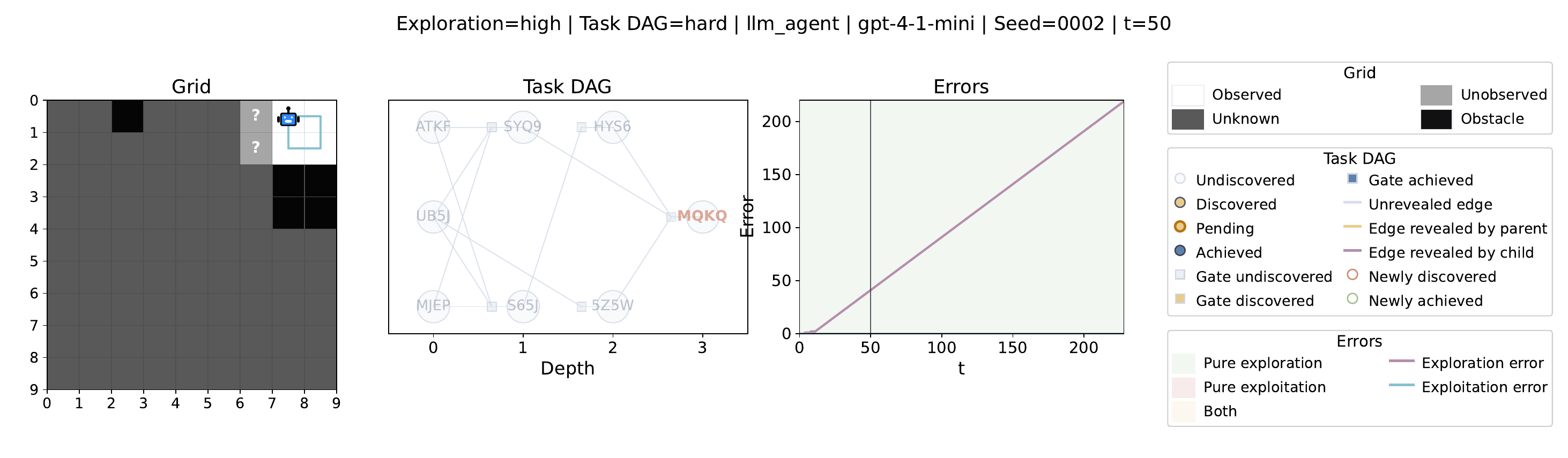}
        \caption{\currentmodel, $t = 50$}
        \end{subfigure}
        \begin{subfigure}[t]{\framewidth}
        \centering
        \includegraphics[width=\textwidth,trim={0 0 0 33pt},clip]{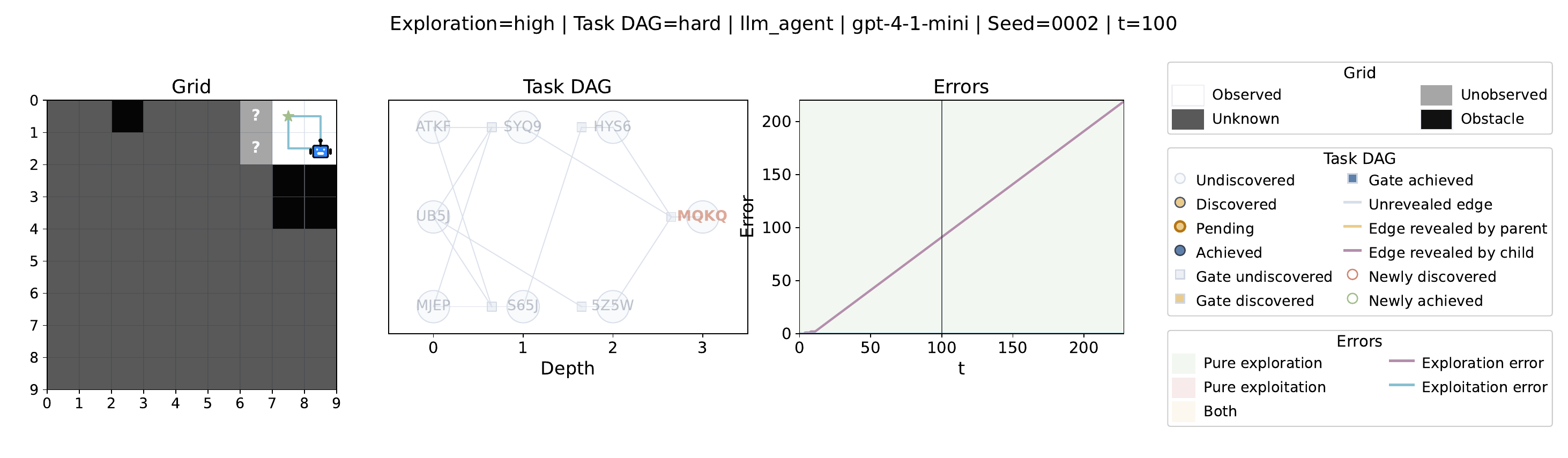}
        \caption{\currentmodel, $t = 100$}
        \end{subfigure}
        \begin{subfigure}[t]{\framewidth}
        \centering
        \includegraphics[width=\textwidth,trim={0 0 0 33pt},clip]{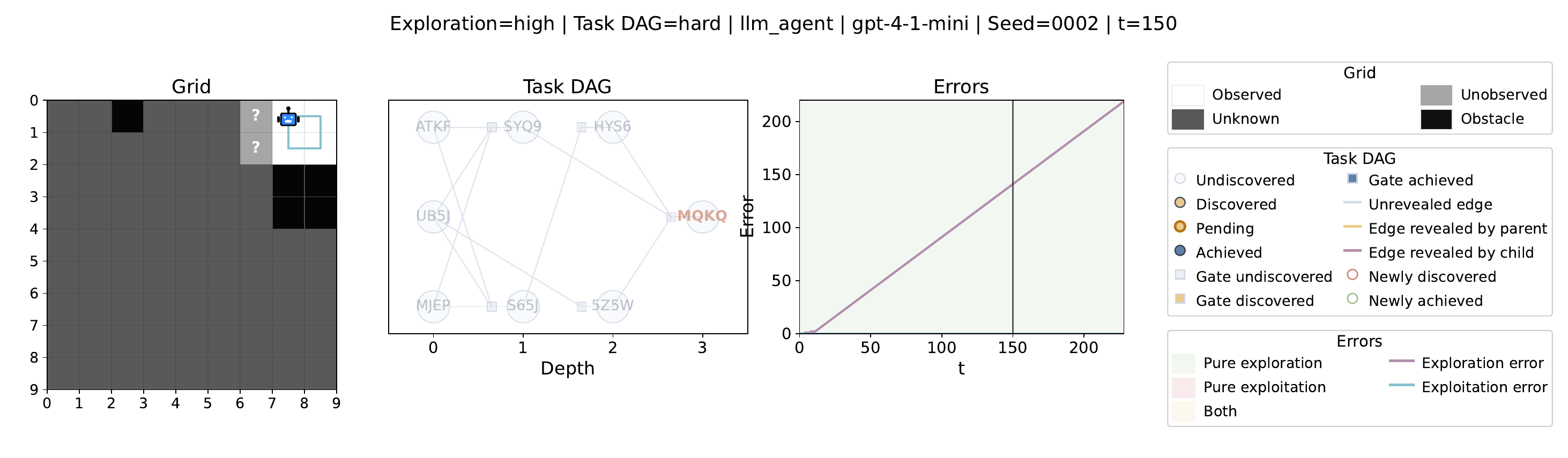}
        \caption{\currentmodel, $t = 150$}
        \end{subfigure}
        \begin{subfigure}[t]{\framewidth}
        \centering
        \includegraphics[width=\textwidth,trim={0 0 0 33pt},clip]{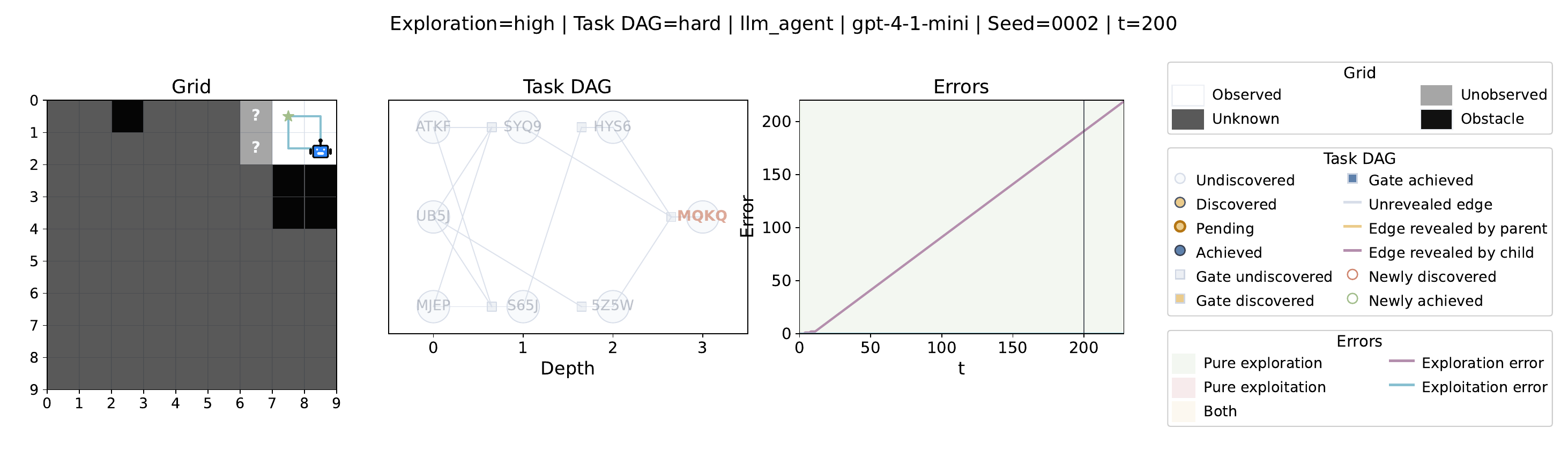}
        \caption{\currentmodel, $t = 200$}
        \end{subfigure}
        \begin{subfigure}[t]{\framewidth}
        \centering
        \includegraphics[width=\textwidth,trim={0 0 0 33pt},clip]{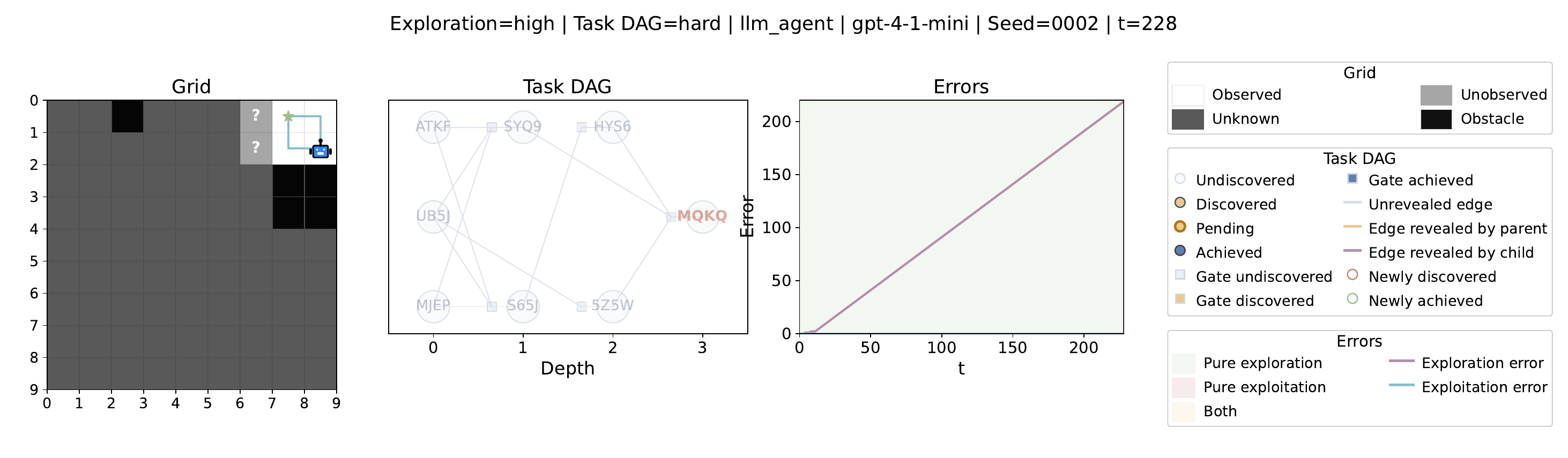}
        \caption{\currentmodel, $t = 228$}
        \end{subfigure}
    \end{center}
    \caption{Results of action trajectories and metric values over each timestep for~\currentmodel.}
\end{figure}

\clearpage
\renewcommand{\currentmodel}{\gptfourone}

\begin{figure}[ht]
    \begin{center}
        \begin{subfigure}[t]{\framewidth}
        \centering
        \includegraphics[width=\textwidth,trim={0 0 0 33pt},clip]{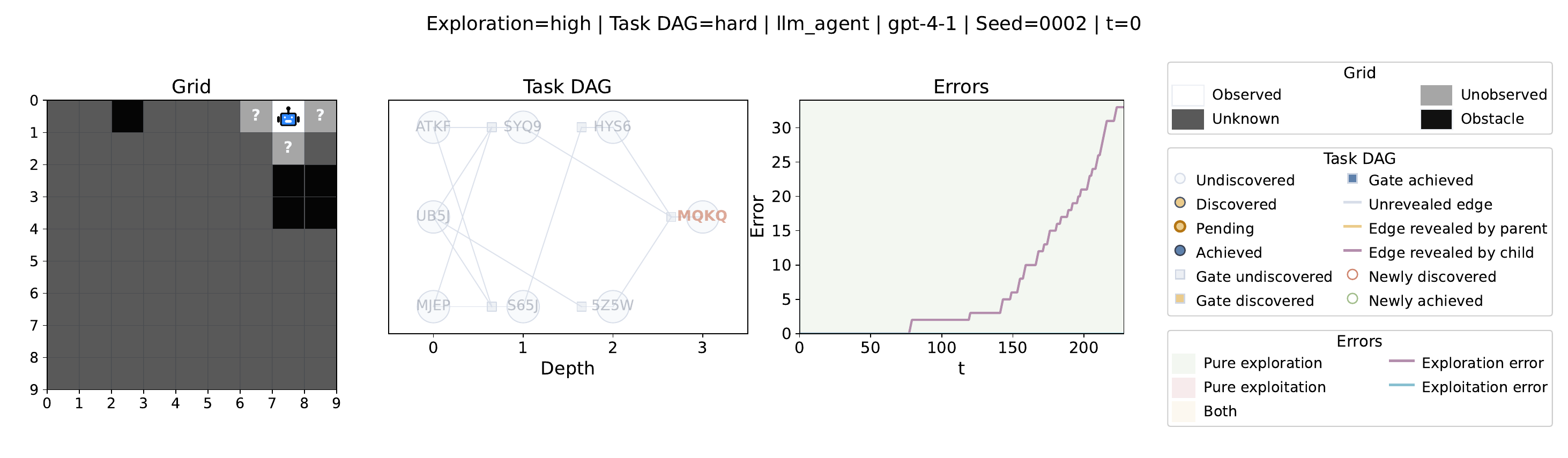}
        \caption{\currentmodel, $t = 0$}
        \end{subfigure}
        \begin{subfigure}[t]{\framewidth}
        \centering
        \includegraphics[width=\textwidth,trim={0 0 0 33pt},clip]{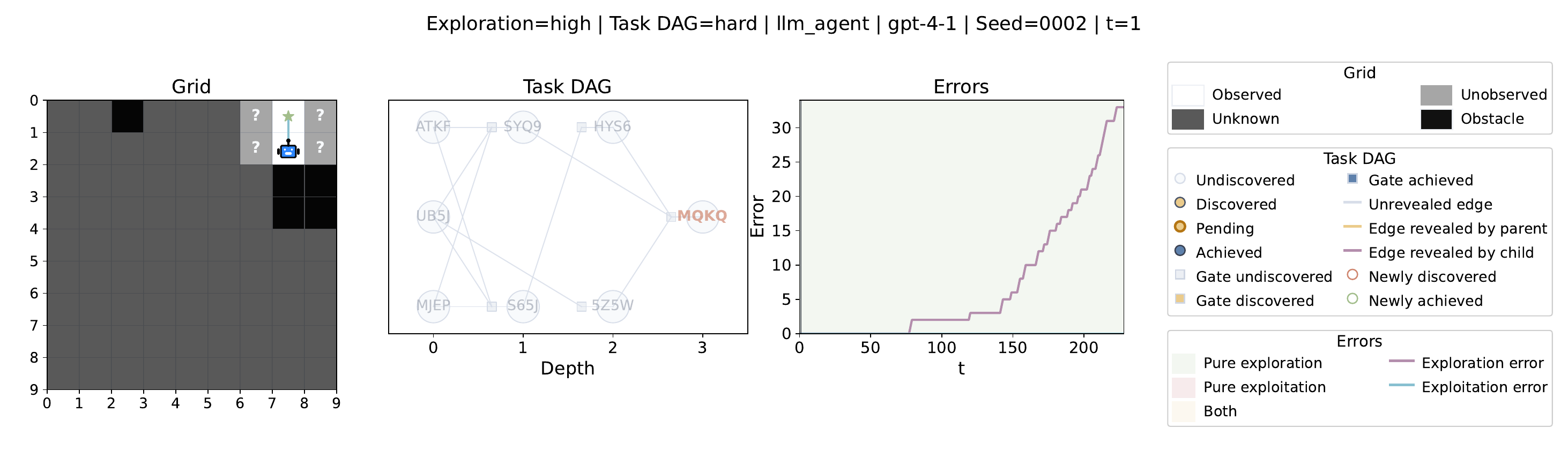}
        \caption{\currentmodel, $t = 1$}
        \end{subfigure}
        \begin{subfigure}[t]{\framewidth}
        \centering
        \includegraphics[width=\textwidth,trim={0 0 0 33pt},clip]{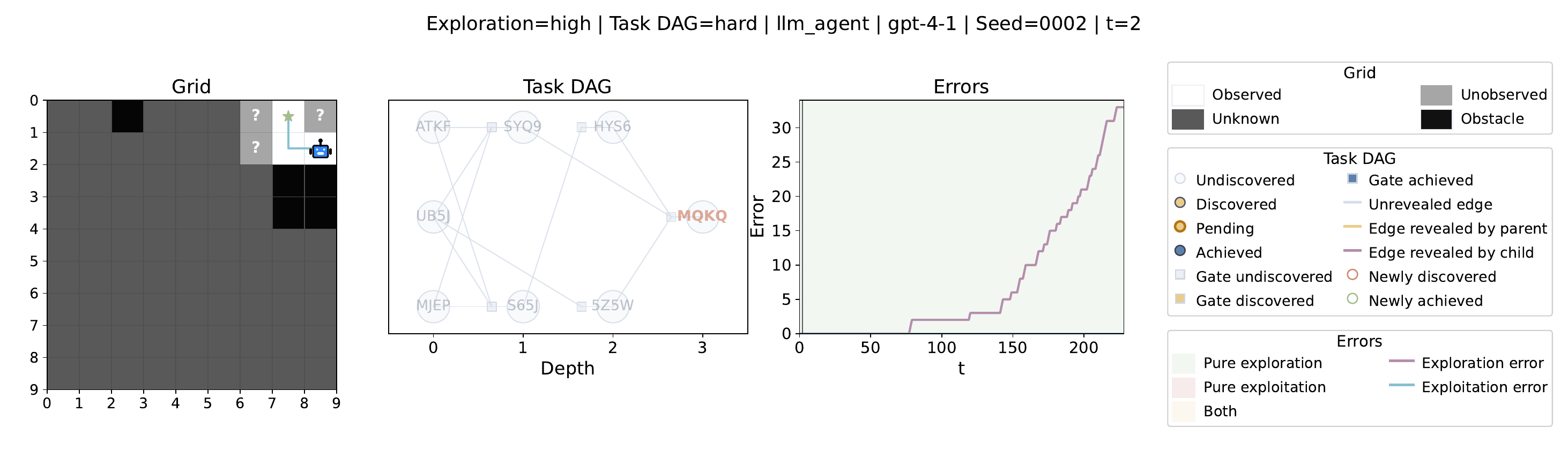}
        \caption{\currentmodel, $t = 2$}
        \end{subfigure}
        \begin{subfigure}[t]{\framewidth}
        \centering
        \includegraphics[width=\textwidth,trim={0 0 0 33pt},clip]{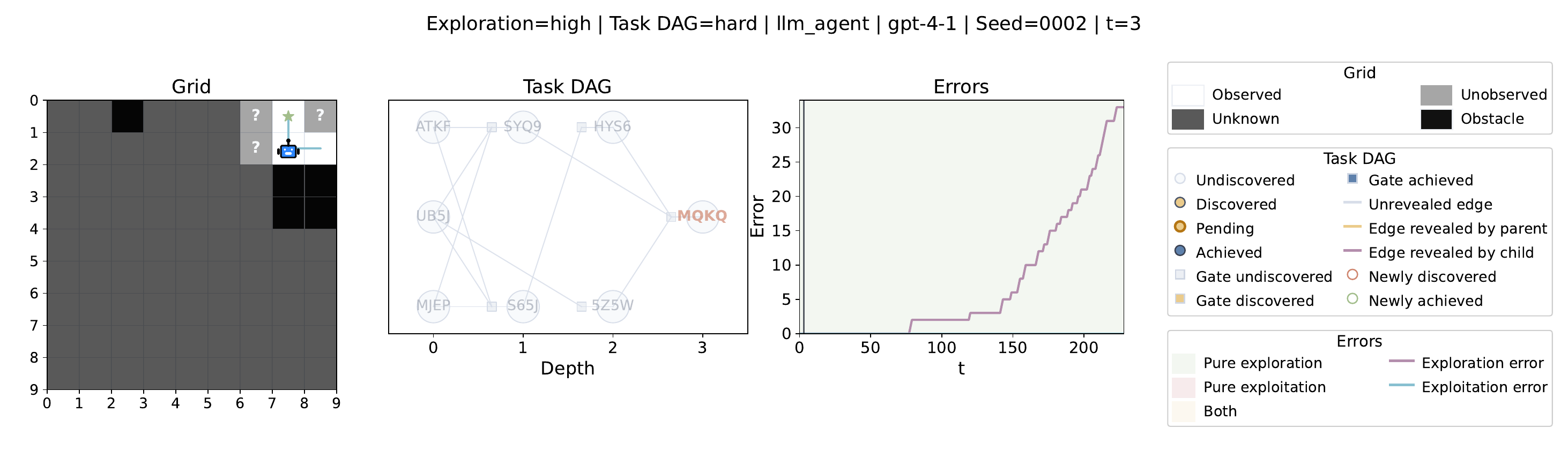}
        \caption{\currentmodel, $t = 3$}
        \end{subfigure}
        \begin{subfigure}[t]{\framewidth}
        \centering
        \includegraphics[width=\textwidth,trim={0 0 0 33pt},clip]{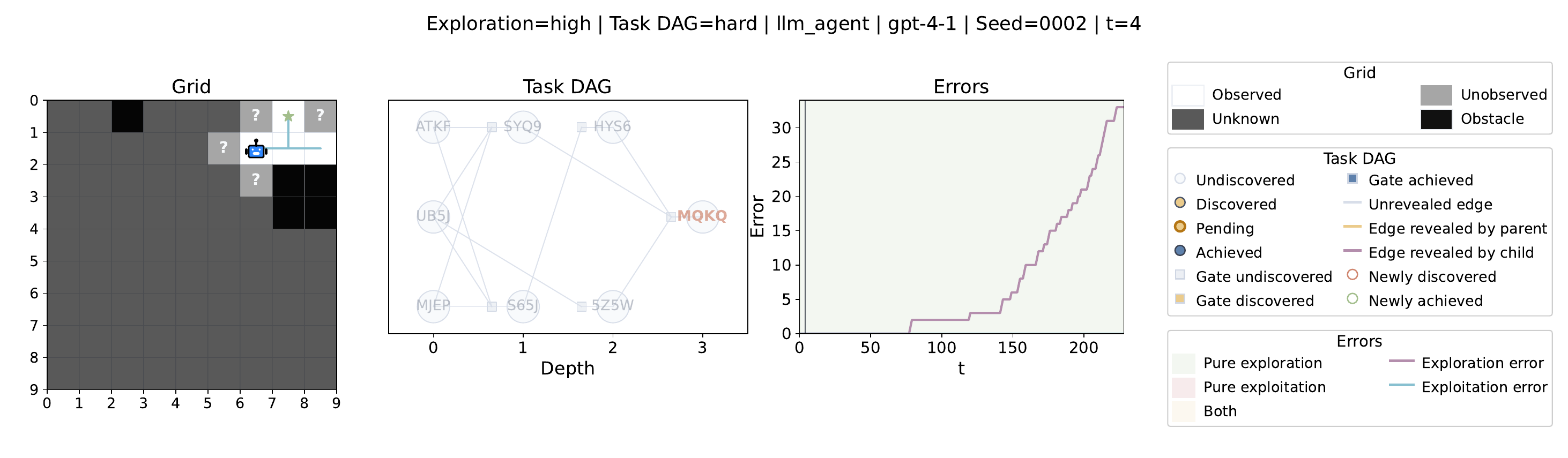}
        \caption{\currentmodel, $t = 4$}
        \end{subfigure}
    \end{center}
    \caption{Results of action trajectories and metric values over each timestep for~\currentmodel.}
\end{figure}

\begin{figure}[ht]
    \begin{center}
        \begin{subfigure}[t]{\framewidth}
        \centering
        \includegraphics[width=\textwidth,trim={0 0 0 33pt},clip]{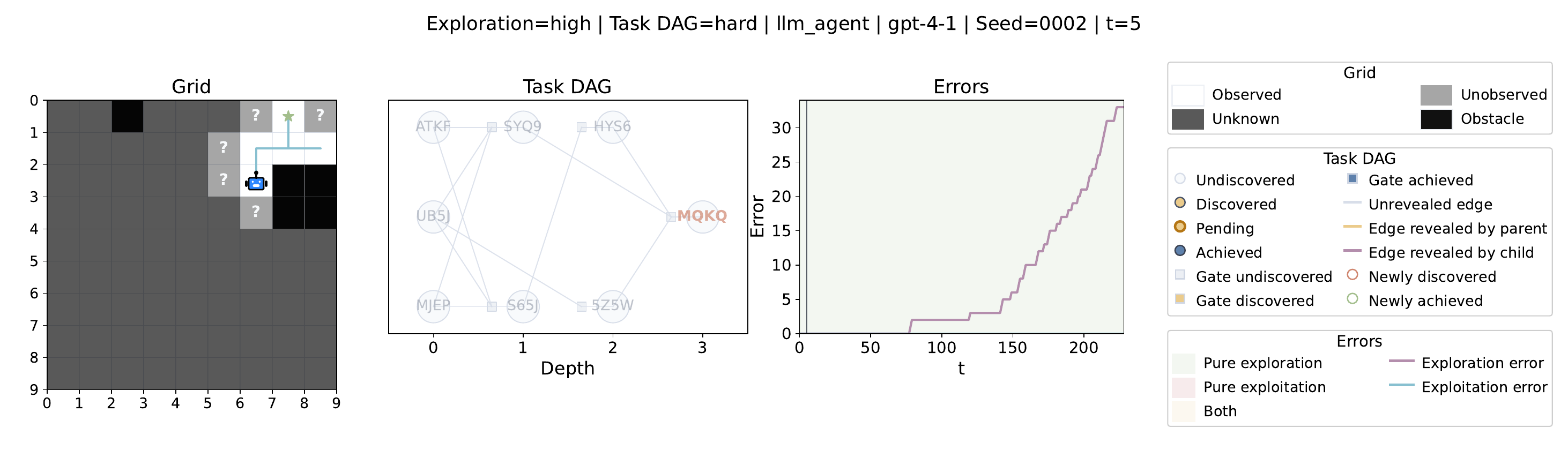}
        \caption{\currentmodel, $t = 5$}
        \end{subfigure}
        \begin{subfigure}[t]{\framewidth}
        \centering
        \includegraphics[width=\textwidth,trim={0 0 0 33pt},clip]{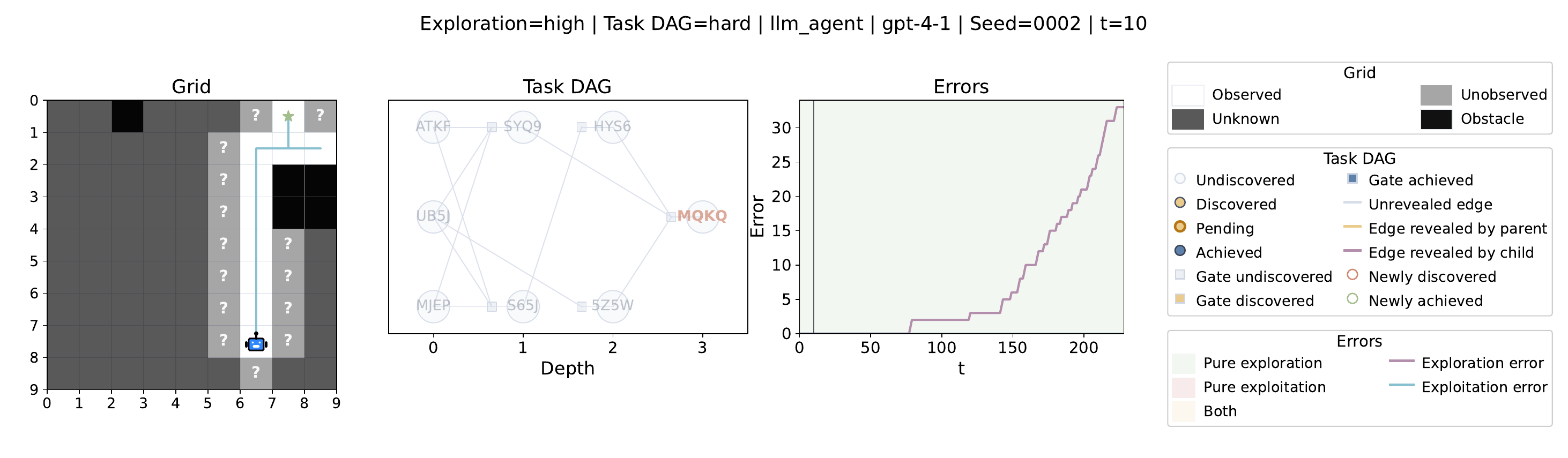}
        \caption{\currentmodel, $t = 10$}
        \end{subfigure}
        \begin{subfigure}[t]{\framewidth}
        \centering
        \includegraphics[width=\textwidth,trim={0 0 0 33pt},clip]{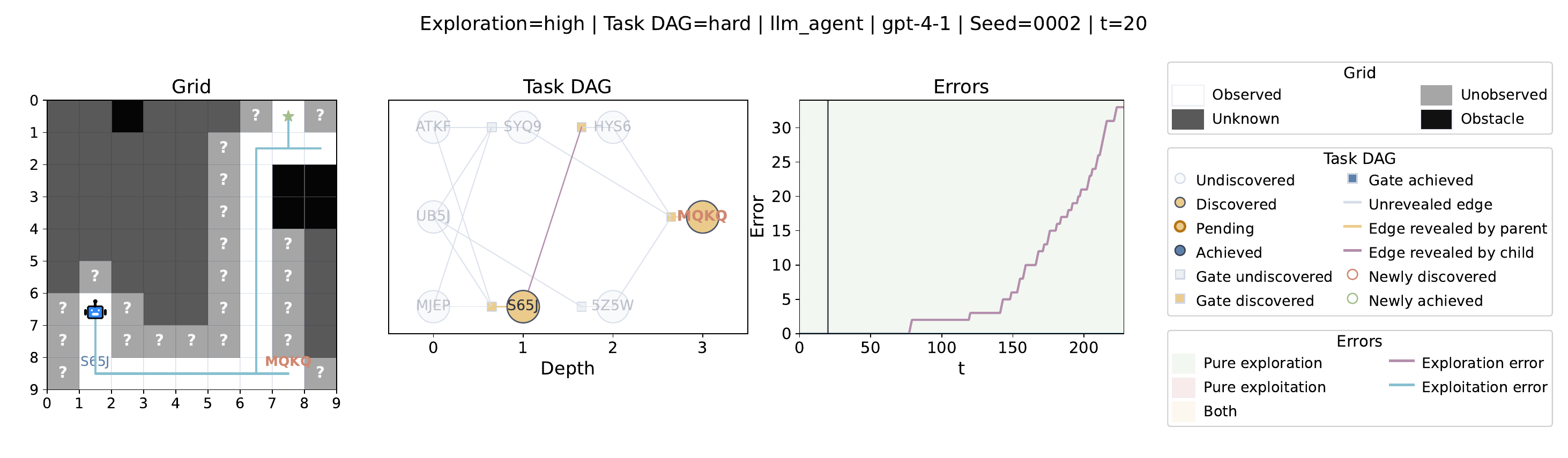}
        \caption{\currentmodel, $t = 20$}
        \end{subfigure}
        \begin{subfigure}[t]{\framewidth}
        \centering
        \includegraphics[width=\textwidth,trim={0 0 0 33pt},clip]{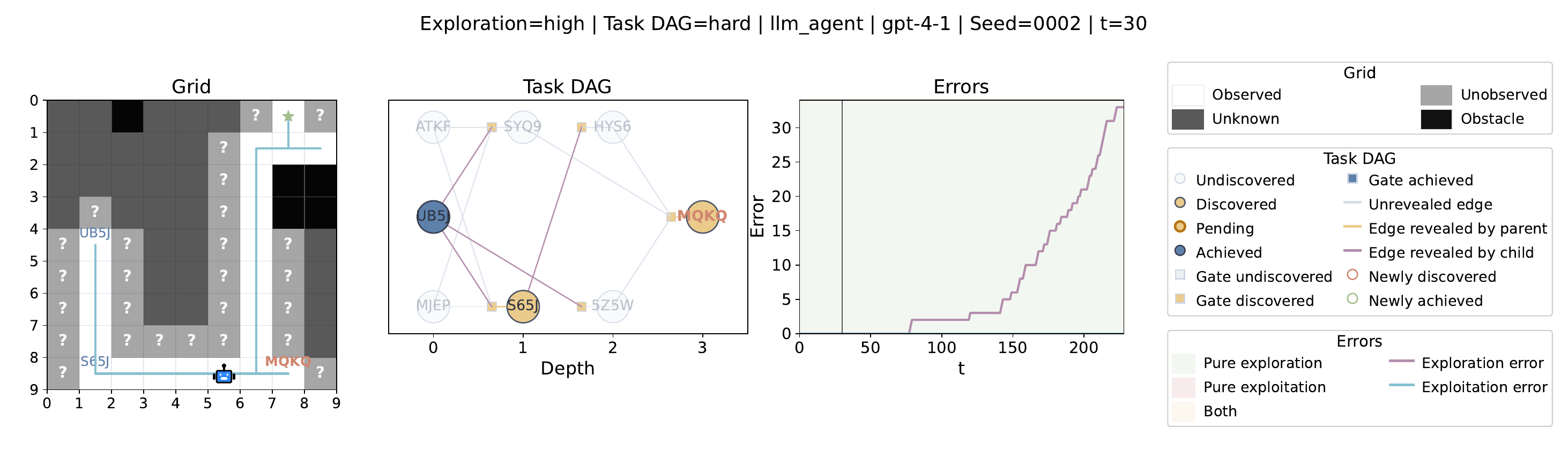}
        \caption{\currentmodel, $t = 30$}
        \end{subfigure}
        \begin{subfigure}[t]{\framewidth}
        \centering
        \includegraphics[width=\textwidth,trim={0 0 0 33pt},clip]{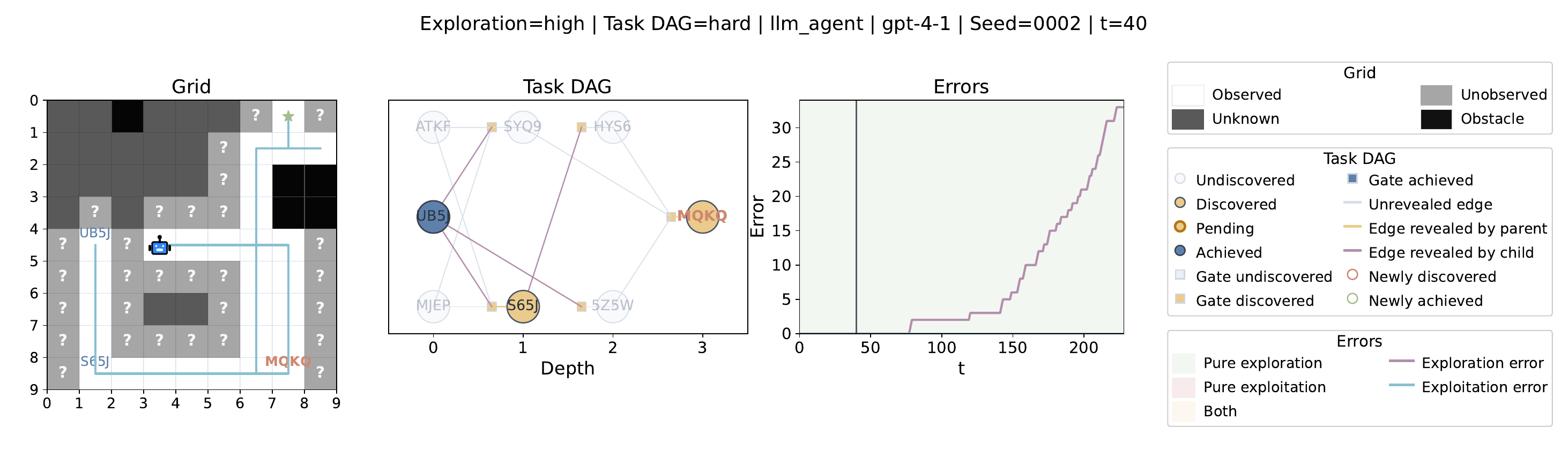}
        \caption{\currentmodel, $t = 40$}
        \end{subfigure}
    \end{center}
    \caption{Results of action trajectories and metric values over each timestep for~\currentmodel.}
\end{figure}

\begin{figure}[ht]
    \begin{center}
        \begin{subfigure}[t]{\framewidth}
        \centering
        \includegraphics[width=\textwidth,trim={0 0 0 33pt},clip]{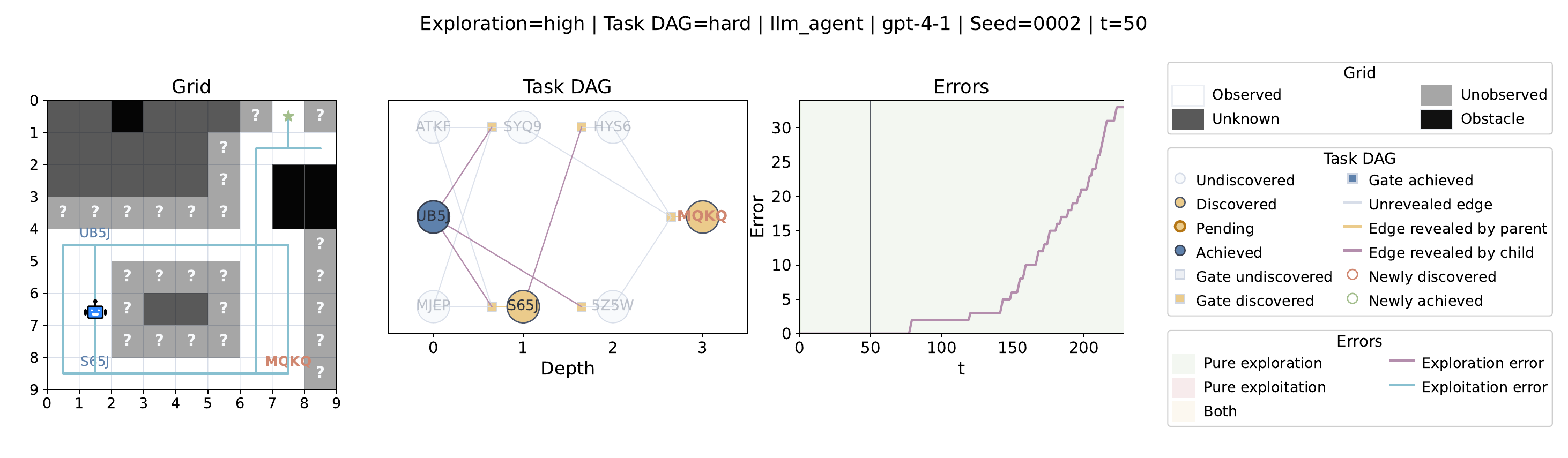}
        \caption{\currentmodel, $t = 50$}
        \end{subfigure}
        \begin{subfigure}[t]{\framewidth}
        \centering
        \includegraphics[width=\textwidth,trim={0 0 0 33pt},clip]{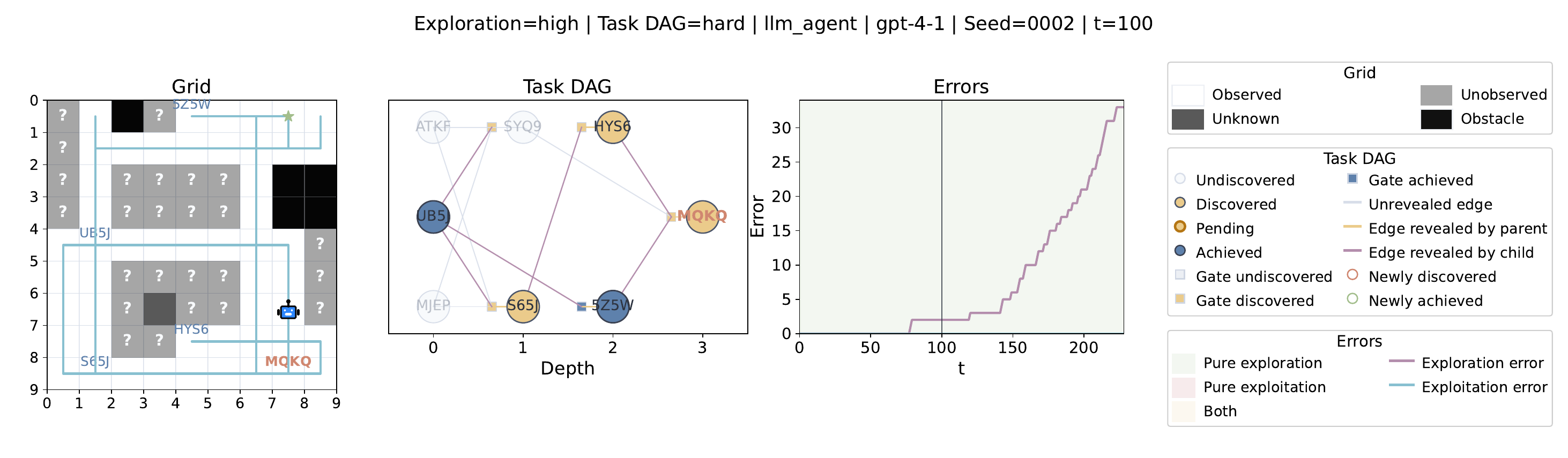}
        \caption{\currentmodel, $t = 100$}
        \end{subfigure}
        \begin{subfigure}[t]{\framewidth}
        \centering
        \includegraphics[width=\textwidth,trim={0 0 0 33pt},clip]{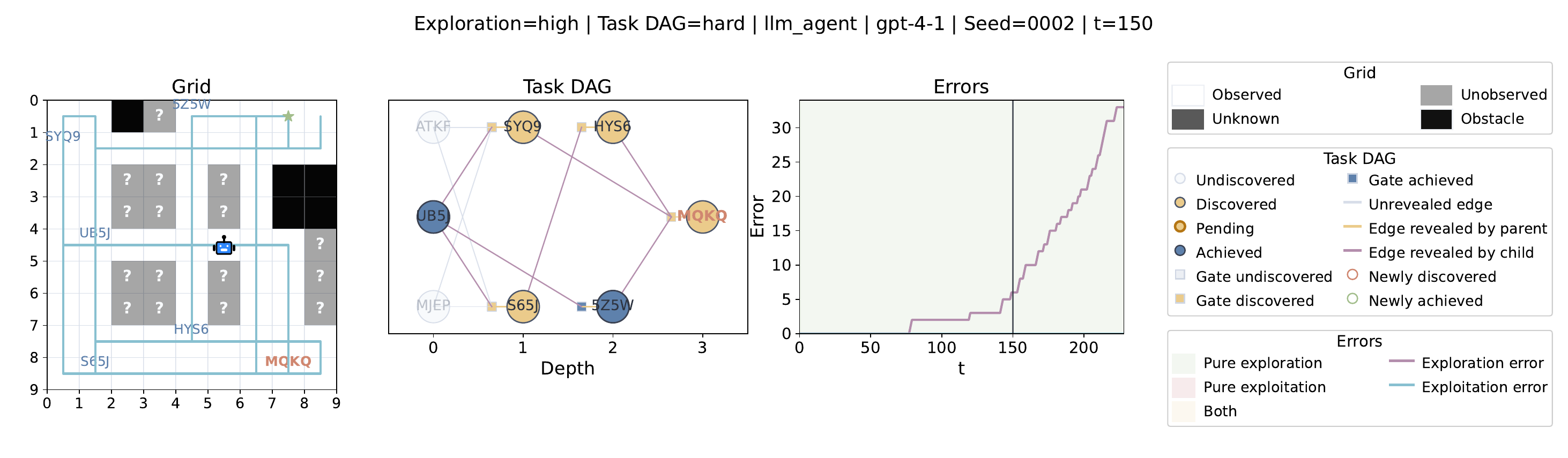}
        \caption{\currentmodel, $t = 150$}
        \end{subfigure}
        \begin{subfigure}[t]{\framewidth}
        \centering
        \includegraphics[width=\textwidth,trim={0 0 0 33pt},clip]{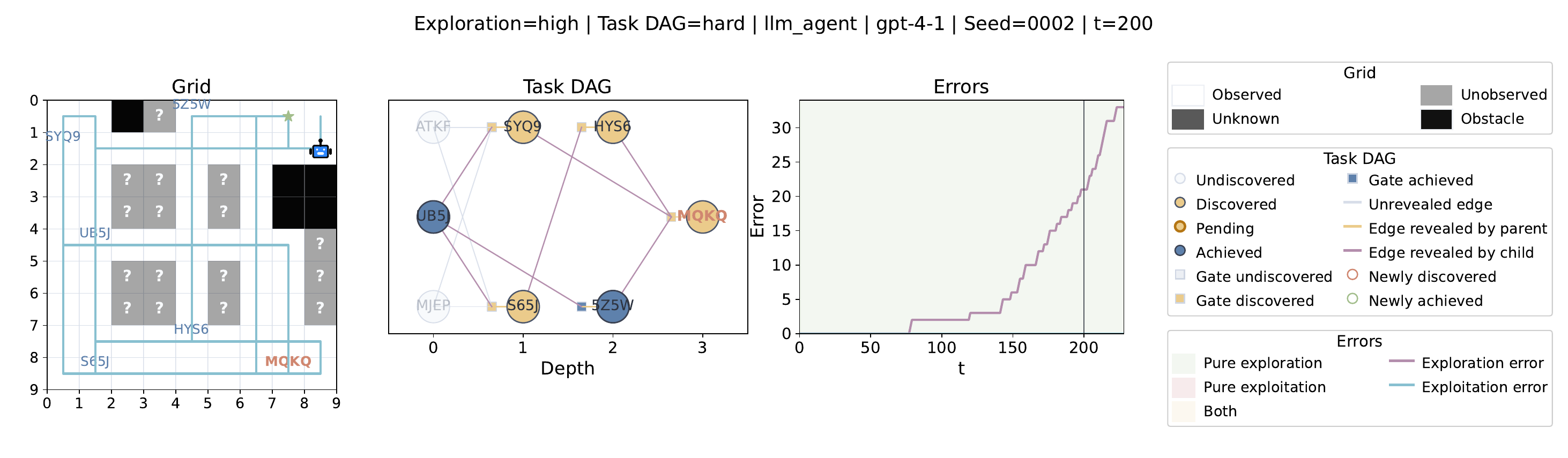}
        \caption{\currentmodel, $t = 200$}
        \end{subfigure}
        \begin{subfigure}[t]{\framewidth}
        \centering
        \includegraphics[width=\textwidth,trim={0 0 0 33pt},clip]{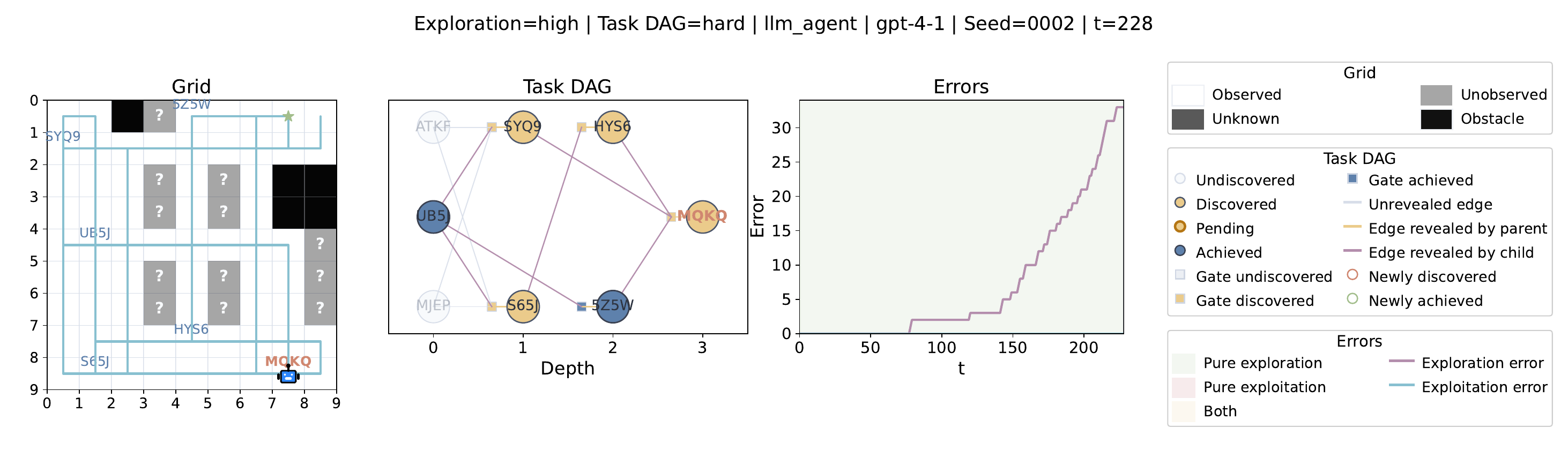}
        \caption{\currentmodel, $t = 228$}
        \end{subfigure}
    \end{center}
    \caption{Results of action trajectories and metric values over each timestep for~\currentmodel.}
\end{figure}

\clearpage

\renewcommand{\currentmodel}{\gptfivefournano}

\begin{figure}[ht]
    \begin{center}
        \begin{subfigure}[t]{\framewidth}
        \centering
        \includegraphics[width=\textwidth,trim={0 0 0 33pt},clip]{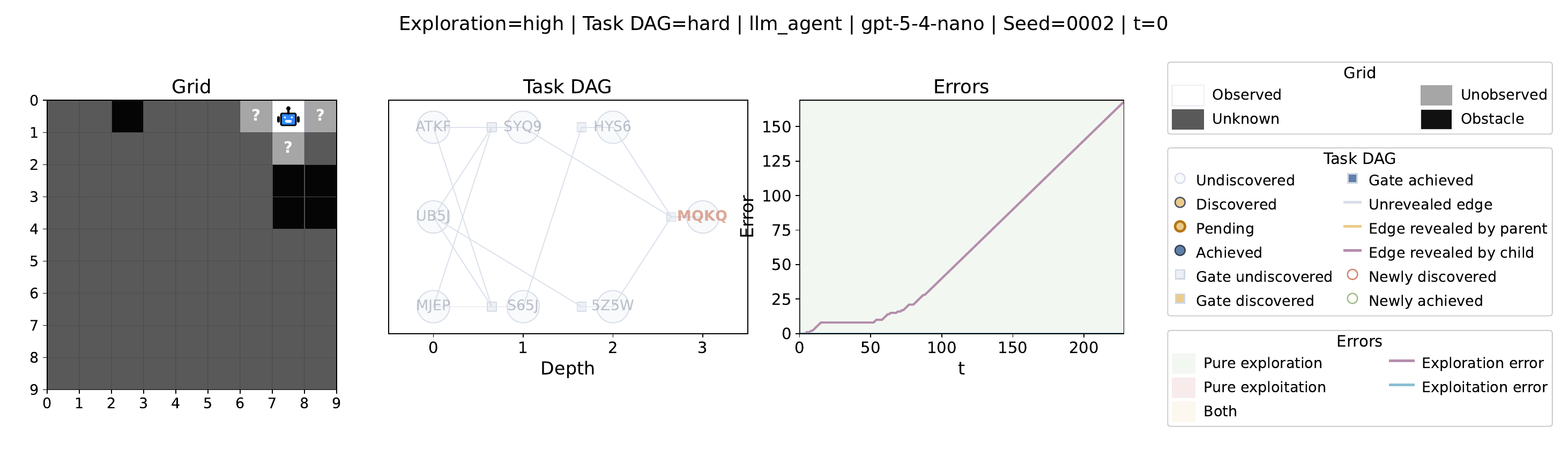}
        \caption{\currentmodel, $t = 0$}
        \end{subfigure}
        \begin{subfigure}[t]{\framewidth}
        \centering
        \includegraphics[width=\textwidth,trim={0 0 0 33pt},clip]{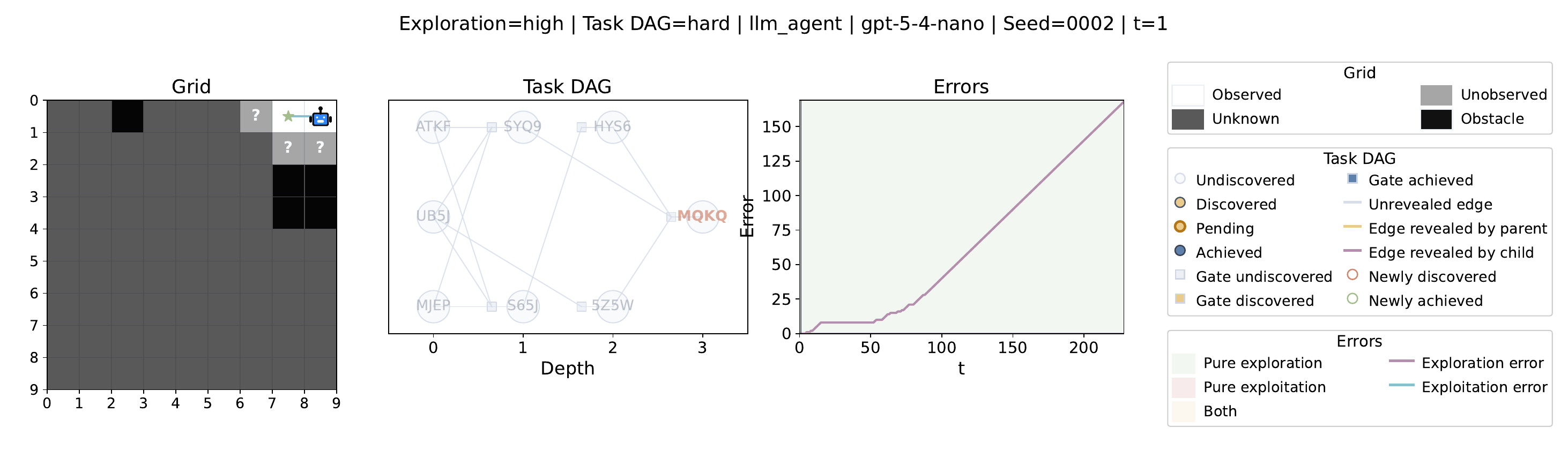}
        \caption{\currentmodel, $t = 1$}
        \end{subfigure}
        \begin{subfigure}[t]{\framewidth}
        \centering
        \includegraphics[width=\textwidth,trim={0 0 0 33pt},clip]{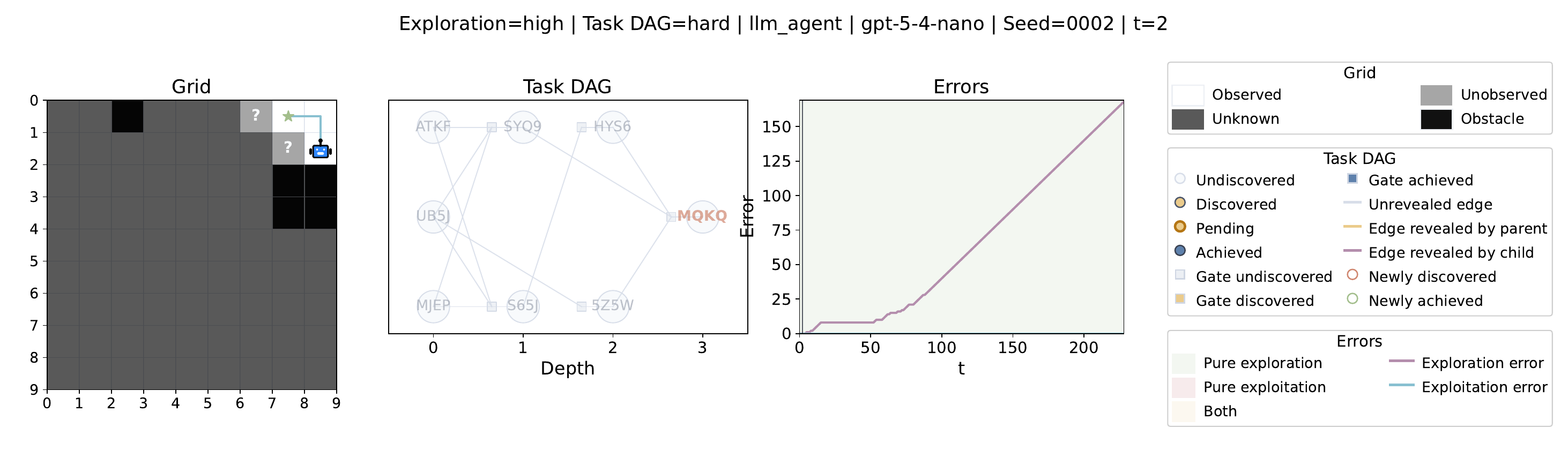}
        \caption{\currentmodel, $t = 2$}
        \end{subfigure}
        \begin{subfigure}[t]{\framewidth}
        \centering
        \includegraphics[width=\textwidth,trim={0 0 0 33pt},clip]{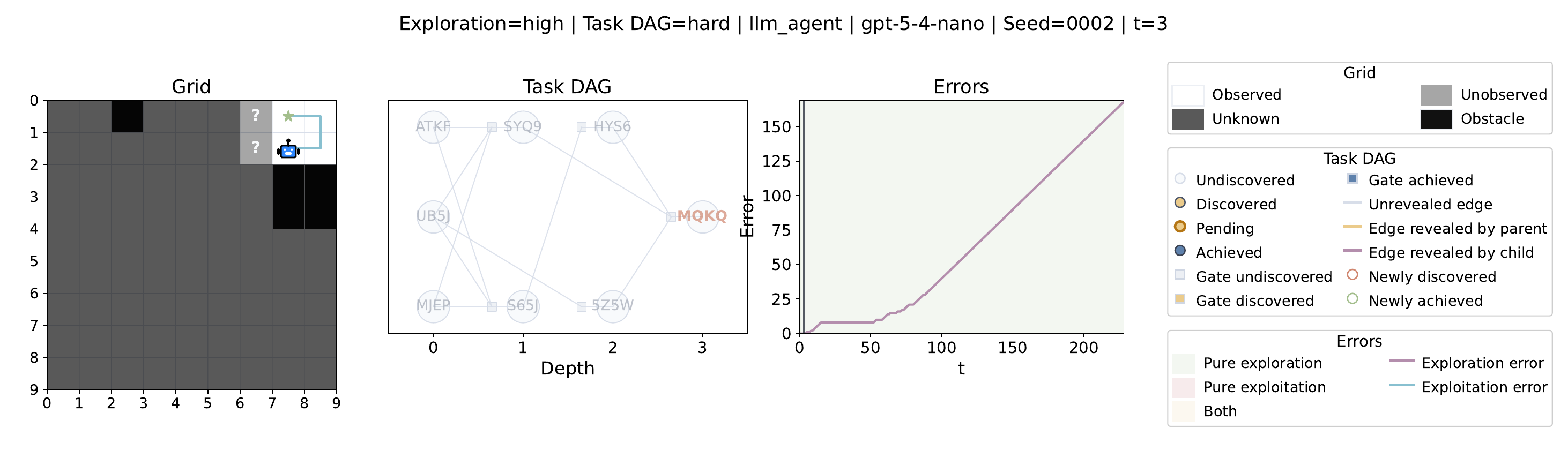}
        \caption{\currentmodel, $t = 3$}
        \end{subfigure}
        \begin{subfigure}[t]{\framewidth}
        \centering
        \includegraphics[width=\textwidth,trim={0 0 0 33pt},clip]{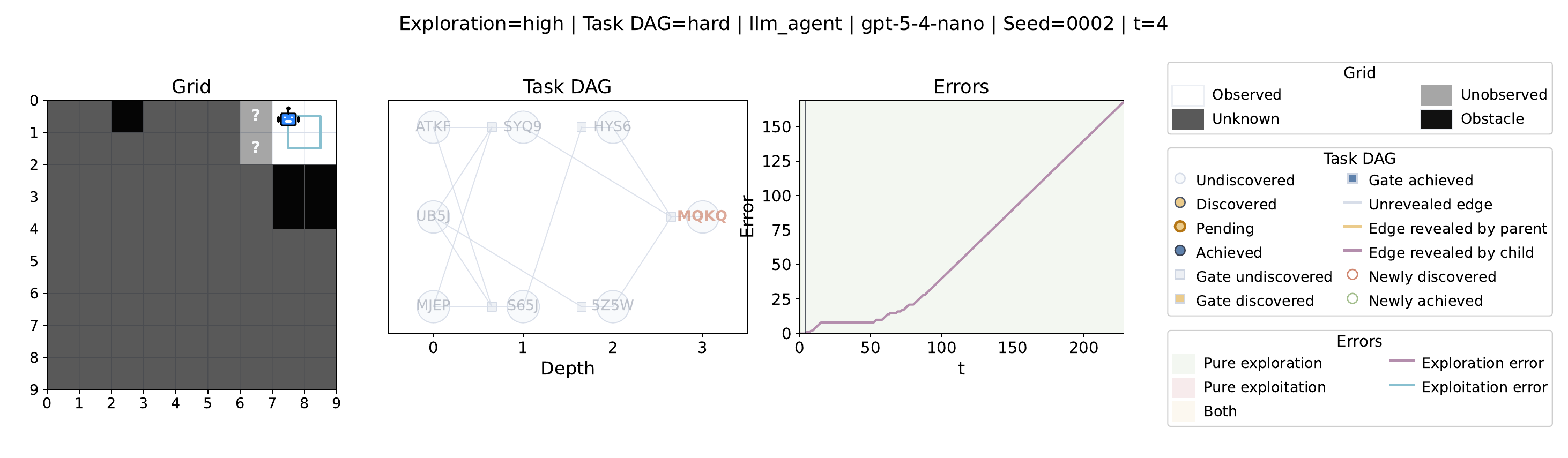}
        \caption{\currentmodel, $t = 4$}
        \end{subfigure}
    \end{center}
    \caption{Results of action trajectories and metric values over each timestep for~\currentmodel.}
\end{figure}

\begin{figure}[ht]
    \begin{center}
        \begin{subfigure}[t]{\framewidth}
        \centering
        \includegraphics[width=\textwidth,trim={0 0 0 33pt},clip]{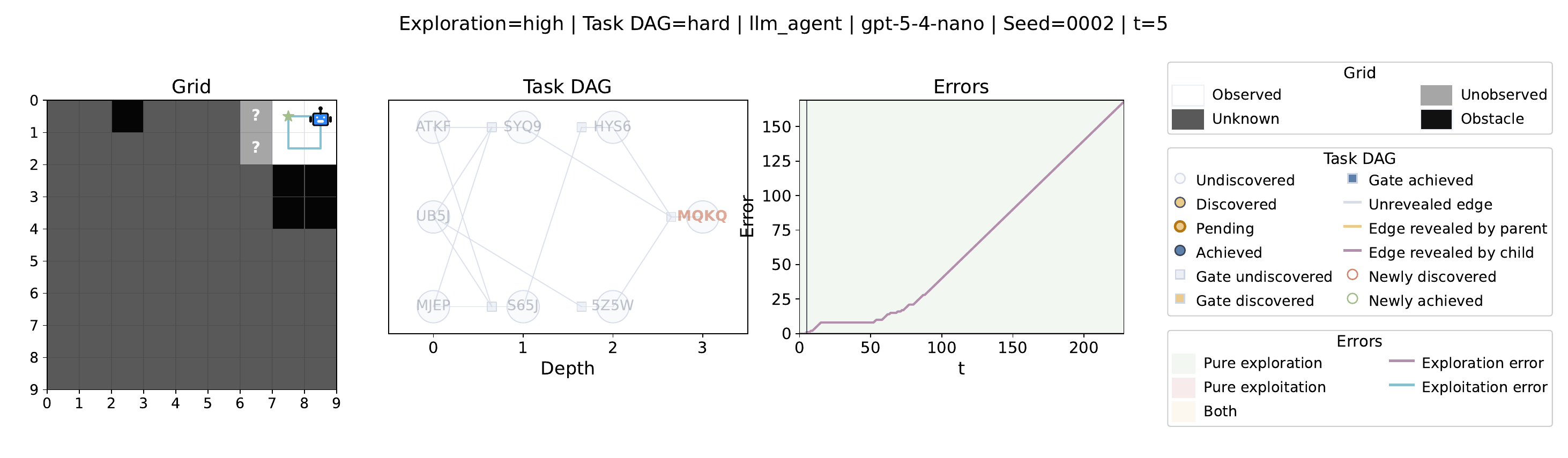}
        \caption{\currentmodel, $t = 5$}
        \end{subfigure}
        \begin{subfigure}[t]{\framewidth}
        \centering
        \includegraphics[width=\textwidth,trim={0 0 0 33pt},clip]{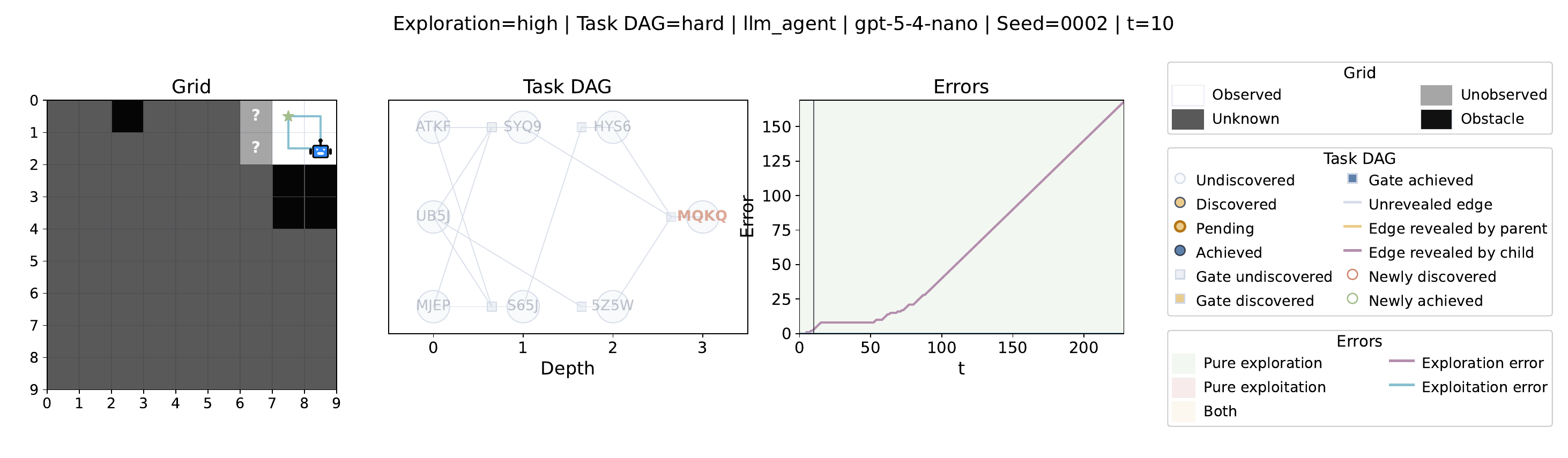}
        \caption{\currentmodel, $t = 10$}
        \end{subfigure}
        \begin{subfigure}[t]{\framewidth}
        \centering
        \includegraphics[width=\textwidth,trim={0 0 0 33pt},clip]{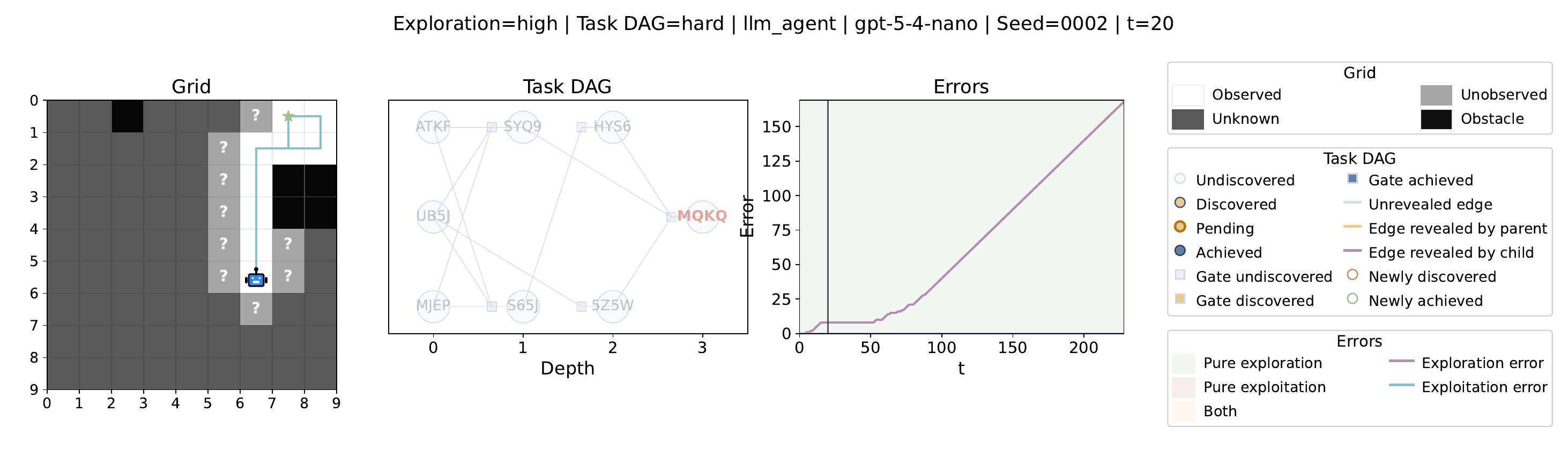}
        \caption{\currentmodel, $t = 20$}
        \end{subfigure}
        \begin{subfigure}[t]{\framewidth}
        \centering
        \includegraphics[width=\textwidth,trim={0 0 0 33pt},clip]{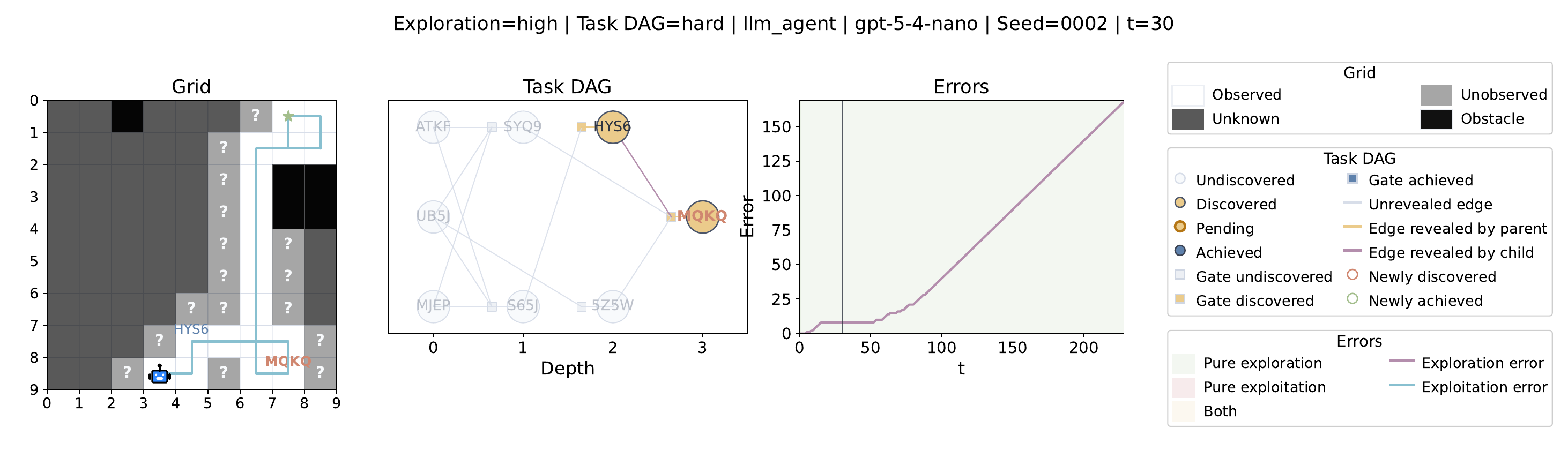}
        \caption{\currentmodel, $t = 30$}
        \end{subfigure}
        \begin{subfigure}[t]{\framewidth}
        \centering
        \includegraphics[width=\textwidth,trim={0 0 0 33pt},clip]{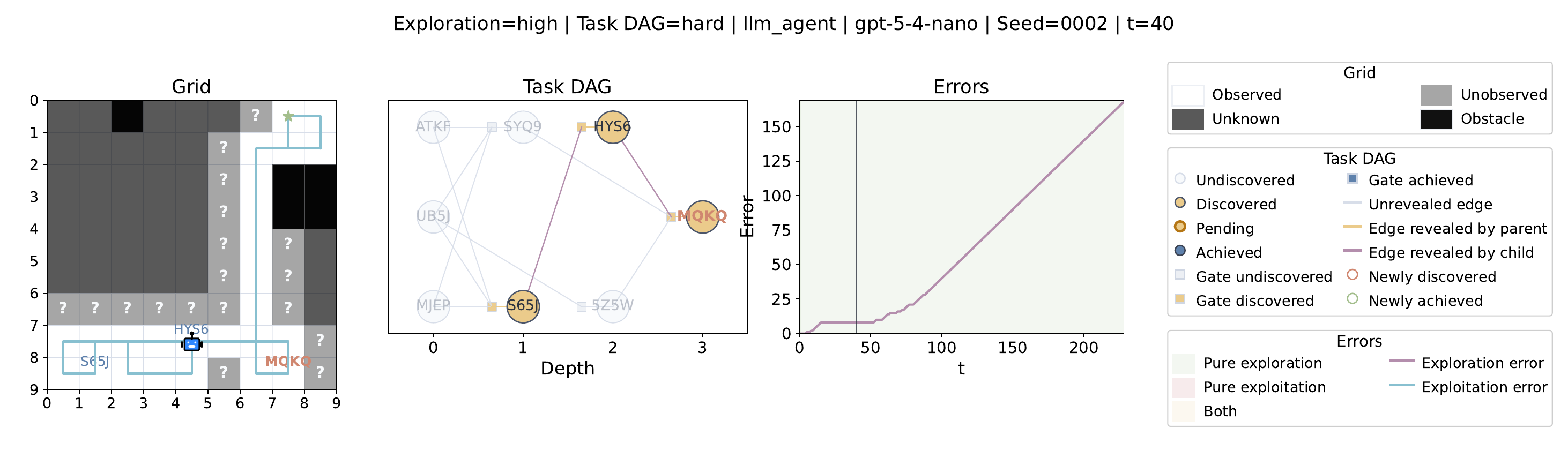}
        \caption{\currentmodel, $t = 40$}
        \end{subfigure}
    \end{center}
    \caption{Results of action trajectories and metric values over each timestep for~\currentmodel.}
\end{figure}

\begin{figure}[ht]
    \begin{center}
        \begin{subfigure}[t]{\framewidth}
        \centering
        \includegraphics[width=\textwidth,trim={0 0 0 33pt},clip]{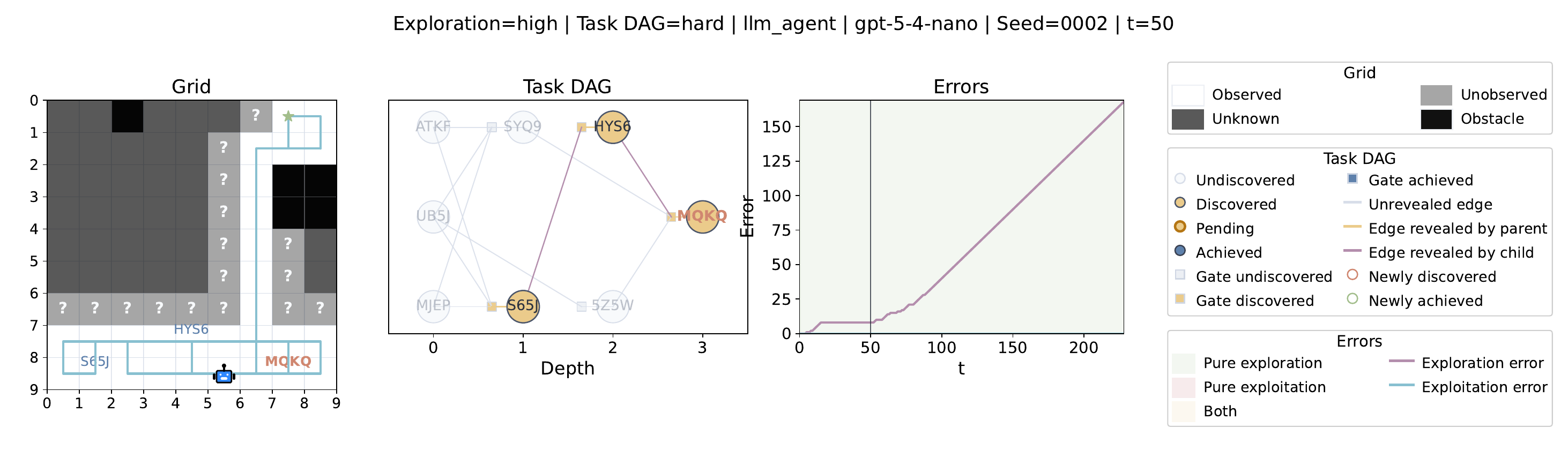}
        \caption{\currentmodel, $t = 50$}
        \end{subfigure}
        \begin{subfigure}[t]{\framewidth}
        \centering
        \includegraphics[width=\textwidth,trim={0 0 0 33pt},clip]{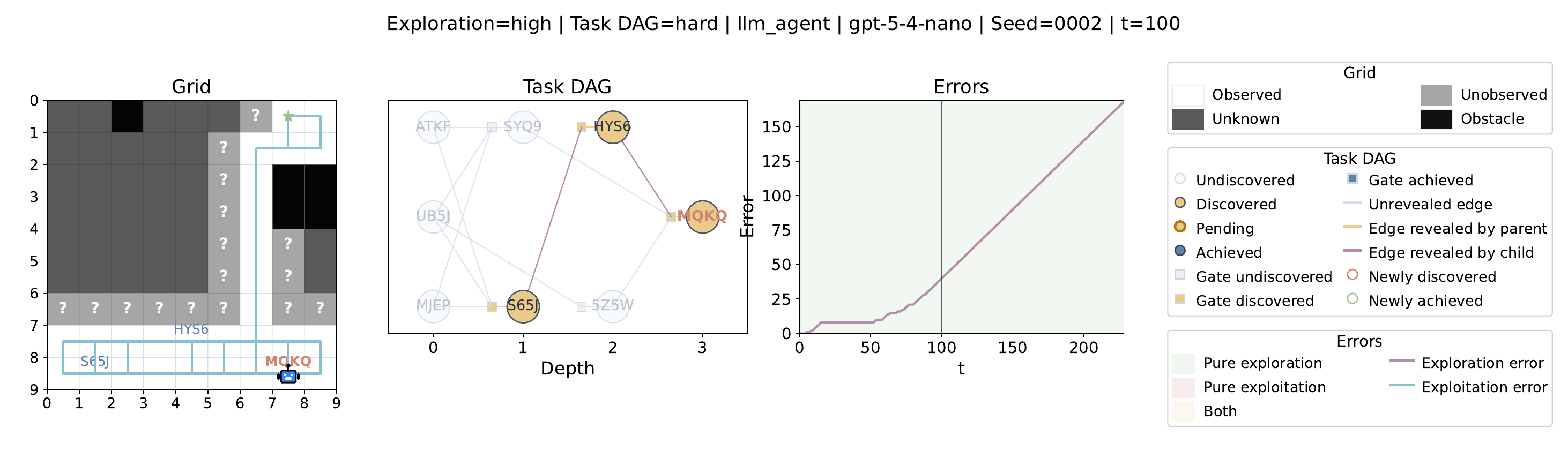}
        \caption{\currentmodel, $t = 100$}
        \end{subfigure}
        \begin{subfigure}[t]{\framewidth}
        \centering
        \includegraphics[width=\textwidth,trim={0 0 0 33pt},clip]{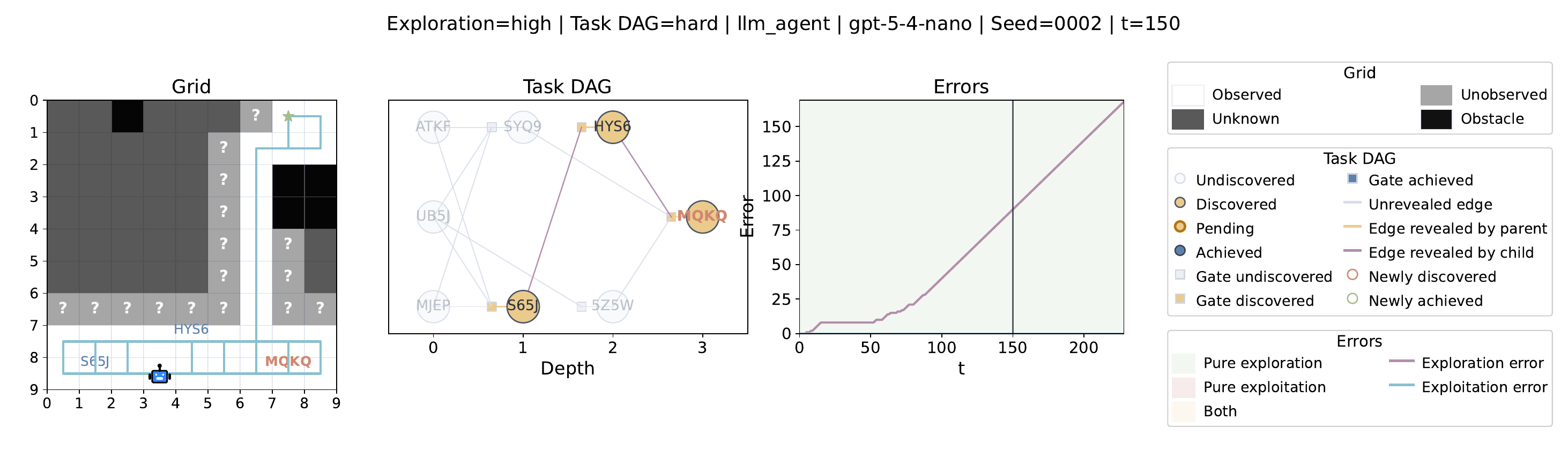}
        \caption{\currentmodel, $t = 150$}
        \end{subfigure}
        \begin{subfigure}[t]{\framewidth}
        \centering
        \includegraphics[width=\textwidth,trim={0 0 0 33pt},clip]{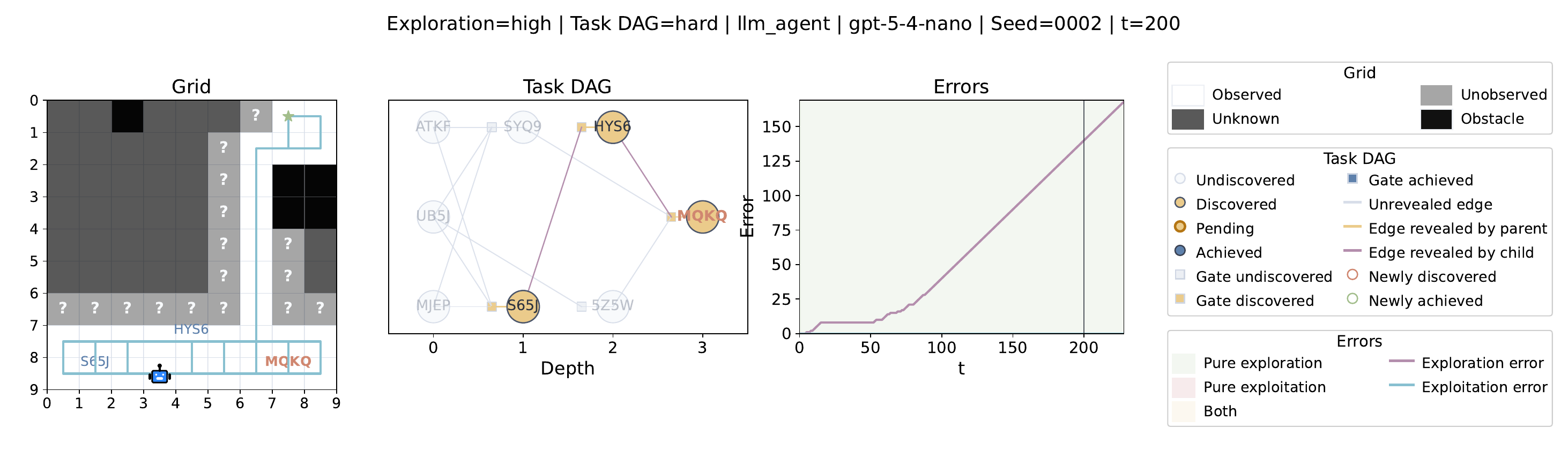}
        \caption{\currentmodel, $t = 200$}
        \end{subfigure}
        \begin{subfigure}[t]{\framewidth}
        \centering
        \includegraphics[width=\textwidth,trim={0 0 0 33pt},clip]{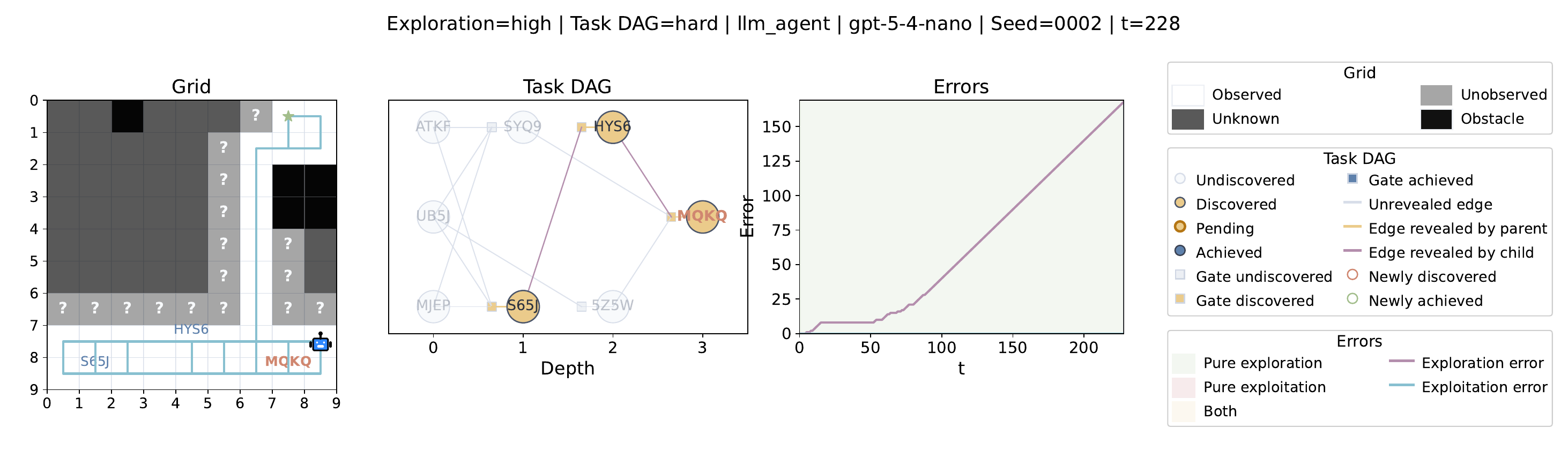}
        \caption{\currentmodel, $t = 228$}
        \end{subfigure}
    \end{center}
    \caption{Results of action trajectories and metric values over each timestep for~\currentmodel.}
\end{figure}

\clearpage
\renewcommand{\currentmodel}{\gptfivefourmini}

\begin{figure}[ht]
    \begin{center}
        \begin{subfigure}[t]{\framewidth}
        \centering
        \includegraphics[width=\textwidth,trim={0 0 0 33pt},clip]{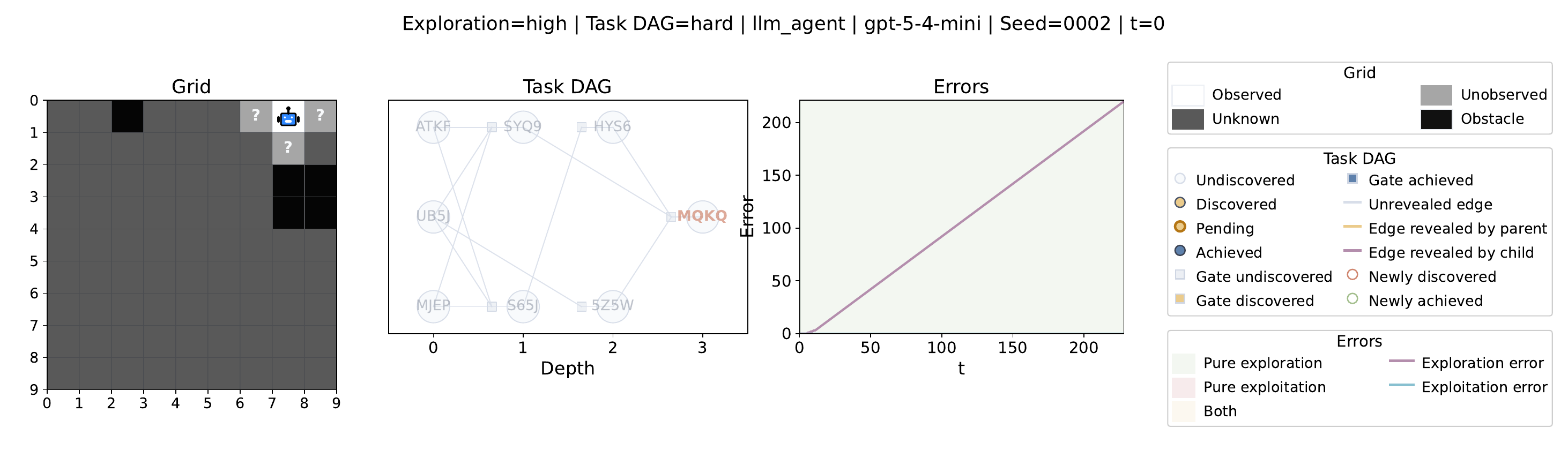}
        \caption{\currentmodel, $t = 0$}
        \end{subfigure}
        \begin{subfigure}[t]{\framewidth}
        \centering
        \includegraphics[width=\textwidth,trim={0 0 0 33pt},clip]{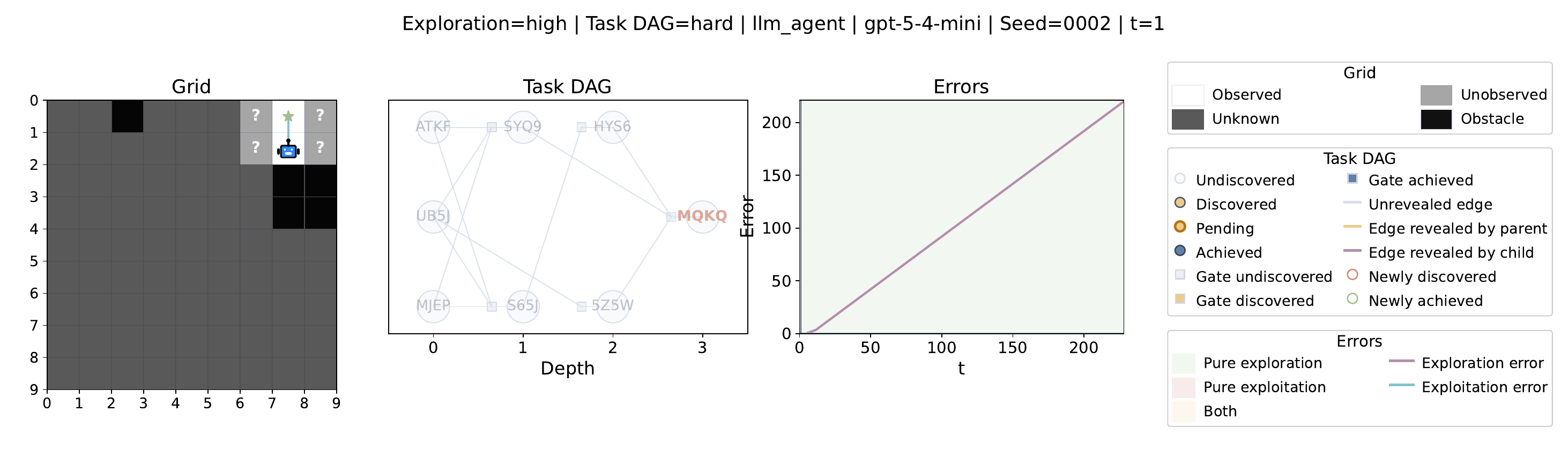}
        \caption{\currentmodel, $t = 1$}
        \end{subfigure}
        \begin{subfigure}[t]{\framewidth}
        \centering
        \includegraphics[width=\textwidth,trim={0 0 0 33pt},clip]{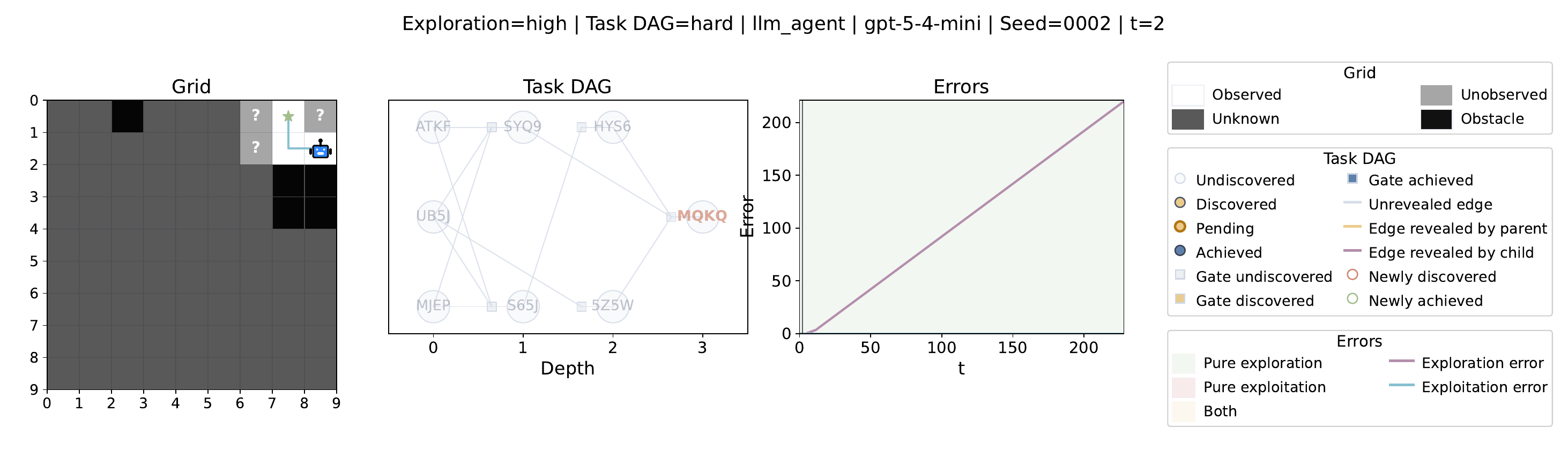}
        \caption{\currentmodel, $t = 2$}
        \end{subfigure}
        \begin{subfigure}[t]{\framewidth}
        \centering
        \includegraphics[width=\textwidth,trim={0 0 0 33pt},clip]{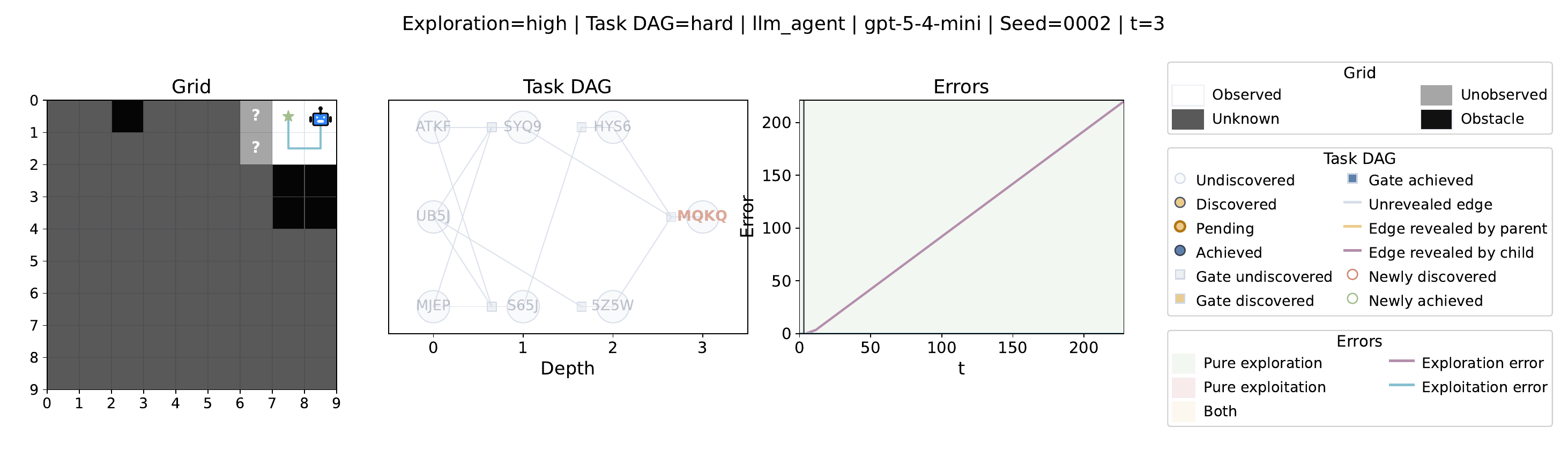}
        \caption{\currentmodel, $t = 3$}
        \end{subfigure}
        \begin{subfigure}[t]{\framewidth}
        \centering
        \includegraphics[width=\textwidth,trim={0 0 0 33pt},clip]{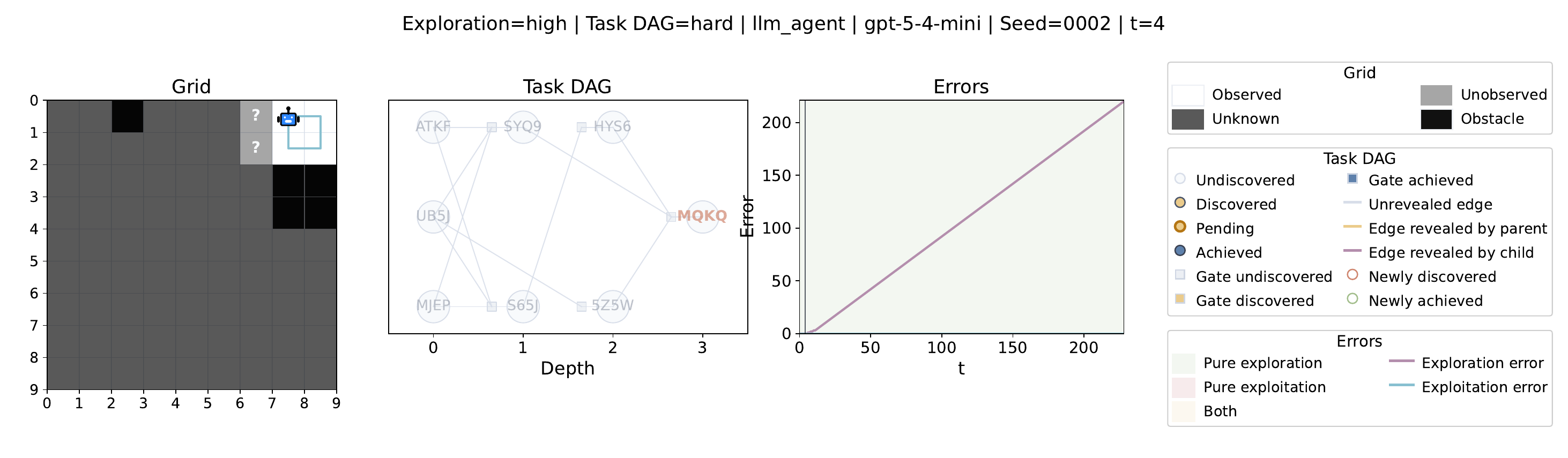}
        \caption{\currentmodel, $t = 4$}
        \end{subfigure}
    \end{center}
    \caption{Results of action trajectories and metric values over each timestep for~\currentmodel.}
\end{figure}

\begin{figure}[ht]
    \begin{center}
        \begin{subfigure}[t]{\framewidth}
        \centering
        \includegraphics[width=\textwidth,trim={0 0 0 33pt},clip]{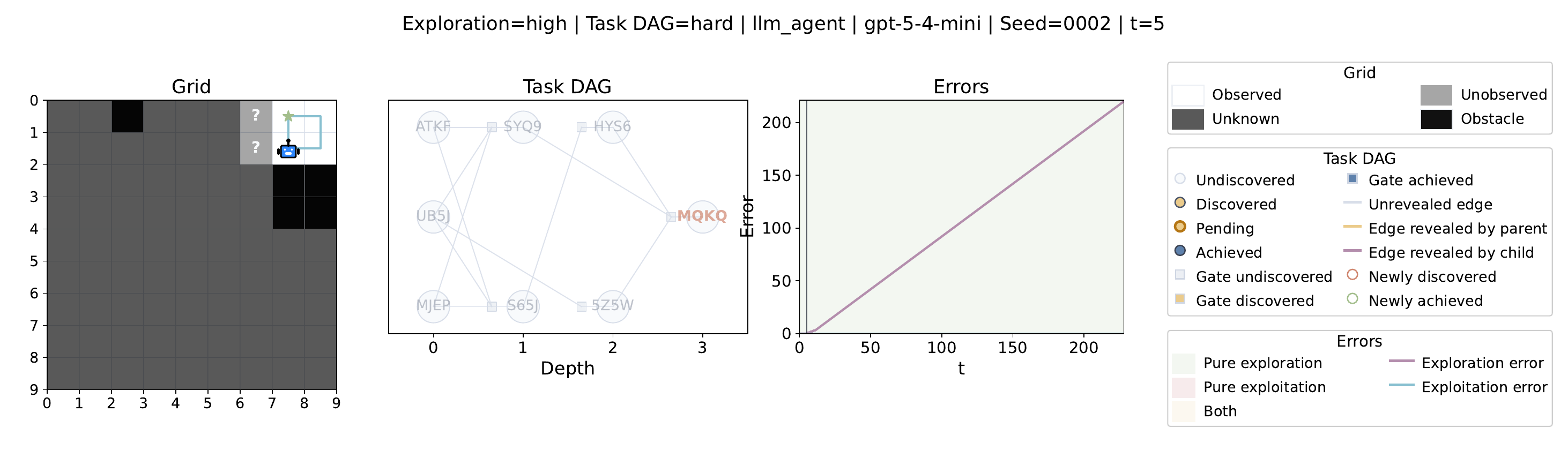}
        \caption{\currentmodel, $t = 5$}
        \end{subfigure}
        \begin{subfigure}[t]{\framewidth}
        \centering
        \includegraphics[width=\textwidth,trim={0 0 0 33pt},clip]{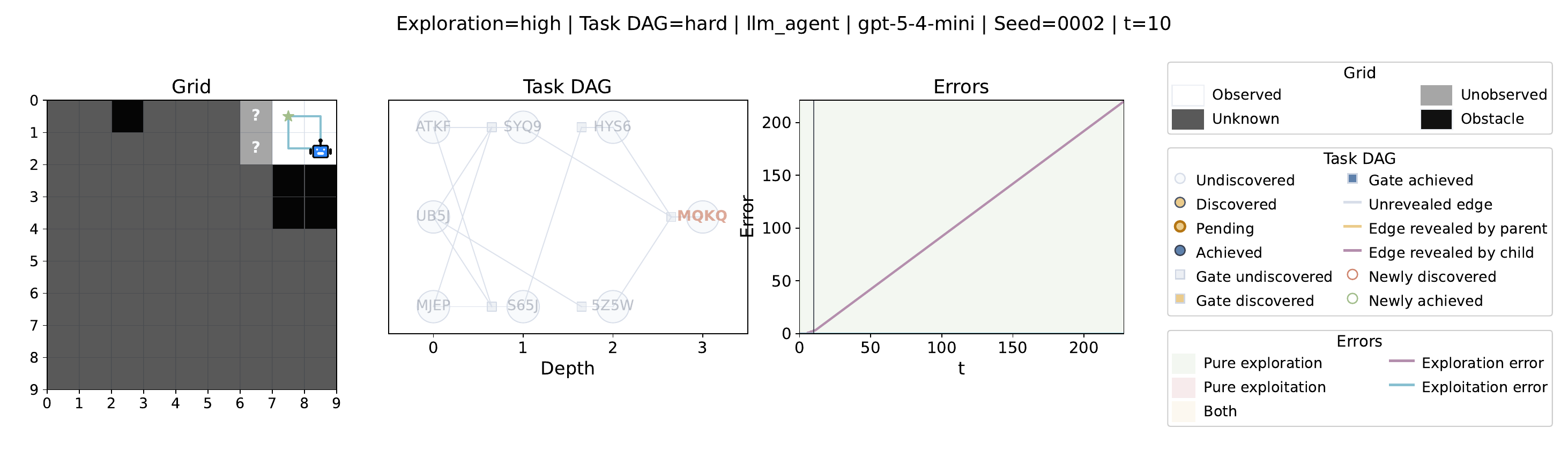}
        \caption{\currentmodel, $t = 10$}
        \end{subfigure}
        \begin{subfigure}[t]{\framewidth}
        \centering
        \includegraphics[width=\textwidth,trim={0 0 0 33pt},clip]{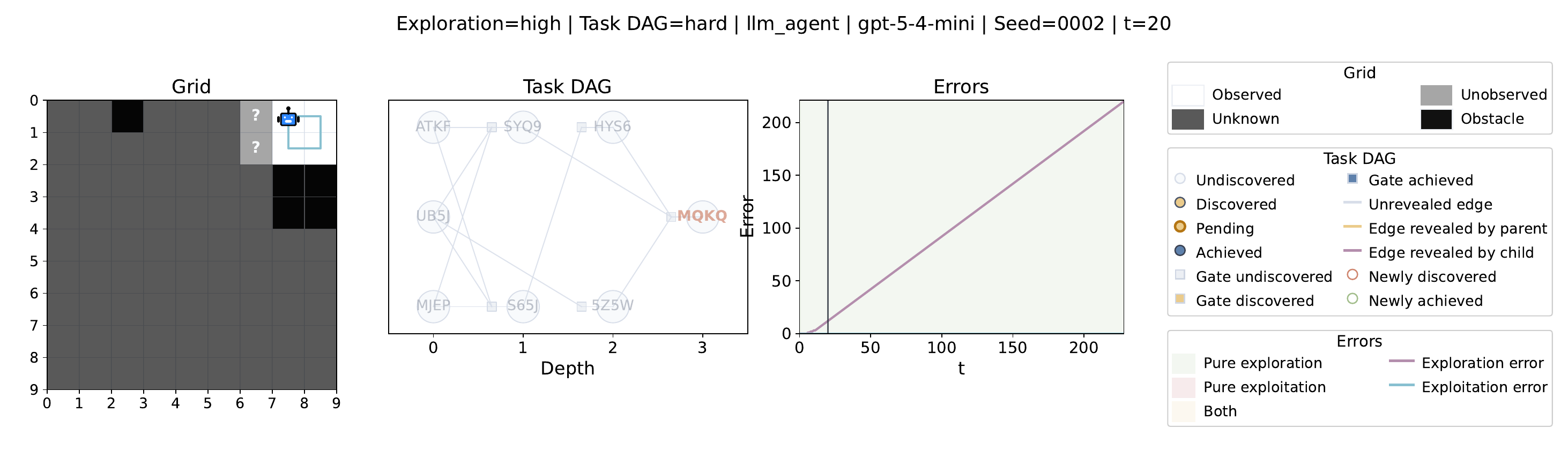}
        \caption{\currentmodel, $t = 20$}
        \end{subfigure}
        \begin{subfigure}[t]{\framewidth}
        \centering
        \includegraphics[width=\textwidth,trim={0 0 0 33pt},clip]{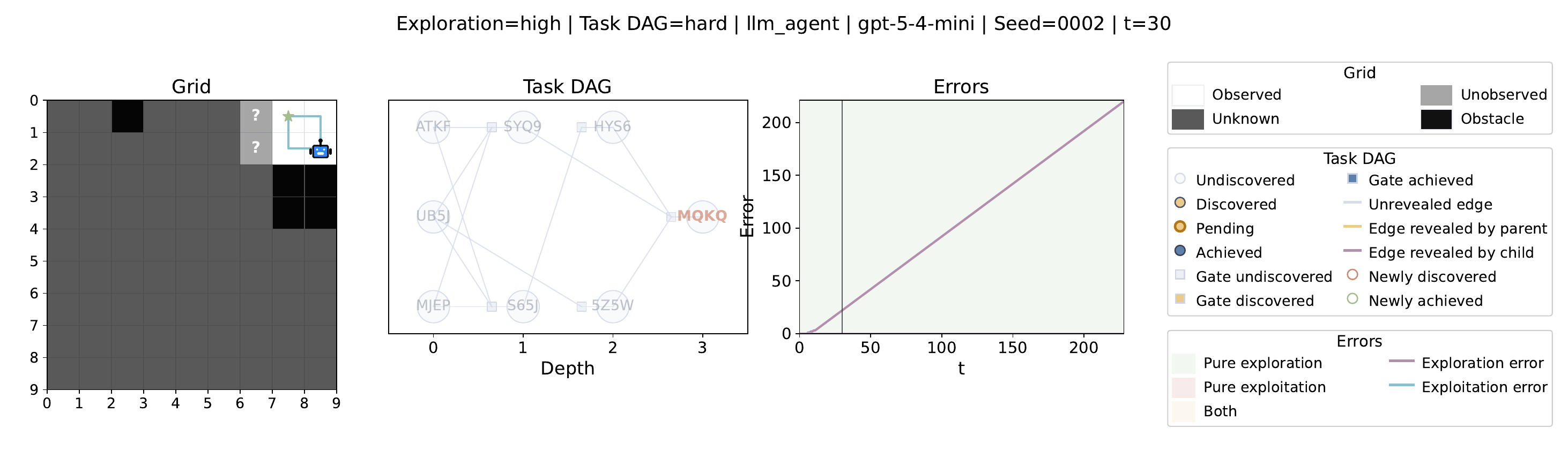}
        \caption{\currentmodel, $t = 30$}
        \end{subfigure}
        \begin{subfigure}[t]{\framewidth}
        \centering
        \includegraphics[width=\textwidth,trim={0 0 0 33pt},clip]{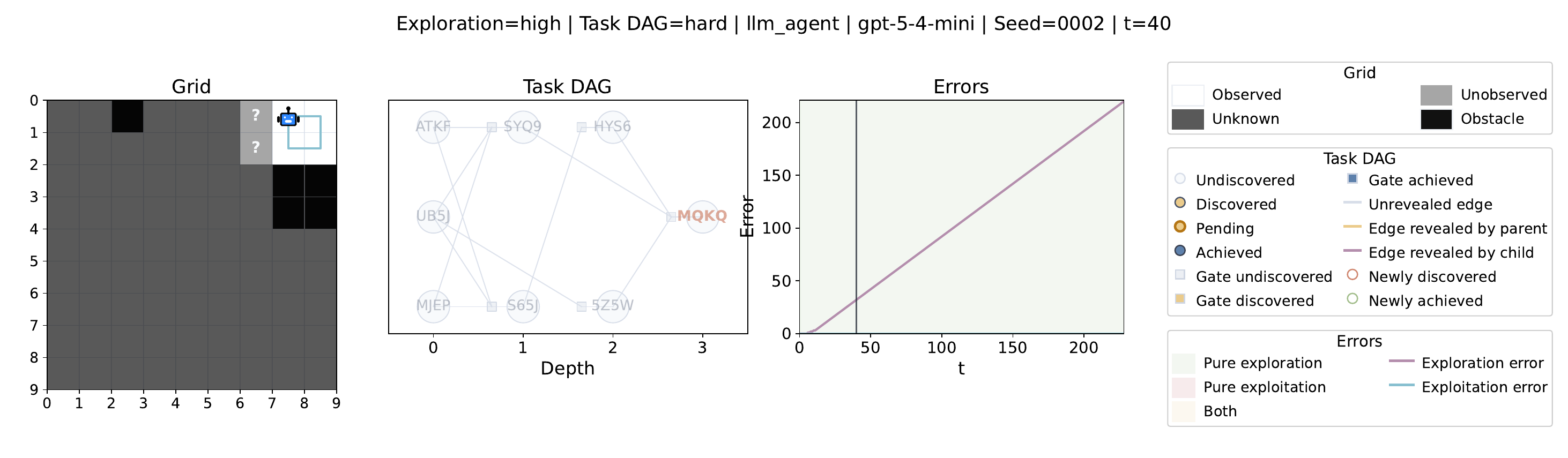}
        \caption{\currentmodel, $t = 40$}
        \end{subfigure}
    \end{center}
    \caption{Results of action trajectories and metric values over each timestep for~\currentmodel.}
\end{figure}

\begin{figure}[ht]
    \begin{center}
        \begin{subfigure}[t]{\framewidth}
        \centering
        \includegraphics[width=\textwidth,trim={0 0 0 33pt},clip]{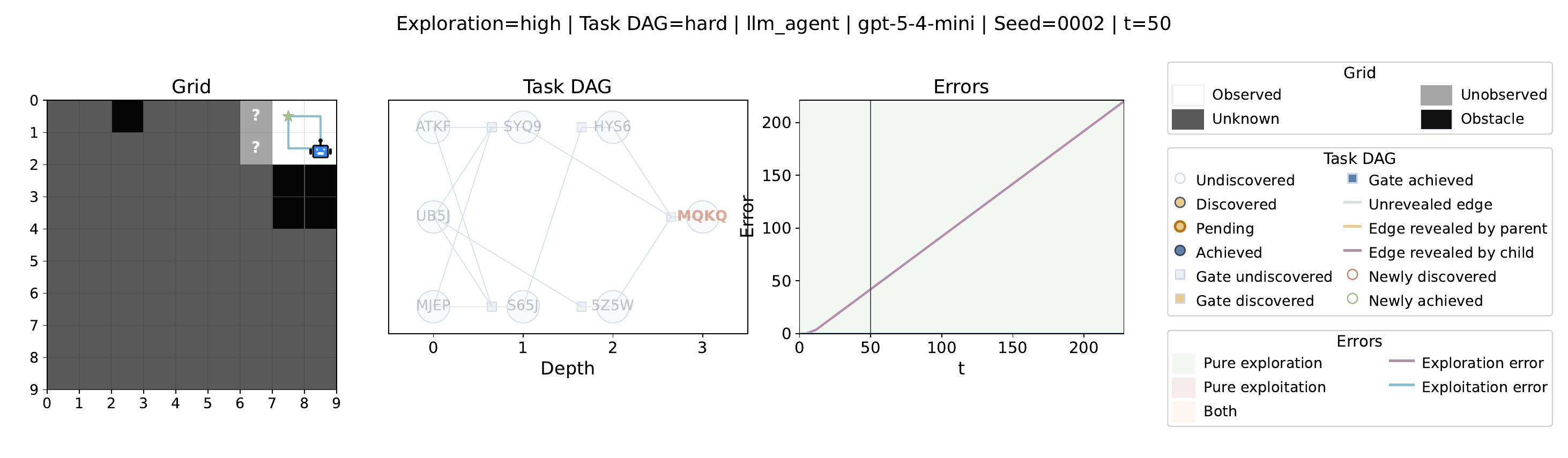}
        \caption{\currentmodel, $t = 50$}
        \end{subfigure}
        \begin{subfigure}[t]{\framewidth}
        \centering
        \includegraphics[width=\textwidth,trim={0 0 0 33pt},clip]{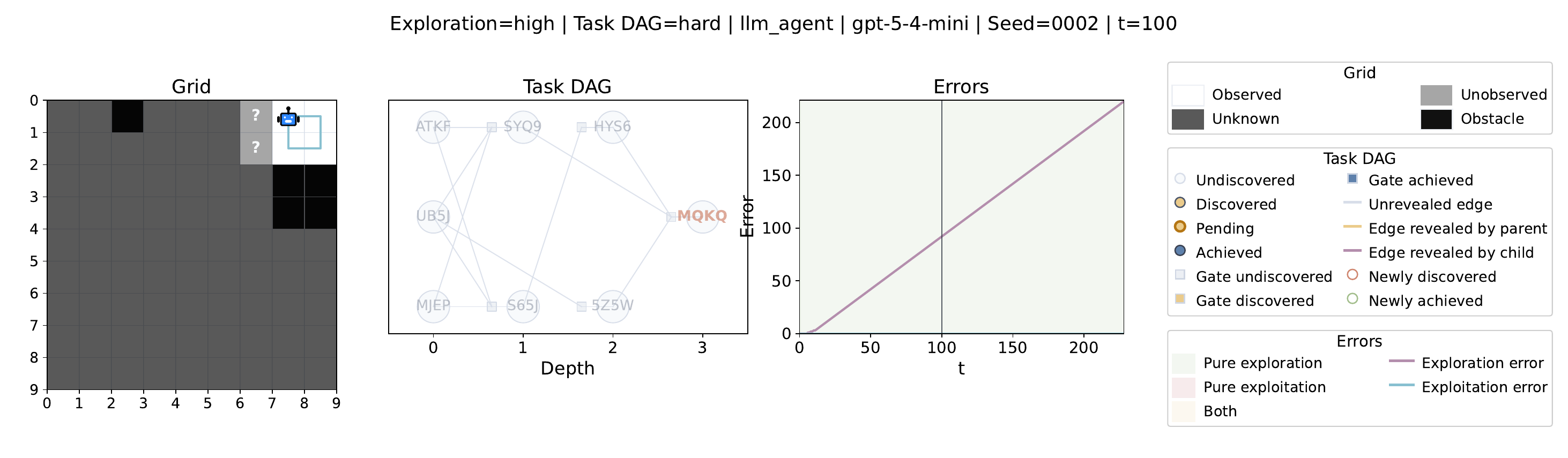}
        \caption{\currentmodel, $t = 100$}
        \end{subfigure}
        \begin{subfigure}[t]{\framewidth}
        \centering
        \includegraphics[width=\textwidth,trim={0 0 0 33pt},clip]{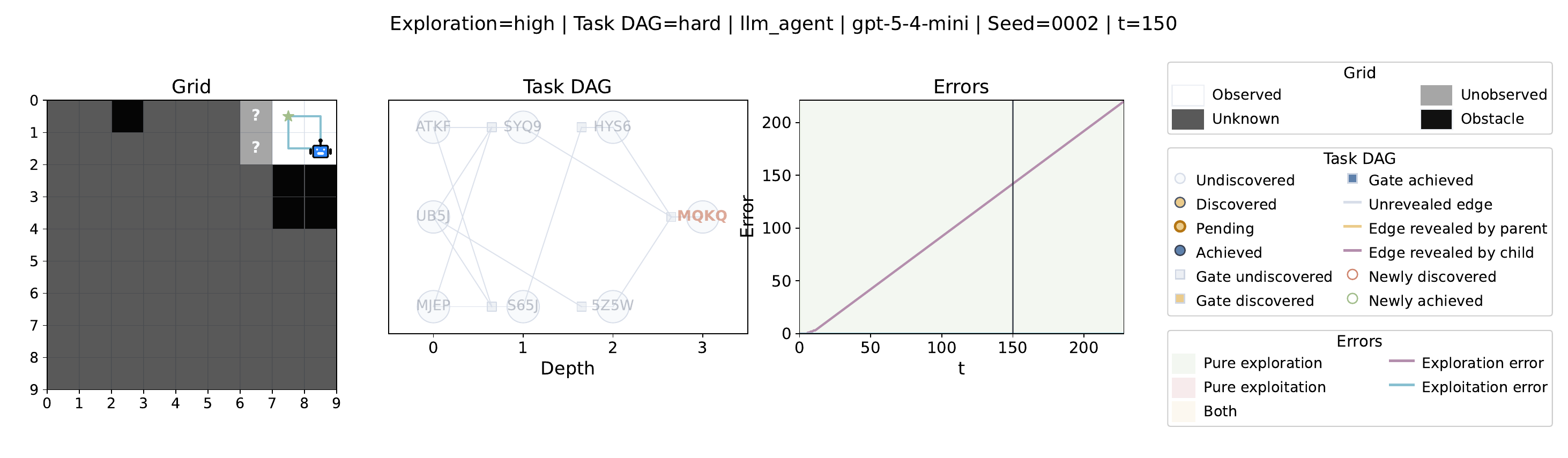}
        \caption{\currentmodel, $t = 150$}
        \end{subfigure}
        \begin{subfigure}[t]{\framewidth}
        \centering
        \includegraphics[width=\textwidth,trim={0 0 0 33pt},clip]{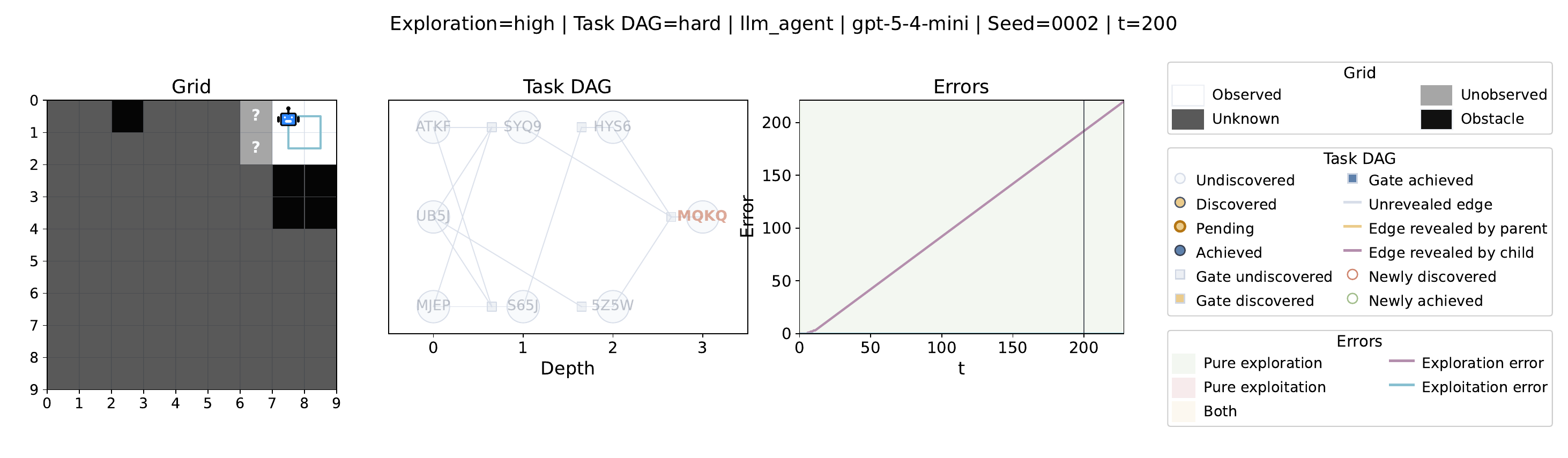}
        \caption{\currentmodel, $t = 200$}
        \end{subfigure}
        \begin{subfigure}[t]{\framewidth}
        \centering
        \includegraphics[width=\textwidth,trim={0 0 0 33pt},clip]{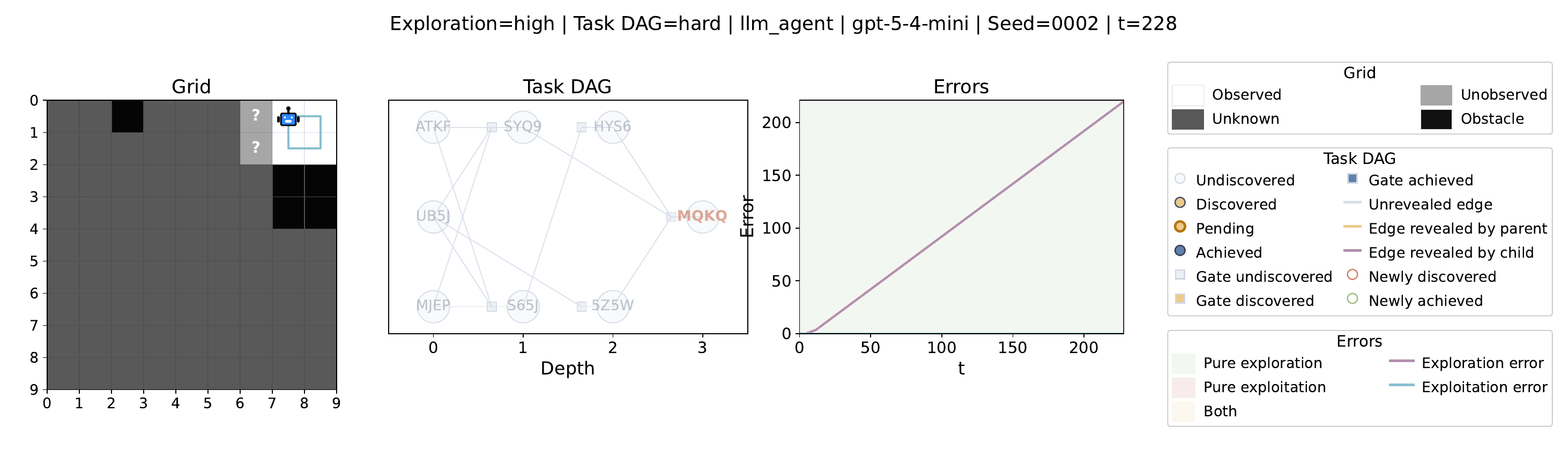}
        \caption{\currentmodel, $t = 228$}
        \end{subfigure}
    \end{center}
    \caption{Results of action trajectories and metric values over each timestep for~\currentmodel.}
\end{figure}

\clearpage
\renewcommand{\currentmodel}{\gptfivefour}

\begin{figure}[ht]
    \begin{center}
        \begin{subfigure}[t]{\framewidth}
        \centering
        \includegraphics[width=\textwidth,trim={0 0 0 33pt},clip]{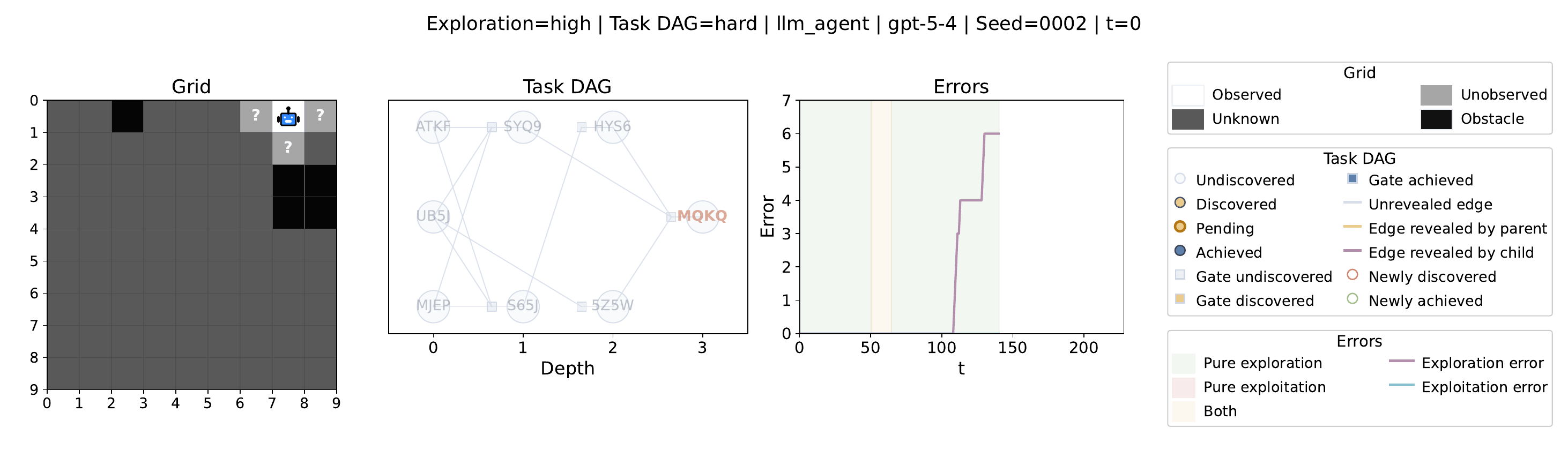}
        \caption{\currentmodel, $t = 0$}
        \end{subfigure}
        \begin{subfigure}[t]{\framewidth}
        \centering
        \includegraphics[width=\textwidth,trim={0 0 0 33pt},clip]{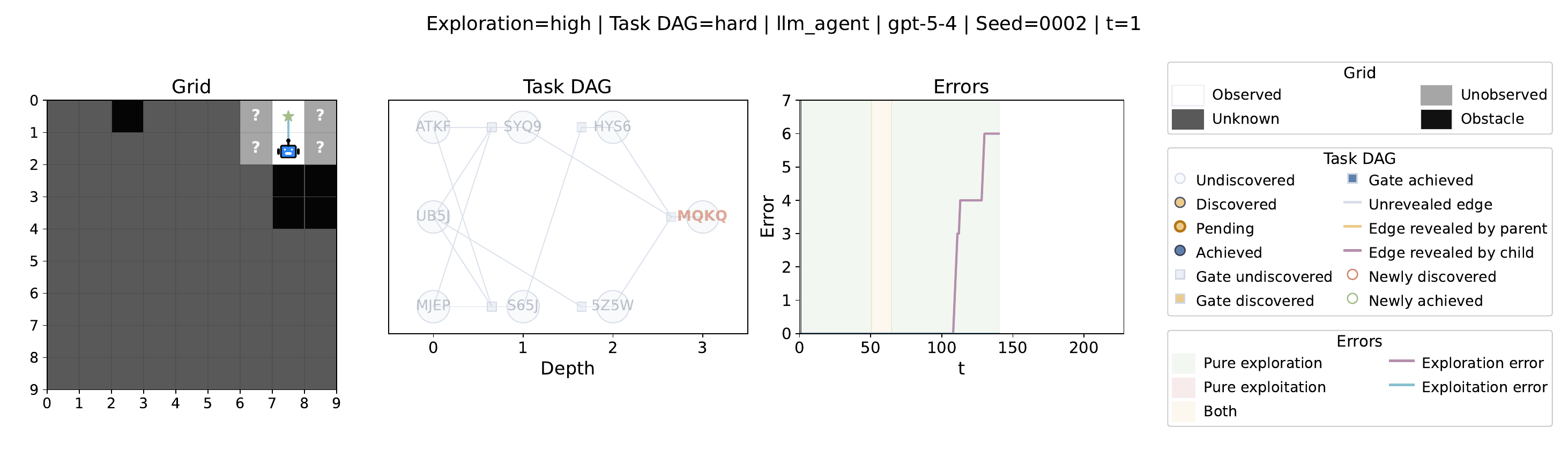}
        \caption{\currentmodel, $t = 1$}
        \end{subfigure}
        \begin{subfigure}[t]{\framewidth}
        \centering
        \includegraphics[width=\textwidth,trim={0 0 0 33pt},clip]{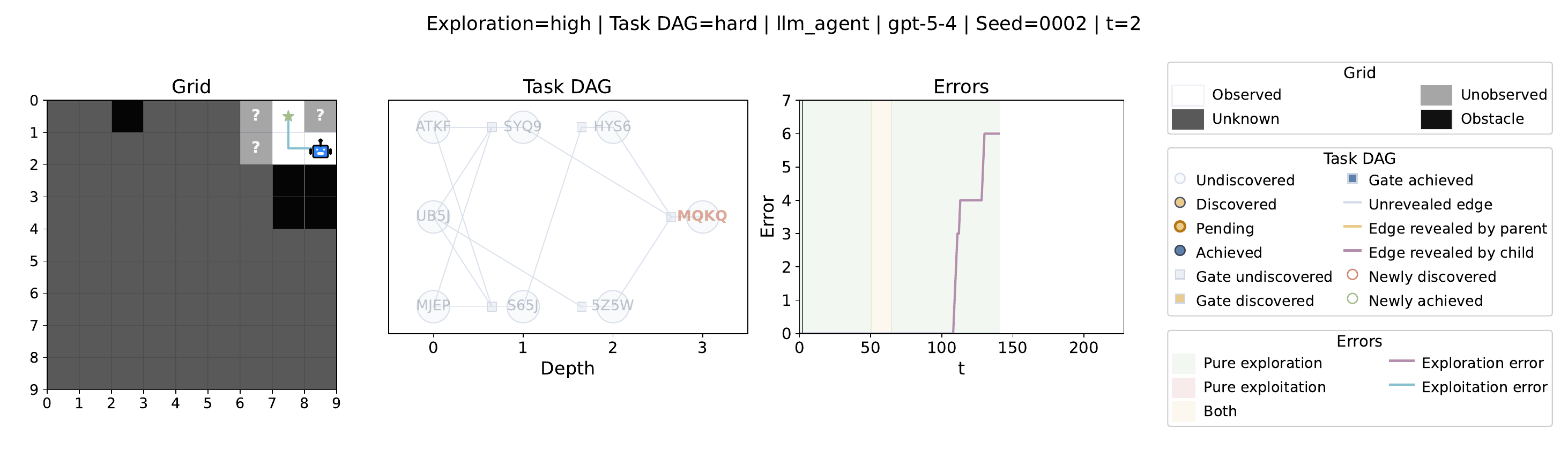}
        \caption{\currentmodel, $t = 2$}
        \end{subfigure}
        \begin{subfigure}[t]{\framewidth}
        \centering
        \includegraphics[width=\textwidth,trim={0 0 0 33pt},clip]{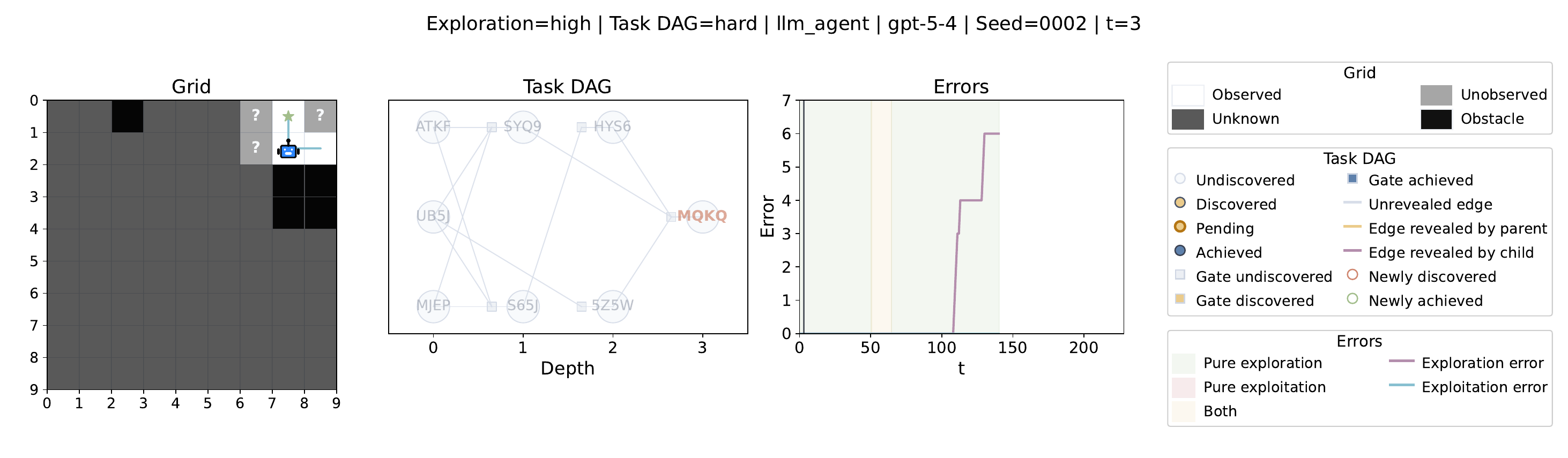}
        \caption{\currentmodel, $t = 3$}
        \end{subfigure}
        \begin{subfigure}[t]{\framewidth}
        \centering
        \includegraphics[width=\textwidth,trim={0 0 0 33pt},clip]{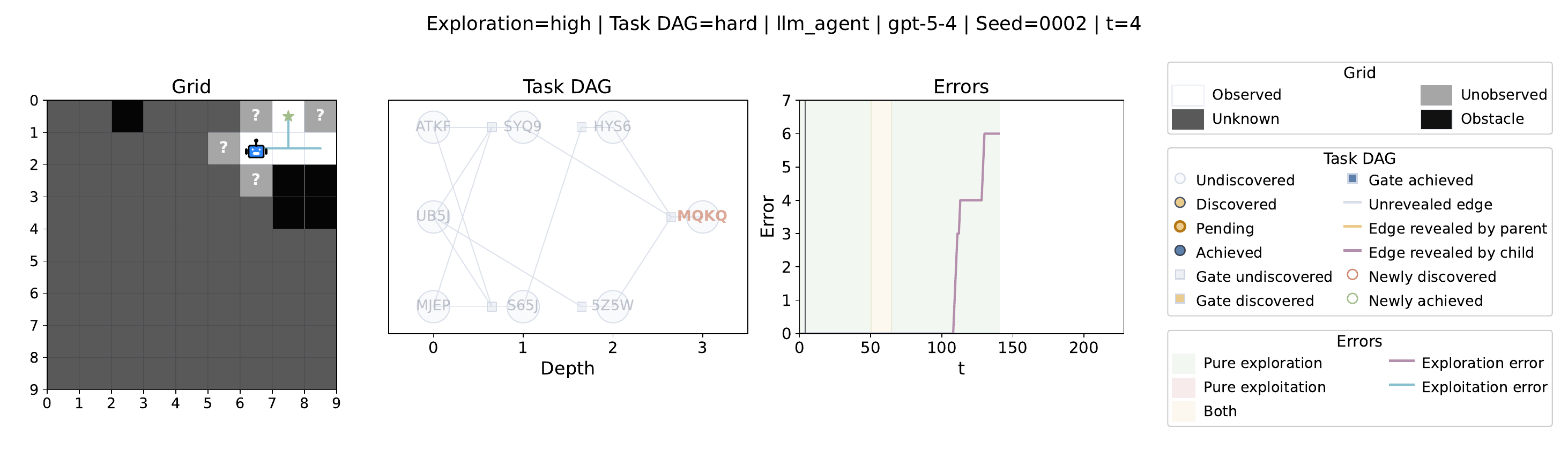}
        \caption{\currentmodel, $t = 4$}
        \end{subfigure}
    \end{center}
    \caption{Results of action trajectories and metric values over each timestep for~\currentmodel.}
\end{figure}

\begin{figure}[ht]
    \begin{center}
        \begin{subfigure}[t]{\framewidth}
        \centering
        \includegraphics[width=\textwidth,trim={0 0 0 33pt},clip]{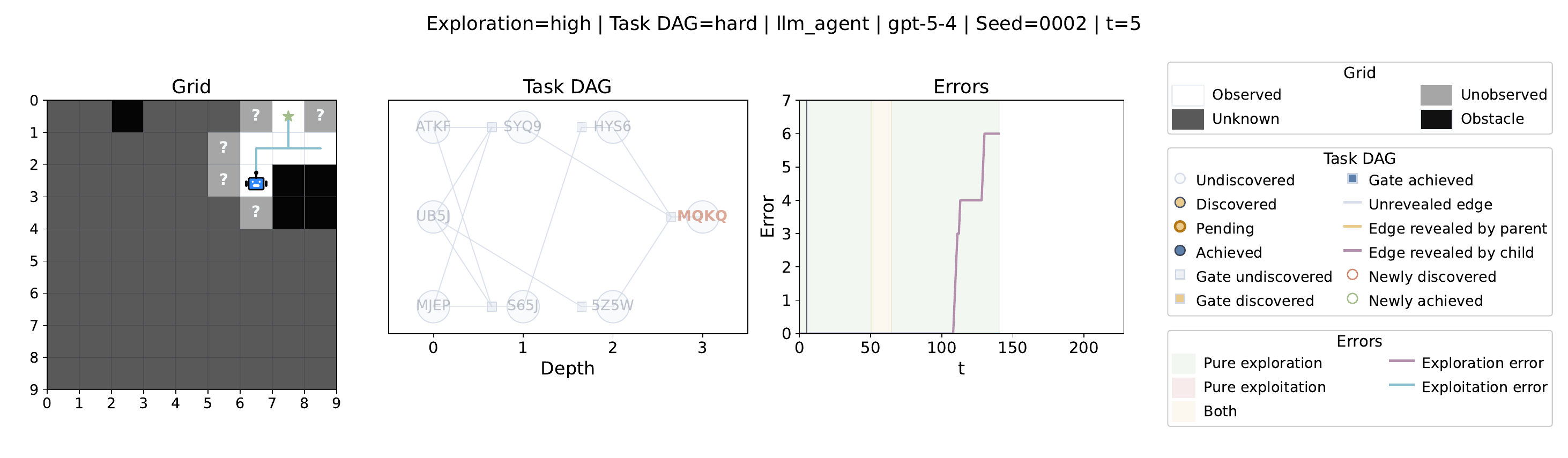}
        \caption{\currentmodel, $t = 5$}
        \end{subfigure}
        \begin{subfigure}[t]{\framewidth}
        \centering
        \includegraphics[width=\textwidth,trim={0 0 0 33pt},clip]{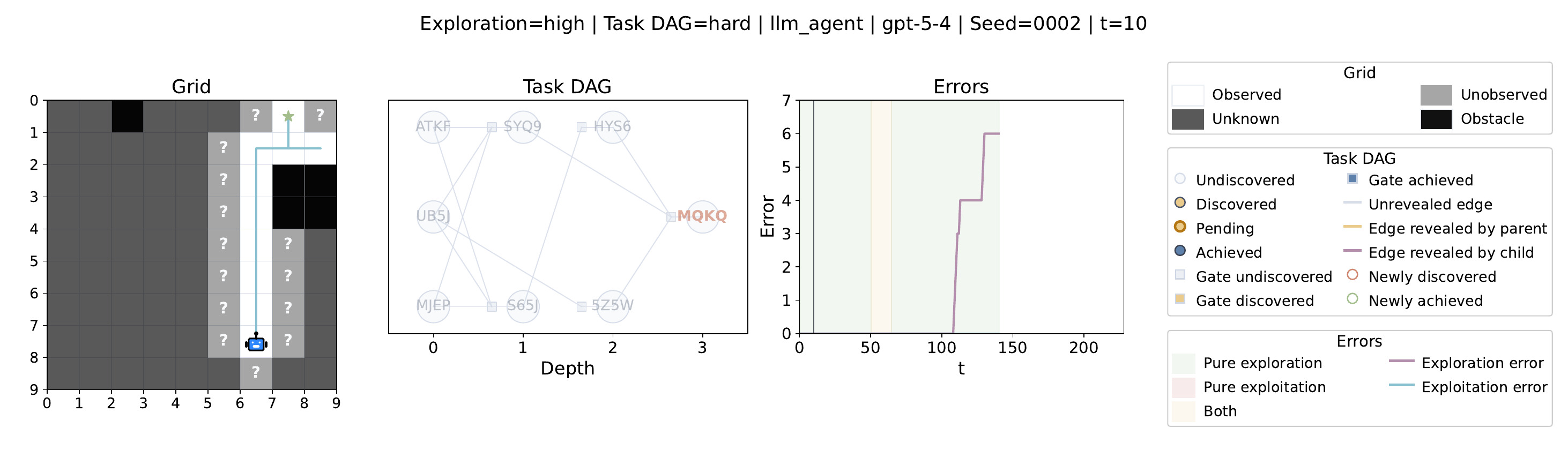}
        \caption{\currentmodel, $t = 10$}
        \end{subfigure}
        \begin{subfigure}[t]{\framewidth}
        \centering
        \includegraphics[width=\textwidth,trim={0 0 0 33pt},clip]{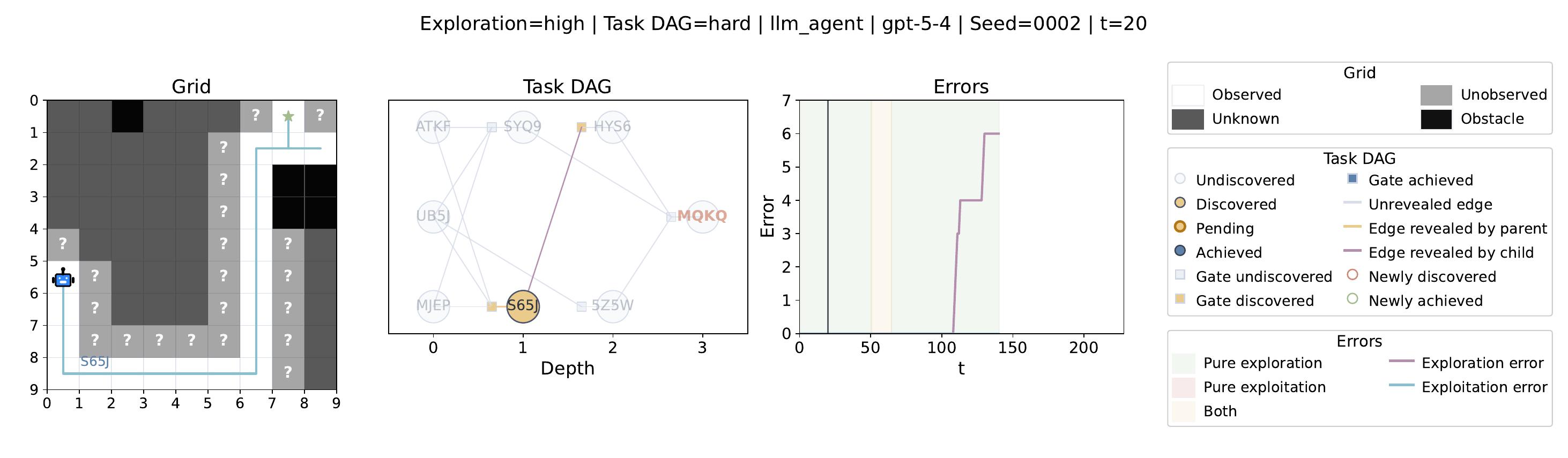}
        \caption{\currentmodel, $t = 20$}
        \end{subfigure}
        \begin{subfigure}[t]{\framewidth}
        \centering
        \includegraphics[width=\textwidth,trim={0 0 0 33pt},clip]{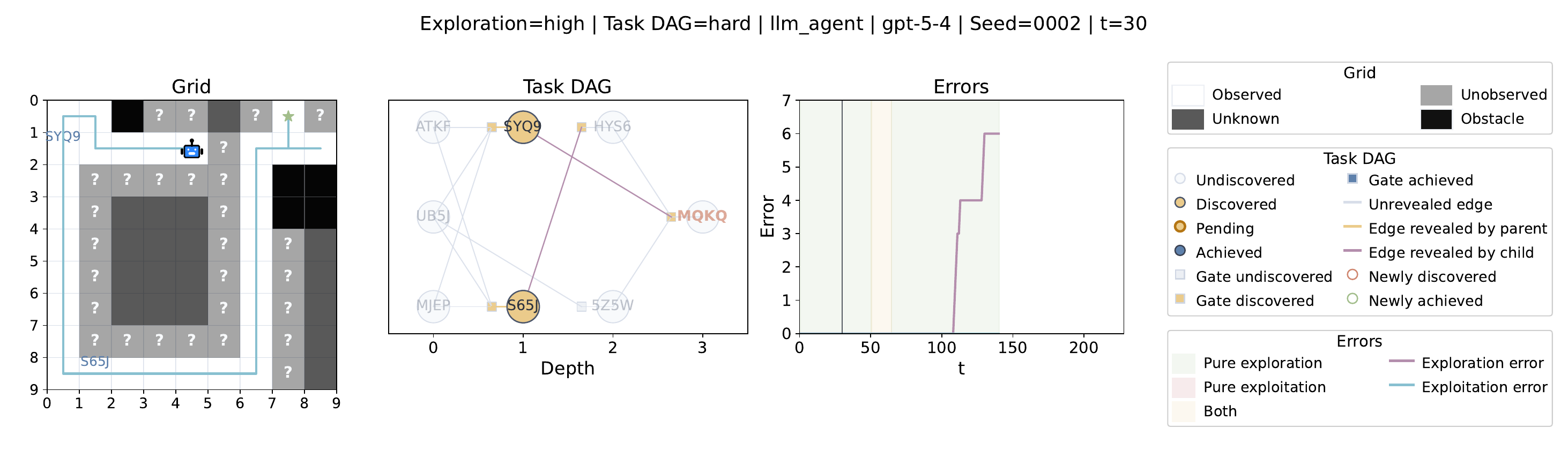}
        \caption{\currentmodel, $t = 30$}
        \end{subfigure}
        \begin{subfigure}[t]{\framewidth}
        \centering
        \includegraphics[width=\textwidth,trim={0 0 0 33pt},clip]{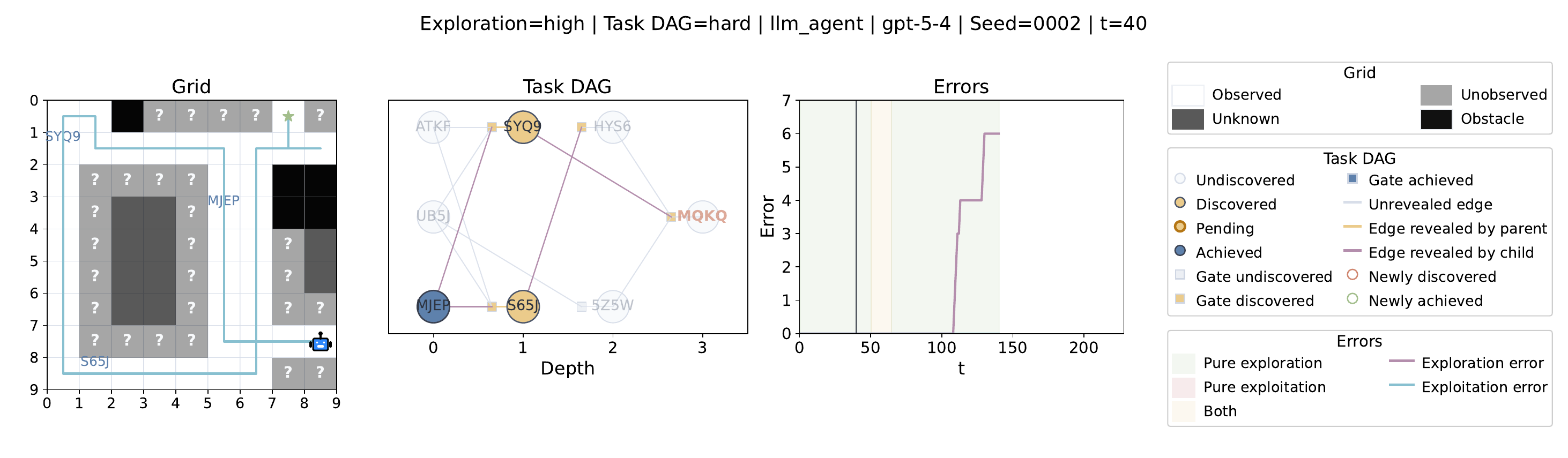}
        \caption{\currentmodel, $t = 40$}
        \end{subfigure}
    \end{center}
    \caption{Results of action trajectories and metric values over each timestep for~\currentmodel.}
\end{figure}

\begin{figure}[ht]
    \begin{center}
        \begin{subfigure}[t]{\framewidth}
        \centering
        \includegraphics[width=\textwidth,trim={0 0 0 33pt},clip]{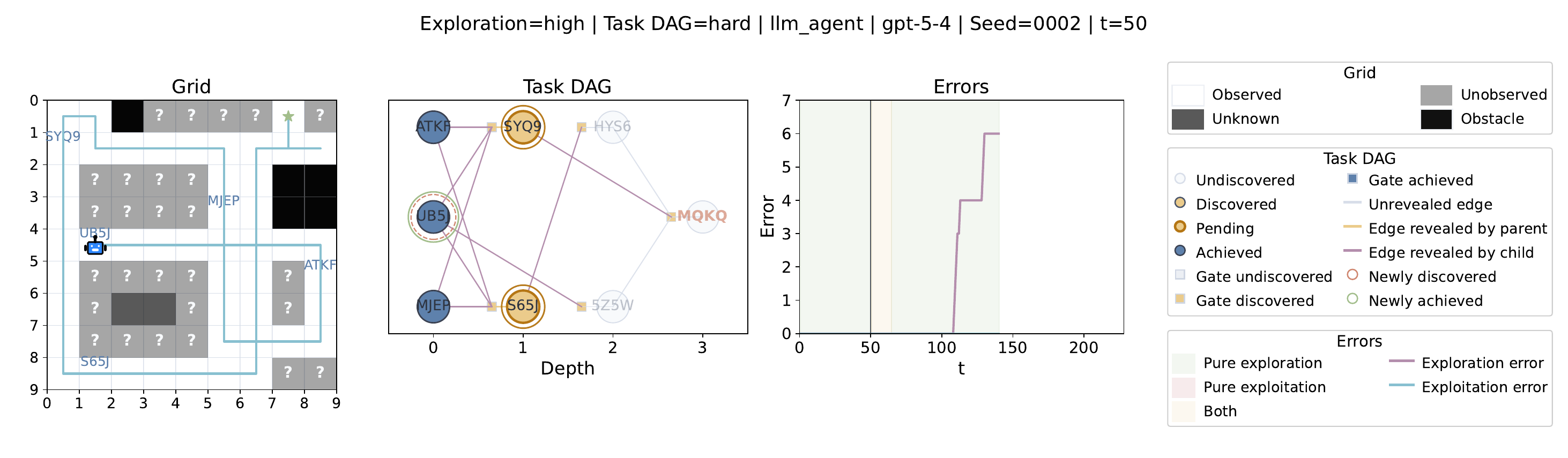}
        \caption{\currentmodel, $t = 50$}
        \end{subfigure}
        \begin{subfigure}[t]{\framewidth}
        \centering
        \includegraphics[width=\textwidth,trim={0 0 0 33pt},clip]{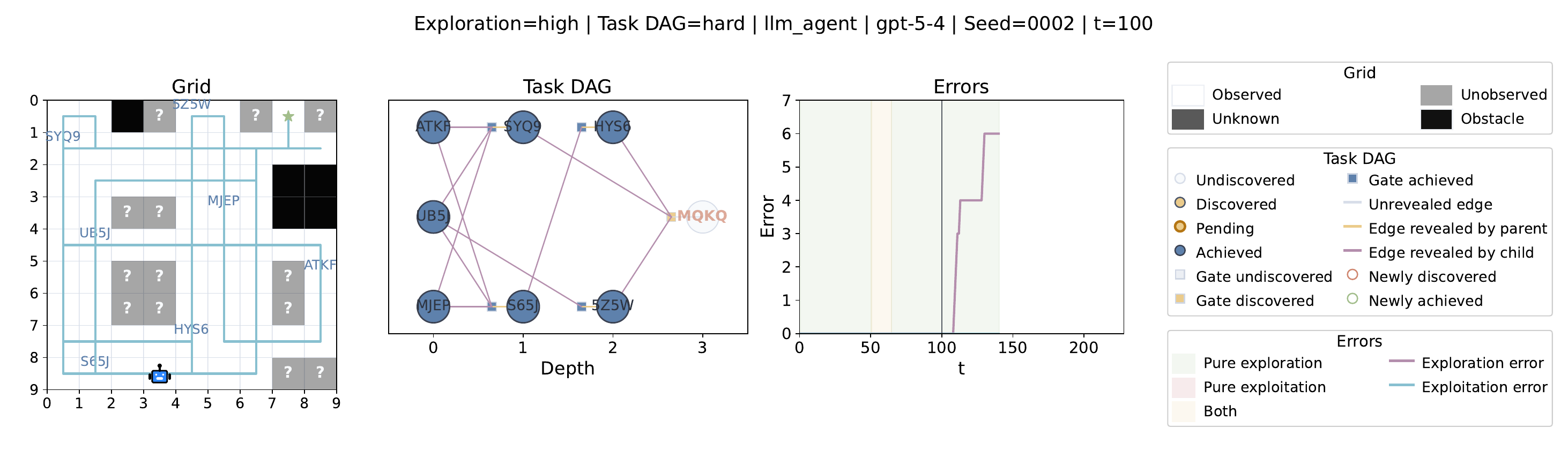}
        \caption{\currentmodel, $t = 100$}
        \end{subfigure}
        \begin{subfigure}[t]{\framewidth}
        \centering
        \includegraphics[width=\textwidth,trim={0 0 0 33pt},clip]{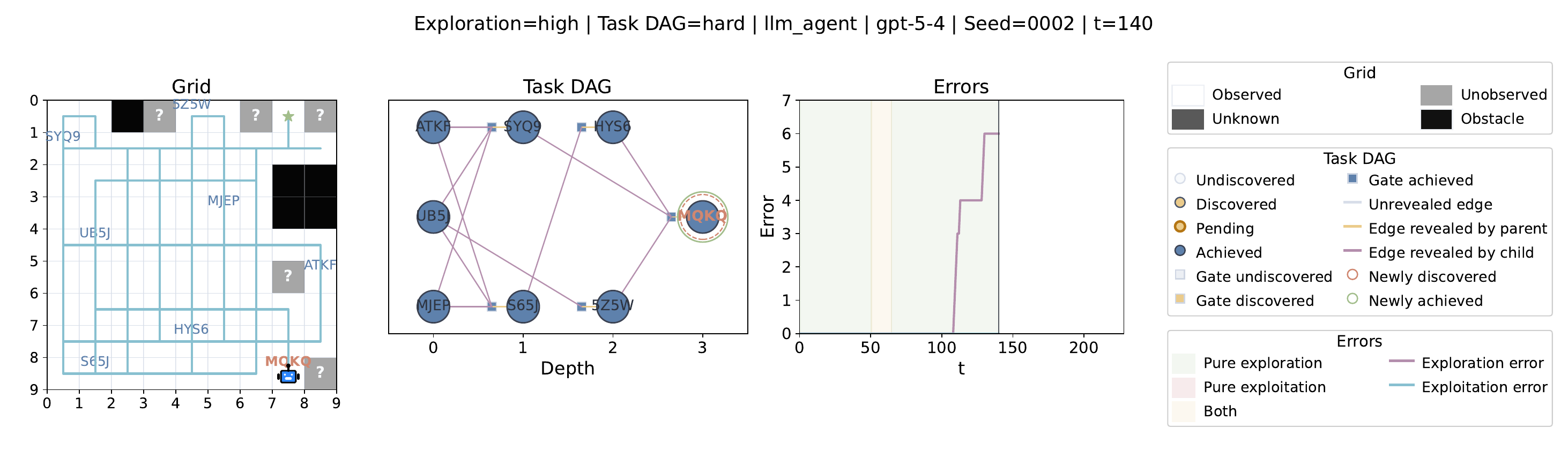}
        \caption{\currentmodel, $t = 140$}
        \end{subfigure}
    \end{center}
    \caption{Results of action trajectories and metric values over each timestep for~\currentmodel.}
\end{figure}

\clearpage

\renewcommand{\currentmodel}{\gptoss}

\begin{figure}[ht]
    \begin{center}
        \begin{subfigure}[t]{\framewidth}
        \centering
        \includegraphics[width=\textwidth,trim={0 0 0 33pt},clip]{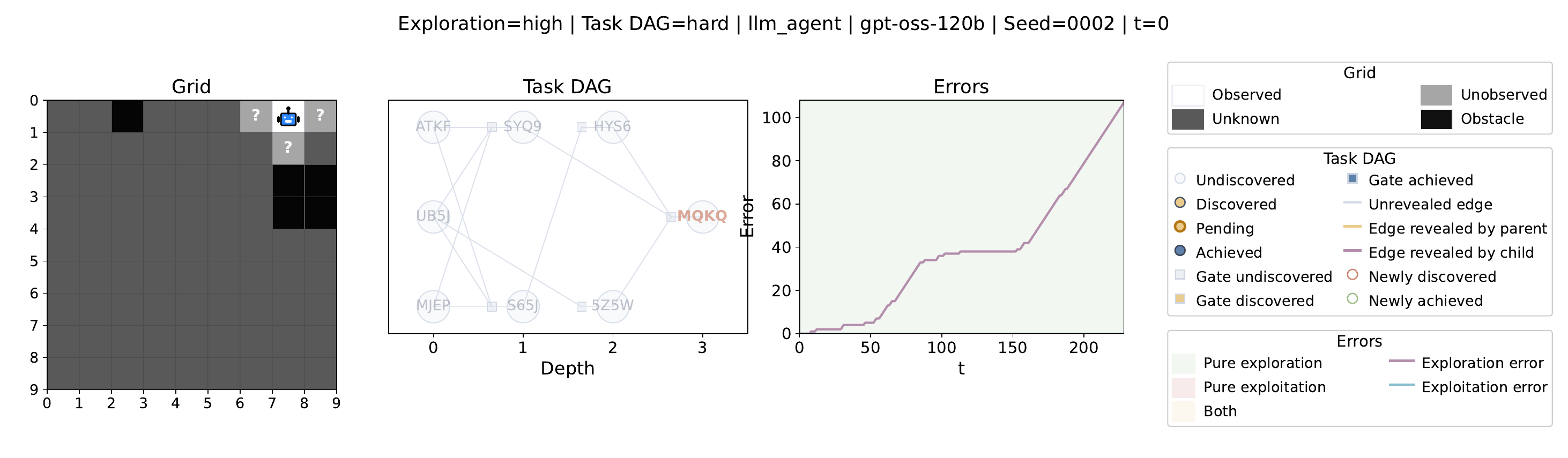}
        \caption{\currentmodel, $t = 0$}
        \end{subfigure}
        \begin{subfigure}[t]{\framewidth}
        \centering
        \includegraphics[width=\textwidth,trim={0 0 0 33pt},clip]{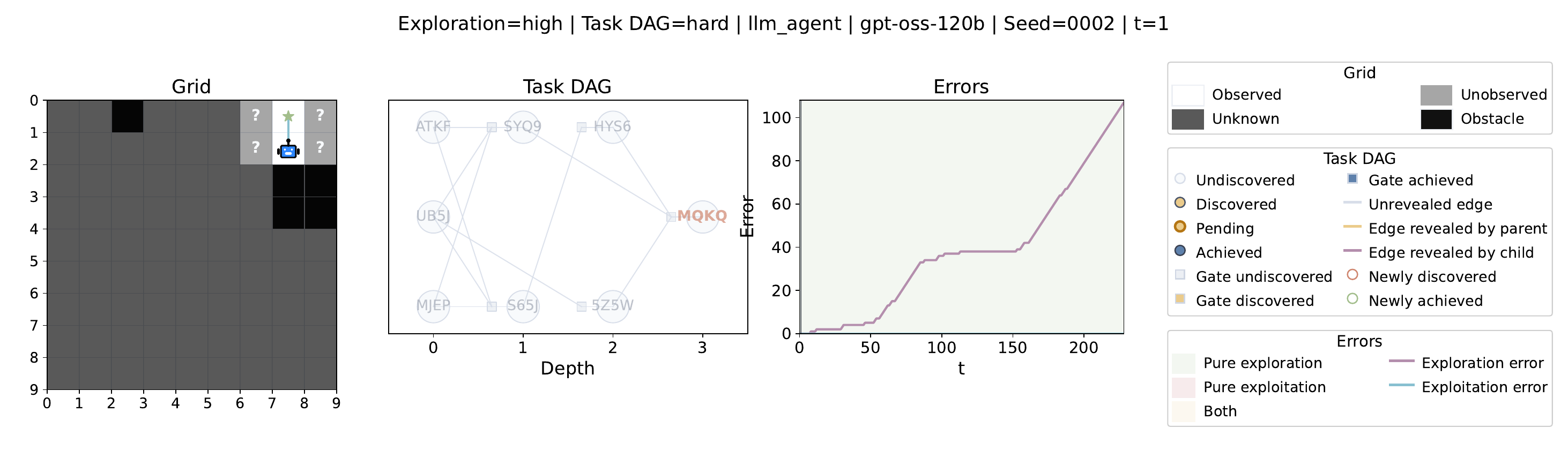}
        \caption{\currentmodel, $t = 1$}
        \end{subfigure}
        \begin{subfigure}[t]{\framewidth}
        \centering
        \includegraphics[width=\textwidth,trim={0 0 0 33pt},clip]{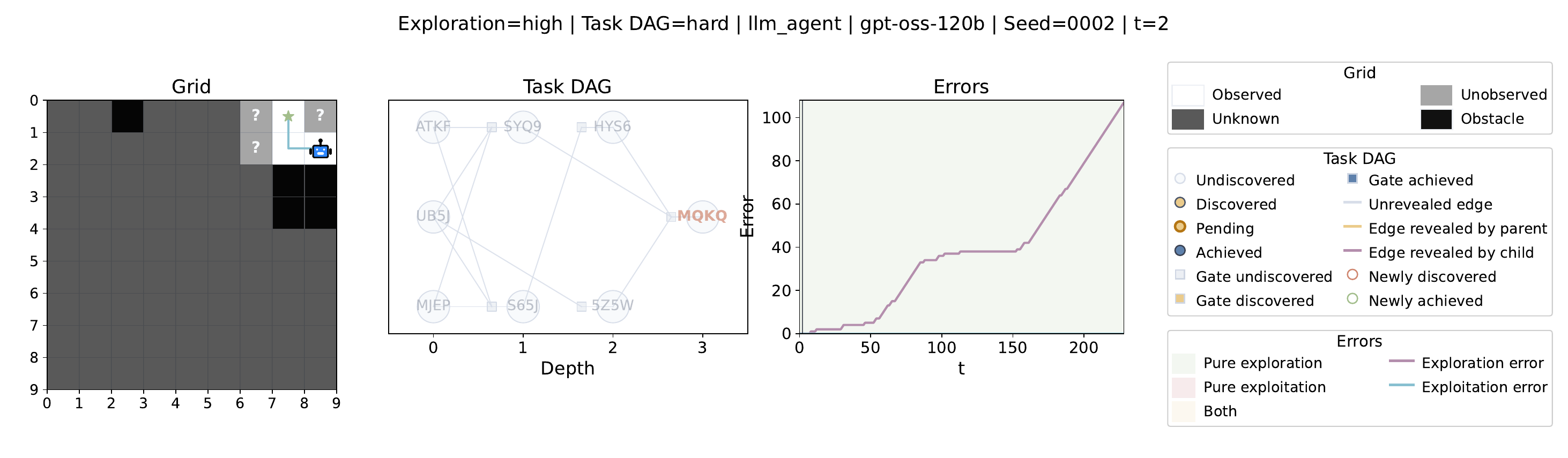}
        \caption{\currentmodel, $t = 2$}
        \end{subfigure}
        \begin{subfigure}[t]{\framewidth}
        \centering
        \includegraphics[width=\textwidth,trim={0 0 0 33pt},clip]{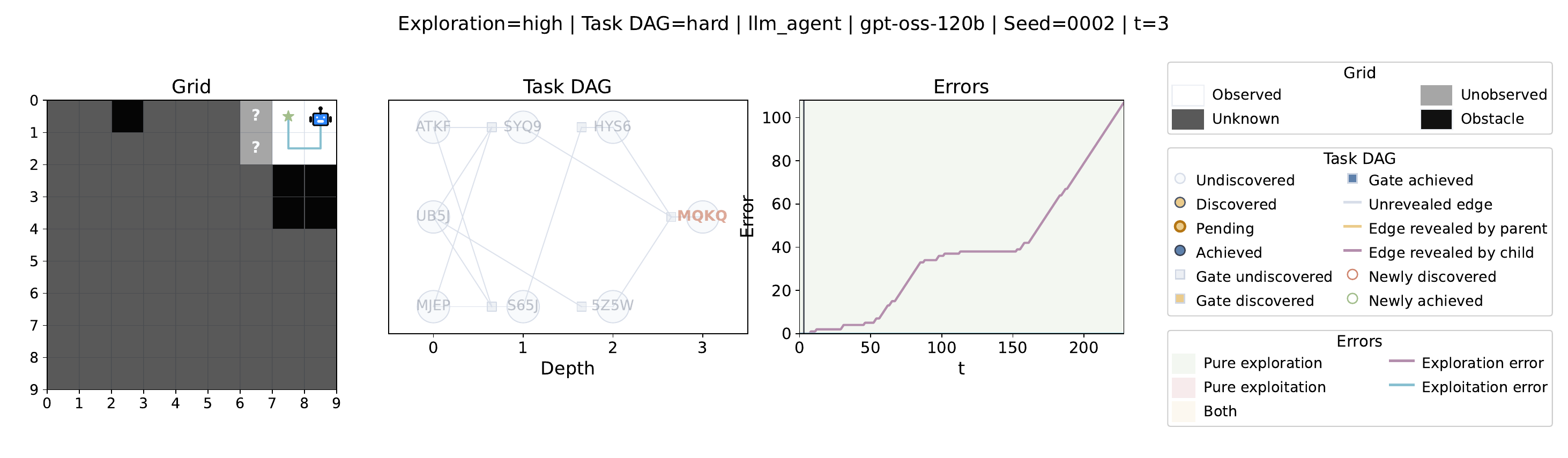}
        \caption{\currentmodel, $t = 3$}
        \end{subfigure}
        \begin{subfigure}[t]{\framewidth}
        \centering
        \includegraphics[width=\textwidth,trim={0 0 0 33pt},clip]{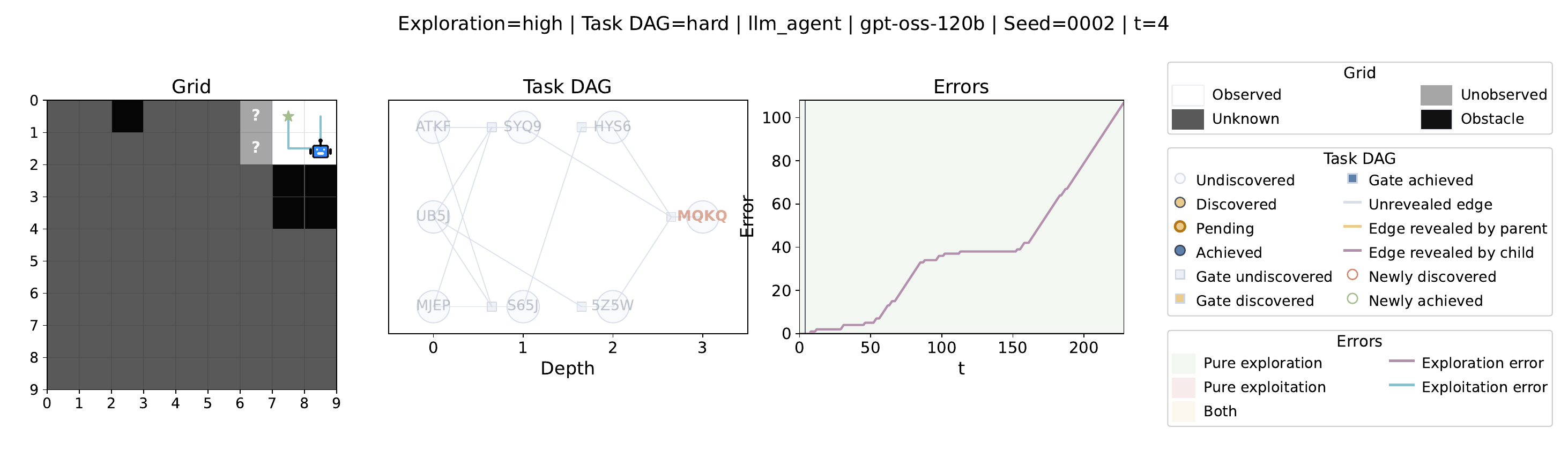}
        \caption{\currentmodel, $t = 4$}
        \end{subfigure}
    \end{center}
    \caption{Results of action trajectories and metric values over each timestep for~\currentmodel.}
\end{figure}

\begin{figure}[ht]
    \begin{center}
        \begin{subfigure}[t]{\framewidth}
        \centering
        \includegraphics[width=\textwidth,trim={0 0 0 33pt},clip]{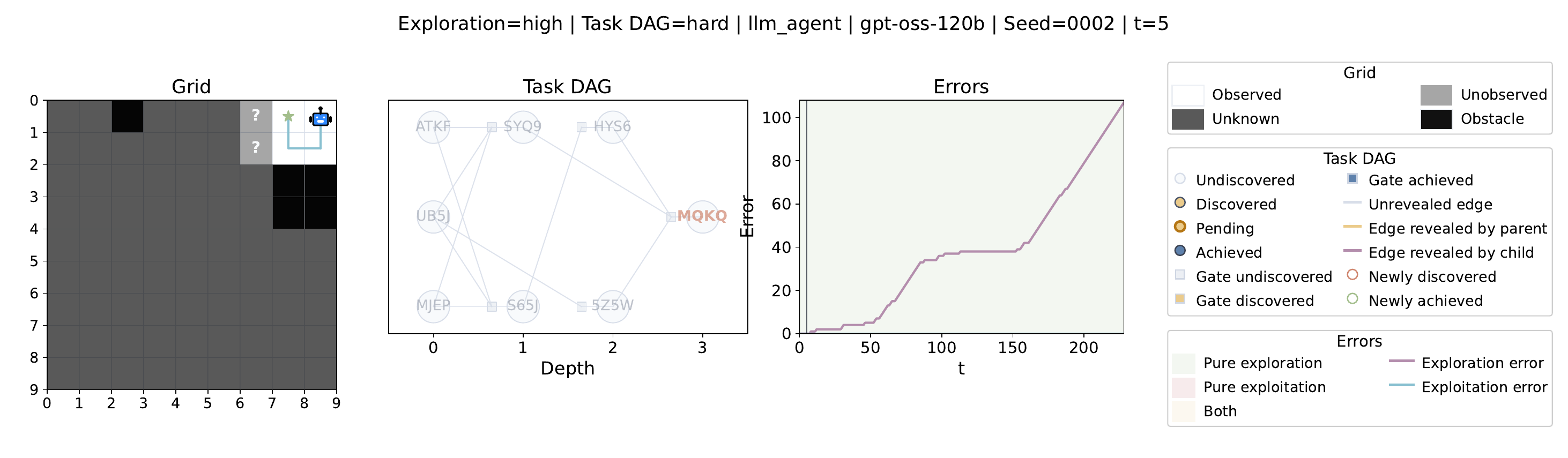}
        \caption{\currentmodel, $t = 5$}
        \end{subfigure}
        \begin{subfigure}[t]{\framewidth}
        \centering
        \includegraphics[width=\textwidth,trim={0 0 0 33pt},clip]{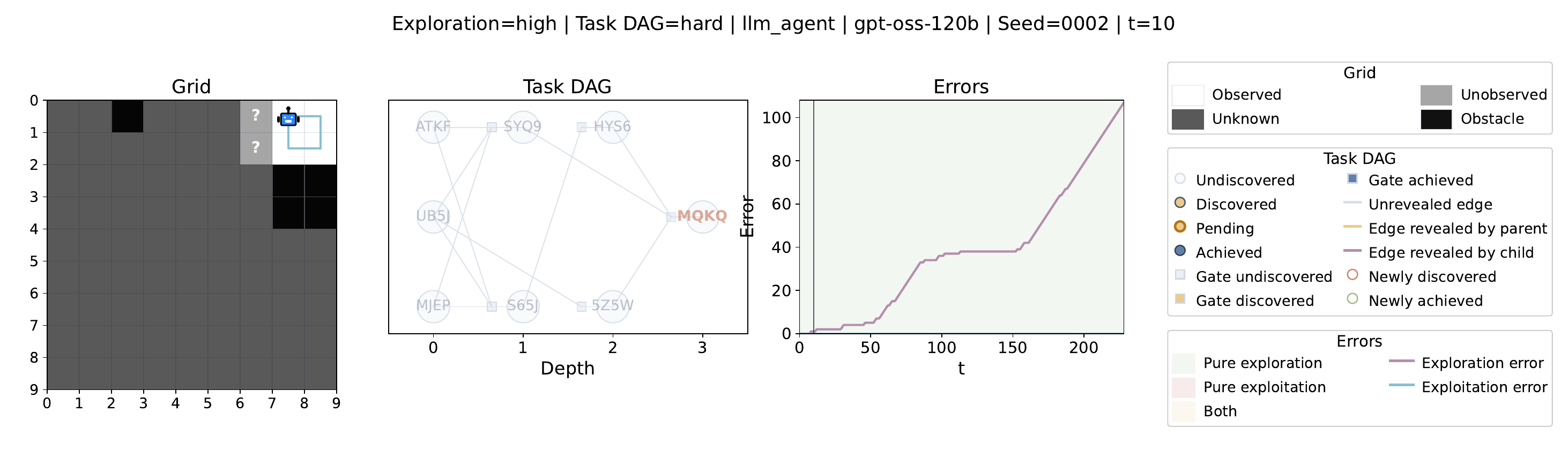}
        \caption{\currentmodel, $t = 10$}
        \end{subfigure}
        \begin{subfigure}[t]{\framewidth}
        \centering
        \includegraphics[width=\textwidth,trim={0 0 0 33pt},clip]{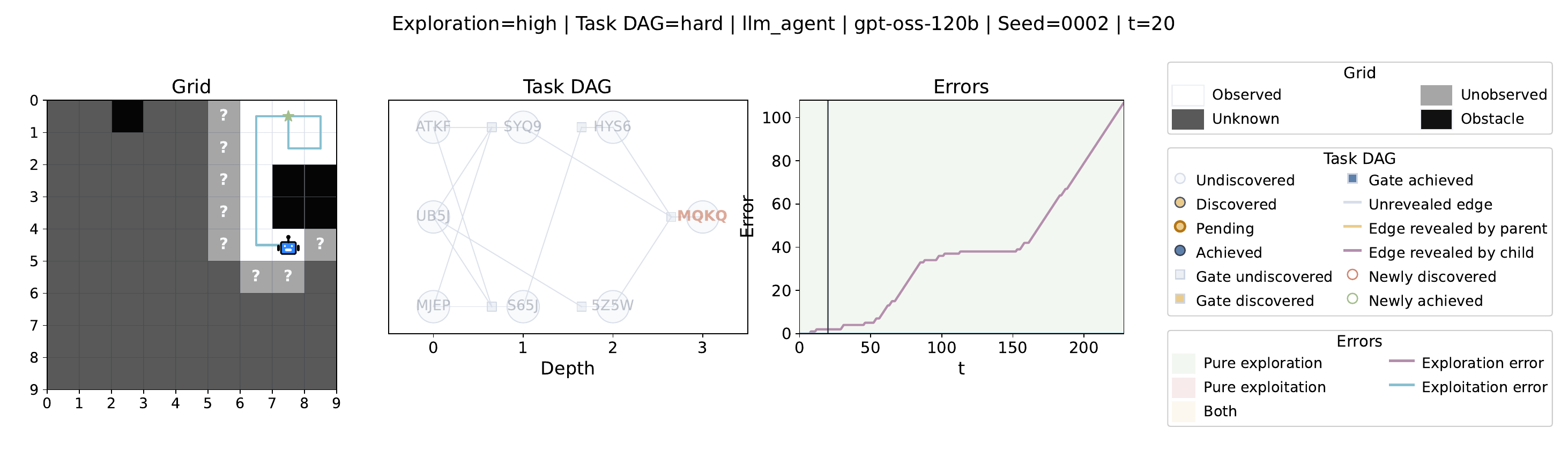}
        \caption{\currentmodel, $t = 20$}
        \end{subfigure}
        \begin{subfigure}[t]{\framewidth}
        \centering
        \includegraphics[width=\textwidth,trim={0 0 0 33pt},clip]{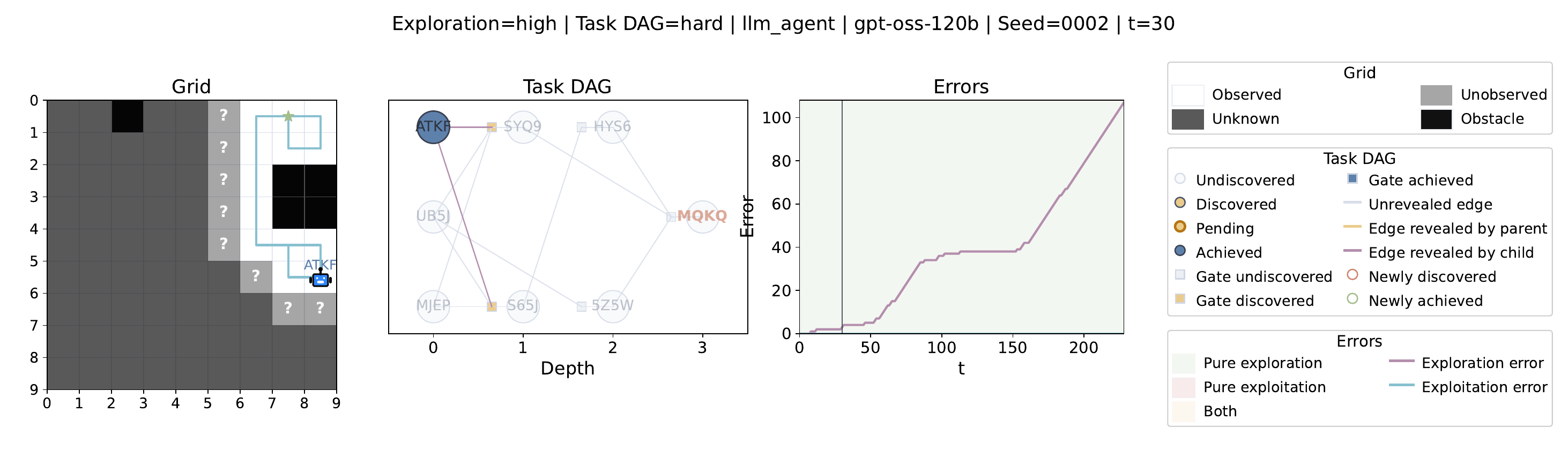}
        \caption{\currentmodel, $t = 30$}
        \end{subfigure}
        \begin{subfigure}[t]{\framewidth}
        \centering
        \includegraphics[width=\textwidth,trim={0 0 0 33pt},clip]{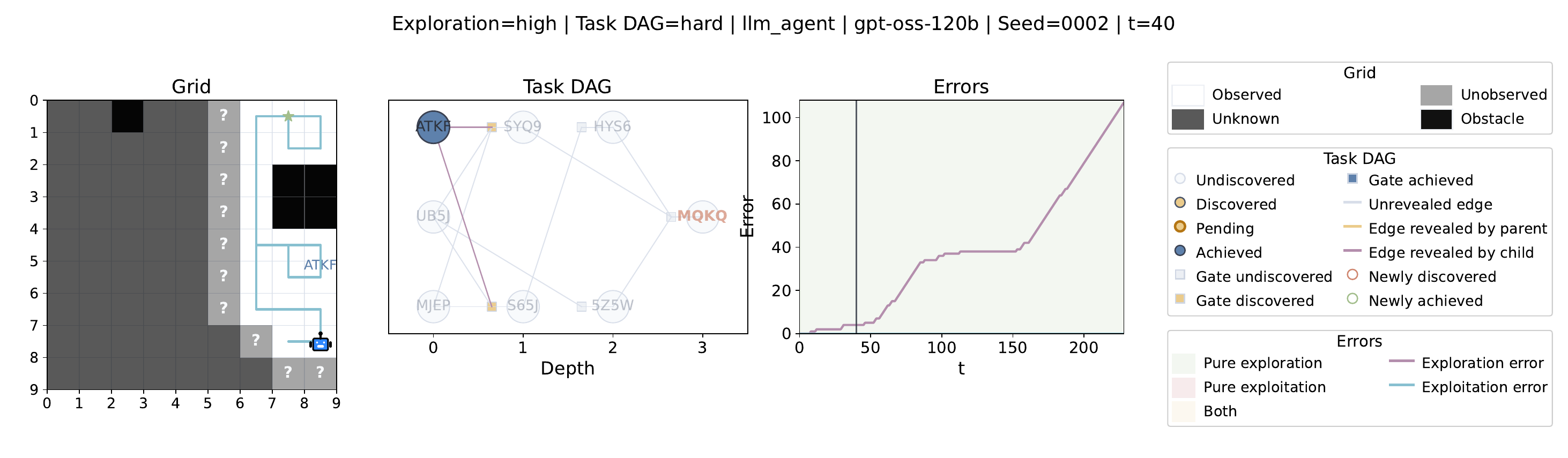}
        \caption{\currentmodel, $t = 40$}
        \end{subfigure}
    \end{center}
    \caption{Results of action trajectories and metric values over each timestep for~\currentmodel.}
\end{figure}

\begin{figure}[ht]
    \begin{center}
        \begin{subfigure}[t]{\framewidth}
        \centering
        \includegraphics[width=\textwidth,trim={0 0 0 33pt},clip]{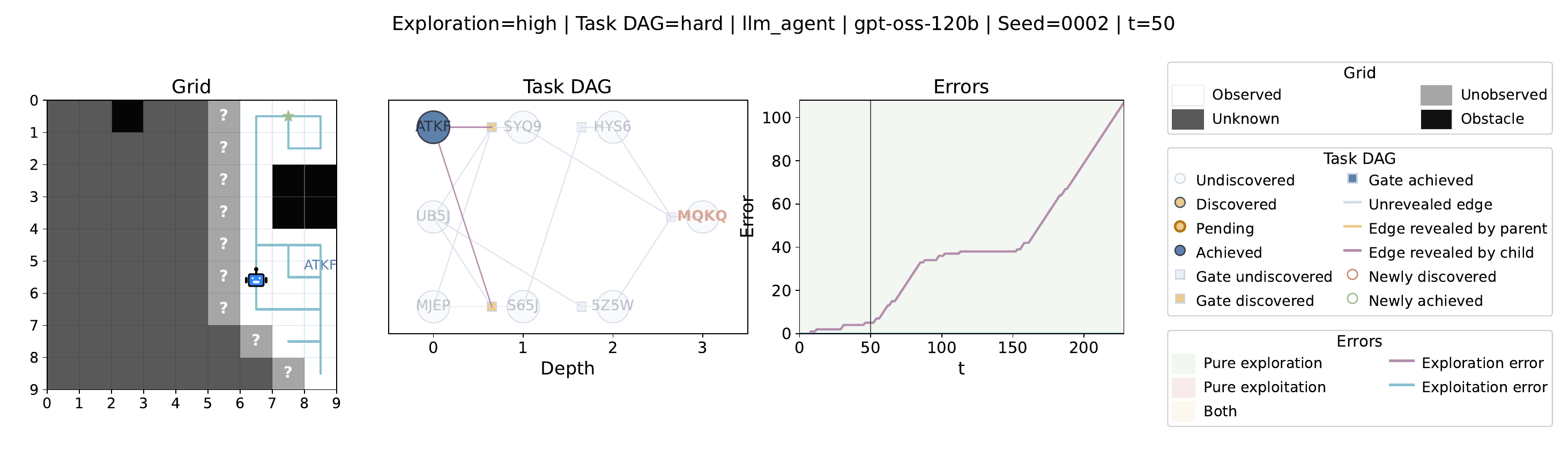}
        \caption{\currentmodel, $t = 50$}
        \end{subfigure}
        \begin{subfigure}[t]{\framewidth}
        \centering
        \includegraphics[width=\textwidth,trim={0 0 0 33pt},clip]{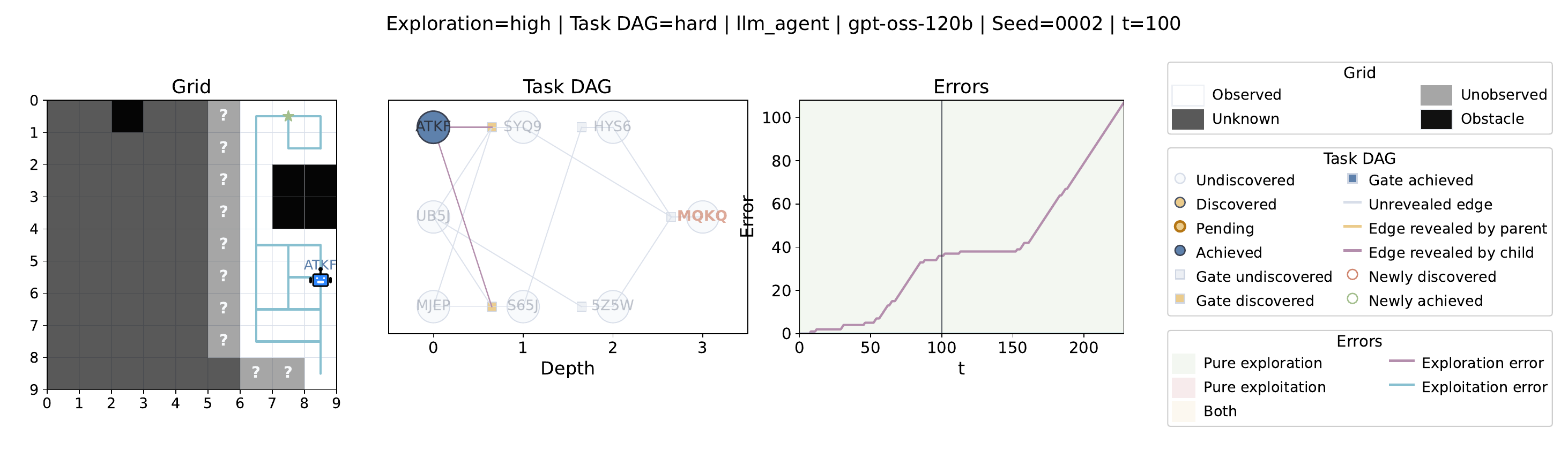}
        \caption{\currentmodel, $t = 100$}
        \end{subfigure}
        \begin{subfigure}[t]{\framewidth}
        \centering
        \includegraphics[width=\textwidth,trim={0 0 0 33pt},clip]{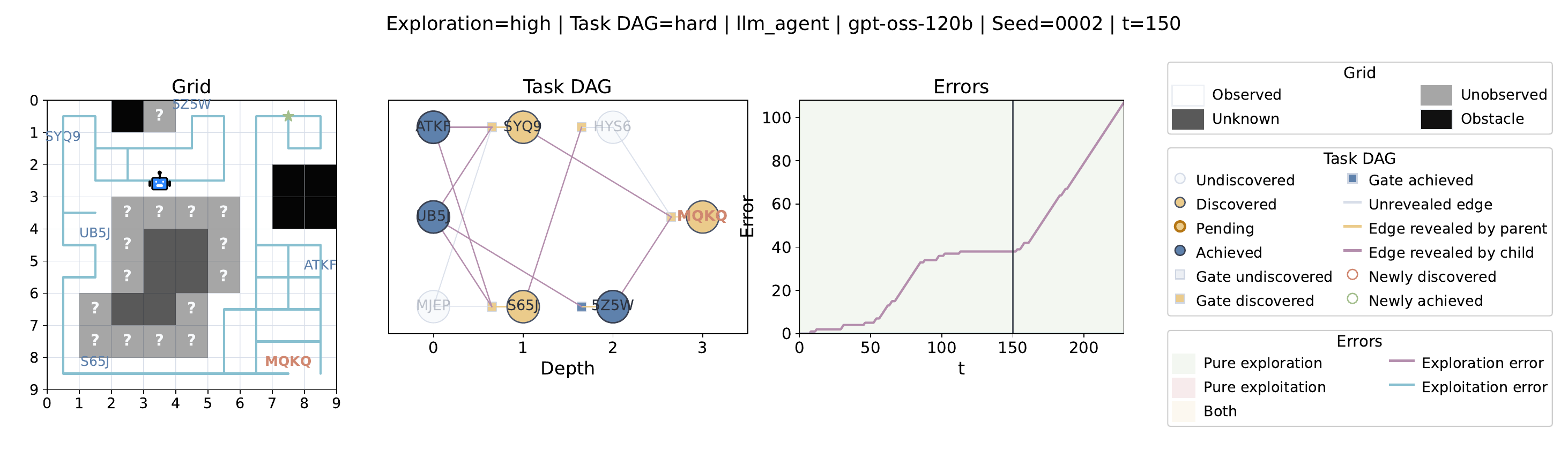}
        \caption{\currentmodel, $t = 150$}
        \end{subfigure}
        \begin{subfigure}[t]{\framewidth}
        \centering
        \includegraphics[width=\textwidth,trim={0 0 0 33pt},clip]{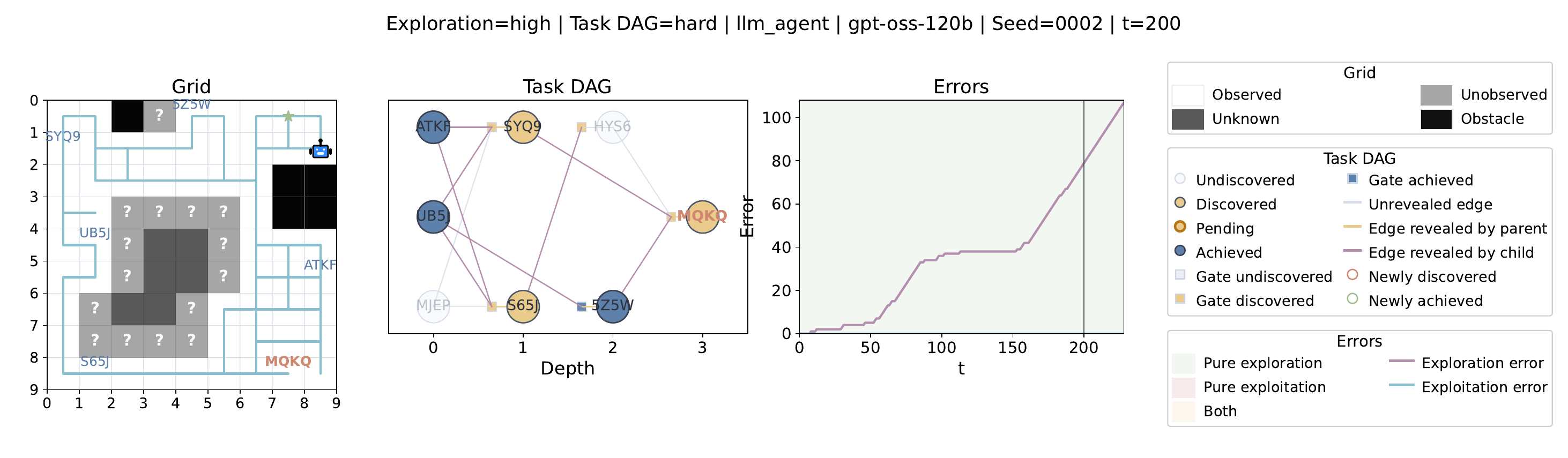}
        \caption{\currentmodel, $t = 200$}
        \end{subfigure}
        \begin{subfigure}[t]{\framewidth}
        \centering
        \includegraphics[width=\textwidth,trim={0 0 0 33pt},clip]{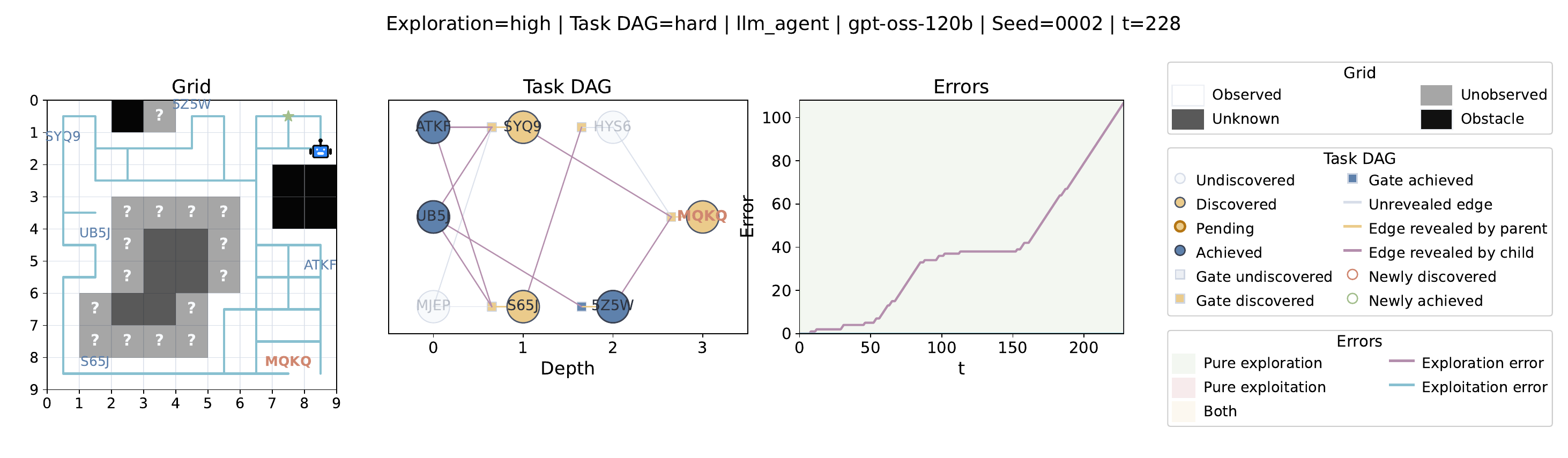}
        \caption{\currentmodel, $t = 228$}
        \end{subfigure}
    \end{center}
    \caption{Results of action trajectories and metric values over each timestep for~\currentmodel.}
    \label{fig:gpt-oss-120b-3}
\end{figure}

\clearpage

\section{Extension to Minecraft Clone}
\label{sec:app_video_games}

\begin{figure}[ht]
  \centering
  \begin{subfigure}[b]{0.49\textwidth}
    \centering
    \includegraphics[height=0.3\textheight]{figures/minecraft_gemini_3_1_flash_lite_step_000001.annotated.png}
    \caption{Seed 0, Gemini 3.1 Flash Lite, Step 1/32}
  \end{subfigure}
  \hfill
  \begin{subfigure}[b]{0.49\textwidth}
    \centering
    \includegraphics[height=0.3\textheight]{figures/minecraft_gemini_3_1_flash_lite_step_000032.annotated.png}
    \caption{Seed 0, Gemini 3.1 Flash Lite, Step 32/32}
  \end{subfigure}
  \vskip 10pt
  \begin{subfigure}[b]{0.49\textwidth}
    \centering
    \includegraphics[height=0.3\textheight]{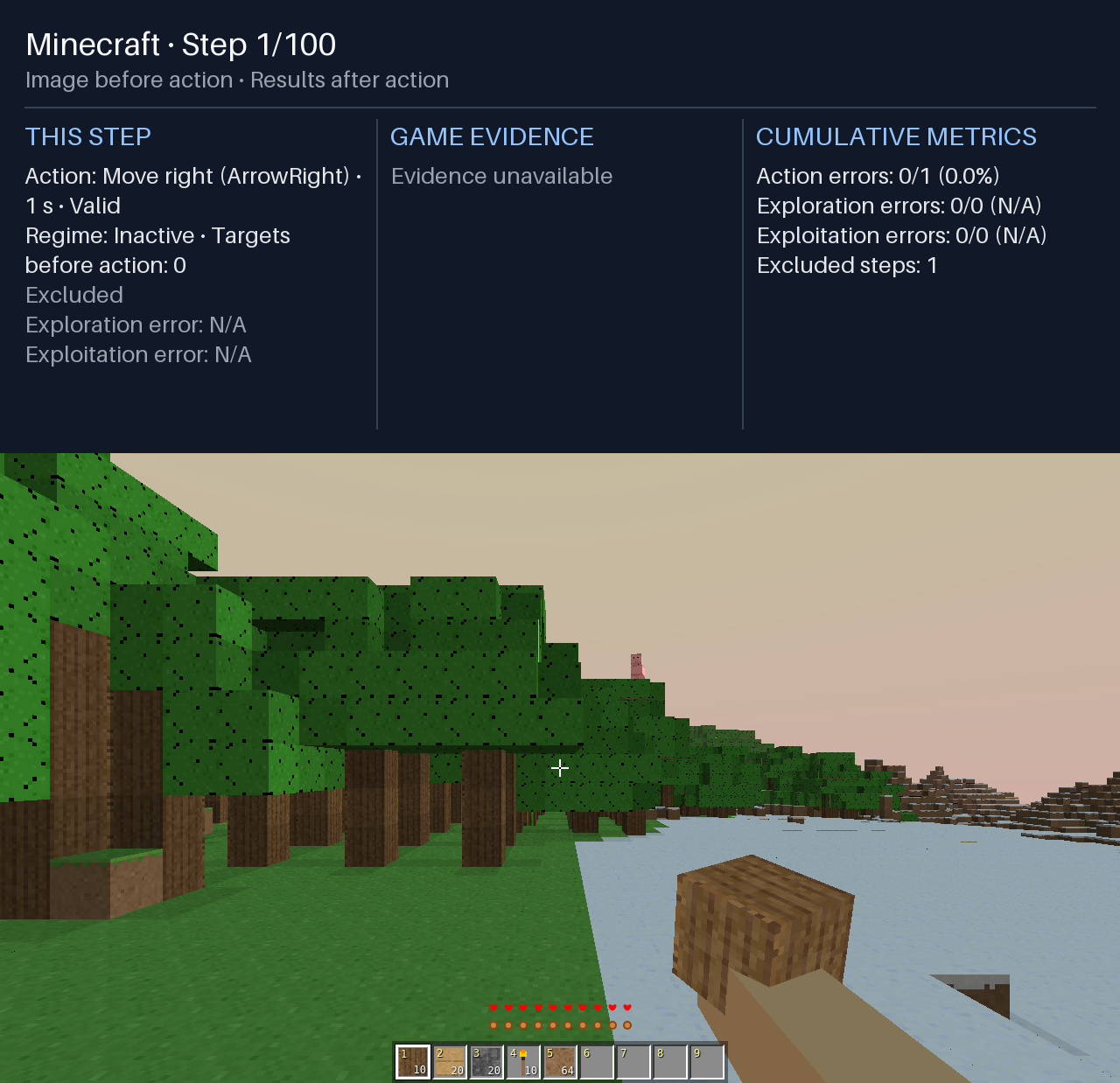}
    \caption{Seed 0, GPT-4.1, Step 1/100}
  \end{subfigure}
  \hfill
  \begin{subfigure}[b]{0.49\textwidth}
    \centering
    \includegraphics[height=0.3\textheight]{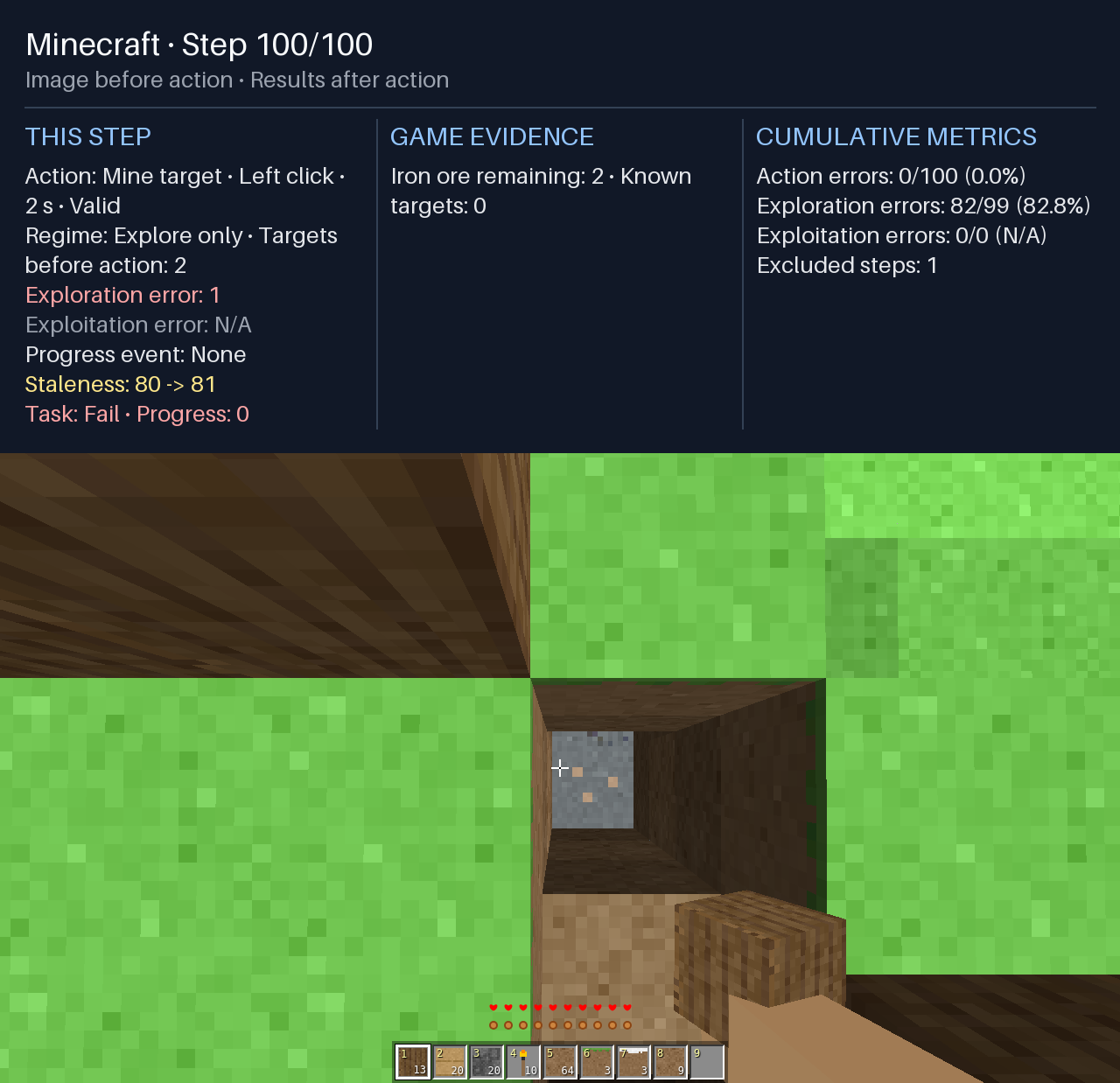}
    \caption{Seed 0, GPT-4.1, Step 100/100}
  \end{subfigure}
  \caption{Observation examples of Minecraft Clone.}
  \label{fig:more_video_game_frames}
\end{figure}

We use Minecraft Clone~\citep{ouyang2026gameworld} as a visually grounded 3D continuous environment to test the applicability of our metric beyond synthetic 2D grid environments. Minecraft Clone requires agents to navigate a three-dimensional block world to discover and collect resources. 

The actions the agents can take are discretized into 21 predefined actions: movement in four directions, camera adjustment in four directions, jumping, mining, block placement, selection of one of nine inventory slots, and waiting. Movements last for 1 second, jumping lasts 0.4 seconds and mining holds the left mouse button for 2 seconds. Camera adjustment applies a fixed mouse displacement with a 0.2 second action budget. The simulation advances only within these action budgets and remains frozen between actions, preventing model inference time from affecting execution.

Through the environment, the agents receive first-person screenshots, game rules, and task instructions but we prevent access to structured game state, hidden resource locations, or evaluation labels. We uses the game's natural starting configuration and require collecting at least two iron ore within 100 action budget, terminating runs early upon success.

We evaluate Gemini 3.1 Flash Lite and GPT-4.1 under two visual memory conditions: with and without visual memory. With visual memory enabled, the agents receive the current frame and up to two previous frames; with visual memory disabled, they receive only the current frame. Both conditions retain the prompts, actions, and rationales from the previous two decisions, isolating the contribution of historical screenshots. All conditions used the same task instructions and guidance to change approach after two attempts without visible progress. We run one episode per model and condition for each of nine environment seeds, 0 to 8, using the same seeds across conditions for 36 episodes in total.

We compute exploration and exploitation errors from the recorded trajectories in an offline manner. An iron block becomes a known target only after it is targeted by the crosshair; hidden resource scans does not contribute to this knowledge.

Exploration remains active while discovered yet uncollected blocks are insufficient to satisfy the remaining collection goal. Exploitation is active whenever a known target remains. Gain with respect to exploration reflects moving to a new position, changing the view point or discovering an ore. Exploitation gain reflects approaching a known ore position, changing view points in a productive manner, mining or acquiring iron. Based on these rules, we compute the stale score which is then used to flag errors. 

We report success rates across the nine seeds by weighting decisions equally regardless of action duration. Invalid actions and transitions without sufficient evidence, including the first recorded interaction, were excluded from these denominators. Rates with no scored opportunities were treated as undefined rather than zero.

\end{document}